\documentclass{article}
\PassOptionsToPackage{numbers}{natbib}

 \usepackage[main, final]{neurips_2026}

\usepackage[utf8]{inputenc} 
\usepackage[T1]{fontenc}    
\usepackage{hyperref}       
\usepackage{url}            
\usepackage{booktabs}       
\usepackage{amsfonts}       
\usepackage{nicefrac}       
\usepackage{microtype}      
\usepackage{xcolor}         
\usepackage{subcaption}
\usepackage{graphicx}
\usepackage{float}
\usepackage{amsmath}

\usepackage{algorithm}
\usepackage{algorithmic}
\usepackage{dsfont}
\usepackage{wrapfig}
\usepackage{needspace}

\usepackage{caption}
\usepackage{subcaption}
\usepackage{enumitem}

\newcommand{\R}{\mathbb{R}}
\newcommand{\A}{\ensuremath{\mathcal{A}}}
\renewcommand{\L}{\ensuremath{\mathcal{L}}}

\newcommand{\cossim}{\text{CosSim}}
\title{Laplacian Heads Improve Transformers \\ by Smoothing Token Representations}

\author{%
Yuchong Zhang, Vardan Papyan \\
University of Toronto, Vector Institute \\
\texttt{yuchongz.zhang@mail.utoronto.ca,}\texttt{vardan.papyan@utoronto.ca}
}

\begin{document}

\maketitle

\begin{abstract}

    Transformers update token representations through multi-head attention and residual connections: $X \leftarrow X + \sum_{i} P^{(i)}\text{Norm}(X)W_{V_i}W_{o_i}$, where $P^{(i)}$ is the softmax probability matrix in head $i$. We propose replacing some $P^{(i)}$'s with $I - P^{(i)}$, giving $X \leftarrow X + \sum_{i \in \mathcal{A}} P^{(i)}\text{Norm}(X)W_{V_i}W_{o_i} + \sum_{i \in \mathcal{L}} (I - P^{(i)})\text{Norm}(X)W_{V_i}W_{o_i}$. We call heads that use $I - P^{(i)}$ Laplacian heads. Our proposal has two motivations. First, it allows attention heads to update the mean of token representations, while Laplacian heads can directly control within-sequence variance. Second, 
    if tokens are viewed as nodes in a graph with normalized edge weights \(P^{(i)}\), then \(I - P^{(i)}\) is a graph Laplacian operator, and the update can be interpreted as one step of heat diffusion on the graph. We show that this simple modification improves performance across supervised learning, causal language modeling, and vision self-supervised learning tasks. To investigate why, we examine the token representations learned with and without Laplacian heads. In supervised learning, Laplacian heads collapse token representations within the same sequence and align the sequence means with the geometry of Neural Collapse. In language modeling, they increase the separability of token representations that share the same next-token prediction. In vision self-supervised learning, they produce token representations whose top principal components better capture segmentation structure. Across modalities, they also lead to faster-decaying spectra, indicating stronger token smoothing. Overall, our findings challenge the prevailing view that token oversmoothing is inherently harmful, showing instead that certain forms of smoothing can be beneficial.

\end{abstract}
\begingroup
\renewcommand{\thefootnote}{}
\footnotetext{Code is available at \url{https://github.com/yuchongzhang/Laplacian-heads}}
\endgroup
\begin{figure}[!h]
\vspace{-1.5em}
\centering 
\begin{subfigure}{0.38\linewidth} 
\centering \includegraphics[ width=\linewidth, trim=60 40 50 70, clip ]{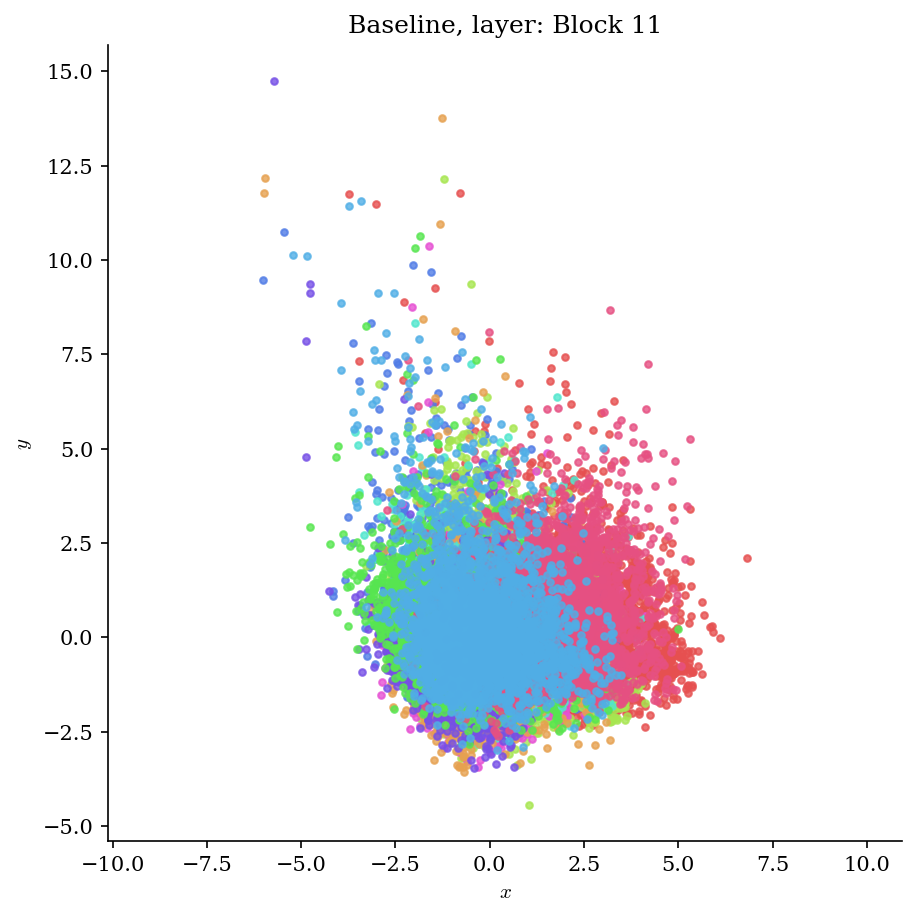} 
\caption{Baseline} 
\label{fig:pca_baseline} 
\end{subfigure} 
\begin{subfigure}{0.38\linewidth} \centering \includegraphics[ width=\linewidth, trim=50 70 20 30, clip ]{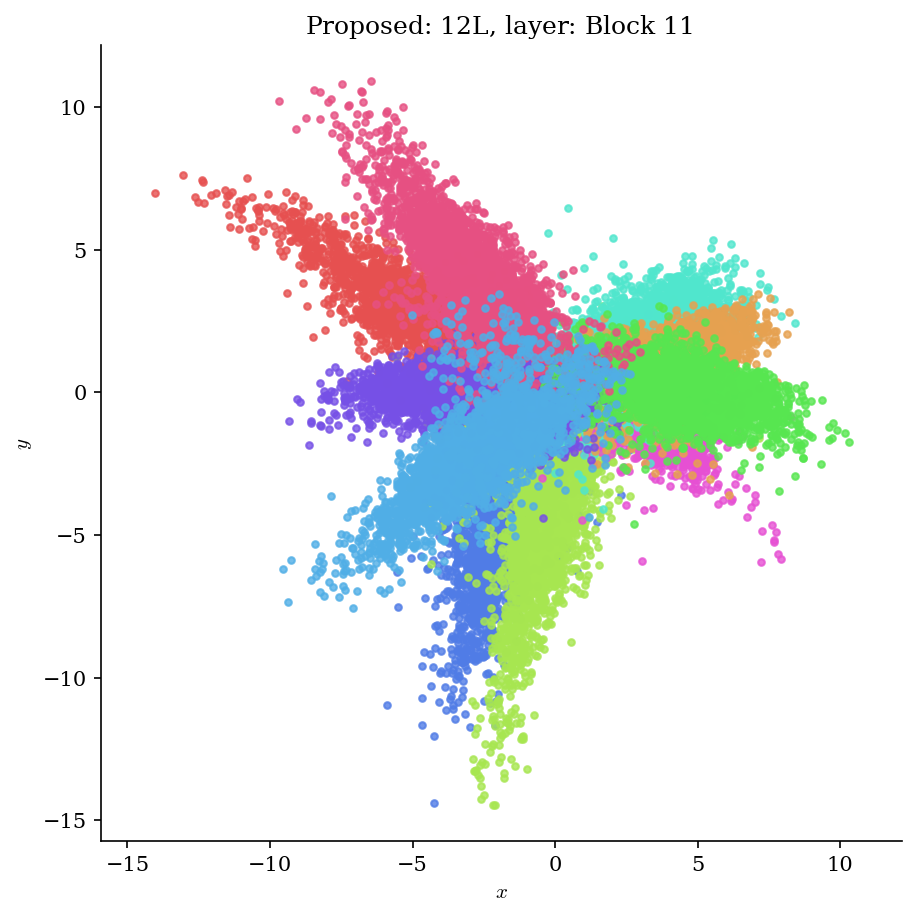} \caption{Proposed} \label{fig:pca_proposed} \end{subfigure} \caption{The proposed modification cleanly separates token representations into distinct clusters by their classes. Each color corresponds to a specific class.} \label{fig:pca_embeddings}
\end{figure}
\clearpage

\section{Introduction}\label{sec:introduction}
The transformer architecture \citep{vaswani2023attentionneed} processes a sequence of token representations $X \in \mathbb{R}^{T \times d}$ through a stack of layers. Each layer contains a multi-head attention module with $H$ heads. Head $i$ computes a softmax attention matrix
\begin{equation}
    P^{(i)} = \mathrm{softmax}\!\left( X W_{Q_i} W_{K_i}^\top X^\top / \sqrt{d_k} \right) \in \R^{T \times T},
\end{equation}
where $W_{Q_i}, W_{K_i} \in \R^{d \times d_k}$ are the query and key projections. The head then outputs $P^{(i)} X W_{V_i}$, where $W_{V_i} \in \R^{d \times d_h}$ is the value projection. The outputs of all $H$ heads are combined through the output projections $W_{o_i} \in \R^{d_h \times d}$ and added back to $X$ via a residual connection \citep{he2015deepresiduallearningimage}\footnote{The equations assume pre-layer normalization \citep{ba2016layernormalization, xiong2020layernormalizationtransformerarchitecture}.}:
\begin{equation}
    X \leftarrow X + \sum_{i=1}^{H} P^{(i)} \text{Norm}(X) W_{V_i} W_{o_i}.\label{eq:baseline}
\end{equation}

We propose a parameter-free modification to the transformer architecture: replace a subset of the $P^{(i)}$'s with the Laplacian operator $I - P^{(i)}$, where $I \in \R^{T\times T}$ is the identity matrix. This changes the update equation into
\begin{equation}
    X \leftarrow X + \sum_{i \in \A} P^{(i)}\text{Norm}(X)W_{V_i}W_{o_i} + \sum_{i \in \L} (I - P^{(i)})\text{Norm}(X)W_{V_i}W_{o_i}.\label{eq:proposed}
\end{equation}
where $\mathcal{A}$ denotes the set of attention heads and $\L$ denotes the set of Laplacian heads. 

Beyond the architectural modification itself, the paper makes three main contributions.
\begin{enumerate}[leftmargin=1.5em]
    \item We provide two motivations for Laplacian heads: they directly control within-sequence variance, and they act as diffusion operators on attention-induced token graphs (Section~\ref{sec:motivation}).
    \item We show that Laplacian heads consistently improve performance across supervised classification, causal language modeling, and vision self-supervised learning tasks (Section~\ref{sec:experiments}).
    \item We characterize the impact of Laplacian heads on token representations. In supervised classification, they collapse tokens within each sequence and move sequence means toward the geometry of Neural Collapse \citep{neuralcollapse} (Section~\ref{sec:analysis_img_cls}). In language modeling, they improve the separability of token representations with the same next-token target, leading to stronger linguistic collapse \citep{wu2024linguistic} (Section~\ref{sec:analysis_language}). In self-supervised learning with DINO \citep{caron2021emergingpropertiesselfsupervisedvision,simeoni2025dinov3}, they produce representations whose top principal components better capture segmentation structure. Across modalities, they also produce representations with faster spectral decay (Sections~\ref{sec:analysis_language} and~\ref{sec:analysis_dino}).
\end{enumerate}

\section{Motivation}\label{sec:motivation}
In this section, we motivate our proposal from two perspectives.
\subsection{Motivation 1: Direct Control of Mean and Within-Sequence Variance}\label{sec:motivation1}
 In Equation~\ref{eq:proposed}, $P^{(i)}$ is row-stochastic. Hence $P^{(i)}XW_{V_i}W_{o_i}$ computes for each token an attention-weighted average of the projected token representations $XW_{V_i}W_{o_i}$. Adding $P^{(i)}XW_{V_i}W_{o_i}$ to $X$ (through the residual connection) then updates the sequence mean. On the other hand, $(I - P^{(i)})XW_{V_i}W_{o_i}$ computes the difference between each projected token representation and its corresponding attention-weighted average. Thus, adding $(I - P^{(i)})XW_{V_i}W_{o_i}$ to $X$ directly controls the within-sequence variance.
 Therefore, using both attention heads $P^{(i)}$ and Laplacian heads $(I - P^{(i)})$ gives the model explicit mechanisms for modifying both the sequence mean and within-sequence variance. In contrast, the standard transformer update in Equation~\ref{eq:baseline} does not explicitly include a term based on each token’s deviation from its attention-weighted average. Empirical evidence supporting this interpretation is in Appendix~\ref{app:out_snr}.

 \begin{figure}[!htbp]
    \centering
    \begin{subfigure}[t]{0.34\textwidth}
        \centering
        \includegraphics[
            width=\linewidth,
            trim={0cm 0cm 1cm 5cm},
            clip
        ]{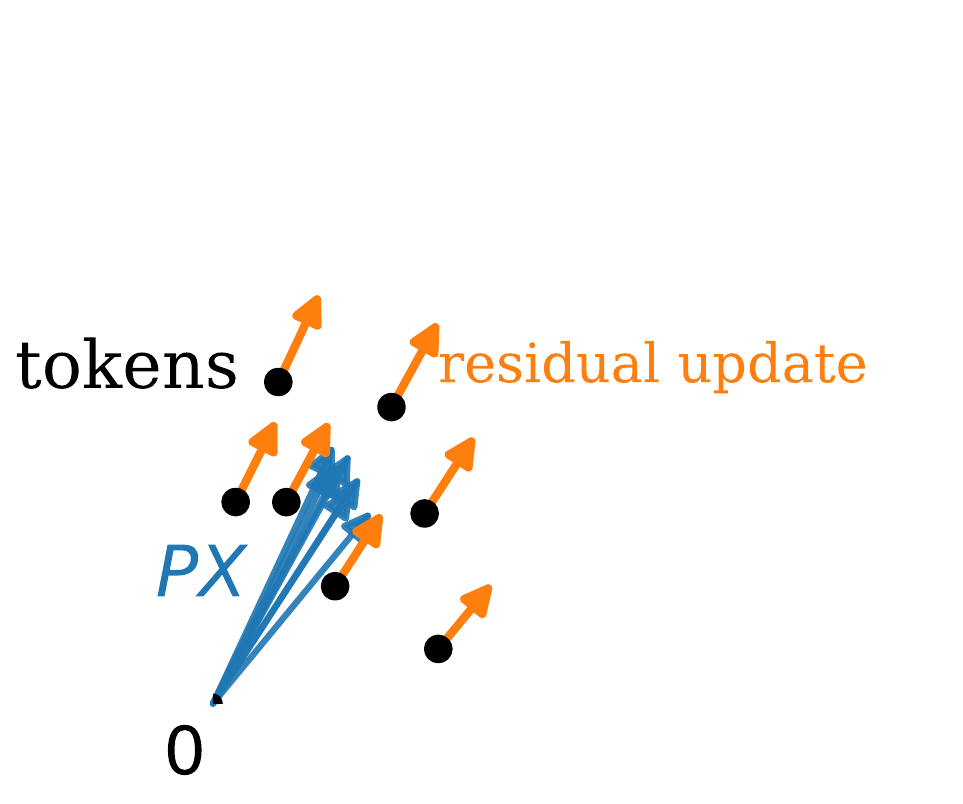}
        \caption{Attention heads update mean.}
        \label{fig:attention-along-mean}
    \end{subfigure}%
    \hspace{0.08\textwidth}%
    \begin{subfigure}[t]{0.34\textwidth}
        \centering
        \includegraphics[
            width=\linewidth,
            trim={0cm 0cm 1cm 5cm},
            clip
        ]{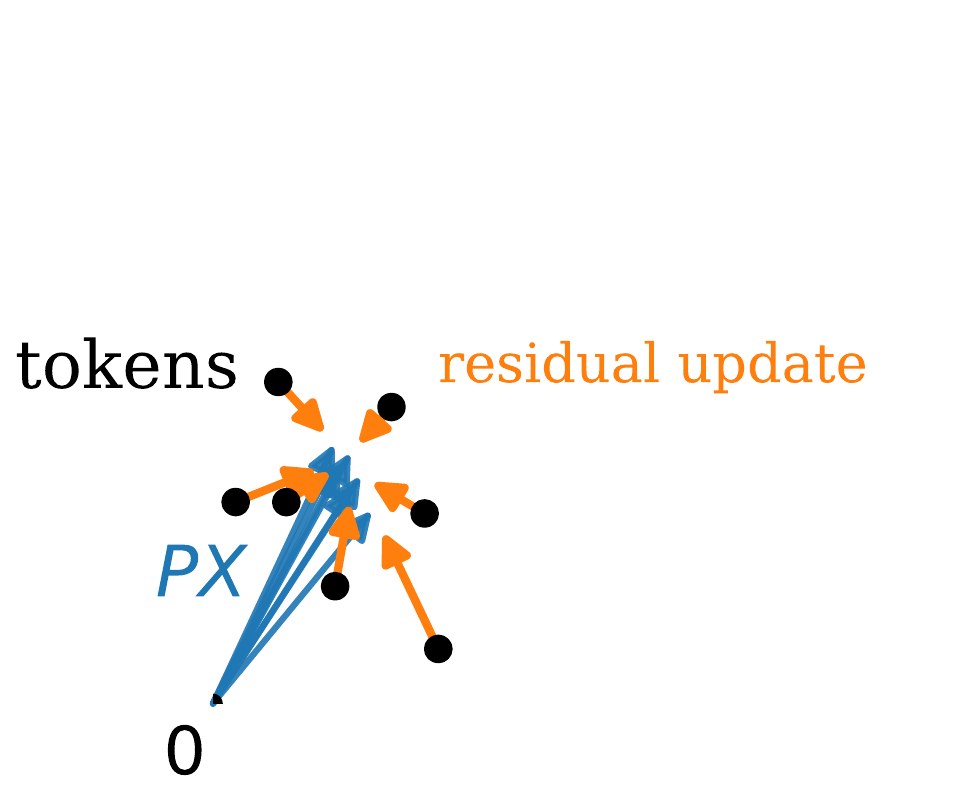}
        \caption{Laplacian heads control variance.}
        \label{fig:laplacian-toward-mean}
    \end{subfigure}
    \caption{Combining attention and Laplacian heads lets the model directly control the mean and within-sequence variance of token representations in each head's subspace.}
    \label{fig:laplacian-cartoon}
\end{figure}

 We illustrate the effects described above in Figure~\ref{fig:laplacian-cartoon}. For attention heads (Figure~\ref{fig:attention-along-mean}), each token (black dots) receives an update (orange arrows) whose direction is parallel to some attention-weighted mean (blue arrows). In contrast, for Laplacian heads (Figure~\ref{fig:laplacian-toward-mean}), each token receives an update pointing towards its corresponding attention-weighted mean.

\subsection{Motivation 2: Diffusion over Graphs}
A second way to motivate Laplacian heads is through diffusion over graphs. Consider the tokens in a sequence as the nodes of a graph $G=(V,E)$ with $|V| = T$. Each attention matrix $P^{(i)} \in \R^{T \times T}$ defines weighted, directed edges, with $P^{(i)}_{jk}$ encoding the influence of token $k$ on token $j$. Since each row of $P^{(i)}$ sums to one, $P^{(i)}$ is a row-stochastic adjacency matrix. $I - P^{(i)}$ is the corresponding random-walk normalized graph Laplacian. Applied to a signal
$x \in \mathbb{R}^T$ on the tokens, this operator measures the deviation of each token value from the
attention-weighted average of its neighbors:
\[
\big((I-P^{(i)})x\big)_j
=
x_j - \sum_{k=1}^T P^{(i)}_{jk}x_k .
\]
This is the exact same quantity that appears in the heat equation over a graph. If $x(t)$ is a time-dependent signal on the vertices, graph heat flow is given by

\[
\frac{d}{dt}x(t) = -(I-P^{(i)})x(t).
\]

The negative Laplacian moves each node value toward the average of its neighbors, progressively
smoothing the signal over the graph. A single Euler step with step size $\Delta t=1$ gives
\[
x_{t+1}
=
x_t - (I-P^{(i)})x_t
=
P^{(i)}x_t .
\]

A Laplacian head applies the same operator $I - P^{(i)}$ to the projected and normalized token representations $\text{Norm}(X)W_{V_i}W_{o_i}$:
\[
(I-P^{(i)})\text{Norm}(X)W_{V_i}W_{o_i}.
\]
This quantity is added to $X$ through the residual connection, and the learned value and output projections determine the sign of this residual update. When they reverse the sign, the update amounts to taking a diffusion step over the attention-induced token graph in the learned subspace determined jointly by $W_{V_i}$ and $W_{O_i}$. Across layers, these diffusion steps reduce the differences between tokens connected by large attention weights, causing their representations to become increasingly similar in the corresponding subspaces \citep{chung1997spectral,kondor2002diffusion}.

Thus, the graph-diffusion perspective predicts that Laplacian heads would induce smoothing among token representations across layers, an effect we examine in detail later.

\section{Experiments}\label{sec:experiments}
In all experiments, we use a simple strategy to incorporate Laplacian heads into transformers. We set the number of Laplacian heads $|\L|$ to be fixed and the number of attention heads $|\A|$ to be $H - |\L|$ in every layer. On each task, we train models with various $|\L|$ using the exact same training recipe. 

\subsection{Image Classification}
We used the DeiT-3 family of vision transformers \citep{Touvron2022DeiTIR}, a strong baseline for image classification. We focused on the ViT-B model, which has 12 attention heads. Following the strategy above, we trained models with $|\L| \in \{0, 3, 6, 9, 11, 12\}$ Laplacian heads in each layer, where $|\L| = 0$ corresponds to the baseline ViT-B. For brevity, we will refer to the baseline as ``ViT-B'' and the model with $k$ Laplacian heads as ``ViT-B-$k$L''. We trained all models on CIFAR10, CIFAR100 \citep{krizhevsky2009cifar10}, and ImageNet-1k \citep{deng2009imagenet}. 

DeiT-3 employs stochastic depth\citep{huang2016deepnetworksstochasticdepth}, a regularization technique where each block is skipped with some probability $0 < p < 1$ (called the drop path rate). For each dataset, we trained models using drop path rates $p \in \{0.1, 0.3, 0.4\}$. Full details of the training setup are in Appendix~\ref{app:exp}.

Figure~\ref{fig:acc-dp-p-head} plots top-1 accuracy against the number of Laplacian heads. Adding Laplacian heads improves performance over the baseline for every choice of $|\L|$, including a $>1\%$ improvement on ImageNet and an almost $2\%$ improvement on CIFAR100. The improvement is also nearly monotonic in the number of Laplacian heads, except for ImageNet under strong stochastic depth regularization (drop path rate = 0.4), where the optimal $|\L|$ is 3. Even then, adding more Laplacian heads improves performance: Table~\ref{tab:top1-accuracy} shows that ViT-B-12L achieves 0.87\% higher accuracy than the baseline.

\begin{figure}[H]
  \centering
  \includegraphics[width=0.95\textwidth]{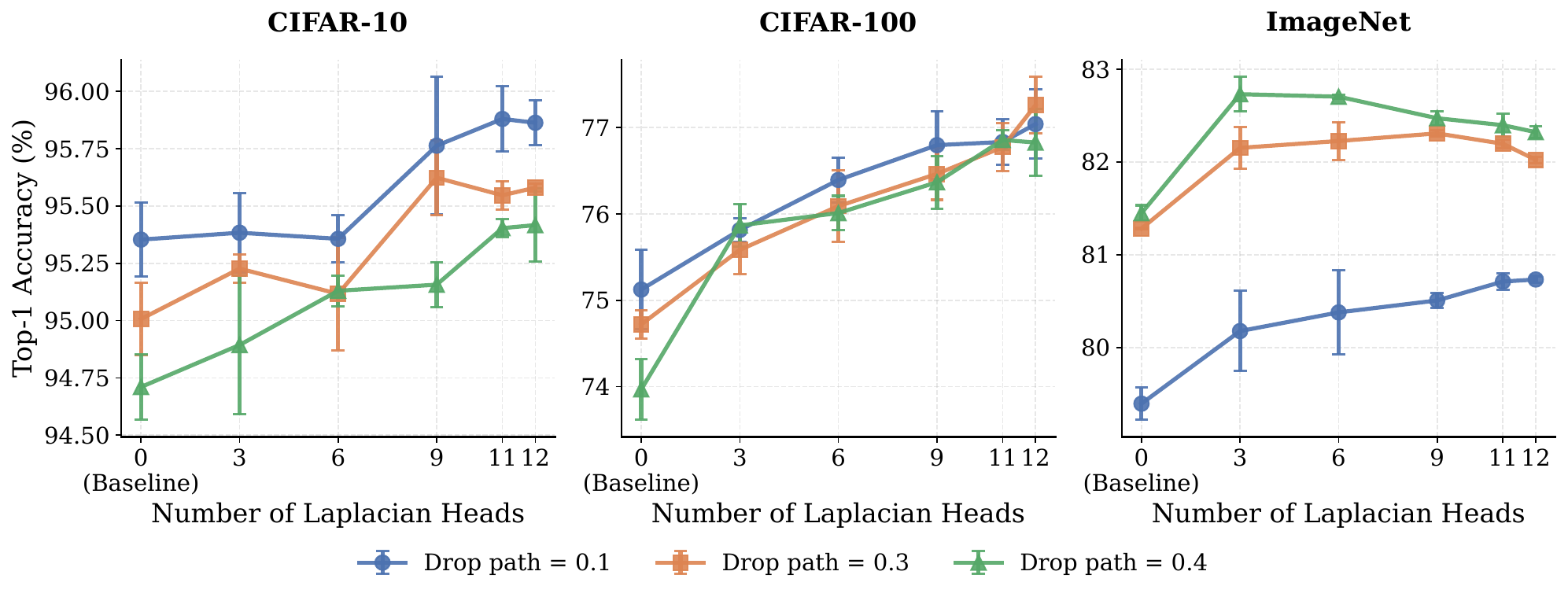}
  \caption{Adding Laplacian heads improves accuracy across all datasets and regularization strengths, with gains nearly monotonic in $|\L|$. Metrics are computed using 3 random seeds.}
  \label{fig:acc-dp-p-head}
\end{figure}

\begin{table}[!htbp]
  \centering
  \caption{Top-1 accuracy (\%, mean ± std) at the drop path rate that maximized baseline performance on each dataset. Best in bold, second best underlined. Metrics are across 3 random seeds.}
  \label{tab:top1-accuracy}
  \vspace{0.3em}
  \setlength{\tabcolsep}{4pt} 
  \begin{tabular}{@{}lccc@{}}
    \toprule
    \textbf{Model} 
    & \textbf{CIFAR10} 
    & \textbf{CIFAR100} 
    & \textbf{ImageNet} \\
    \midrule
    Baseline (no Laplacian)
      & $95.35 \pm 0.16$ 
      & $75.12 \pm 0.46$ 
      & $81.45 \pm 0.08$ \\

    3 Laplacian heads
      & $95.38 \pm 0.17$ 
      & $75.81 \pm 0.14$ 
      & $\mathbf{82.73 \pm 0.19}$ \\

    6 Laplacian heads
      & $95.36 \pm 0.10$ 
      & $76.39 \pm 0.26$ 
      & $\underline{82.70 \pm 0.02}$ \\

    9 Laplacian heads  
      & $95.76 \pm 0.30$ 
      & $76.80 \pm 0.39$ 
      & $82.47 \pm 0.08$ \\

    11 Laplacian heads 
      & $\mathbf{95.88 \pm 0.14}$ 
      & $\underline{76.83 \pm 0.26}$ 
      & $82.40 \pm 0.12$ \\

    12 Laplacian heads
      & $\underline{95.86 \pm 0.10}$ 
      & $\mathbf{77.04 \pm 0.40}$ 
      & $82.32 \pm 0.07$ \\
    \bottomrule
  \end{tabular}
  \vspace{-0.6em}
\end{table}

\subsection{Causal Language Modeling}
We trained GPT-2 style transformers with 135 million, 477 million, and 1.45 billion parameters (which we refer to as d12, d20, and d30 models, respectively) on autoregressive next-token prediction \citep{radford2019language}. The three model sizes use 6, 10, and 15 attention heads per block and were trained on 1.32, 5.22, and 16.68 billion tokens from ClimbMix \citep{diao2025climb}, respectively. We trained models with varying numbers of Laplacian heads $|\L|$ using the same training recipe following \cite{Karpathy2025nanochat}. We then evaluated the pre-trained models on validation data and downstream tasks. We report validation bits per byte (BPB), where lower values indicate better next-token prediction, and CORE, the mean chance-adjusted accuracy across 22 downstream evaluation tasks, where higher values indicate better performance. More details of the experiments are in Appendix~\ref{app:exp}.

\begin{figure}[!htbp]
    \centering
    \includegraphics[width=\textwidth]{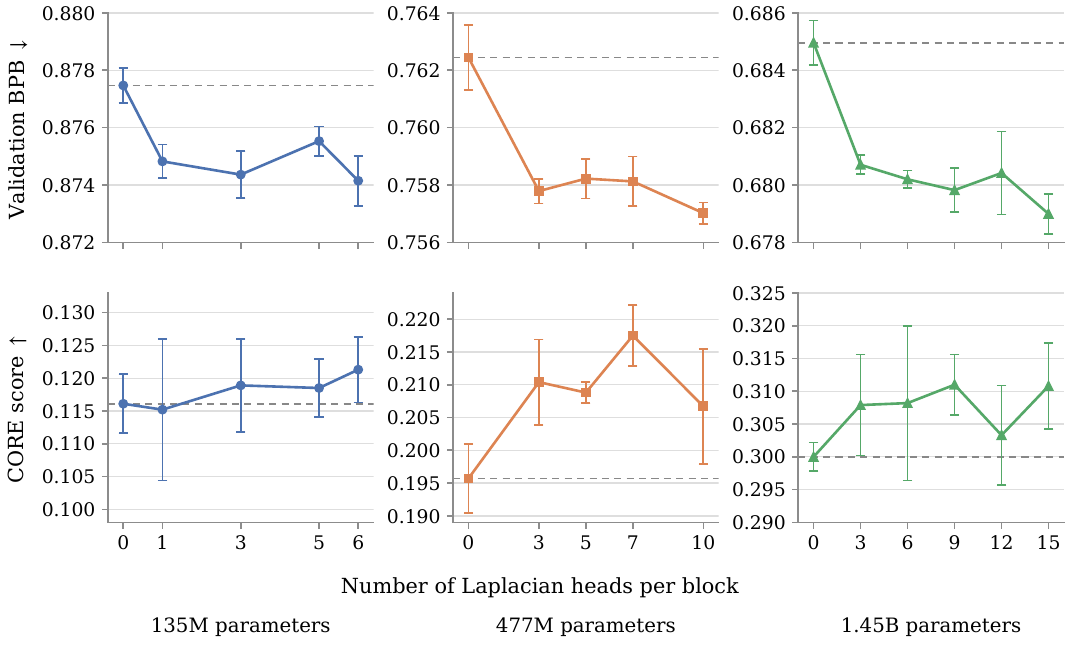}
    \caption{Models with Laplacian heads achieve lower validation BPB (top, $\downarrow$) and higher CORE scores (bottom, $\uparrow$) than the baseline, with the improvement being roughly monotonic in the number of Laplacian heads. Mean metrics and standard deviation are across 3-4 random seeds.}
    \label{fig:language_bpb}
\end{figure}

The top row of Figure~\ref{fig:language_bpb} shows that adding Laplacian heads consistently improves validation BPB across all three model sizes. For each model size, the improvement is roughly monotonic in the number of Laplacian heads, and lowest mean BPB is achieved when all heads are Laplacian. The d12, d20, and d30 models improve from 0.8775, 0.7624, and 0.6850 BPB in the baseline to 0.8741, 0.7570, and 0.6790 BPB, respectively. The bottom row of Figure~\ref{fig:language_bpb} further shows that models with Laplacian heads generally achieve higher CORE scores than the baseline.

\subsection{Vision Self-supervised Learning}
We trained ViT-S models on ImageNet-1k using the DINO self-supervised learning method \citep{caron2021emergingpropertiesselfsupervisedvision}. ViT-S has 6 heads, and we trained models with $k \in \{1, 3, 5, 6\}$ Laplacian heads per block, which we refer to as ``ViT-S-$k$L''. We evaluated the learned token representations using  linear probing on ImageNet classification and ADE20K semantic segmentation \citep{zhou2018semanticunderstandingscenesade20k}. 

Table~\ref{tab:dino_linear_probe} shows that models with Laplacian heads generally outperform the baseline on both tasks. ViT-S-3L achieves the highest mean performance on both tasks, improving ImageNet linear-probing accuracy by 1.26 percentage points and ADE20K mIoU by 1.84 points over the baseline.

\begin{table}[!htbp]
\centering
\caption{ImageNet linear-probing Top-1 accuracy and ADE20K segmentation mIoU (\%), reported as mean $\pm$ sample standard deviation across three random seeds. Bold indicates best.}
\vspace{0.3em}
\small
\setlength{\tabcolsep}{4pt}
\begin{tabular}{lccccc}
\hline
\textbf{Task} & \textbf{ViT-S} & \textbf{1L} & \textbf{3L} & \textbf{5L} & \textbf{6L} \\
\hline
ImageNet Top-1 & $72.08 \pm 0.18$ & $73.05 \pm 0.27$ & $\mathbf{73.34 \pm 0.20}$ & $73.12 \pm 0.13$ & $72.94 \pm 0.09$ \\
ADE20K mIoU & $30.27 \pm 0.70$ & $31.05 \pm 0.57$ & $\mathbf{32.11 \pm 0.54}$ & $31.83 \pm 0.82$ & $30.04 \pm 1.08$ \\
\hline
\end{tabular}
\label{tab:dino_linear_probe}
\end{table}

\section{Analysis of Token Representations}\label{sec:analysis}
In this section, we investigate the effect of Laplacian heads on learned token representations, with each subsection dedicated to a different paradigm.
\subsection{Image Classification}\label{sec:analysis_img_cls}
Consider a dataset $D$ with $C$ classes, where class $c$ contains $N_c$ examples and each example is represented by $T_i$ tokens. Let $X_{t,i,c}$ denote the representation of token $t$ in example $i$ from class $c$. 
 \paragraph{Principal Component Analysis} We apply PCA to a batch $\{X_{t,i,c}\}_{t=1}^B$ and project them onto \(\mathbb{R}^2\) using the top-two principal components. Tokens in the same class are visualized in the same color.

Figure~\ref{fig:pca_embeddings} illustrates the results for ViT-B and ViT-B-12L trained on CIFAR-10.  
For ViT-B, tokens from different classes exhibit no clear structure.  
In contrast, tokens of ViT-B-12L form well-separated clusters by classes. Models trained on CIFAR-100 and ImageNet show similar patterns (Appendix~\ref{app:additional_results}).

\paragraph{Analysis of Variance (ANOVA)}

The PCA visualizations show that token representations exhibit substantially more structure in models with Laplacian heads, but there is still noticeable variability within each class. This raises a natural question: does this variability mainly come from differences within the same sequence, or across sequences within the same class? To investigate this, we perform analysis of variance on token representations.

First, we define the \emph{sequence mean}, \emph{class mean}, and \emph{global mean} as
\[
\mu_{i,c}=\mathbb{E}_t[X_{t,i,c}],
\qquad
\mu_c=\mathbb{E}_i[\mu_{i,c}]=\mathbb{E}_{t,i}[X_{t,i,c}],
\qquad
\mu_G=\mathbb{E}_c[\mu_c]=\mathbb{E}_{t,i,c}[X_{t,i,c}].
\]

We then define the within-sequence, within-class, between-class, and total variances, each measuring the variability within a certain group:
\begin{align*}
  \operatorname{WithinSeqVar}    &= \mathbb{E}_{t,i,c}\!\left[\|X_{t,i,c}-\mu_{i,c}\|_2^2\right], &
  \operatorname{WithinClassVar}  &= \mathbb{E}_{i,c}\!\left[\|\mu_{i,c}-\mu_c\|_2^2\right], \\
  \operatorname{BetweenClassVar} &= \mathbb{E}_c\!\left[\|\mu_c-\mu_G\|_2^2\right], &
  \operatorname{TotalVar}        &= \mathbb{E}_{t,i,c}\!\left[\|X_{t,i,c}-\mu_G\|_2^2\right].
\end{align*}
By expanding the squared norm, the total variance decomposes additively into three components:
\[
\operatorname{TotalVar}
=
\operatorname{BetweenClassVar}
+
\operatorname{WithinClassVar}
+
\operatorname{WithinSeqVar}.
\]

\begin{figure}[!htbp]
\centering
    \setlength{\parskip}{0pt}
    \includegraphics[width=\textwidth]{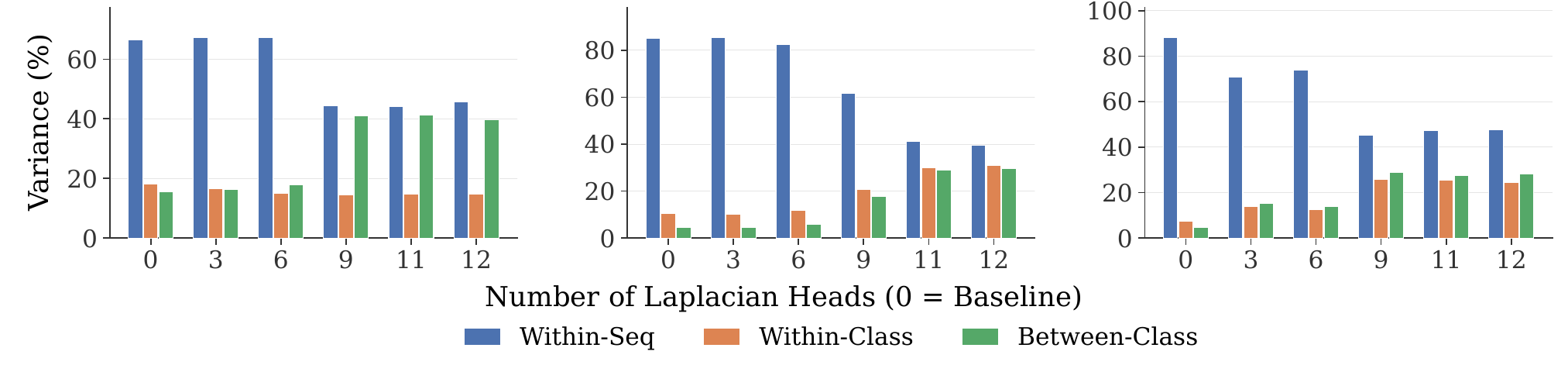}\par\nointerlineskip
    \captionsetup[subfigure]{skip=0pt}
    \makebox[\textwidth][l]{%
        \hspace*{0.07\textwidth}%
        \begin{subfigure}[t]{0.26\textwidth}
            \caption{CIFAR 10.}
        \end{subfigure}\hspace{0.07\textwidth}%
        \begin{subfigure}[t]{0.26\textwidth}
            \caption{CIFAR100.}
        \end{subfigure}\hspace{0.07\textwidth}%
        \begin{subfigure}[t]{0.26\textwidth}
            \caption{ImageNet-1k.}
        \end{subfigure}%
    }\par
\caption{
 Laplacian heads shift variance from within-sequence to between-class, inducing stronger between-class separability.}
\label{fig:anova_decomp}
\end{figure}

Following these definitions, we use the last-layer token representations to compute the total variance and its decomposition into the three components, each reported as a fraction of TotalVar. Figure~\ref{fig:anova_decomp} shows that as the number of Laplacian heads increases, the WithinSeqVar fraction decreases sharply, while the BetweenClassVar fraction increases substantially, indicating more within-sequence collapse and stronger separability between classes. For models with fewer Laplacian heads ($|\L| = 0, 3, 6$), WithinClassVar constitutes a much larger portion of the total variance than BetweenClassVar. This gap narrows as more Laplacian heads are added, and the BetweenClassVar fraction becomes nearly equal to or exceeds the WithinClassVar fraction for large $k$. These results imply that Laplacian heads indeed control within-sequence variance and smooth token representations, as explained in Section~\ref{sec:motivation}.

\begin{figure}[!htbp]
    \centering

    \includegraphics[width=0.8\linewidth, trim=0 30 45 20, clip]{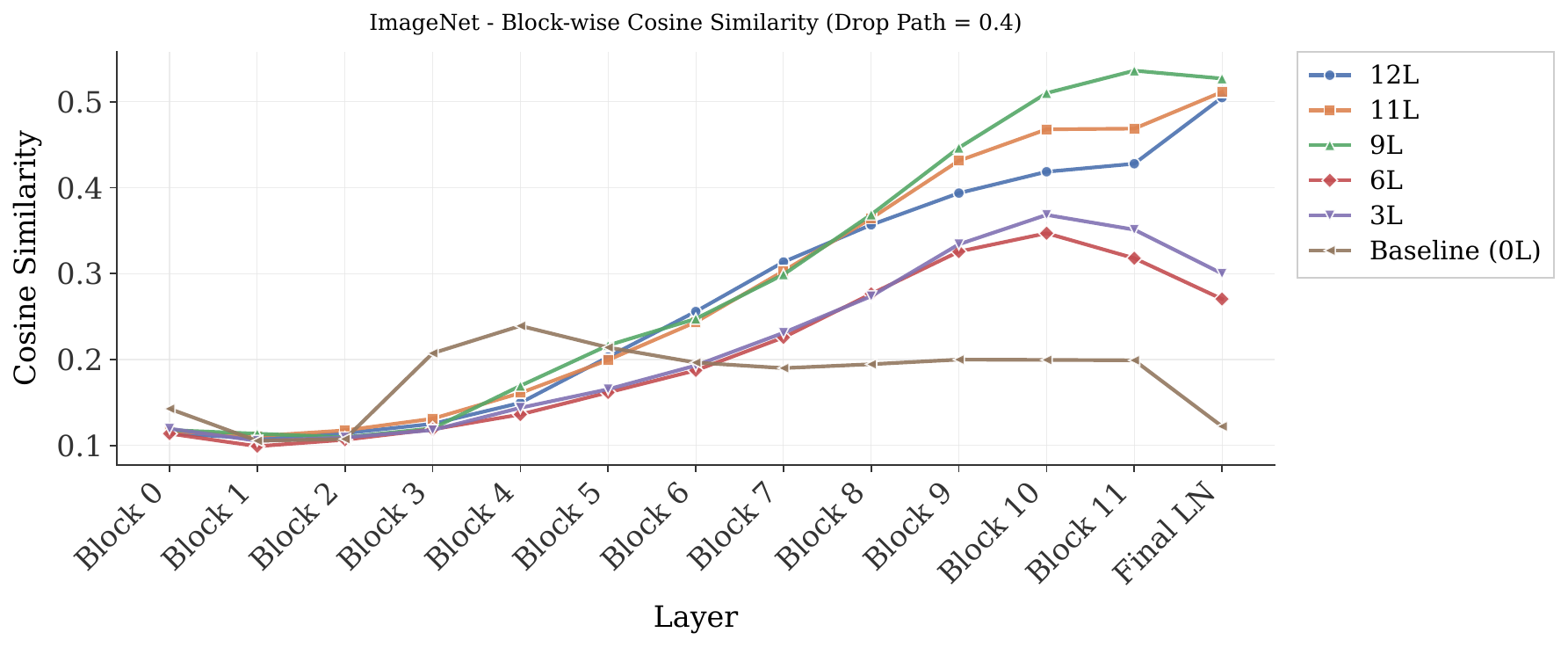}
    \caption{Laplacian heads induce more smoothing across depth. ``L'': the number of Laplacian heads.}
    \label{fig:cos_sim}
\end{figure}

\paragraph{Layerwise Within-sequence Collapse}
To investigate the decrease in within-sequence variability observed in the ANOVA results, we measure the average pairwise cosine similarity, a standard metric in the oversmoothing literature:
\[
\text{CosSim}(X) = \frac{1}{B} \sum_{b=1}^{B}
  \frac{1}{T(T-1)}
  \sum_{\substack{i,j=1\\ i\neq j}}^{T}
  \frac{\langle X_{b,i,c_b},\, X_{b,j,c_b}\rangle}
       {\|X_{b,i,c_b}\|\,\|X_{b,j,c_b}\|}.\,
\]
Figure~\ref{fig:cos_sim} reports \(\cossim\) for the output tokens of each layer in models trained on ImageNet. For ViT-B, \(\cossim\) remains low across layers, indicating limited within-sequence collapse. Adding more Laplacian heads produces a steeper increase in \(\cossim\) across depth, reaching higher values in deeper layers. This suggests that Laplacian heads induce more token smoothing in deeper layers.

\Needspace{16\baselineskip}
\paragraph{Neural Collapse Visualization and Metrics}\leavevmode\par\nobreak
\nointerlineskip
\noindent\begin{minipage}[t]{0.58\textwidth}
    \vspace{0pt}
ANOVA shows that Laplacian heads increase between-class variance, even exceeding within-class variance for some models, which indicates a tendency towards Neural Collapse.
We use a visualization technique introduced by \cite{fisher2024pushingboundariesmixupsinfluence} (Algorithm~\ref{alg:simplex_projection}) and the Neural Collapse metrics from \cite{han2022neuralcollapsemseloss} (see Appendix~\ref{appendix_nc}) to investigate this.
Figure~\ref{fig:simplex_projections} visualizes the token representations projected onto the classifier for ViT-B and ViT-B-12L, trained on CIFAR10. While the baseline displays overlapping clouds of tokens from different classes, ViT-B-12L produces well-separated clusters with a clear simplex-like structure. Across all models, the projections reveal a clearer simplex ETF structure as the number of Laplacian heads increases (Appendix~\ref{app:simplex_proj}).
\end{minipage}\hfill
\begin{minipage}[t]{0.4\textwidth}
    \vspace{0pt}
    \captionsetup{type=figure}
\centering
\begin{subfigure}{0.5\linewidth}
  \centering
  \includegraphics[
    width=\linewidth,
    trim=10 10 70 60,clip
  ]{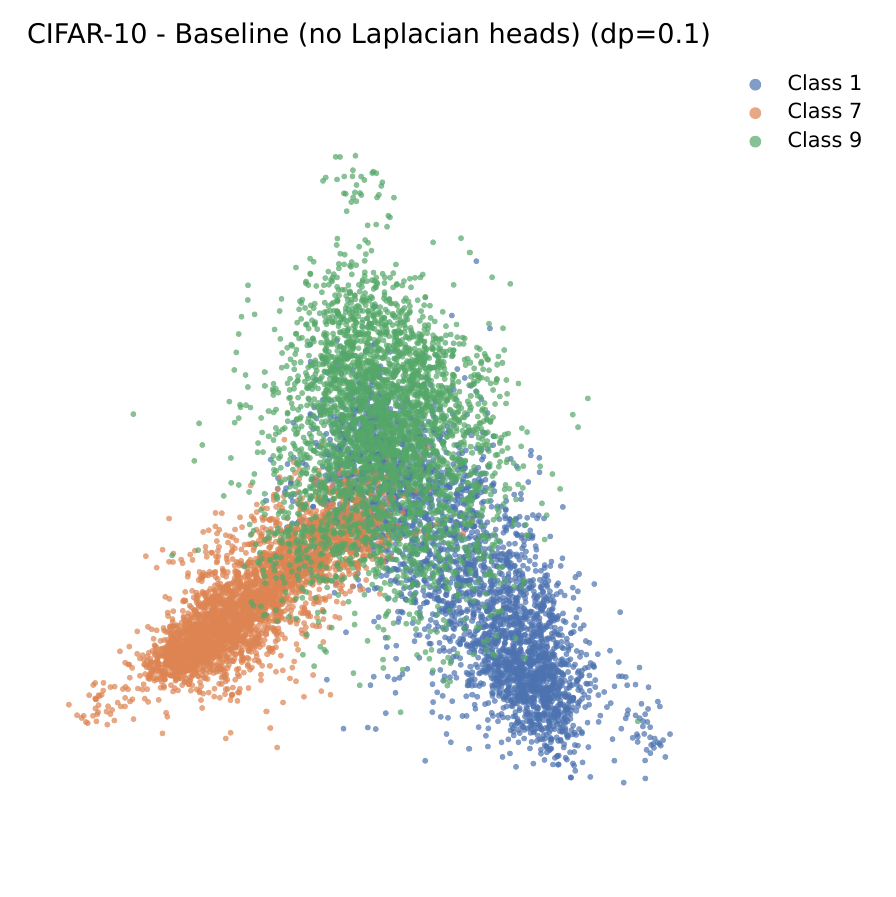}
  \caption{Baseline}
  \label{fig:simplex_baseline}
\end{subfigure}%
\begin{subfigure}{0.5\linewidth}
  \centering
  \includegraphics[
    width=\linewidth,
    trim=10 10 70 60,clip
  ]{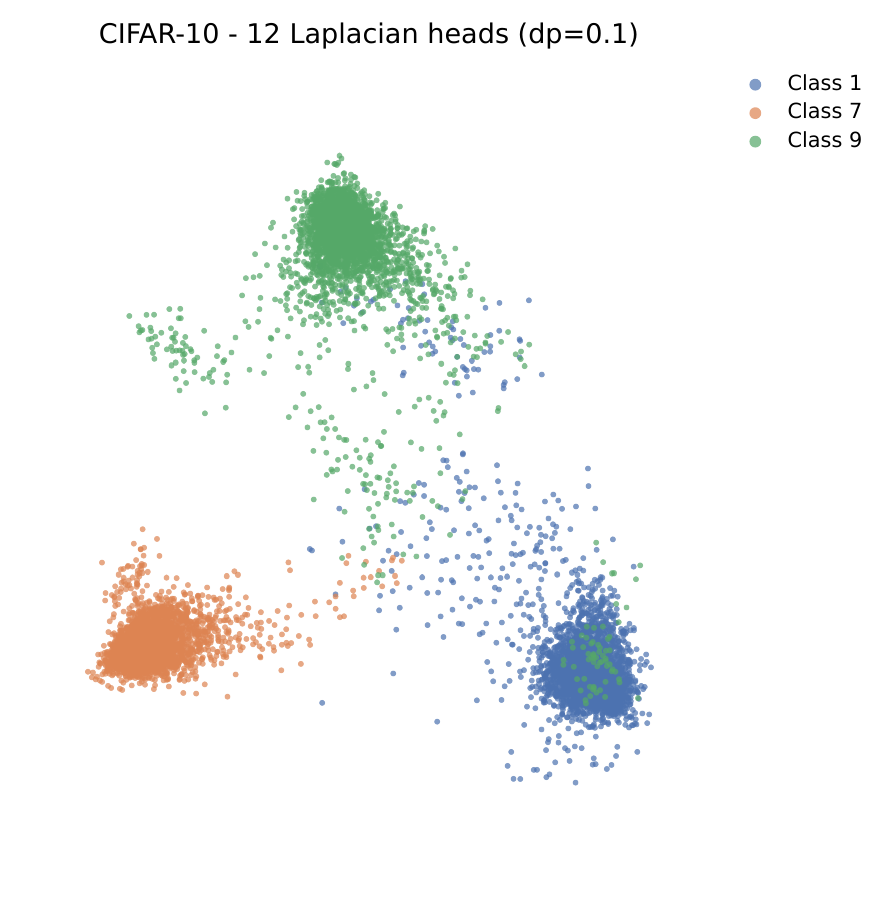}
  \caption{Proposed}
  \label{fig:simplex_proposed}
\end{subfigure}
\caption{Laplacian heads promote a simplex ETF in token representations across classes.}
\label{fig:simplex_projections}
\end{minipage}\par

\subsection{Causal Language Modeling}\label{sec:analysis_language}
Causal language modeling can be viewed as classification: each token representation belongs to the class defined by its next-token target. But unlike image classification, where all tokens in a sequence share the same class label, tokens within a language sequence generally have different next tokens, so we do not expect full within-sequence collapse. Nevertheless, prior work on token geometry in language models shows that token representations concentrate in low-dimensional subspaces \citep{zhang2025attentionsinkscatchtag,zhao2025implicitgeometrynexttokenprediction}, suggesting some form of within-sequence smoothing. Since Laplacian heads directly control within-sequence variance, we expect them to promote this smoothing to a greater extent.

\paragraph{Spectrum Measurements} To test this hypothesis, we measure the spectral decay of token representations within each sequence. For sequence $i$, let $X_i \in \R^{T_i\times d}$ be its $T_i$ token representations and $\Tilde{X}_i = X_i - \vec{1}\mu_{i}^\top$ be the centered representations, where $\mu_i$ is the sequence mean. Let $\{\lambda_1(i),\dots,\lambda_d(i)\}$ be the eigenvalues of $\Tilde{X}_i^\top\Tilde{X}_i$. We use these eigenvalues to measure spectral decay, where faster decay indicates that token representations concentrate in fewer directions. This provides a relaxed notion of within-sequence smoothing. First, we measure the average and standard deviation of the $j$-th eigenvalue across sequences: $\bar{\lambda}_j = \frac{1}{N}\sum_{i=1}^N \lambda_j(i)$. Second, for a fixed $0 <\alpha < 1$, we define $k_{\alpha}(i)$ as the smallest number of principal components that explain an $\alpha$-fraction of the total variance: $
    k_\alpha(i) = \min \left\{ k \,:\, \frac{\sum_{j=1}^k \lambda_j(i)}{\sum_{j=1}^d \lambda_j(i)} \geq \alpha \right\}$. We then average this value across sequences, giving $k_{\alpha} = \frac{1}{N}\sum_{i=1}^N k_\alpha(i)$.

\begin{figure}[!htbp]
    \centering
    \begin{subfigure}[t]{0.48\textwidth}
        \centering
        \includegraphics[width=\textwidth]{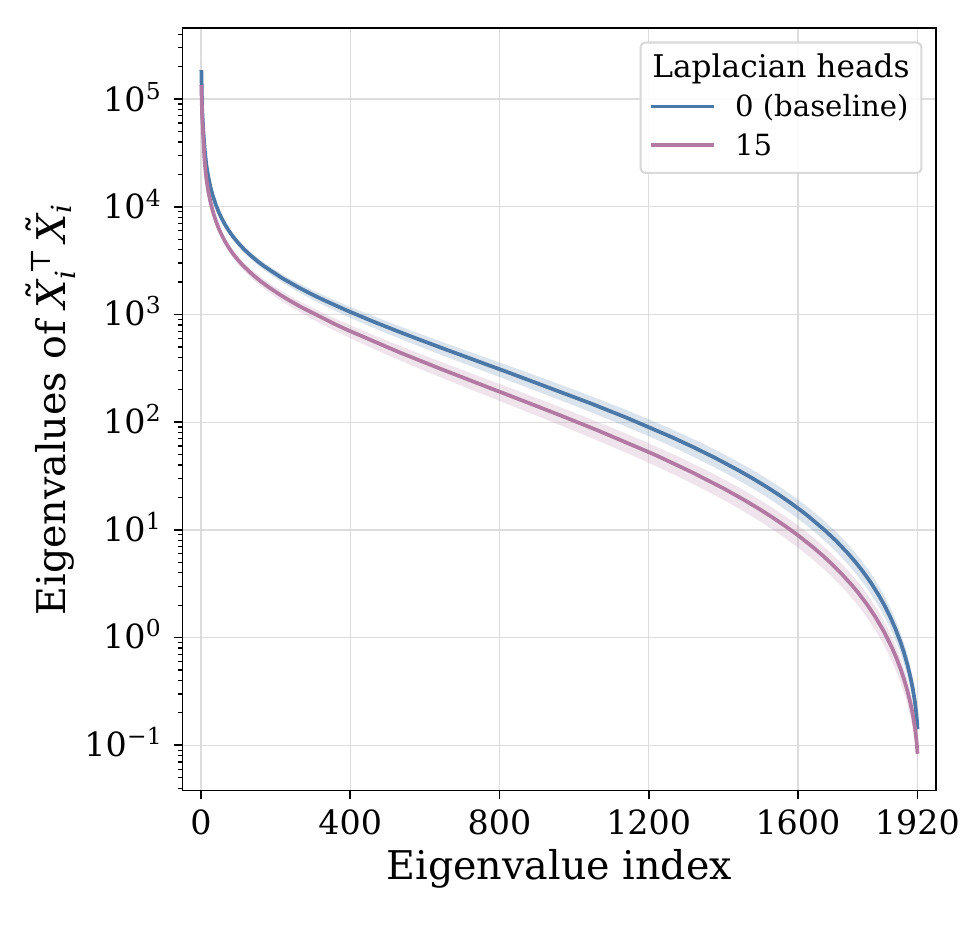}
        \caption{The d30 model with 15 Laplacian heads learn representations with a faster-decaying spectrum.}
        \label{fig:spectrum_language_3_models}
    \end{subfigure}\hfill
    \begin{subfigure}[t]{0.48\textwidth}
        \centering
        \includegraphics[width=\textwidth]{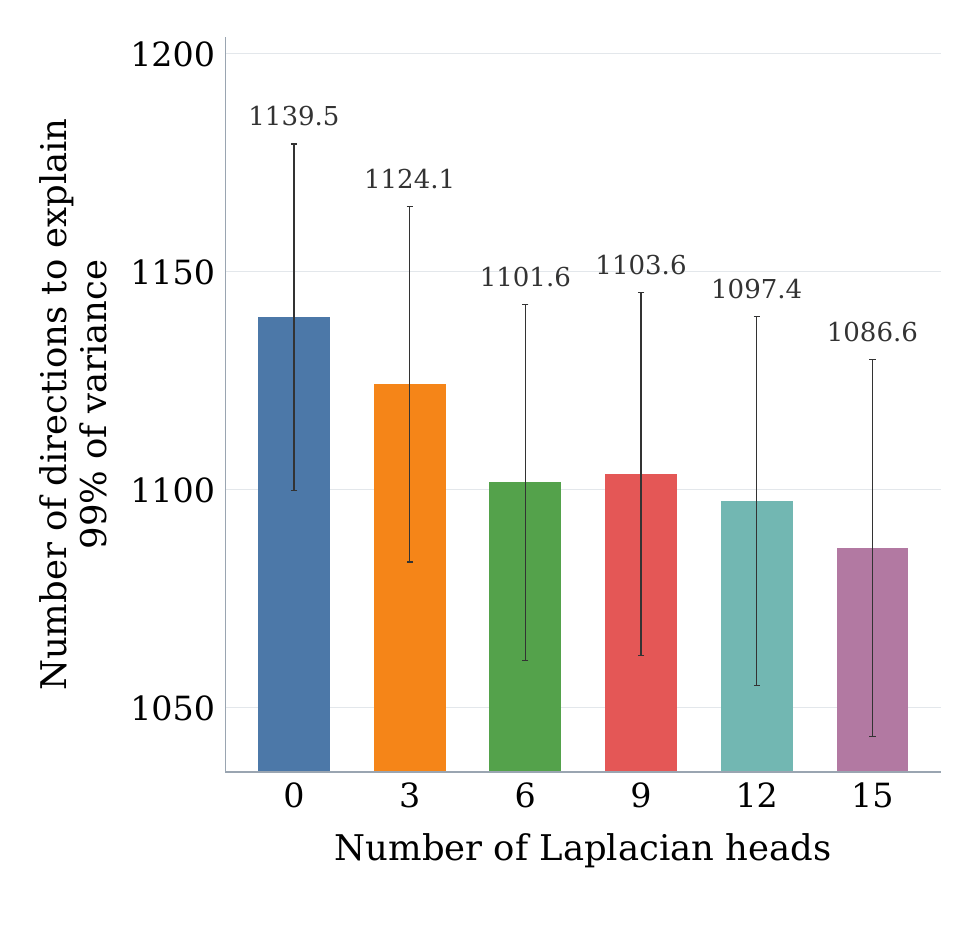}
        \caption{d30 models with Laplacian heads require fewer directions to capture 99\% of the variance.}
        \label{fig:explained_variance_language_99}
    \end{subfigure}

    \caption{Laplacian heads lead to stronger within-sequence smoothing in causal language modeling.}
    \label{fig:language_spectrum_measurements}
\end{figure}

Figure~\ref{fig:spectrum_language_3_models} shows that Laplacian heads produce a faster-decaying spectrum: the leading eigenvalues are comparable to the baseline, while the bulk and tail eigenvalues are smaller. Figure~\ref{fig:explained_variance_language_99} further shows that the number of principal components required to capture 99\% of the variance is substantially smaller in models with Laplacian heads, and that overall, this number decreases as the number of Laplacian heads increases. Together, these results indicate that Laplacian heads lead to more within-sequence smoothing, and that this effect generally grows with the number of Laplacian heads.

\Needspace{7\baselineskip}
\paragraph{ANOVA}
For the token representation at each position $t$ within a given sequence $i$, its class $c$ is defined by the next-token target. We can then compute ANOVA, analogous to Section~\ref{sec:analysis_img_cls}:
\begin{align*}
  &\mu_c=\mathbb{E}_{t,i}[X_{t,i,c}],\qquad
  \mu_G=\mathbb{E}_{t,i,c}[X_{t,i,c}],\qquad
  \operatorname{TotalVar}
  =\mathbb{E}_{t,i,c}\!\left[\|X_{t,i,c}-\mu_G\|_2^2\right],\\
  &\operatorname{WithinClassVar}
  =\mathbb{E}_{t,i,c}\!\left[\|X_{t,i,c}-\mu_c\|_2^2\right],\qquad
  \operatorname{BetweenClassVar}
  =\mathbb{E}_c\!\left[\|\mu_c-\mu_G\|_2^2\right].
\end{align*}

The decomposition of the total variance becomes $\text{WithinClassVar} + \text{BetweenClassVar} = \text{TotalVar}.$
\noindent
\begin{minipage}[t]{0.49\textwidth}
    \vspace{0pt}
 Figure~\ref{fig:language_anova} shows that adding Laplacian heads decreases the within-class fraction and increases the between-class fraction, with the effect strengthening as more Laplacian heads are added. This indicates that Laplacian heads increase the separability among token representations that share the same next-token prediction. This also suggests that the proposed modification induces stronger linguistic collapse---the language-model analogue of neural collapse---than the standard transformer~\citep{wu2024linguistic}.
\end{minipage}\hfill
\begin{minipage}[t]{0.48\textwidth}
    \vspace{0pt}
    \captionsetup{type=figure}
    \centering
    \includegraphics[width=\linewidth]{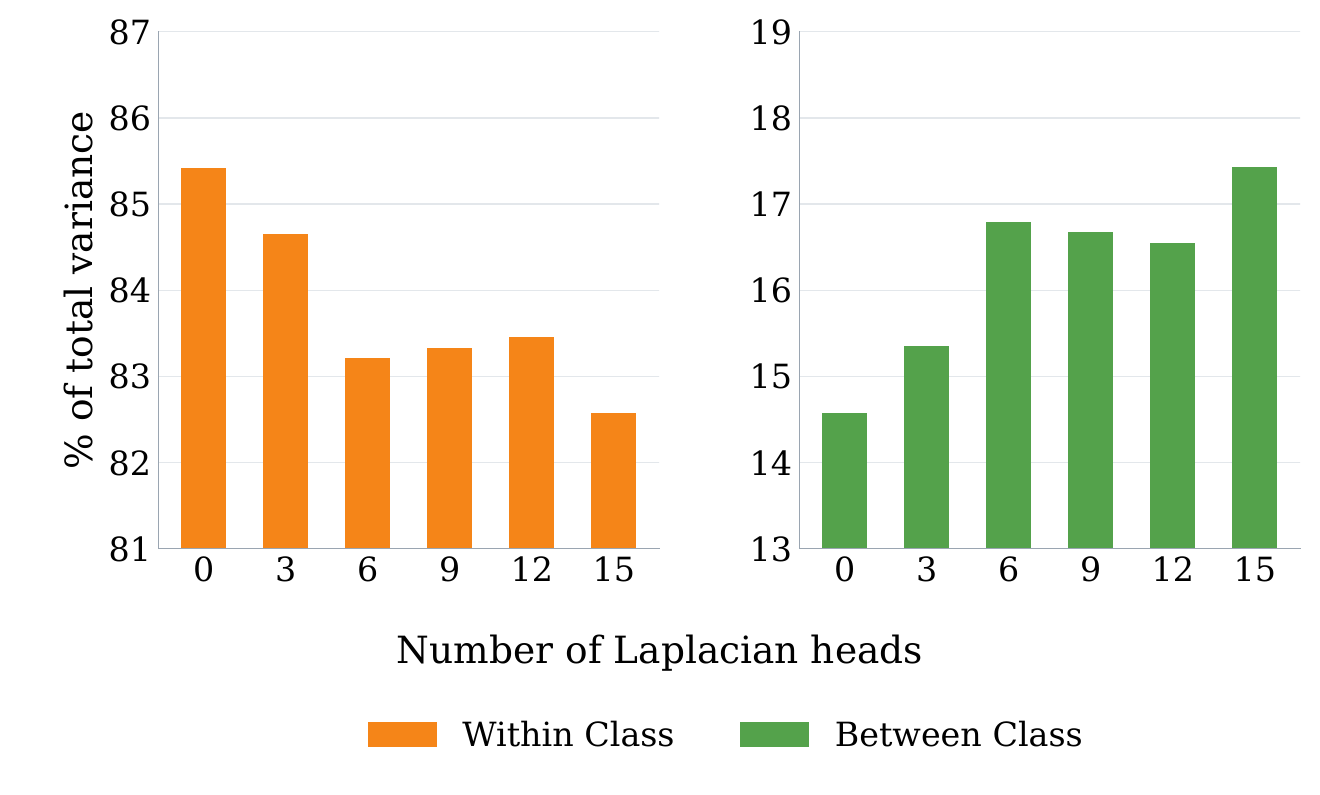}
    \caption{Laplacian heads lead to greater between-class separability for d30 models.}
    \label{fig:language_anova}
\end{minipage}\par

\subsection{Self-supervised Learning}\label{sec:analysis_dino}
Because DINO-style self-supervised learning does not use class labels in its training objective, we examine the information captured in the top principal components of token representations and apply the same spectrum measurements from Section~\ref{sec:analysis_language}.

\begin{figure}[H]
  \centering
  \includegraphics[width=\linewidth]{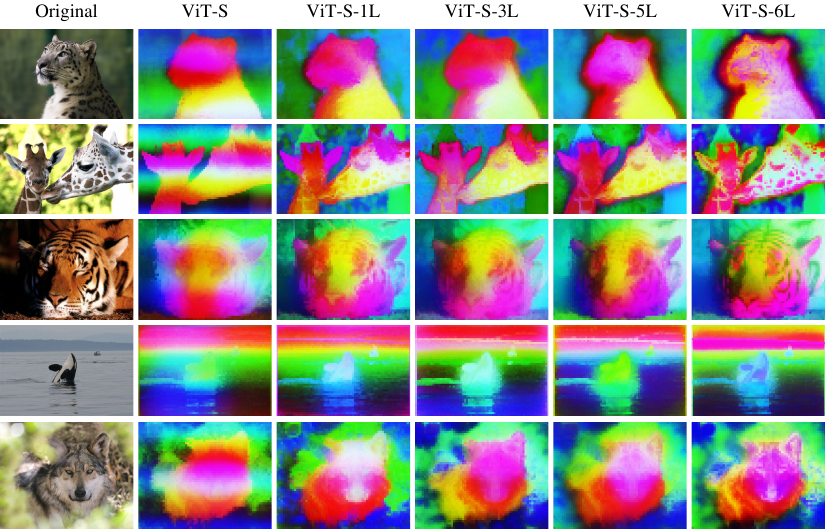}
  \caption{PCA maps of image patch representations across models. Laplacian heads preserve more segmentation structure in the top-three principal components of patch representations.}
  \label{fig:dino_pca}
\end{figure}

\paragraph{PCA Visualizations}
We use the PCA visualization technique from \citep{simeoni2025dinov3} to examine the segmentation structure of representations. We project each image's token representations onto the top-three principal components and map the result to RGB color values. Figure~\ref{fig:dino_pca} compares the PCA feature maps across models with varying numbers of Laplacian heads. It shows that for models with Laplacian heads, the top-three principal components of representations capture more segmentation structure, where boundaries and finer details are preserved more faithfully. More examples are in Appendix~\ref{app:additional_results}.

\paragraph{Spectrum Measurements}

\begin{figure}[H]
  \centering

  \begin{subfigure}[t]{0.49\textwidth}
      \centering
      \vspace{0pt}
      \includegraphics[width=\textwidth]{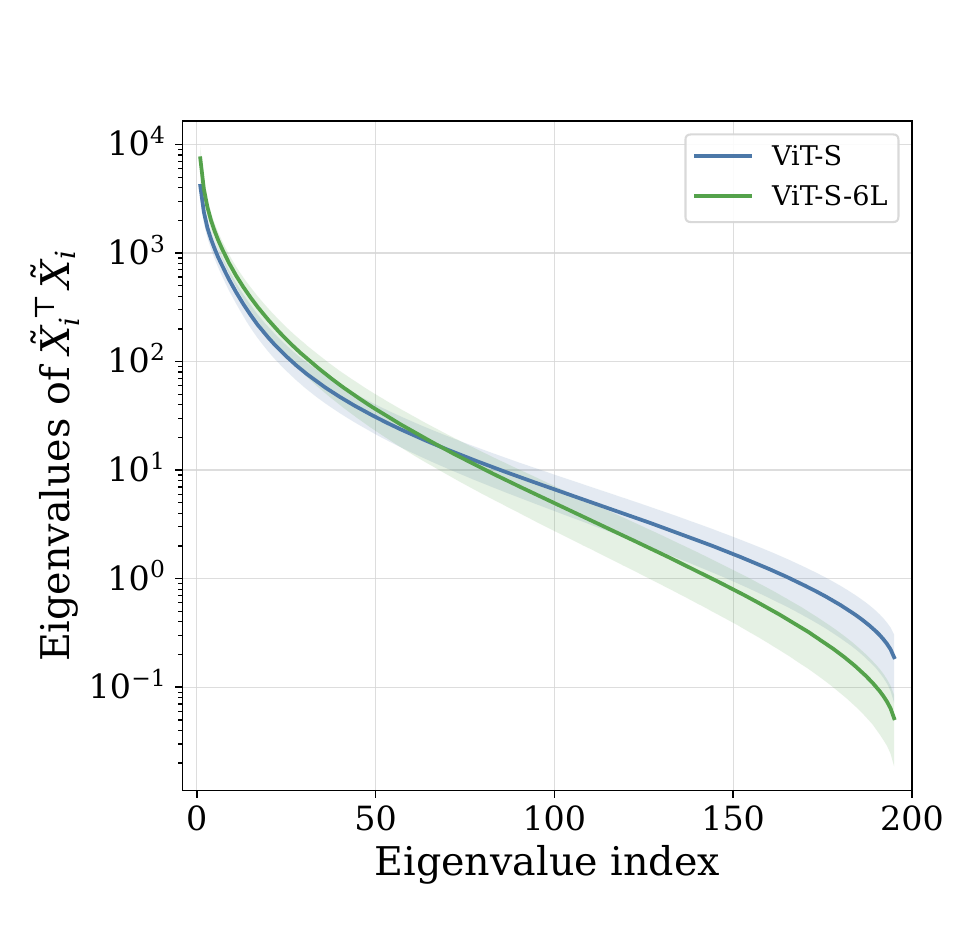}
      \caption{ViT-S-6L (proposed) has a faster-decaying spectrum than ViT-S (baseline).}
      \label{fig:spectrum_two_models}
  \end{subfigure}
  \hfill
  \begin{subfigure}[t]{0.49\textwidth}
      \centering
      \vspace{0pt}
      \includegraphics[
          width=\textwidth,
          trim=0 0 0 0,
          clip
      ]{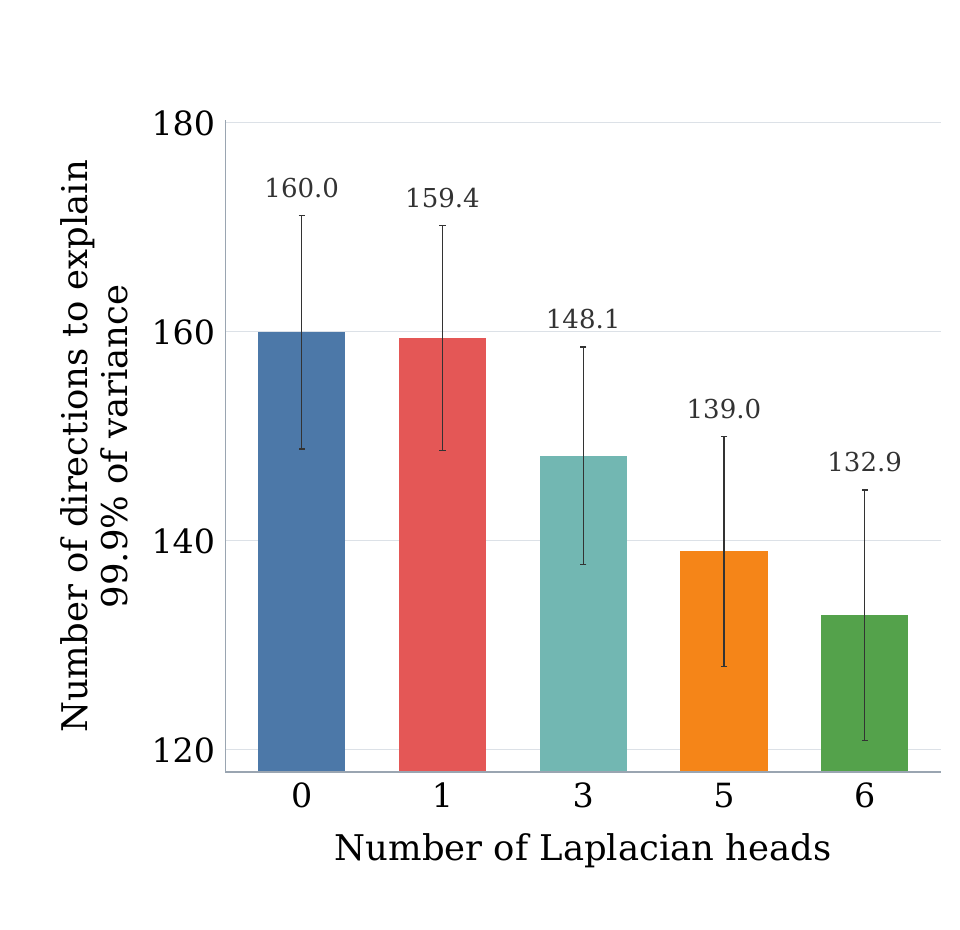}
      \caption{Models with Laplacian heads require fewer directions to capture 99.9\% of the variance.}
      \label{fig:rank_explained_variance_999}
  \end{subfigure}

  \caption{Laplacian heads lead to stronger within-sequence smoothing in DINO SSL.}
  \label{fig:spectrum_measurements}
\end{figure}

Figure~\ref{fig:spectrum_two_models} shows that the token representations of ViT-S-6L have faster-decaying spectra than those of ViT-S: the leading eigenvalues are larger, and the tail eigenvalues drop more sharply. Figure~\ref{fig:rank_explained_variance_999} further shows that in models with Laplacian heads, fewer principal directions are required to capture 99.9\% of the variance. Together, these results suggest that Laplacian heads concentrate within-image token representations into a lower-dimensional subspace.

\section{Related Works}\label{related_works}
\paragraph{Oversmoothing.}
Oversmoothing was first studied in GNNs, where message passing was viewed as smoothing and deep layers made node representations too similar, losing expressive power \citep{li2018deeper,oono2020graph,cai2020note,chen2020measuring,rusch2023surveyoversmoothinggraphneural}. This connection is relevant to transformers because self-attention can also be viewed as message passing on a graph formed by token representations \citep{joshi2025transformersgraphneuralnetworks}. Most prior work treats oversmoothing as a problem. In GNNs, this has led to normalization, edge dropping, residual and normalization schemes, oscillator-based dynamics, sheaf diffusion, state-space formulations, and other methods that seek to directly prevent oversmoothing \citep{zhao2020pairnorm,rong2020dropedge,scholkemper2025residual,rusch2022graphcoupledoscillatornetworks,bodnar2023neuralsheafdiffusiontopological,arroyo2025vanishinggradientsoversmoothingoversquashing,roth2024rankcollapsecausesoversmoothing,roth2024preventingrepresentationalrankcollapse}. Although some people question whether oversmoothing is inherent to transformers \citep{dovonon2024setting}, most works still treat oversmoothing as a failure mode, with attempts to mitigate it via diversity losses, Fourier-domain corrections, graph-based layer fusion, or regularized nonlocal objectives \citep{wu2023demystifying,gong2021vision,wang2022antioversmoothingdeepvisiontransformers,shi2022revisiting,nguyen2023mitigatingoversmoothingtransformersregularized}. 

A smaller line of work gives a more nuanced view. Wu et al. argue that smoothing has a desirable denoising effect and a harmful mixing effect, and oversmoothing is only harmful when the mixing effect dominates the denoising effect \citep{wu2023nonasymptotic}. Our work follows this latter view. Rather than treating smoothing as merely a failure mode, we introduce Laplacian heads and make graph diffusion explicitly a part of the architecture, which the model can use to control within-sequence variance.

\paragraph{Rank collapse.}
Rank collapse is the phenomenon where token representations become increasingly low-rank across depth. There is a wide body of theoretical and empirical works that study and try to mitigate rank collapse in transformers. \cite{dong2023attentionneedpureattention} shows that pure attention collapses representations doubly exponentially to rank one. \cite{noci2022signalpropagationtransformerstheoretical} links collapse to poor signal propagation and mitigates it with residual scaling. \cite{noci2023shaped} avoids degeneracy in the infinite-depth-and-width limit by centering softmax attention at the identity and scaling logits. \cite{wu2024masks} shows that sparse masks slow down collapse and LayerNorm permits higher-rank equilibria. \cite{joseph2024lambda} prevents collapse via lambda-skip connections. \cite{saada2025mindgapspectralanalysis} attributes width-induced collapse to a softmax spectral gap and removes the outlier mode. \cite{giorlandino2025two} identifies initializations that avoid rank and entropy collapse. \cite{barbero2024transformersneedglassesinformation} studies final-token collapse and mitigates it with input-level cues. Related dynamical and mean-field analyses show that self-attention forms limiting clusters, low-rank attention patterns, or metastable multi-cluster states \cite{geshkovski2024emergenceclustersselfattentiondynamics,geshkovski2025mathematicalperspectivetransformers,bruno2025emergencemetastableclusteringmeanfield}. Several empirical works link language-model geometry to prompting, sequence length, token geometry, and next-token classes \cite{kirsanov2025geometrypromptingunveilingdistinct,zhou2025lengthinduced,viswanathan2025geometrytokensinternalrepresentations,wu2024linguistic}. More broadly, collapse has different implications across learning settings. In supervised classification, Neural Collapse describes within-class variability collapse while class means form a simplex ETF, a configuration that has been shown to be the global optimum of standard training objectives \cite{neuralcollapse,zhou2022optimization,sukenik2025global}. In self-supervised learning, collapse is viewed as a failure mode, motivating objectives that preserve variance or reduce redundancy \cite{jing2021dimensional,zbontar2021barlow,bardes2022vicreg,ziyin2023ssl,zhuo2023rdm}. Our work suggests that collapse is neither always good nor always bad: Laplacian heads can collapse token representations into geometric configurations that benefit the task at hand.

\section{Conclusion}\label{sec:conclusion}

We proposed Laplacian heads, a parameter-free modification to the transformer architecture. We motivated our
proposal from two perspectives: giving the model direct control
over within-sequence variance and the ability to perform diffusion steps over the attention graph of token representations.

Incorporating Laplacian heads into the transformer improved supervised classification, language modeling,
and vision self-supervised learning. Across all three settings, Laplacian heads
induced more smoothing in token representations, and this smoothing improved
representations rather than degrading them. 

Taken together, our findings challenge the prevailing view that token
oversmoothing is universally bad. Prior work has largely treated
smoothing and collapse as failure modes to be prevented. Our results show
instead that smoothing can take task-specific forms that benefit
representation learning. In supervised classification, tokens collapsed within
sequences and sequence means converged towards Neural Collapse; in language
modeling and DINO, token representations exhibited faster spectral decay
while becoming better suited to their respective tasks. Across all three paradigms, Laplacian heads
smoothed token representations, driving them into task-beneficial configurations.

Our simple strategy of using a fixed number of Laplacian heads across layers
is likely suboptimal, so a future direction is to investigate better
strategies. Another important direction is to further investigate the geometric configurations that
token representations should tend toward in language modeling and self-supervised
learning, which the present work does not characterize. More broadly, our work points to smoothing as a mechanism for
representation learning that deserves further study.


\section*{Acknowledgements and Disclosures}
We would like to thank David Emerson and Chris Maddison for their discussions and feedback that helped improve our work. We acknowledge the support of the Natural Sciences and Engineering Research Council of Canada (NSERC). This
research was supported in part by the Province of Ontario, the Government of Canada through CIFAR, and industry sponsors of the Vector Institute (\url{www.vectorinstitute.ai/partnerships/current-partners/}). This research was also enabled in part by support provided by Compute Ontario (\url{https://www.computeontario.ca/}) and the Digital Research Alliance of Canada
(\url{https://alliancecan.ca/}). We declare no competing interests.

\bibliography{references}
\bibliographystyle{unsrt}
\newpage
\appendix
\section{Experiment Details}\label{app:exp}
The image classification and DINO experiments were performed using 4 L40S GPUs (48GB RAM each). The language modeling experiments used 4 H100 GPUs (80GB RAM each).
\subsection{Image Classification}
To produce results in Table ~\ref{tab:top1-accuracy}, we trained the ViT-B model from \cite{Touvron2022DeiTIR}.
The model consists of 12 blocks, with 12 attention heads in each block, and an embedding dimension of 768. The number of trainable parameters is around 86.6 million. For each dataset, we swept the peak learning rate over the set $\{4e-5, 3e-4, 5e-4, 3e-3, 4e-3\}$ and weight decay over the set $\{0.01, 0.02, 0.05\}$ and selected whichever combination that worked the best. Many other hyperparameters (such as the mixup \citep{zhang2018mixupempiricalriskminimization} $\alpha)$ were selected following the training recipe detailed in \cite{Touvron2022DeiTIR}. Depending on the dataset, we used RandAugment \citep{cubuk2019randaugmentpracticalautomateddata} or 3-Aug \cite{Touvron2022DeiTIR} for data augmentation and AdamW \citep{kingma2015adam,loshchilov2019adamw} or LAMB \citep{you2020lamb} as the optimizer. All models in Table~\ref{tab:top1-accuracy} were trained for 300 epochs using 3 random seeds.  Full details of our training setup, including the hyperparameters that were eventually selected for the experiments, are provided in Table~\ref{tab:training_recipe}.
\begin{table}[h]
\centering
\renewcommand{\arraystretch}{1.2}
\begin{tabular}{|l|c|c|c|}
\hline
\textbf{} & \textbf{CIFAR-10} & \textbf{CIFAR-100} & \textbf{ImageNet} \\
\hline
Loss & Cross Entropy & Cross Entropy & Binary Cross Entropy\\
Optimizer & AdamW  & AdamW & LAMB \\
AdamW $\beta_1$ & 0.9 & 0.9 & 0.9\\
AdamW $\beta_2$ & 0.999 & 0.999 & 0.999\\
Starting Learning Rate & 1e-6 & 1e-6 & 1e-6\\
Peak Learning Rate & 3e-4 & 3e-4 & 3e-3 \\
Minimum Learning Rate & 1e-6 & 1e-6 & 1e-6 \\
Weight Decay & 0.05 & 0.05 & 0.02 \\
Drop Path Rate & 0.1 & 0.1 & 0.4 \\
Batch Size & 512 & 512 & 2048 \\
Gradient Clipping & None & None & 1.0 \\
LR Scheduler & Cosine Annealing & Cosine Annealing & Cosine Annealing \\
Warmup Epochs & 10 & 10 & 5 \\
Data Augmentation & RandAugment & RandAugment & 3-Aug  \\
Mixup $\alpha$ & 0.8 & 0.8 & 0.8 \\
Mixup Probability & 1.0 & 1.0 & 1.0 \\
Input Size & 32$\times$32 & 32$\times$32 & 224$\times$224 \\
Patch Size & 4$\times$4 & 4$\times$4 & 16$\times$16 \\
Precision & float16 AMP & float16 AMP & bfloat16\\
\hline
\end{tabular}
\caption{Training setup for CIFAR-10, CIFAR-100, and ImageNet.}
\label{tab:training_recipe}
\end{table}

\subsection{Autoregressive Next-token Prediction}
The model architecture we used was based on \citep{Karpathy2025nanochat}. The model is a decoder-only transformer similar to GPT-2, with the following architectural modifications:
\begin{itemize}
    \item Parameter-free RMSNorm \citep{zhang2019rmsnorm}.
    \item Query-key normalization \citep{henry2020querykeynormalizationtransformers, yang2025qwen3technicalreport}.
    \item Rotary positional encodings \citep{su2023roformerenhancedtransformerrotary}.
    \item Squared ReLU activation \citep{nvidia2024nemotron4340btechnicalreport}.
    \item Untied weights for the vocabulary embedding and unembedding matrices.
    \item Soft logits clipping \citep{gemmateam2024gemma2improvingopen}.
\end{itemize}
The 12-, 20-, and 30-block models (d12, d20, and d30) have embedding dimensions 768, 1280, and 1920, with 6, 10, and 15 heads per block, respectively. We tried $|\L|\in\{0,1,3,5,6\}$ for d12, $|\L|\in\{0,3,5,7,10\}$ for d20, and $|\L|\in\{0,3,6,9,12,15\}$ for d30. All models used a context length of 2048 and a vocabulary of 32,768 tokens. Attention followed a repeated pattern of three sliding-window blocks and one full-context block, with full-context attention in the final block.

We pre-trained the models on 1.32, 5.22, and 16.68 billion ClimbMix tokens for d12, d20, and d30, respectively, using FP8 matrix multiplications during training. The training budget was 12 tokens per scaling parameter (transformer matrices plus the unembedding matrix, excluding input embeddings). At each model size, the parameter count, training-token budget, batch size, and learning-rate schedule were the same for every choice of $|\L|$. Each configuration used four random seeds, except for the d12 model with $|\L|=6$ and the d20 model with $|\L|=3$, which used three seeds. We optimize transformer matrices with Muon~\citep{jordan2024muon} and the embedding and unembedding matrices with AdamW. Table~\ref{tab:nanochat_recipe} summarizes the pretraining recipe.

\begin{table}[!htbp]
\centering
\small
\setlength{\tabcolsep}{4pt}
\renewcommand{\arraystretch}{1.05}
\caption{Nanochat pretraining recipe.}
\label{tab:nanochat_recipe}
\begin{tabular}{@{}lccc@{}}
\toprule
\textbf{Hyperparameter} & \makebox[0.19\textwidth]{\textbf{135M}} & \makebox[0.19\textwidth]{\textbf{477M}} & \makebox[0.19\textwidth]{\textbf{1.45B}} \\
\midrule
Blocks / width / heads & $12/768/6$ & $20/1280/10$ & $30/1920/15$ \\
Training tokens (billions) & $1.32$ & $5.22$ & $16.68$ \\
Optimizer steps & $2{,}520$ & $4{,}980$ & $15{,}907$ \\
Global batch size (tokens) & $524{,}288$ & $1{,}048{,}576$ & $1{,}048{,}576$ \\
Peak LR: transformer (Muon) & $0.0200$ & $0.02828$ & $0.02828$ \\
Peak LR: embedding (AdamW) & $0.3000$ & $0.3286$ & $0.2683$ \\
Peak LR: unembedding (AdamW) & $0.008000$ & $0.008764$ & $0.007155$ \\
Initial Muon weight decay & $0.2800$ & $0.1002$ & $0.03136$ \\
\midrule
Data / objective & \multicolumn{3}{l}{ClimbMix / next-token cross-entropy} \\
Context / vocabulary size & \multicolumn{3}{l}{$2{,}048$ / $32{,}768$} \\
LR schedule & \multicolumn{3}{p{0.62\textwidth}@{}}{40-step linear warmup, constant, then linear decay over the final 65\% of steps to 5\% of peak LR} \\
Muon settings & \multicolumn{3}{p{0.62\textwidth}@{}}{5 orthogonalization iterations; $\beta_2=0.9$; momentum $0.85\to0.97$ over 400 steps, then $\to0.90$ during LR decay} \\
Weight decay & \multicolumn{3}{p{0.62\textwidth}@{}}{Muon: cautious decay\footnotemark, cosine schedule to zero; AdamW: $0.001$ (embedding), $0.01$ (unembedding)} \\
AdamW $(\beta_1,\beta_2)$; $\epsilon$ & \multicolumn{3}{p{0.62\textwidth}@{}}{$(0.8,0.995)$ (embedding), $(0.8,0.96)$ (unembedding); $\epsilon=10^{-10}$} \\
Dropout / gradient clipping & \multicolumn{3}{l}{None / none} \\
Hardware / precision & \multicolumn{3}{p{0.62\textwidth}@{}}{4 H100 GPUs; BF16 activations with tensorwise FP8 linear matrix multiplications; BF16 evaluation} \\
\bottomrule
\end{tabular}
\end{table}
\footnotetext{Cautious weight decay skips decay for parameters whose optimizer update points away from zero.}

We evaluated the final pretrained checkpoints, without any post training. Validation BPB was computed using 20,971,520 tokens from the same held-out ClimbMix shard, and CORE used all examples in each of the 22 downstream tasks and averages their chance-adjusted accuracies.

\subsection{DINO}
We trained ViT-S/16 models from scratch on ImageNet-1k using the DINOv3 implementation \citep{simeoni2025dinov3}, with DINO, iBOT, and KoLeo losses and no Gram loss. Each backbone has 12 blocks, embedding dimension 384, six heads per block, and no register tokens. We used $|\L|\in\{0,1,3,5,6\}$ and three training seeds for each model. We trained for 100 epochs with a global batch size of 896. Table~\ref{tab:dino_hparams} gives the training hyperparameters.

The learning-rate schedule starts at zero, warms up linearly for 10 epochs, and then decays with a cosine schedule. The configured peak and minimum rates, $1.5\times10^{-3}$ and $10^{-6}$, are both multiplied by $4\sqrt{896/1024}$ before applying the layerwise and patch-embedding multipliers. The teacher EMA momentum is fixed at $0.992$ throughout training. Weight decay increases linearly from $0.04$ to $0.1$ over the first 10 epochs and remains at $0.1$ thereafter.
\begin{table}[H]
\centering
\small
\caption{Key hyperparameters for DINO pretraining on ImageNet-1k. Learning rates are after batch-size scaling and before parameter-group multipliers.}
\label{tab:dino_hparams}
\begin{tabular}{ll}
\toprule
\textbf{Hyperparameter} & \textbf{Value} \\
\midrule
\multicolumn{2}{l}{\textit{Architecture}} \\
Backbone & ViT-S/16 \\
Position embedding & RoPE (base $100$) \\
Drop path rate & $0.1$ \\
LayerScale initialization & $10^{-5}$ \\
Precision & BF16 parameters / FP32 gradient reduction \\
\midrule
\multicolumn{2}{l}{\textit{Optimization}} \\
Optimizer & AdamW ($\beta_1{=}0.9$, $\beta_2{=}0.999$) \\
Epochs / optimizer steps & $100$ / $143{,}000$ \\
Batch size (per GPU / global) & $224$ / $896$ \\
Peak learning rate & $5.6125 \times 10^{-3}$ \\
LR warmup epochs & $10$ \\
Minimum learning rate & $3.7417 \times 10^{-6}$ \\
Weight decay & $0.04 \to 0.1$ over 10 epochs, then constant \\
Layerwise LR decay & $0.95$ \\
Patch-embedding LR multiplier & $0.2$ \\
Head last-layer freeze & First epoch \\
Gradient clipping & $3.0$ \\
\midrule
\multicolumn{2}{l}{\textit{Teacher}} \\
EMA momentum & $0.992$ (constant) \\
Teacher temperature (start $\to$ end) & $0.04 \to 0.07$ \\
Teacher temp warmup epochs & $30$ \\
Teacher target normalization & Sinkhorn--Knopp (3 iterations) \\
Student temperature & $0.1$ \\
\midrule
\multicolumn{2}{l}{\textit{DINO head}} \\
Loss weight & $1.0$ \\
Prototypes & $65{,}536$ \\
KoLeo loss weight & $0.1$ \\
\midrule
\multicolumn{2}{l}{\textit{iBOT head}} \\
Loss weight & $1.0$ \\
Mask probability & $0.5$ \\
Mask ratio range & $[0.1, 0.5]$ \\
Prototypes & $65{,}536$ \\
\midrule
\multicolumn{2}{l}{\textit{Multi-crop augmentation}} \\
Global crops & $2 \times 224^2$, scale $[0.32, 1.0]$ \\
Local crops & $8 \times 96^2$, scale $[0.05, 0.32]$ \\
\bottomrule
\end{tabular}
\end{table}

We evaluated the final EMA teacher checkpoint with a frozen backbone. ImageNet linear probing concatenates the final-block CLS token and mean patch features and selects the best of 16 learning rates (batch size 128; 100,100 steps). ADE20K segmentation uses a linear segmentation head (batch size 16; 40,000 steps; FP32). Reported means and sample standard deviations are across the three backbone training seeds.

\newpage
\section{Additional Results}\label{app:additional_results}

\subsection{Image Classification}\label{app:add_results_img_cls}
\Needspace{0.45\textheight}
\paragraph{PCA (CIFAR10)}\leavevmode

\begin{figure}[H]
  \centering  
    \includegraphics[width=0.44\linewidth]{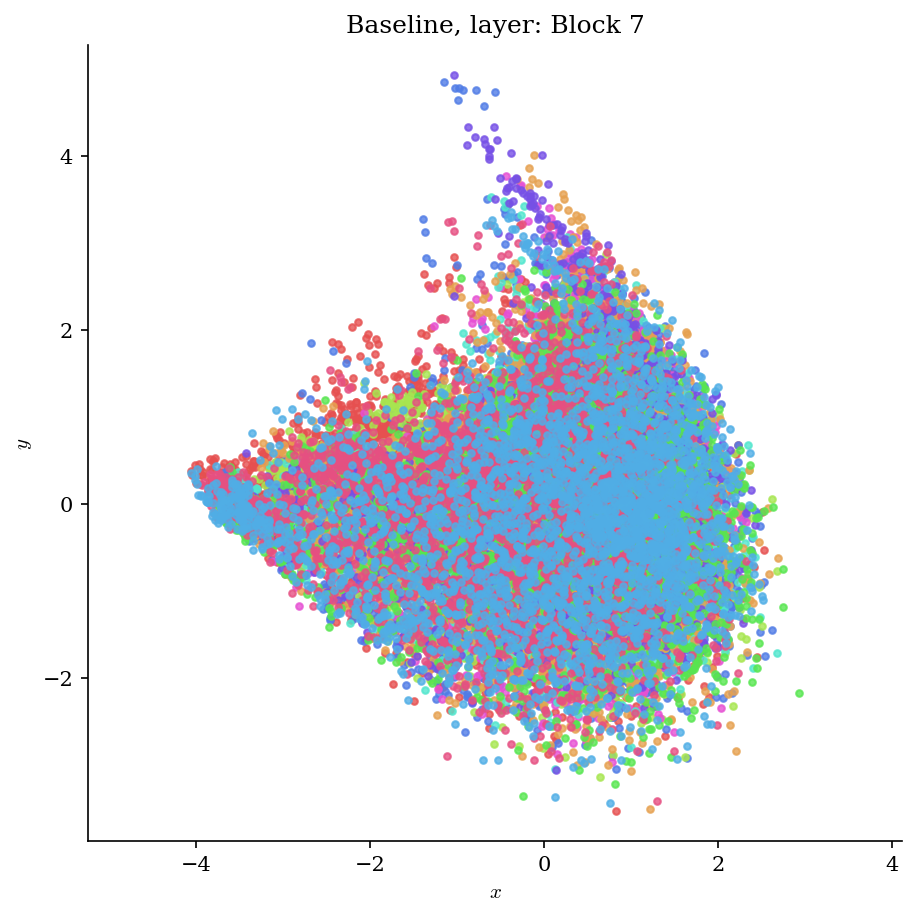}
  \hfill
    \includegraphics[width=0.44\linewidth]{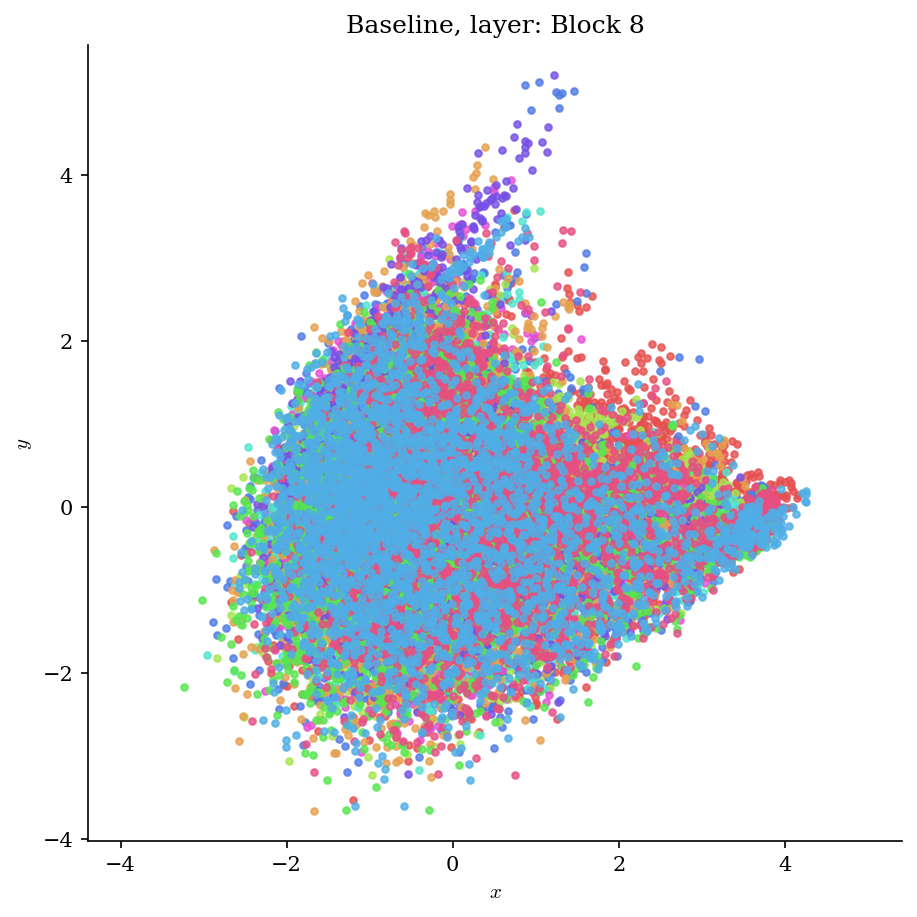}
  
  \vspace{-0.3em}
  
    \includegraphics[width=0.44\linewidth]{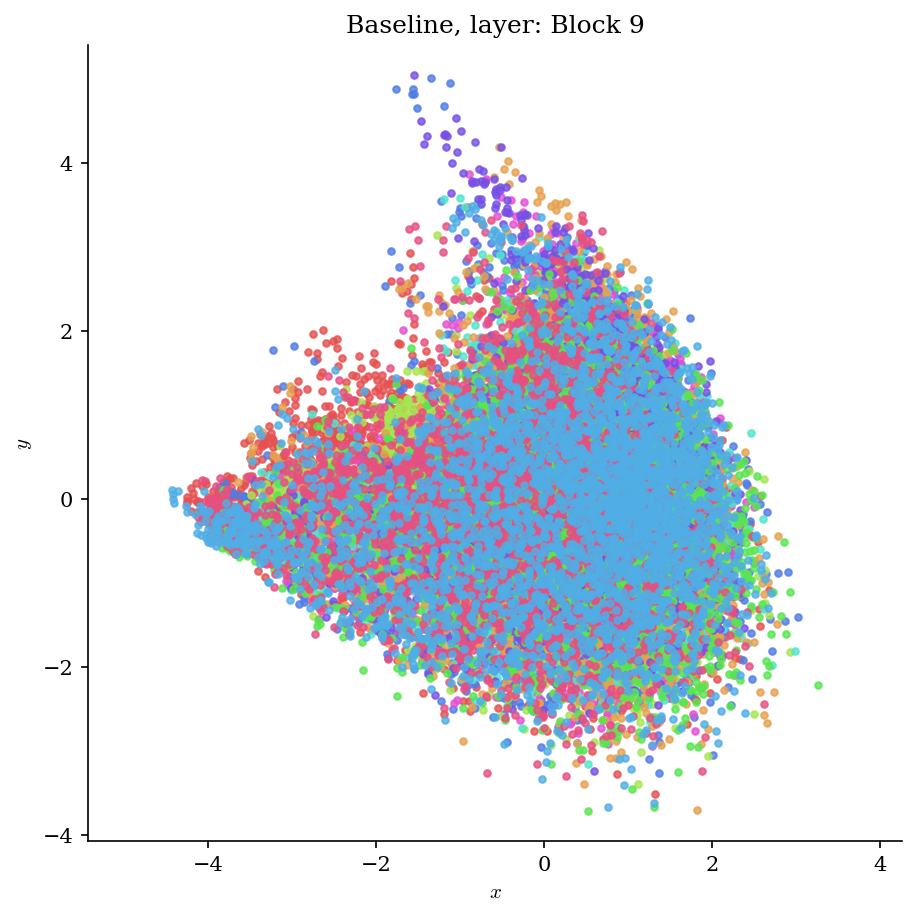}
  \hfill
    \includegraphics[width=0.44\linewidth]{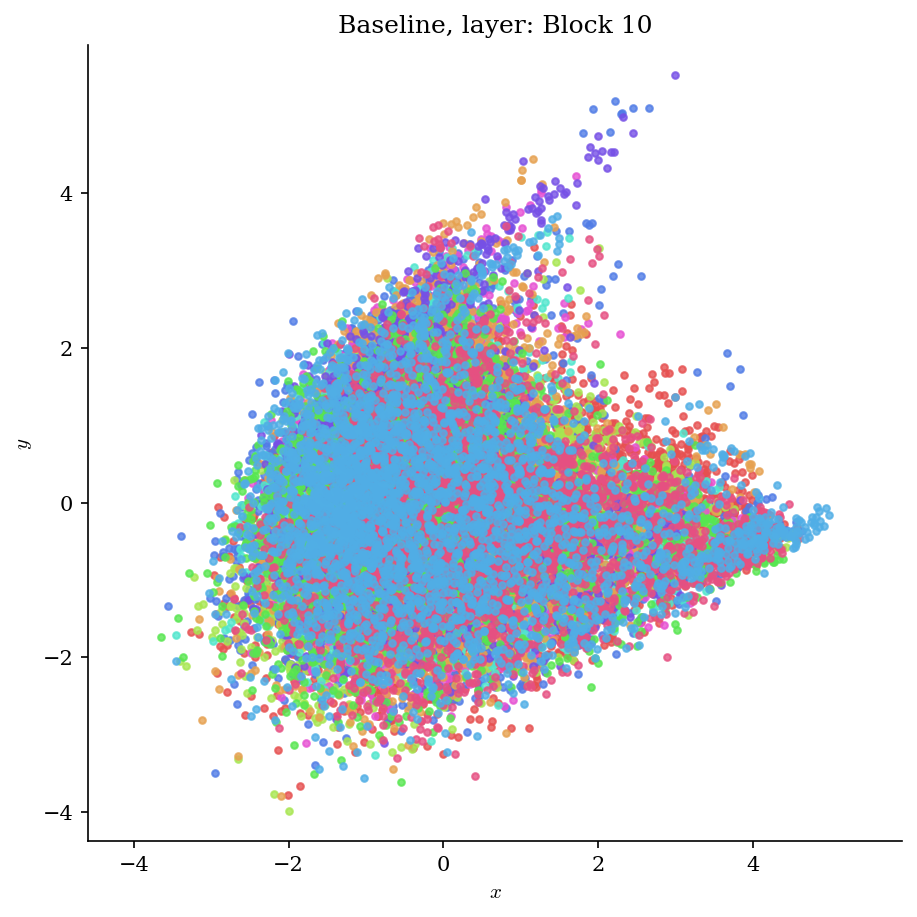}
  
  \vspace{-0.3em}
  
    \includegraphics[width=0.44\linewidth]{figures/pca/cifar10/deit_facebook_layer_scale_block_base_patch4_LS/block_11.png}
  \hfill
    \includegraphics[width=0.44\linewidth]{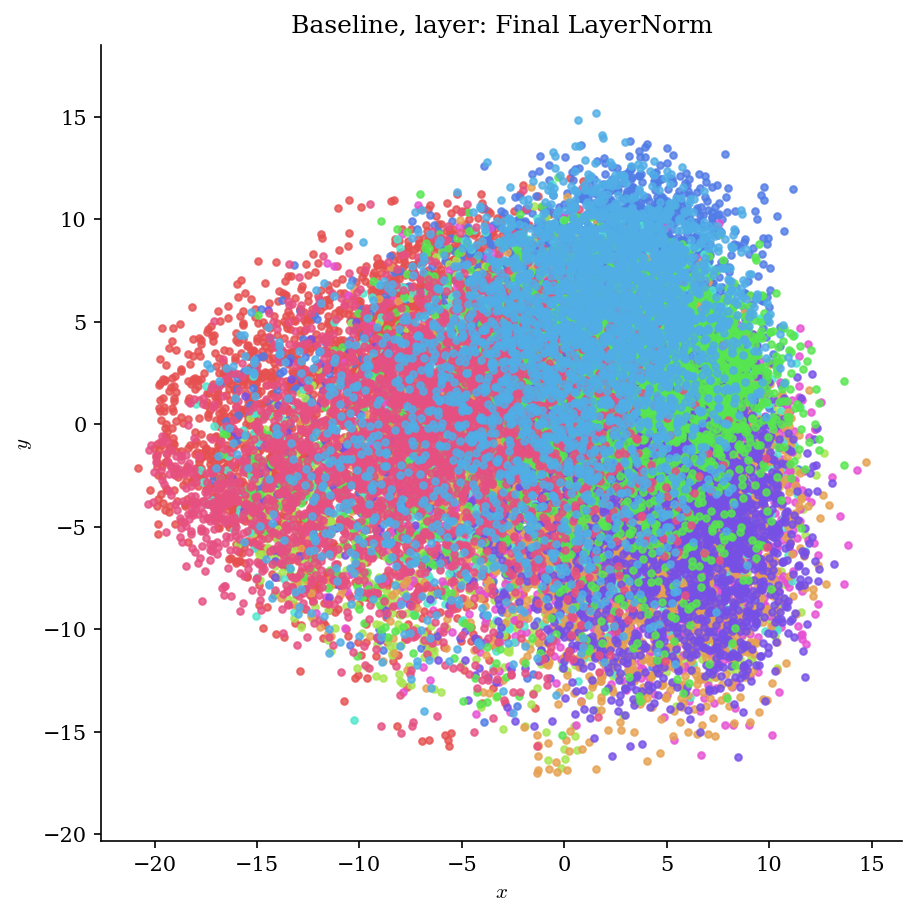}
    \caption{ViT-B (Baseline)}
\end{figure}

\newpage
\begin{figure}[H]
  \centering  
    \includegraphics[width=0.49\linewidth]{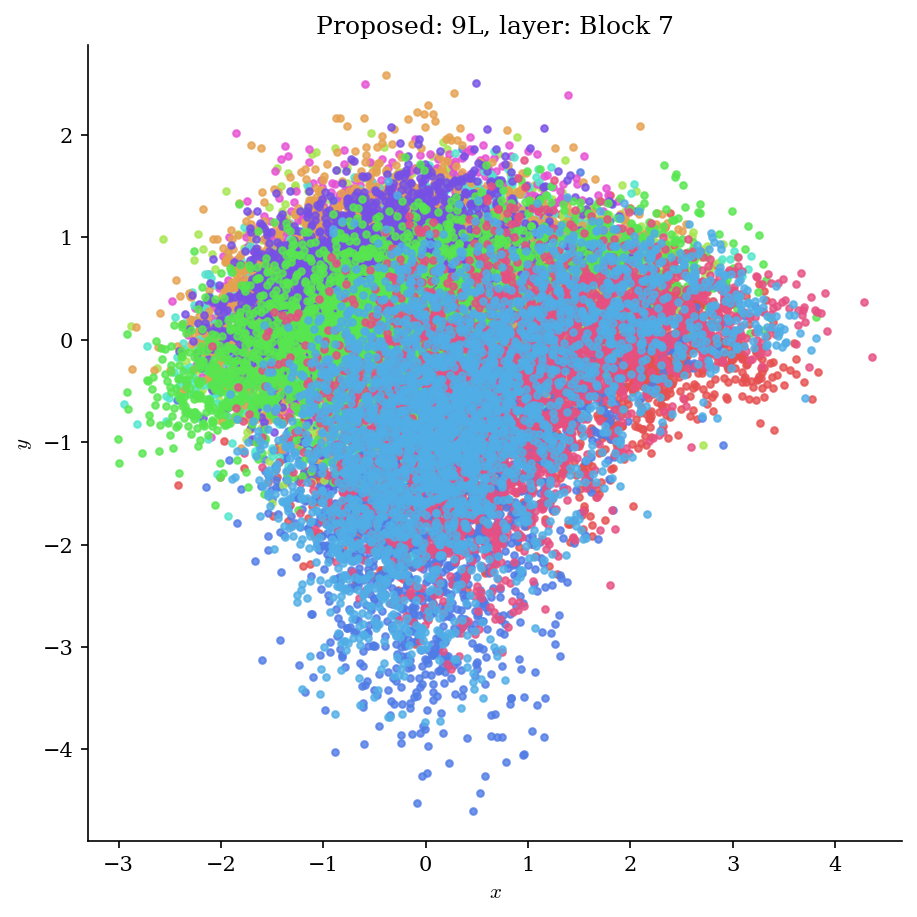}
  \hfill
    \includegraphics[width=0.49\linewidth]{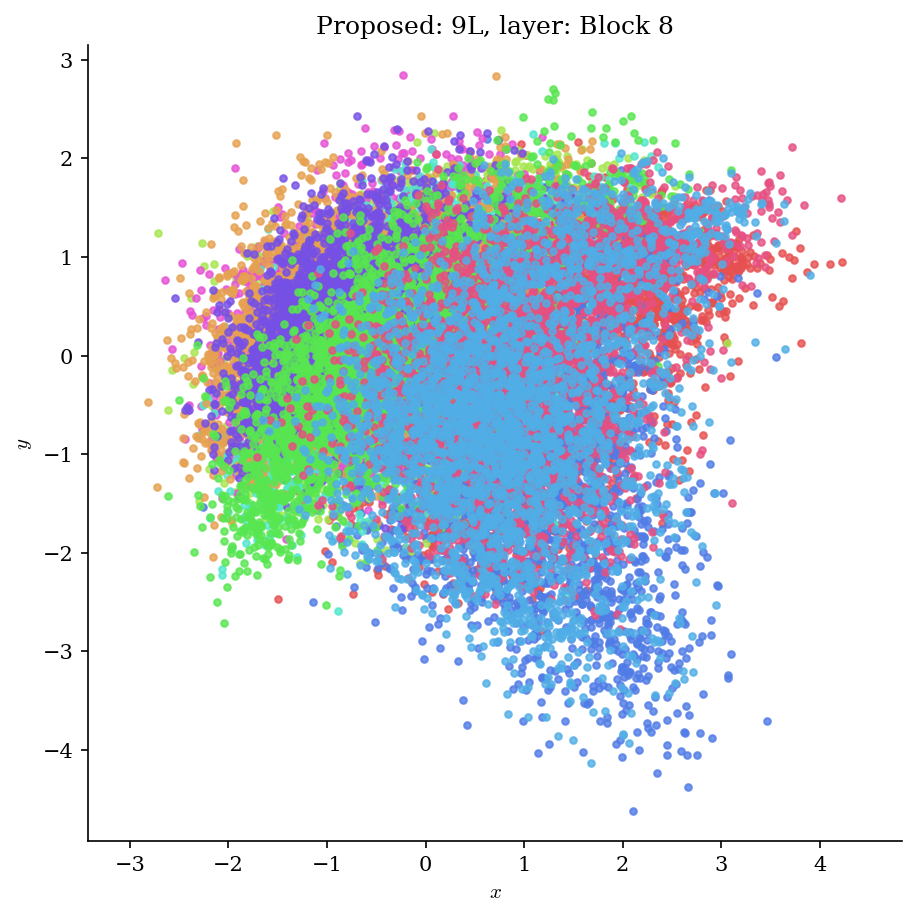}
  
  \vspace{-0.3em}
  
    \includegraphics[width=0.49\linewidth]{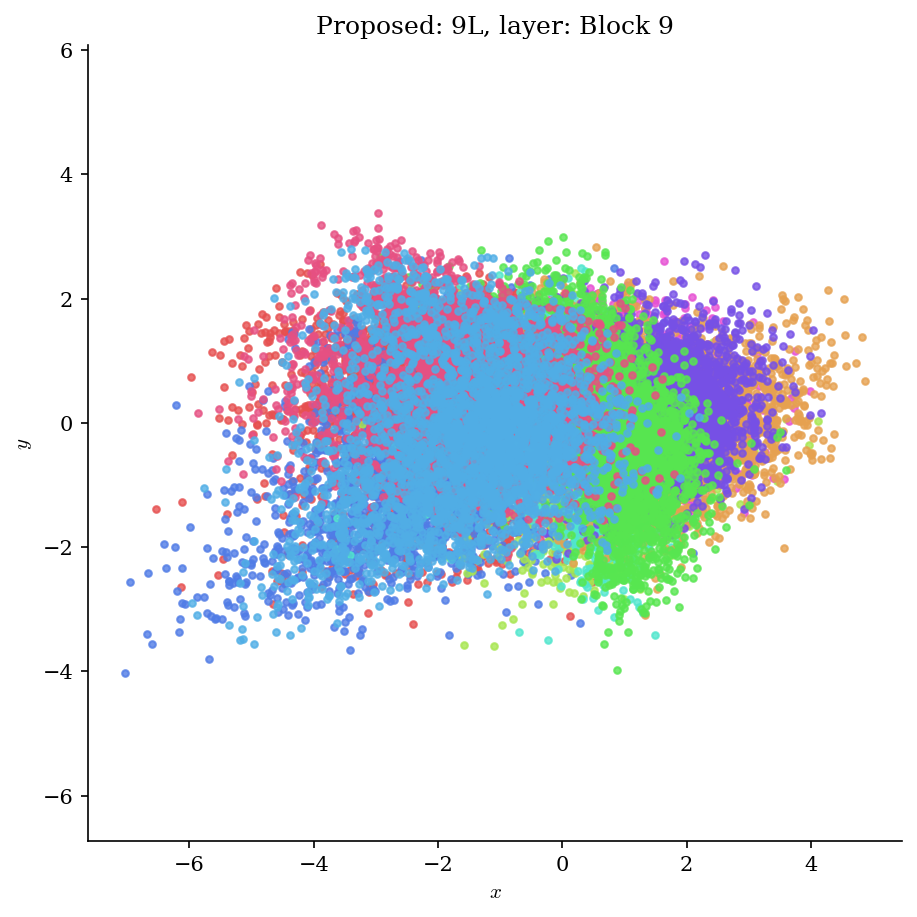}
  \hfill
    \includegraphics[width=0.49\linewidth]{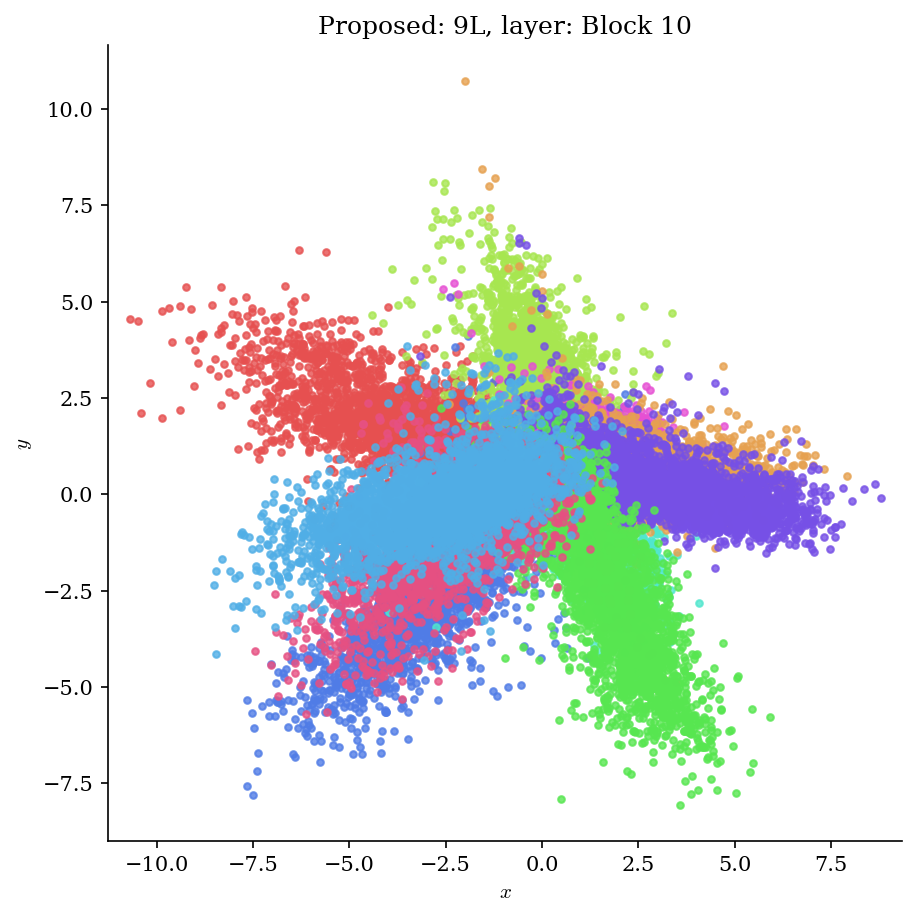}
  
  \vspace{-0.3em}
  
    \includegraphics[width=0.49\linewidth]{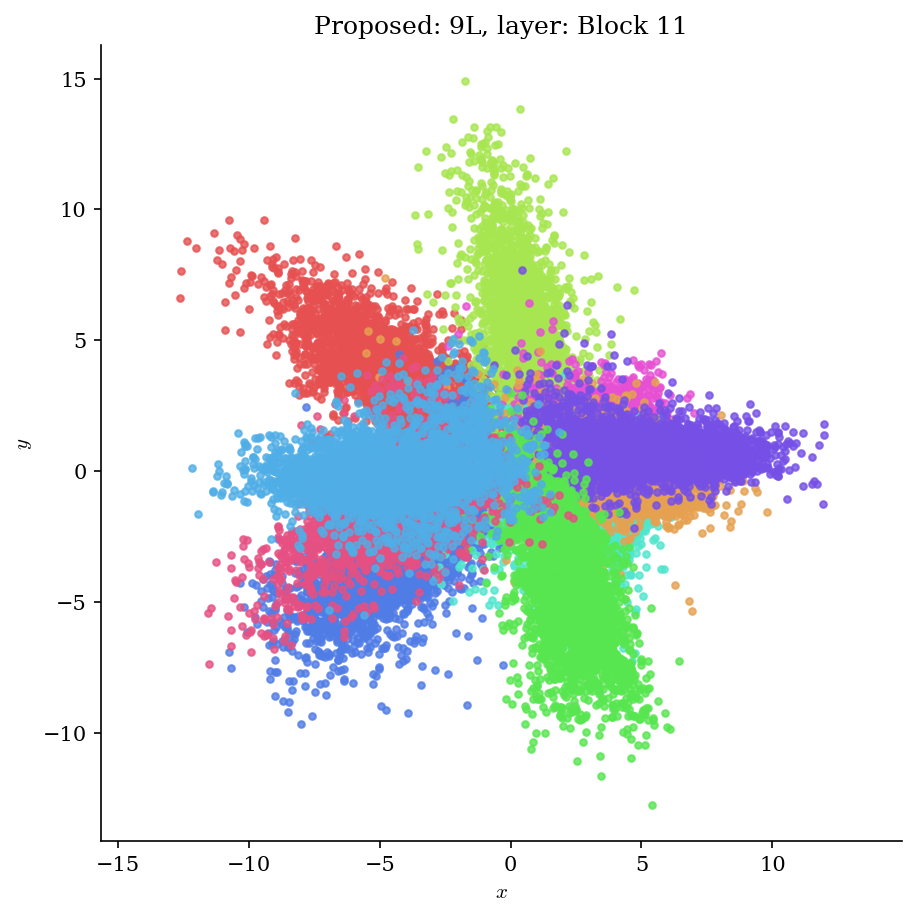}
  \hfill
    \includegraphics[width=0.49\linewidth]{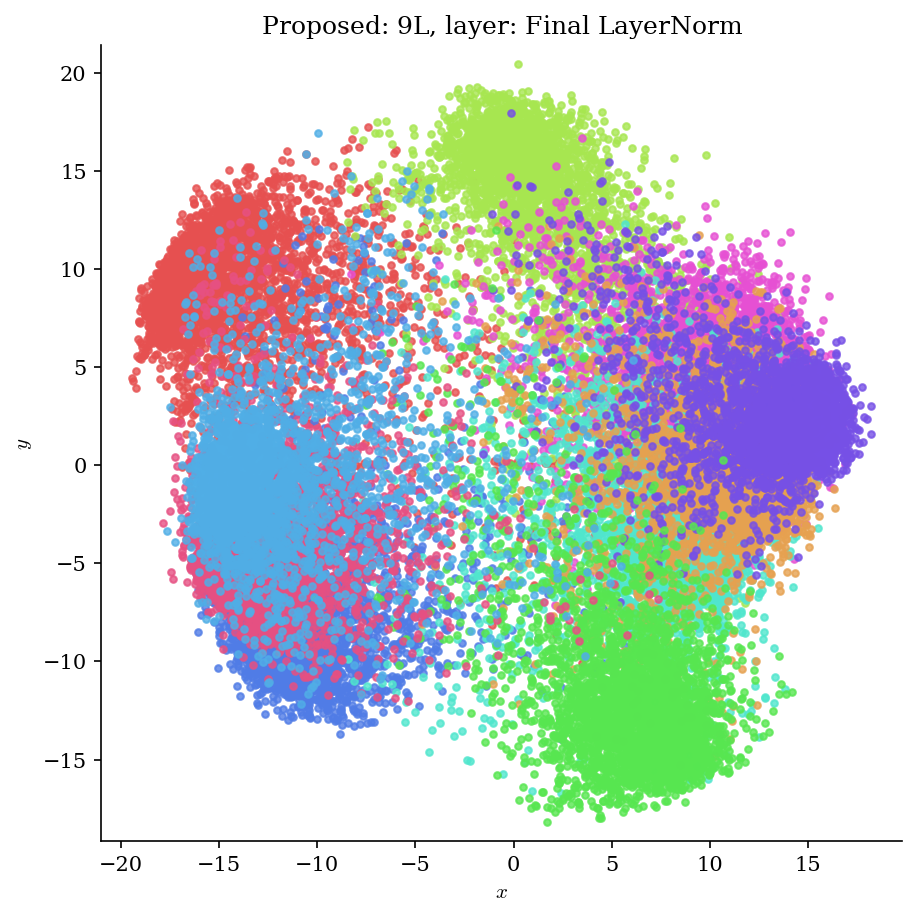}
    \caption{ViT-B-9L (Proposed)}
\end{figure}
\newpage
\begin{figure}[H]
  \centering  
    \includegraphics[width=0.49\linewidth]{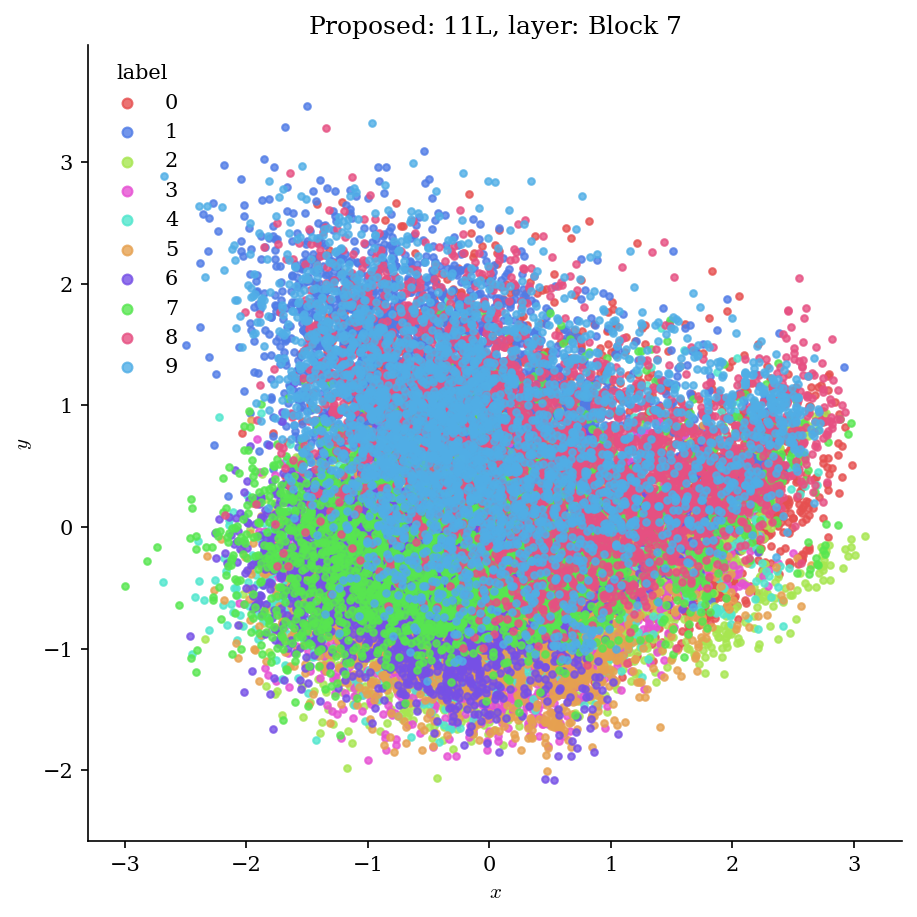}
  \hfill
    \includegraphics[width=0.49\linewidth]{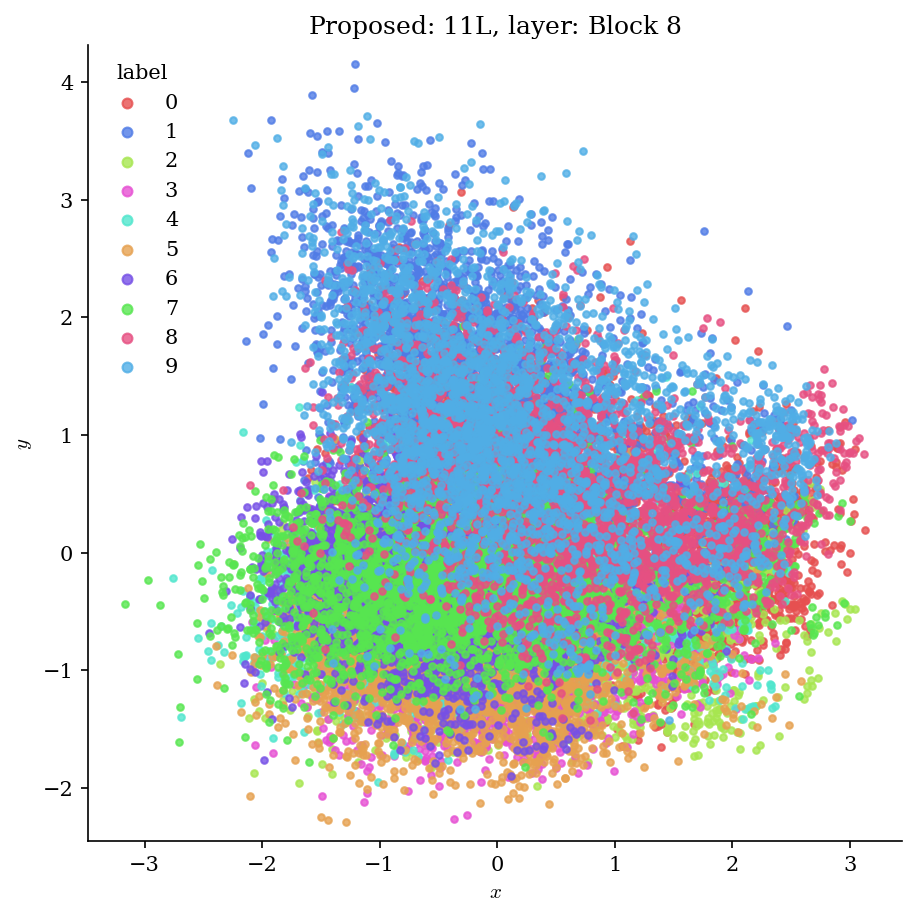}
  
  \vspace{-0.3em}
  
    \includegraphics[width=0.49\linewidth]{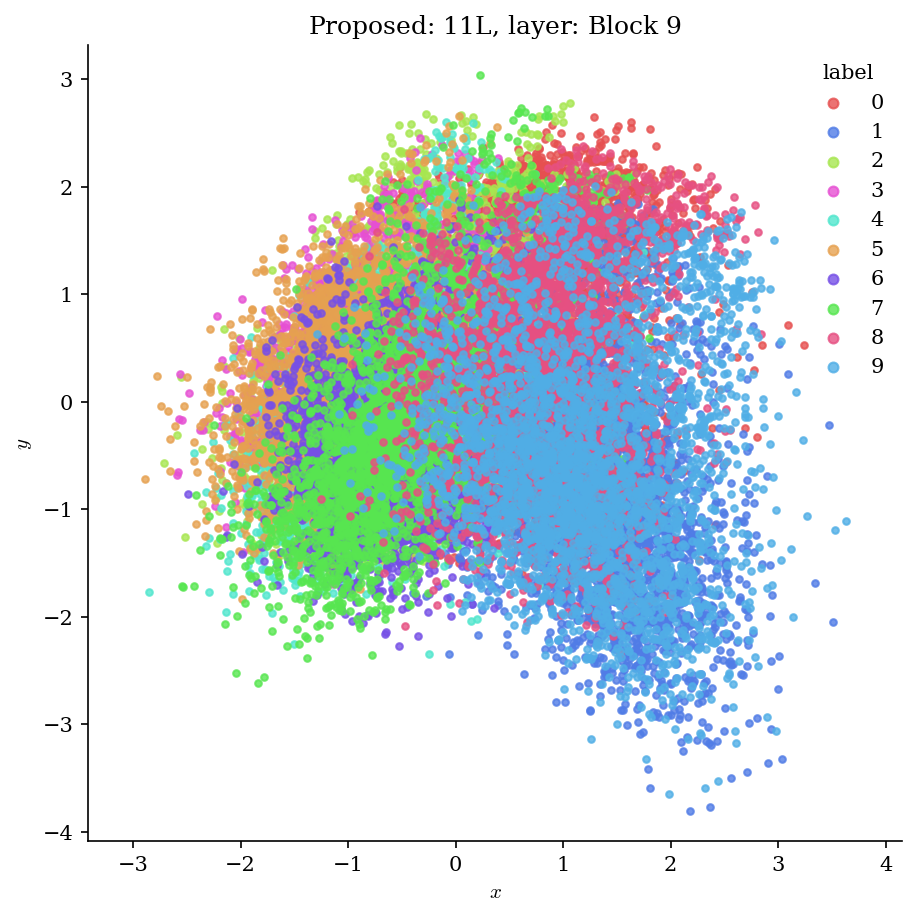}
  \hfill
    \includegraphics[width=0.49\linewidth]{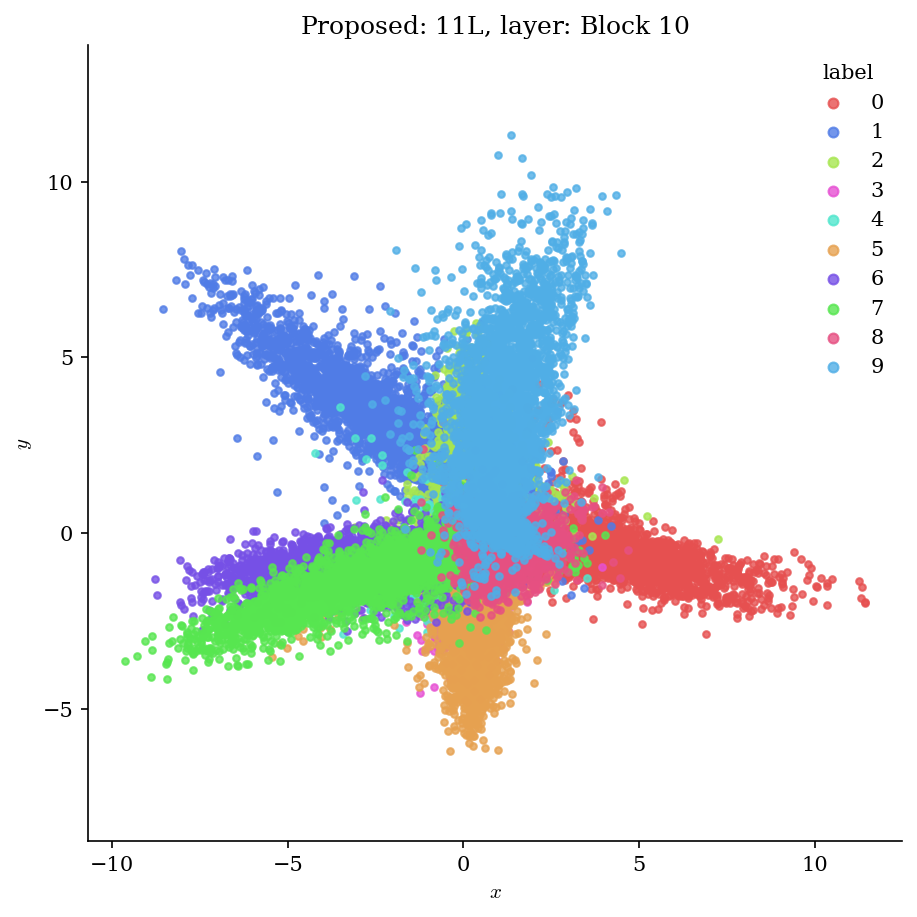}
  
  \vspace{-0.3em}
  
    \includegraphics[width=0.49\linewidth]{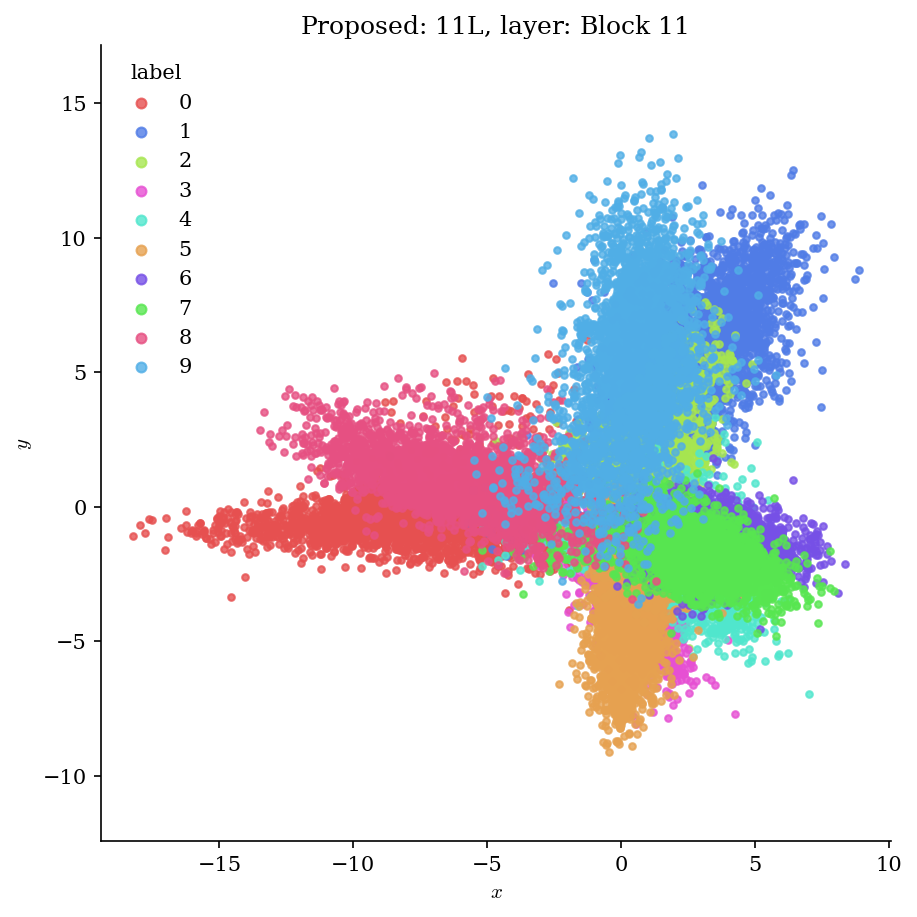}
  \hfill
    \includegraphics[width=0.49\linewidth]{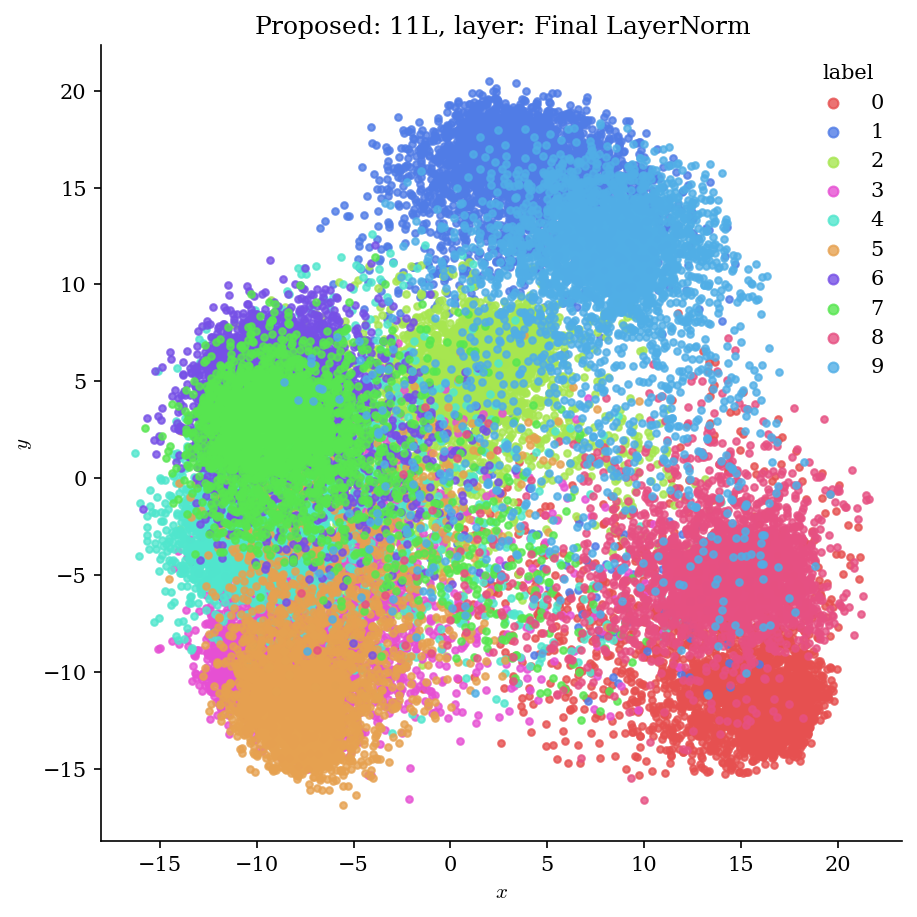}
    \caption{ViT-B-11L (Proposed)}
\end{figure}

\newpage
\begin{figure}[H]
  \centering  
    \includegraphics[width=0.49\linewidth]{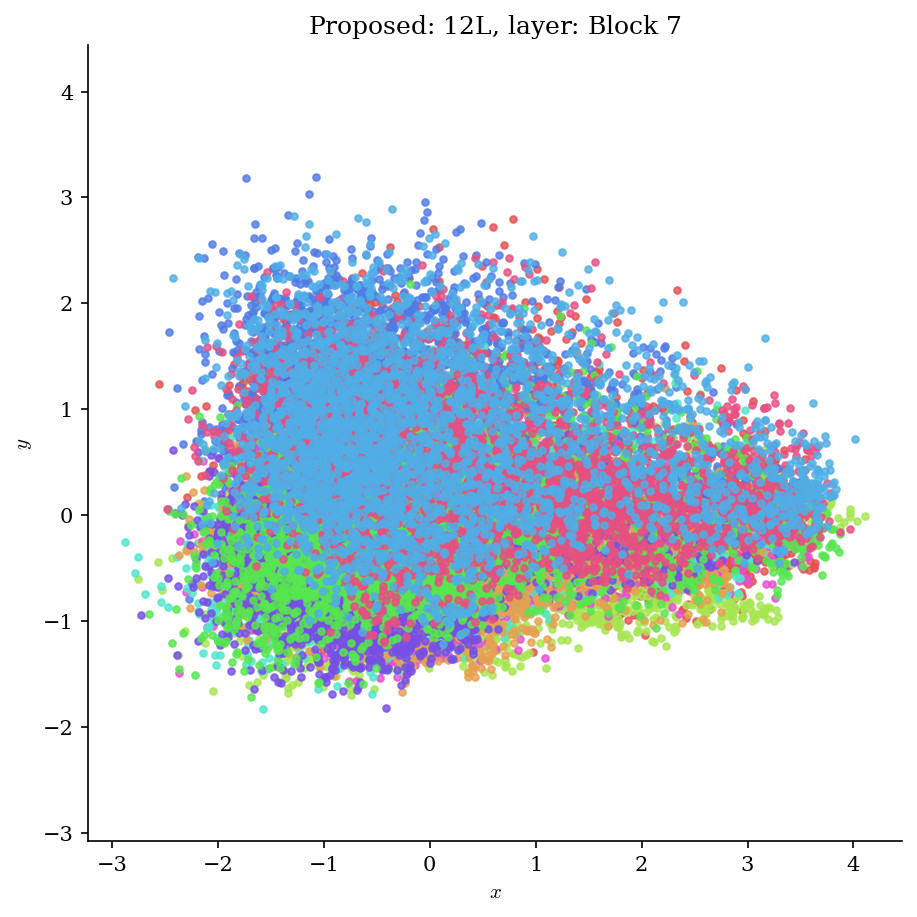}
  \hfill
    \includegraphics[width=0.49\linewidth]{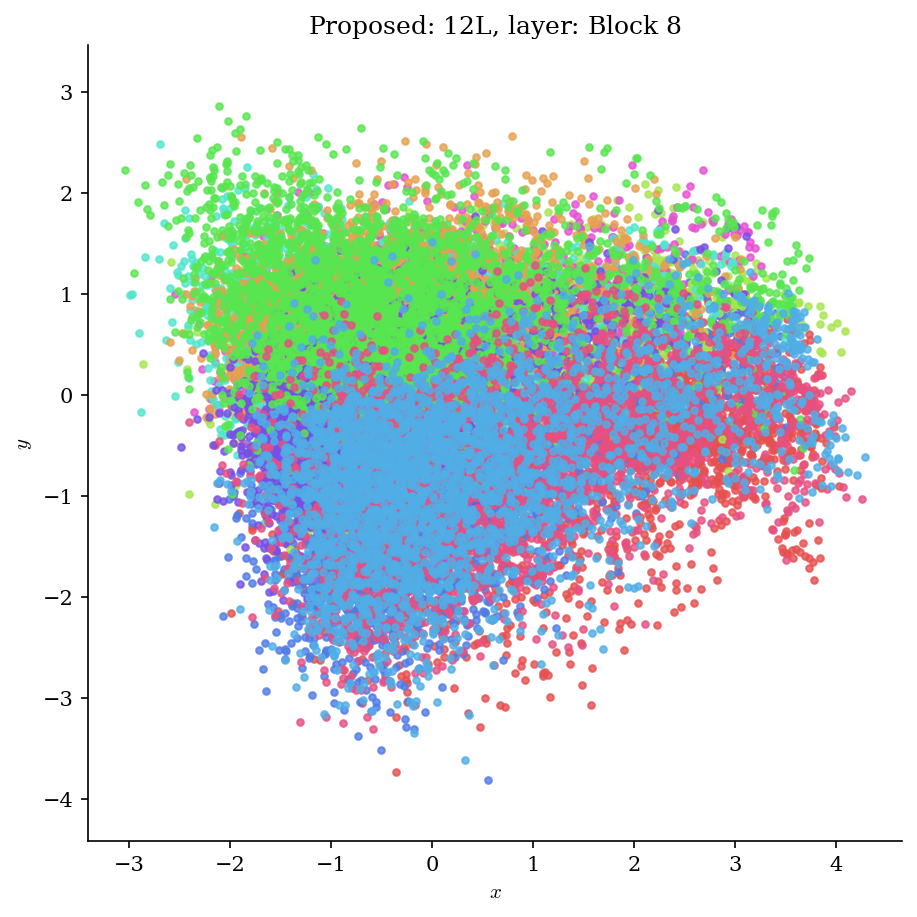}
  
  \vspace{-0.3em}
  
    \includegraphics[width=0.49\linewidth]{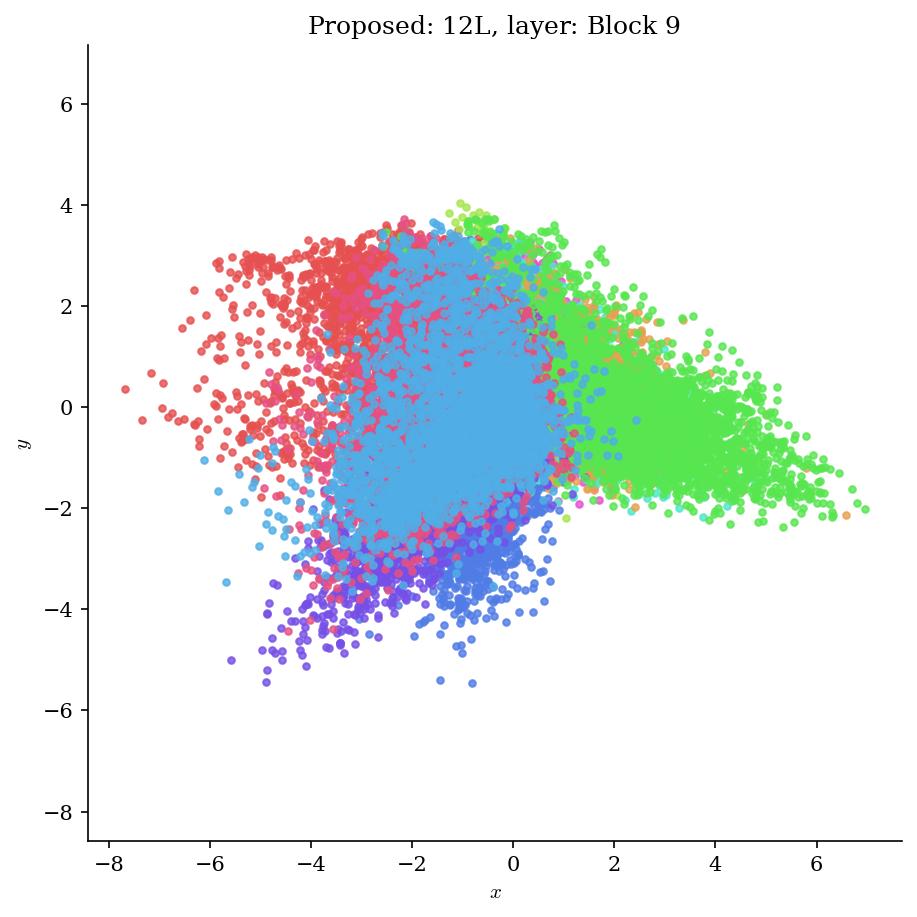}
  \hfill
    \includegraphics[width=0.49\linewidth]{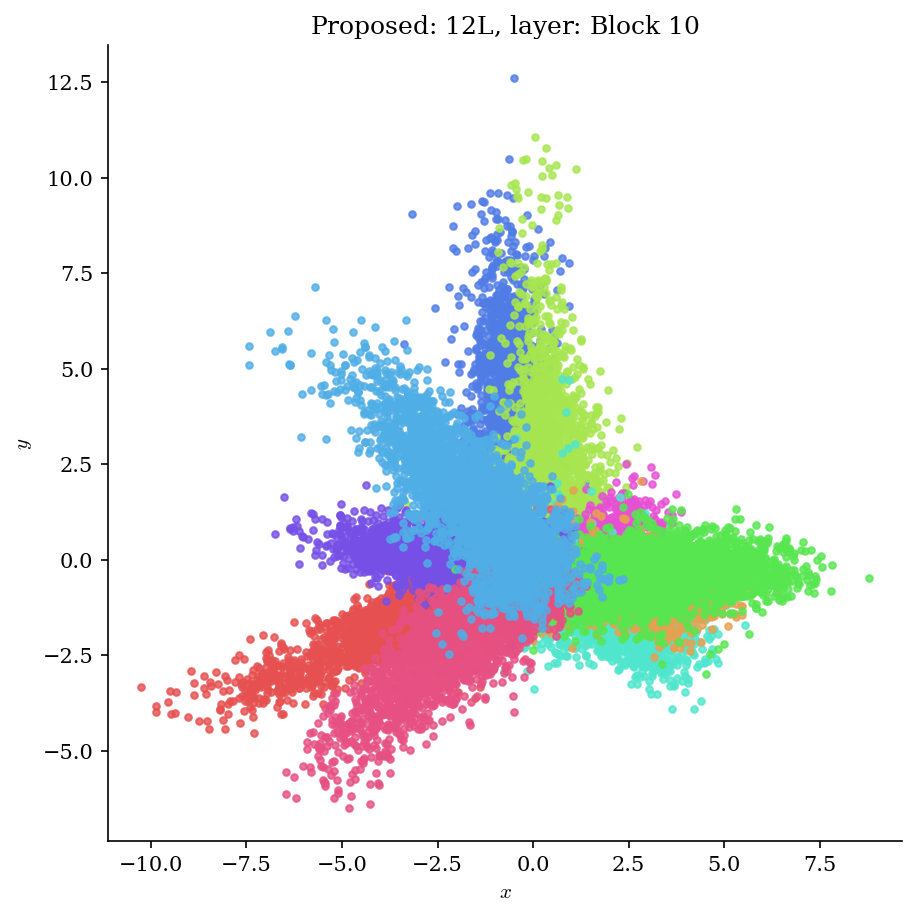}
  
  \vspace{-0.3em}
  
    \includegraphics[width=0.49\linewidth]{figures/pca/cifar10/deit_all_I_minus_layer_scale_block_base_patch4_LS/block_11_L.png}
  \hfill
    \includegraphics[width=0.49\linewidth]{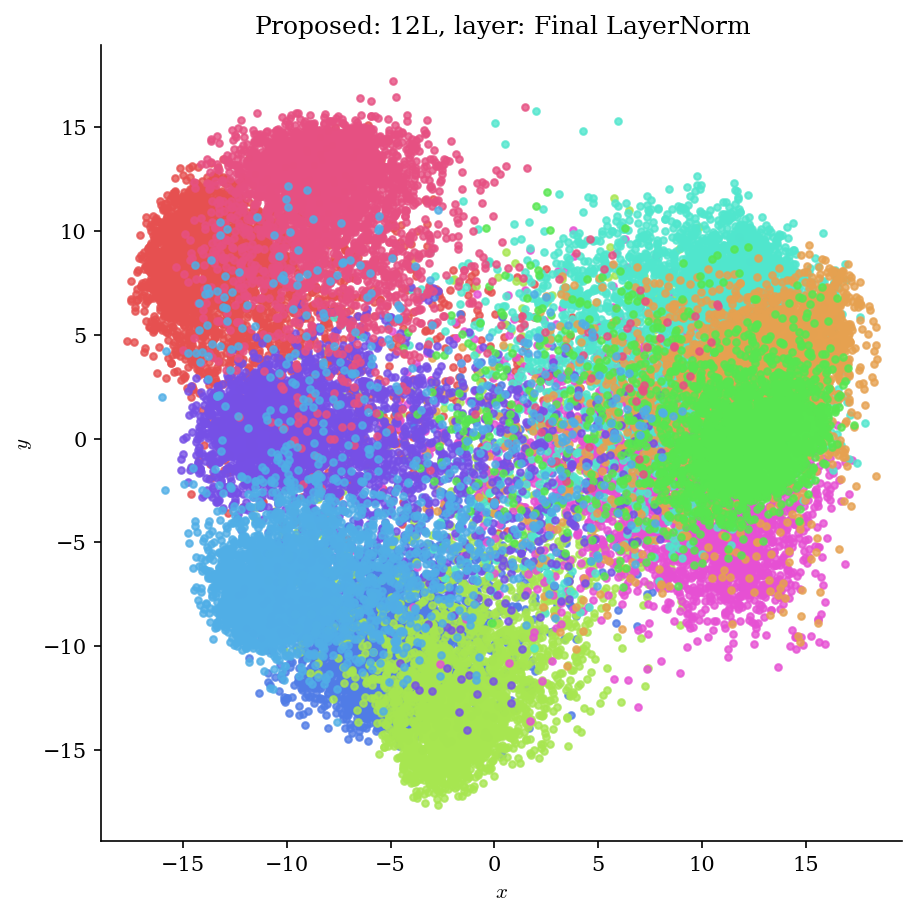}
    \caption{ViT-B-12L (Proposed)}
\end{figure}

\Needspace{0.50\textheight}
\paragraph{PCA (CIFAR100)}\leavevmode
\begin{figure}[H]
  \centering
  \setlength{\intextsep}{6pt}  
    \includegraphics[width=0.36\linewidth]{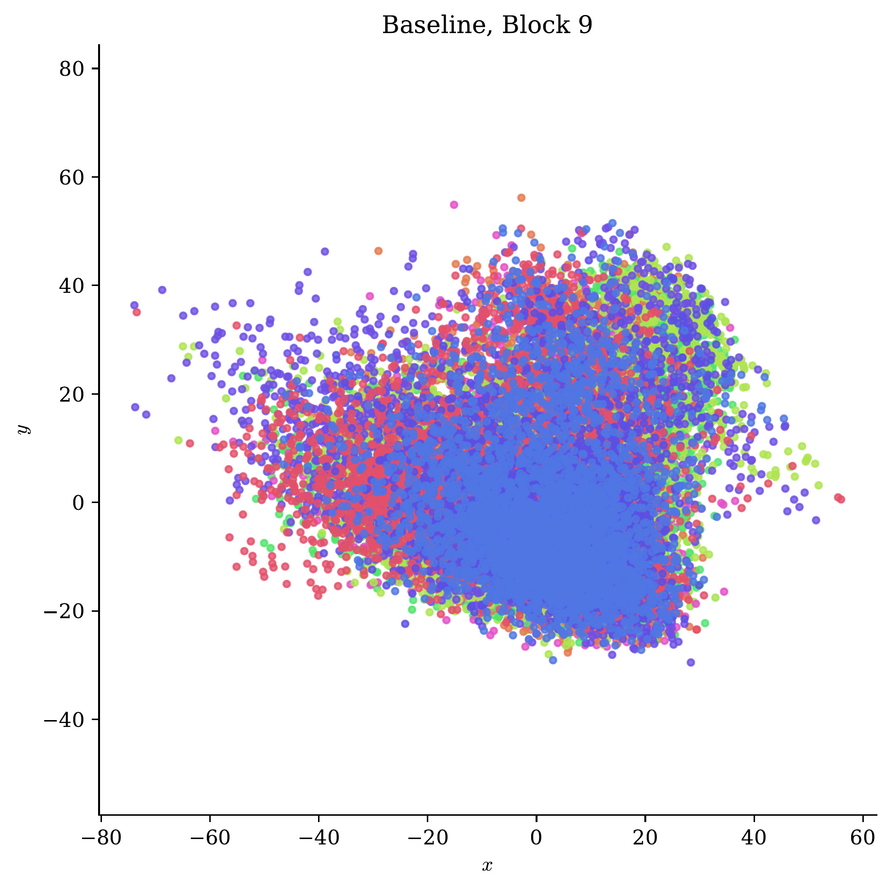}
  \hfill
    \includegraphics[width=0.36\linewidth]{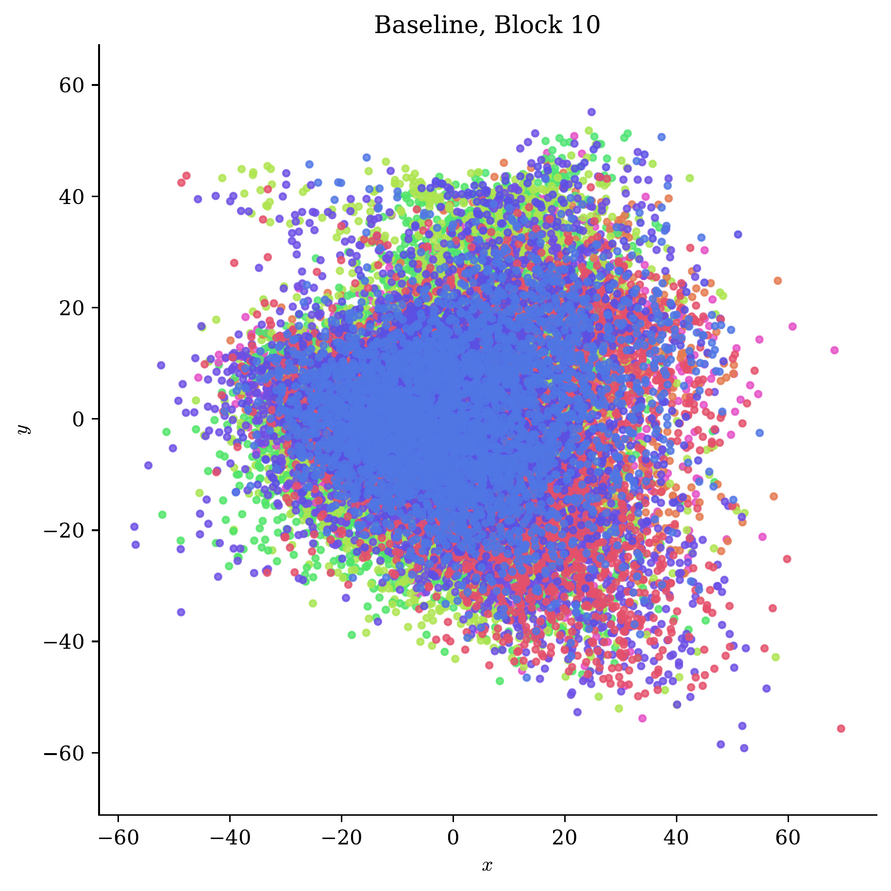}
  
  \vspace{-0.3em}
  
    \includegraphics[width=0.36\linewidth]{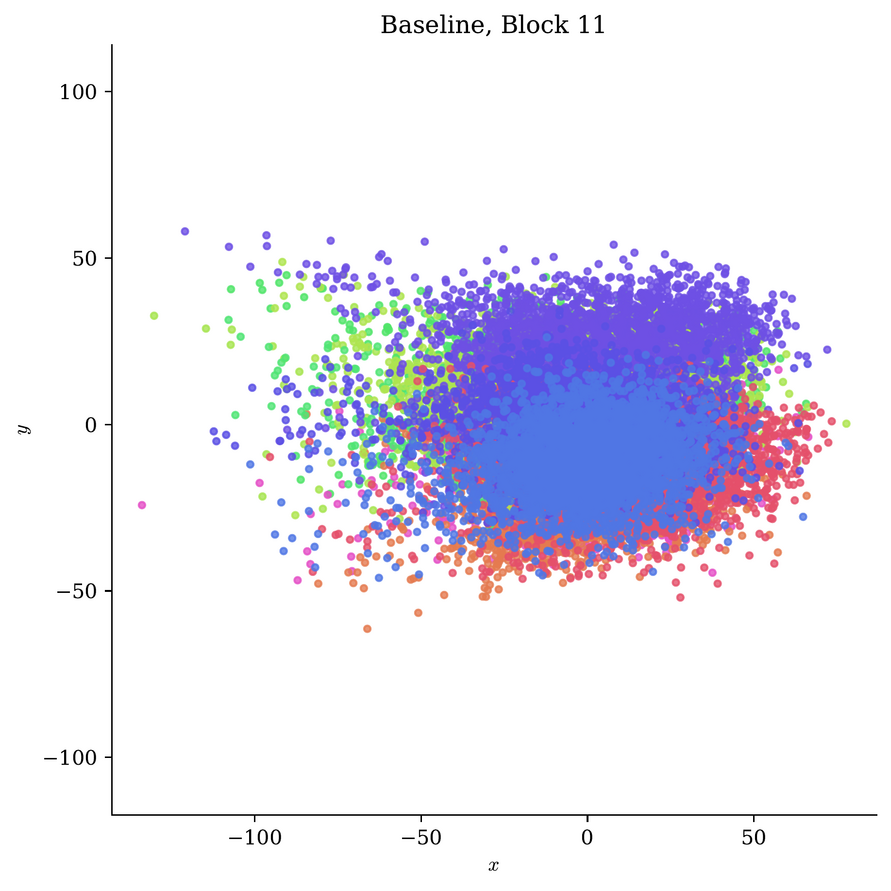}
  \hfill
    \includegraphics[width=0.36\linewidth]{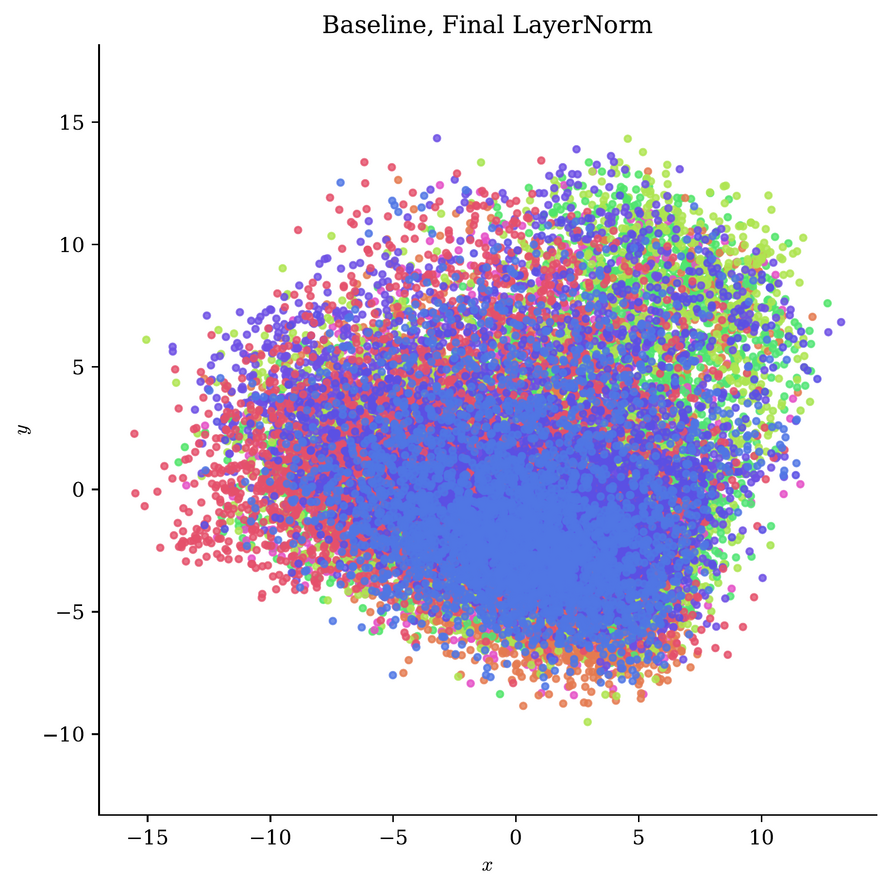}
    \caption{ViT-B (Baseline)}
\end{figure}
\begin{figure}[H]
  \centering
  \setlength{\intextsep}{6pt}  
    \includegraphics[width=0.36\linewidth]{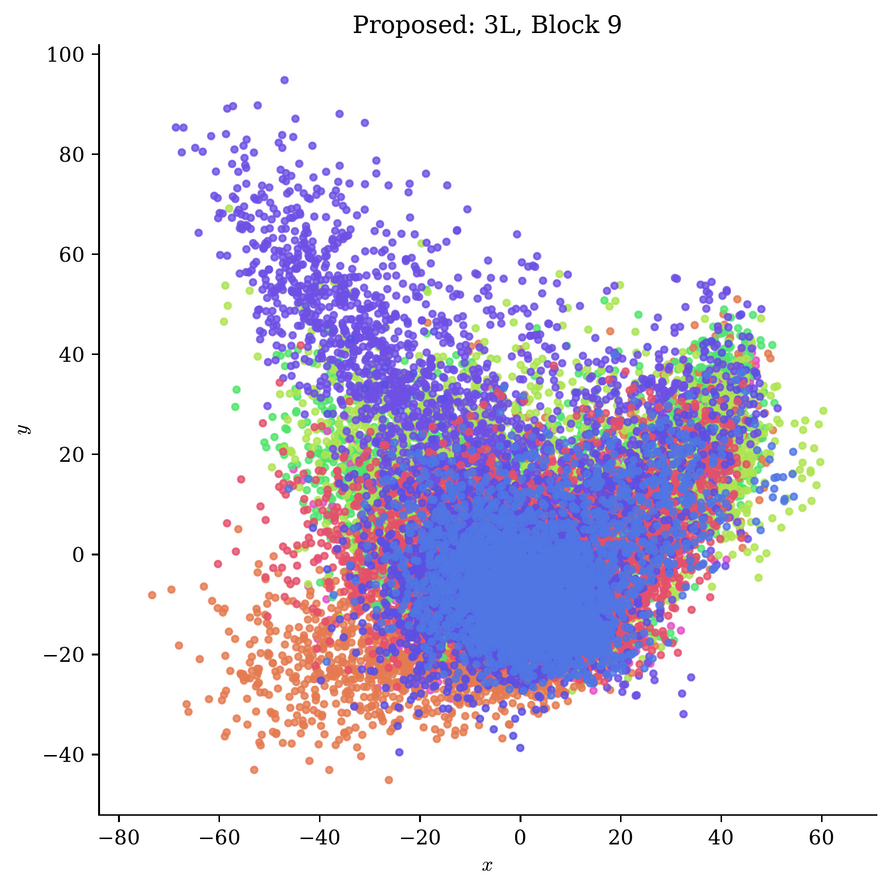}
  \hfill
    \includegraphics[width=0.36\linewidth]{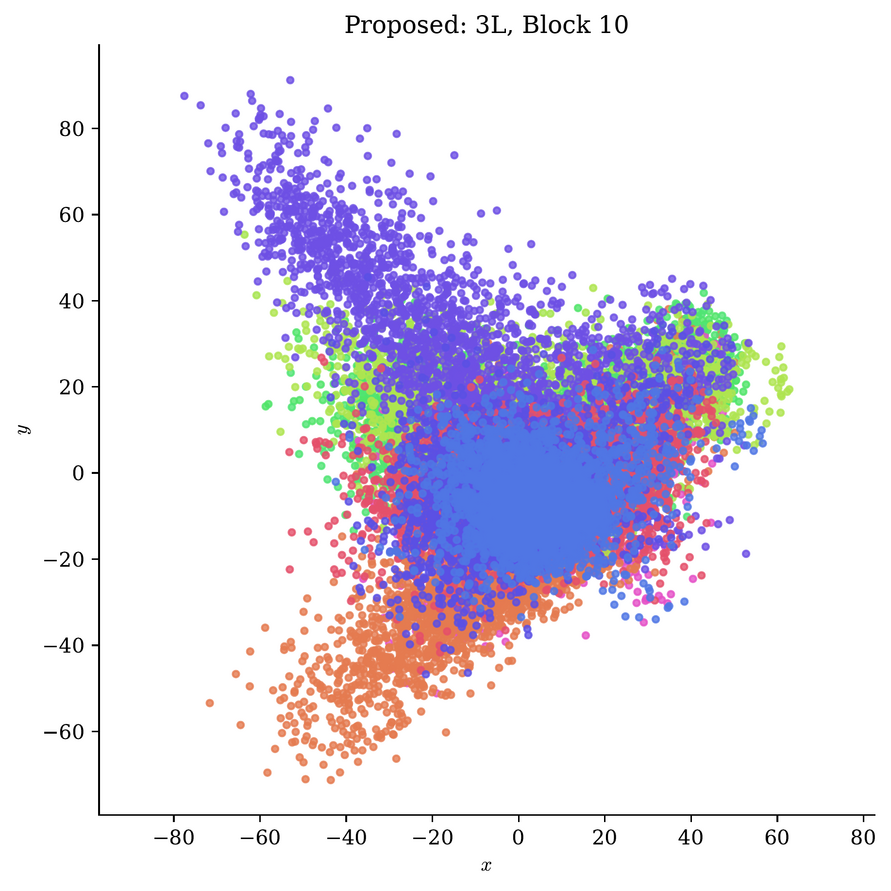}
  
  \vspace{-0.3em}
  
    \includegraphics[width=0.36\linewidth]{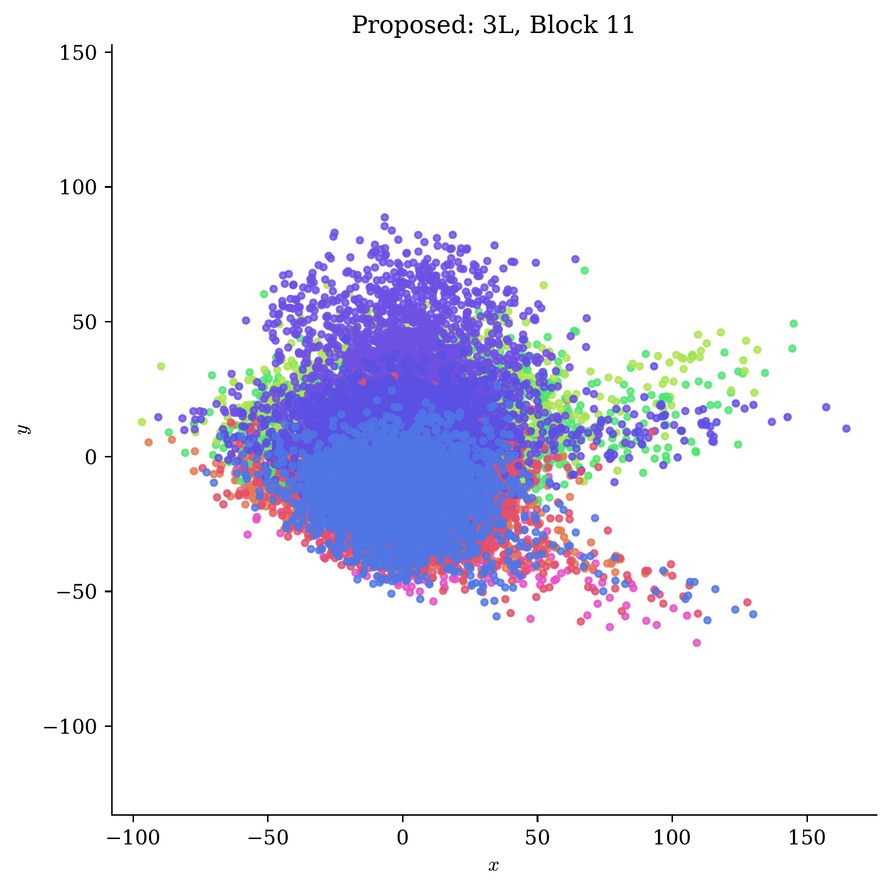}
  \hfill
    \includegraphics[width=0.36\linewidth]{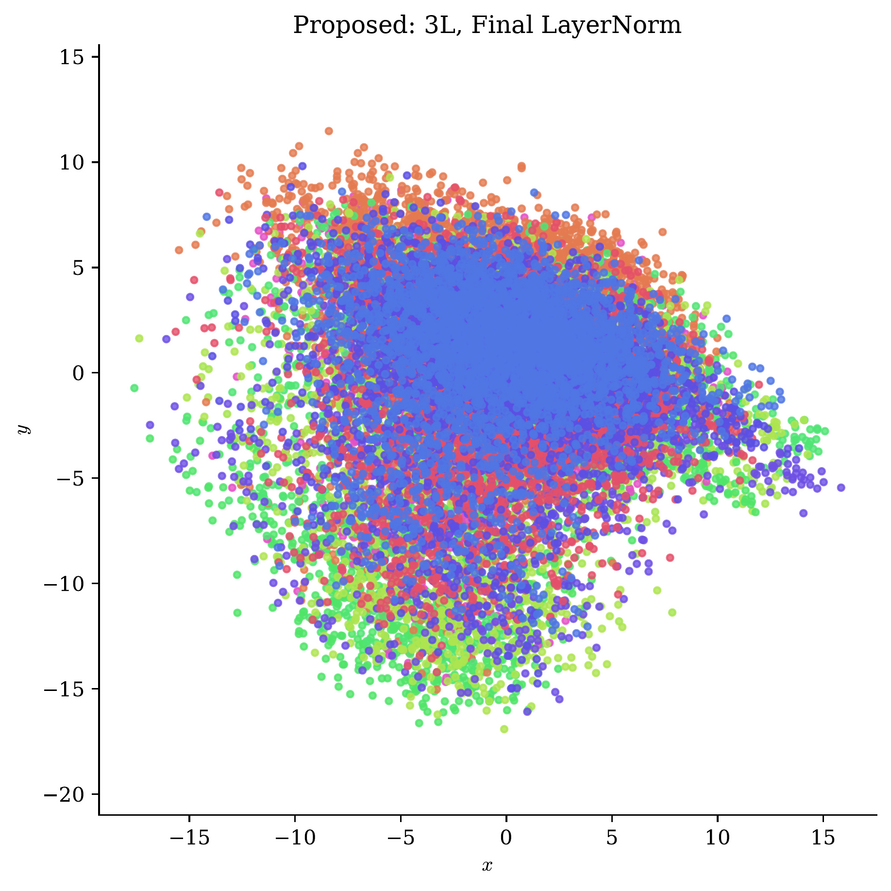}
    \caption{ViT-B-3L (Proposed)}
\end{figure}

\begin{figure}[H]
  \centering
  \setlength{\intextsep}{6pt}  
    \includegraphics[width=0.36\linewidth]{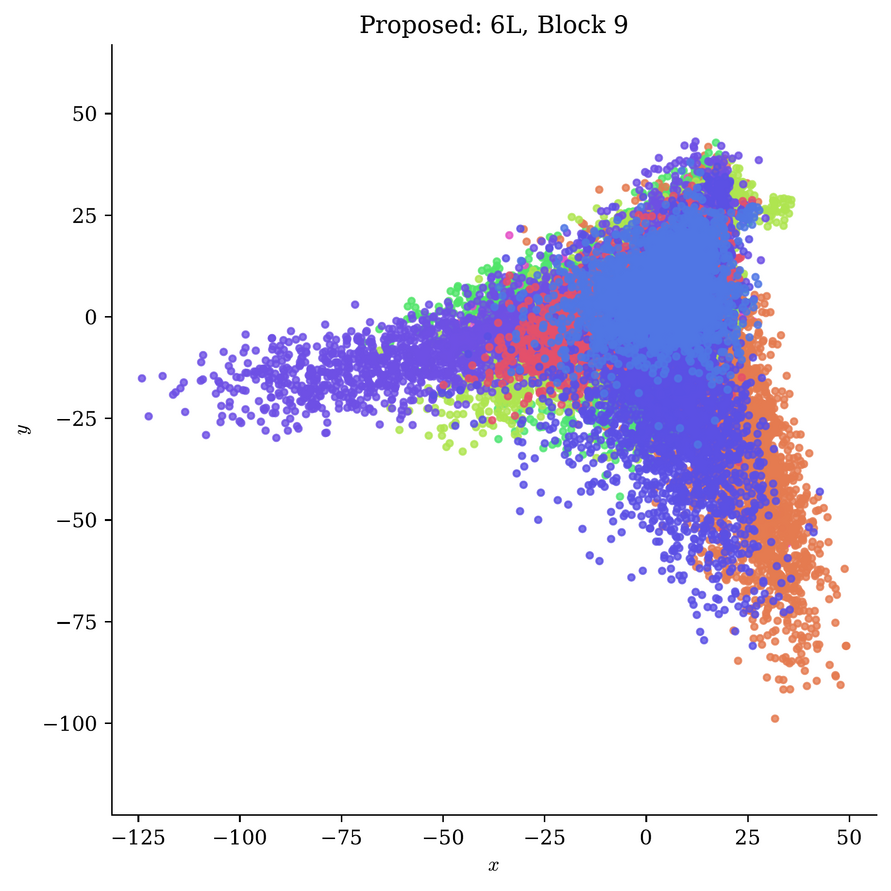}
  \hfill
    \includegraphics[width=0.36\linewidth]{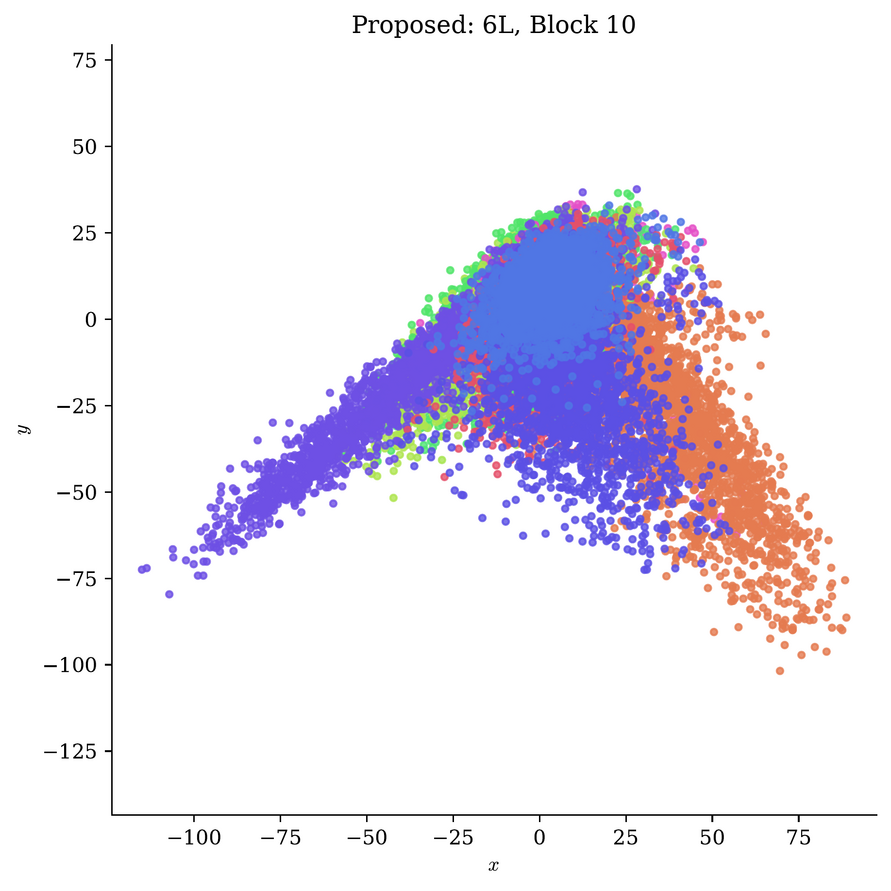}
  
  \vspace{-0.3em}
  
    \includegraphics[width=0.36\linewidth]{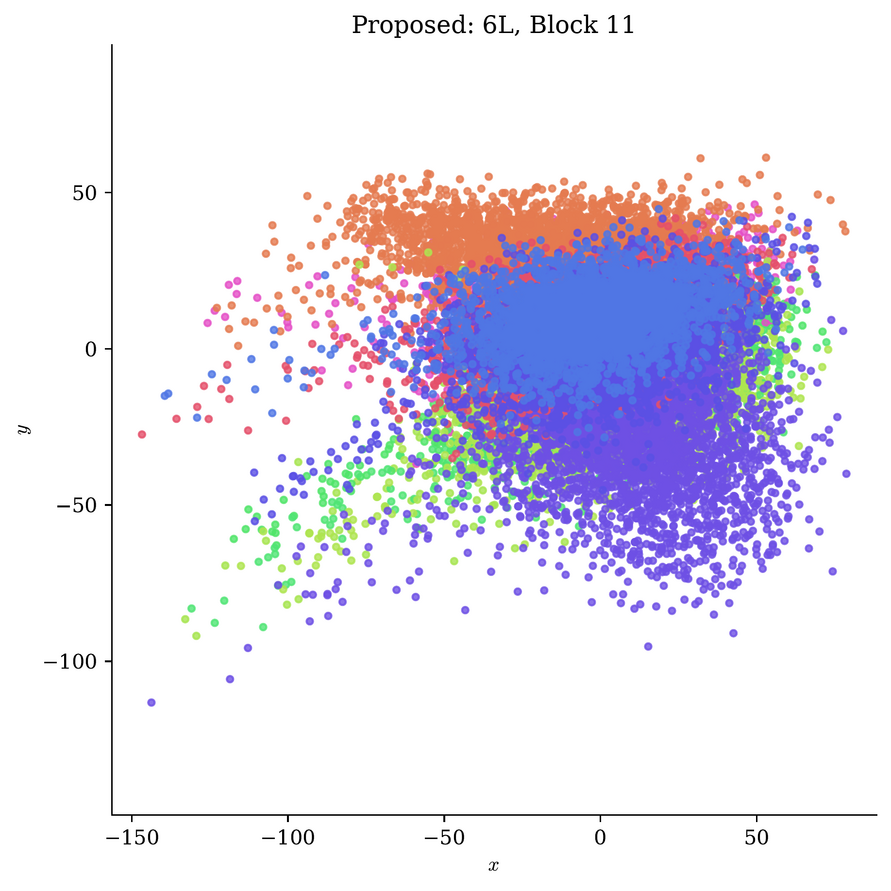}
  \hfill
    \includegraphics[width=0.36\linewidth]{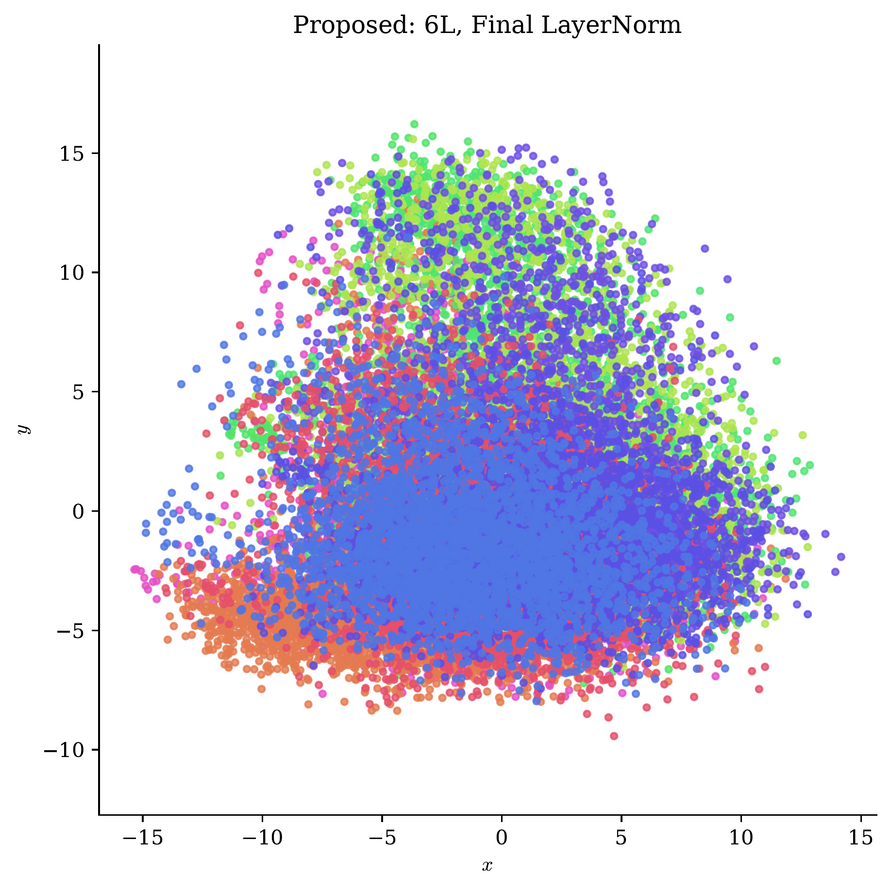}
    \caption{ViT-B-6L (Proposed)}
\end{figure}
\begin{figure}[H]
  \centering
  \setlength{\intextsep}{6pt}  
    \includegraphics[width=0.36\linewidth]{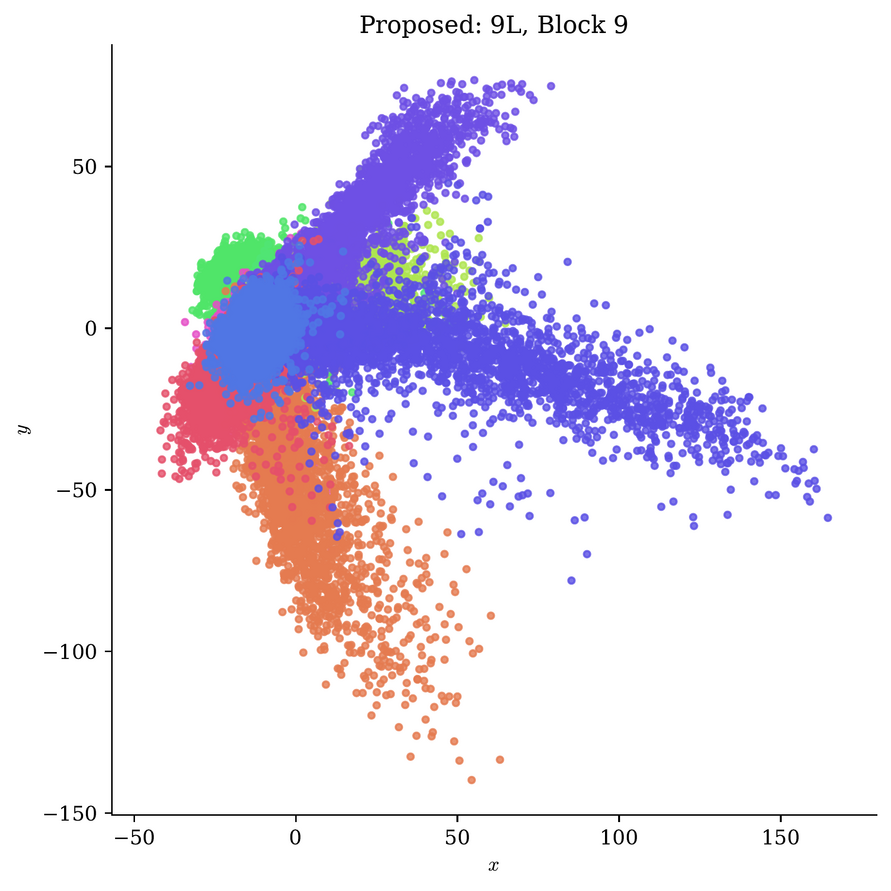}
  \hfill
    \includegraphics[width=0.36\linewidth]{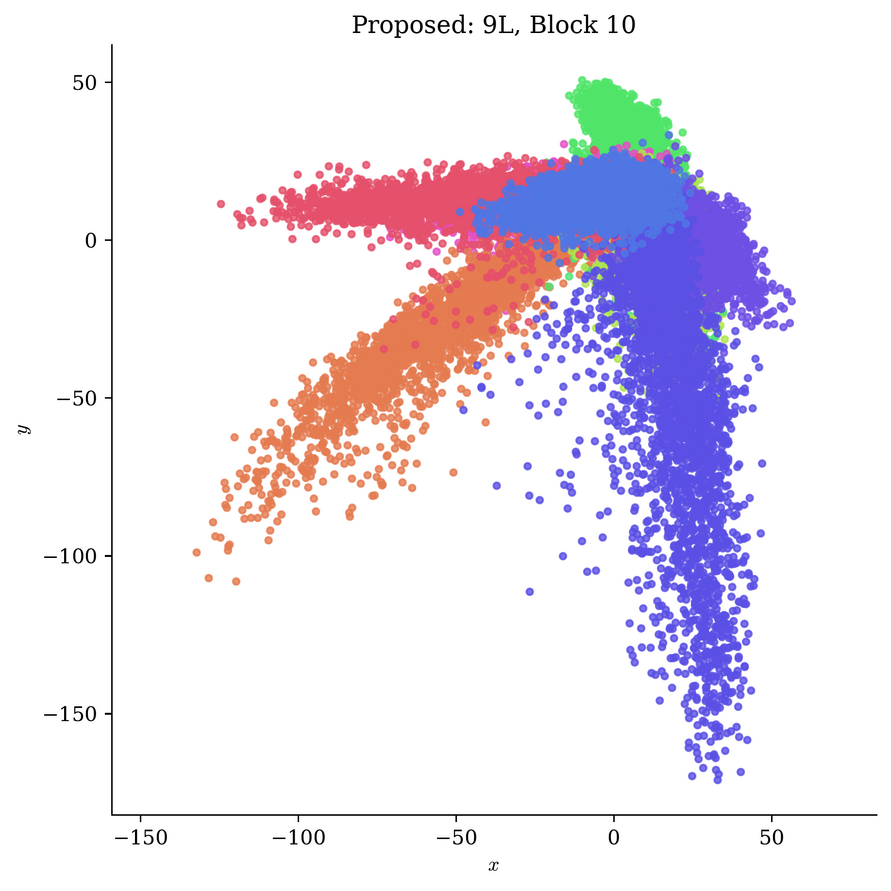}
  
  \vspace{-0.3em}
  
    \includegraphics[width=0.36\linewidth]{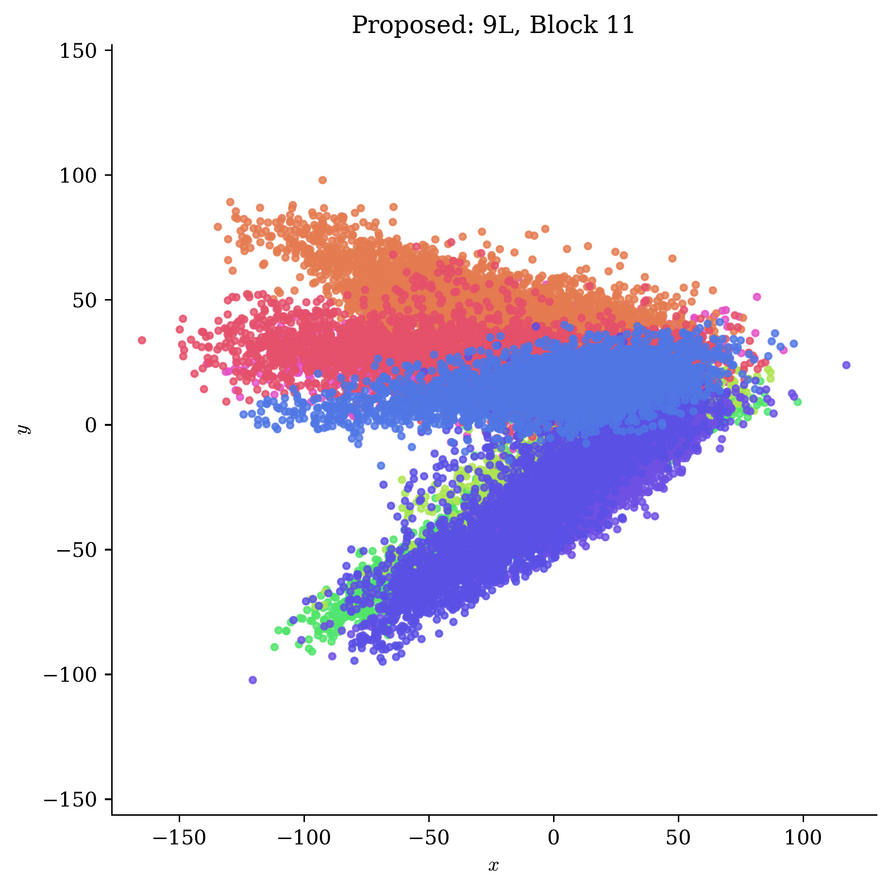}
  \hfill
    \includegraphics[width=0.36\linewidth]{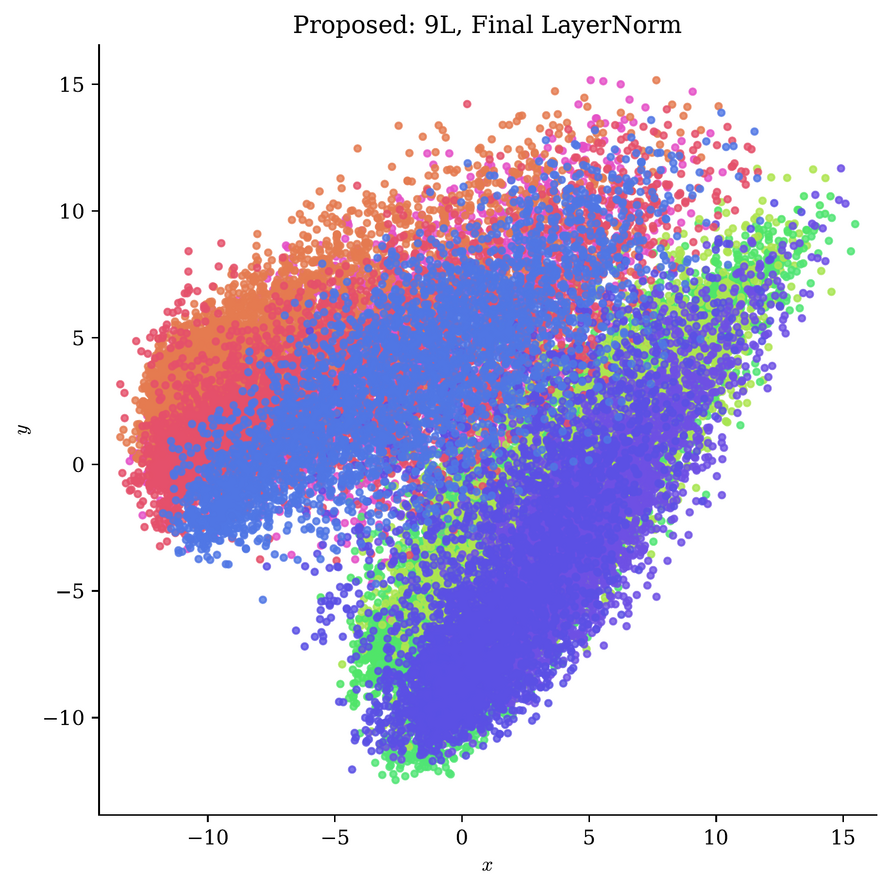}
    \caption{ViT-B-9L (Proposed)}
\end{figure}
\begin{figure}[H]
  \centering
  \setlength{\intextsep}{6pt}  
    \includegraphics[width=0.36\linewidth]{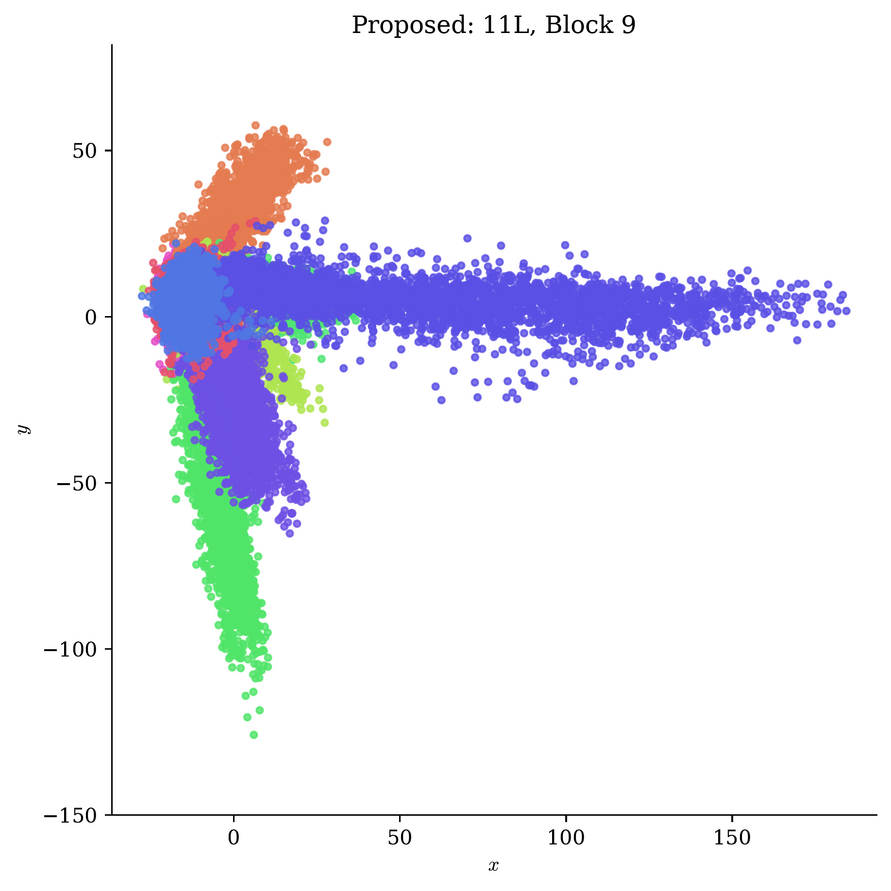}
  \hfill
    \includegraphics[width=0.36\linewidth]{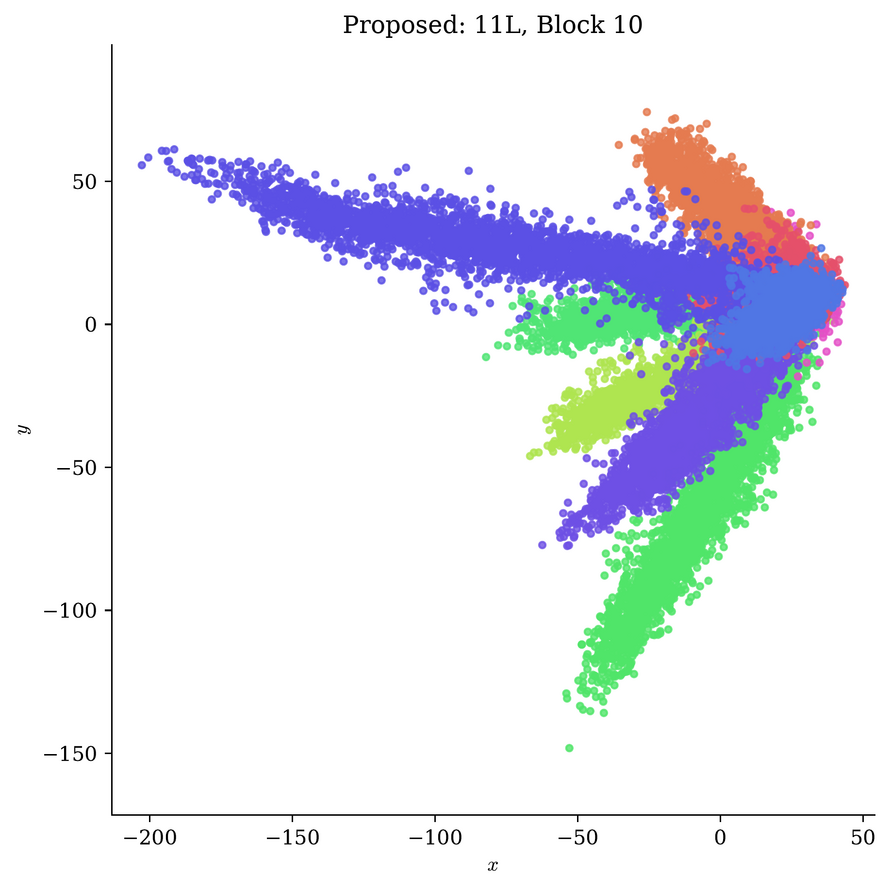}
  
  \vspace{-0.3em}
  
    \includegraphics[width=0.36\linewidth]{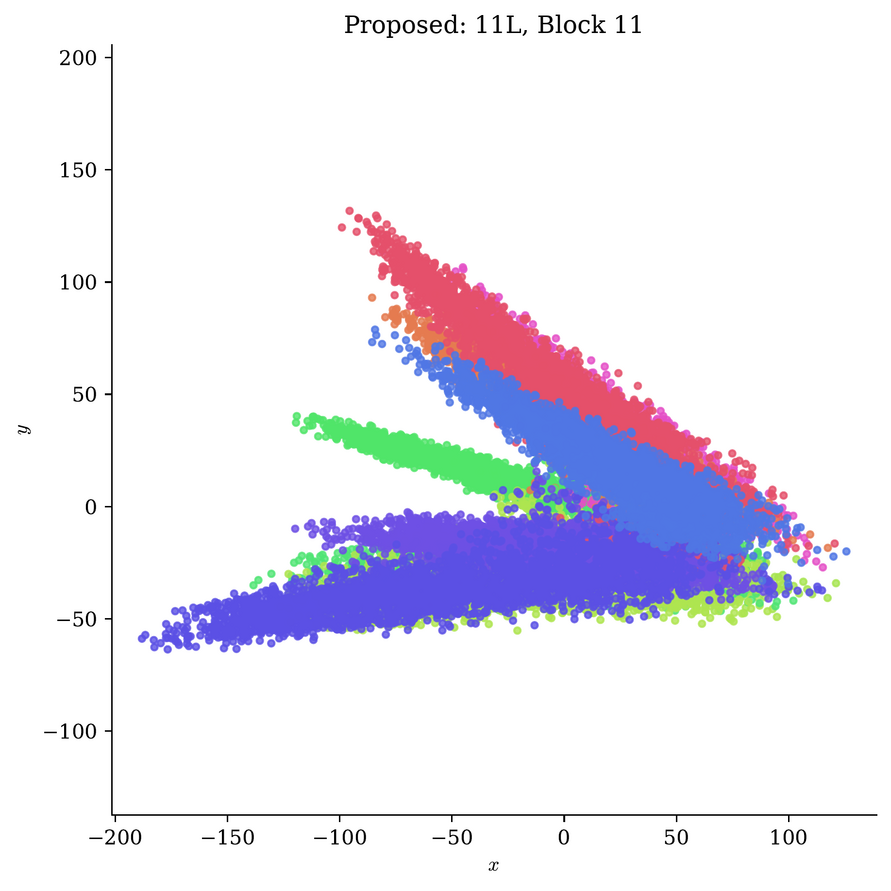}
  \hfill
    \includegraphics[width=0.36\linewidth]{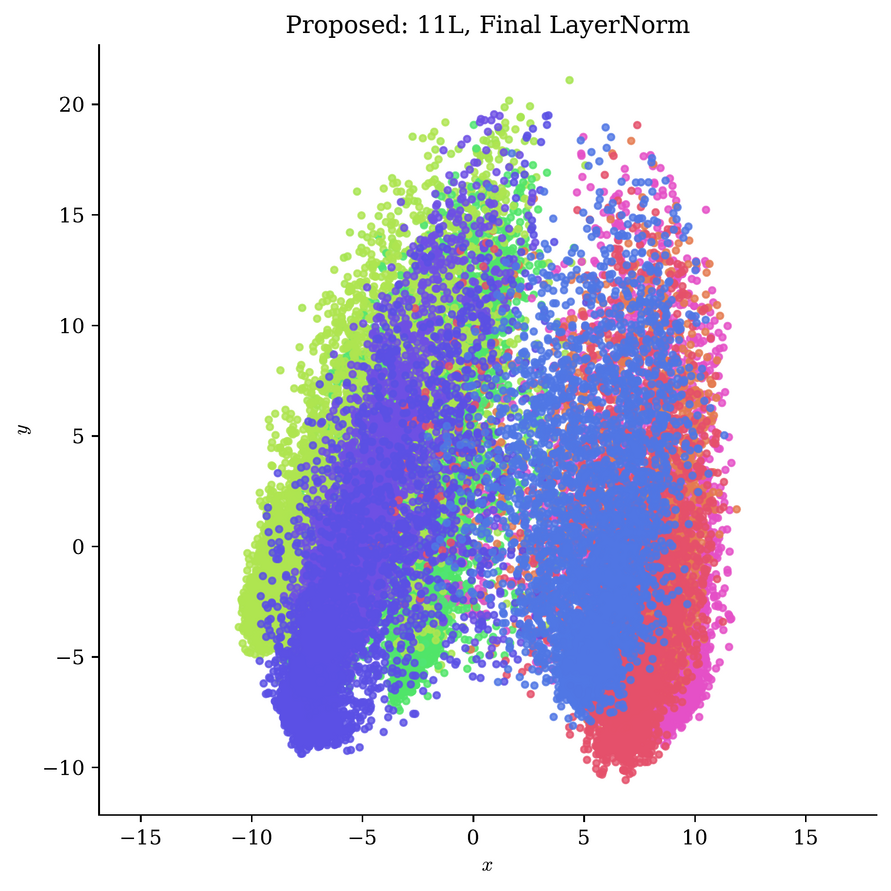}
    \caption{ViT-B-11L (Proposed)}
\end{figure}

\begin{figure}[H]
  \centering
  \setlength{\intextsep}{6pt}  
    \includegraphics[width=0.36\linewidth]{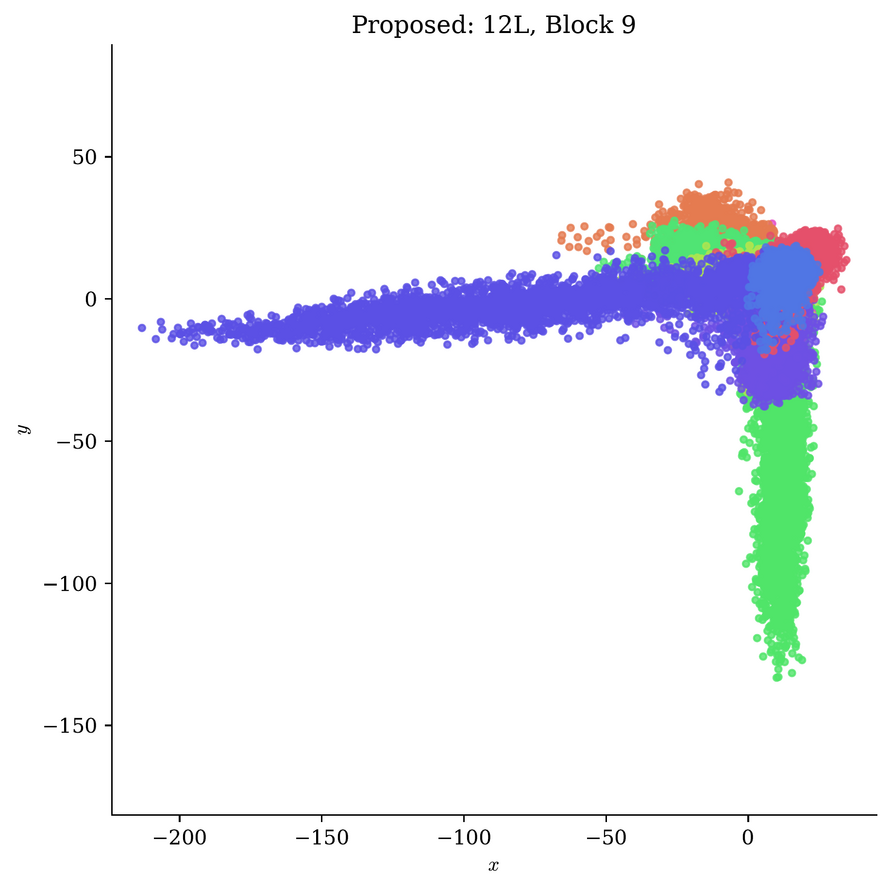}
  \hfill
    \includegraphics[width=0.36\linewidth]{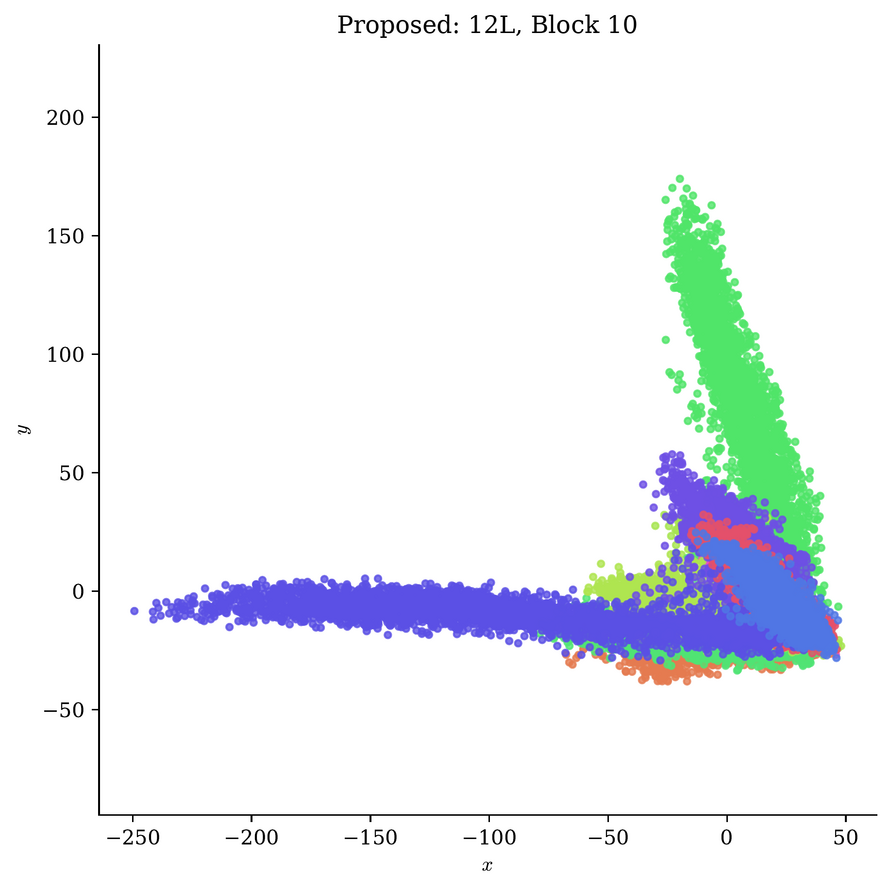}
  
  \vspace{-0.3em}
  
    \includegraphics[width=0.36\linewidth]{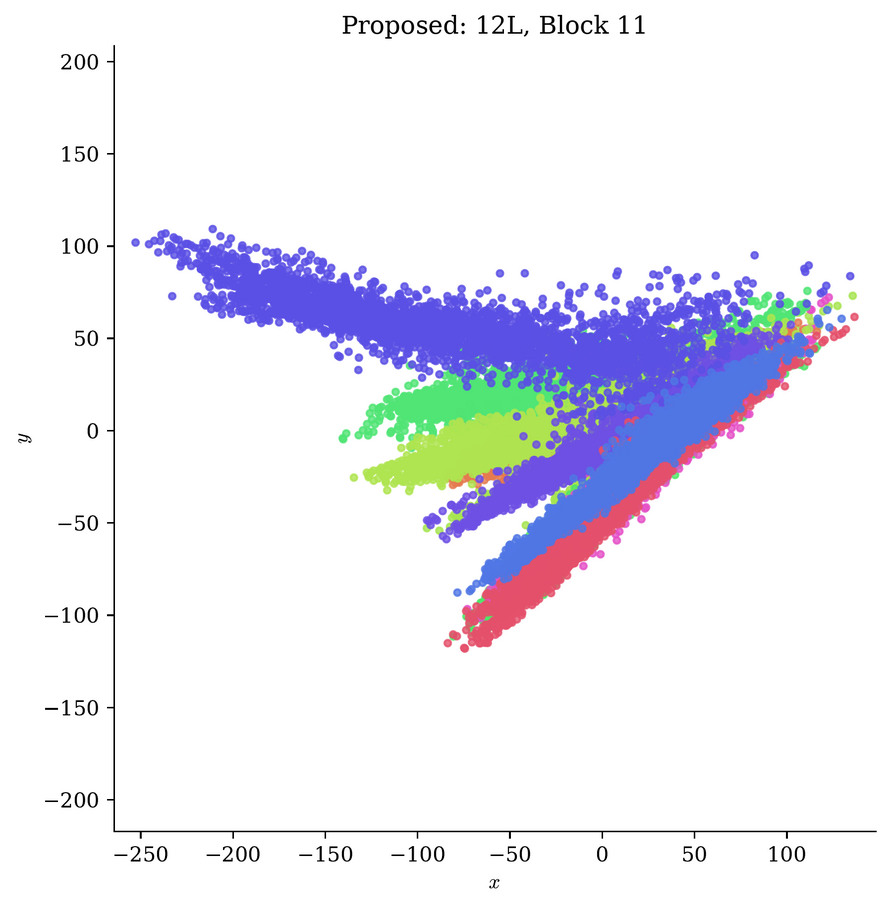}
  \hfill
    \includegraphics[width=0.36\linewidth]{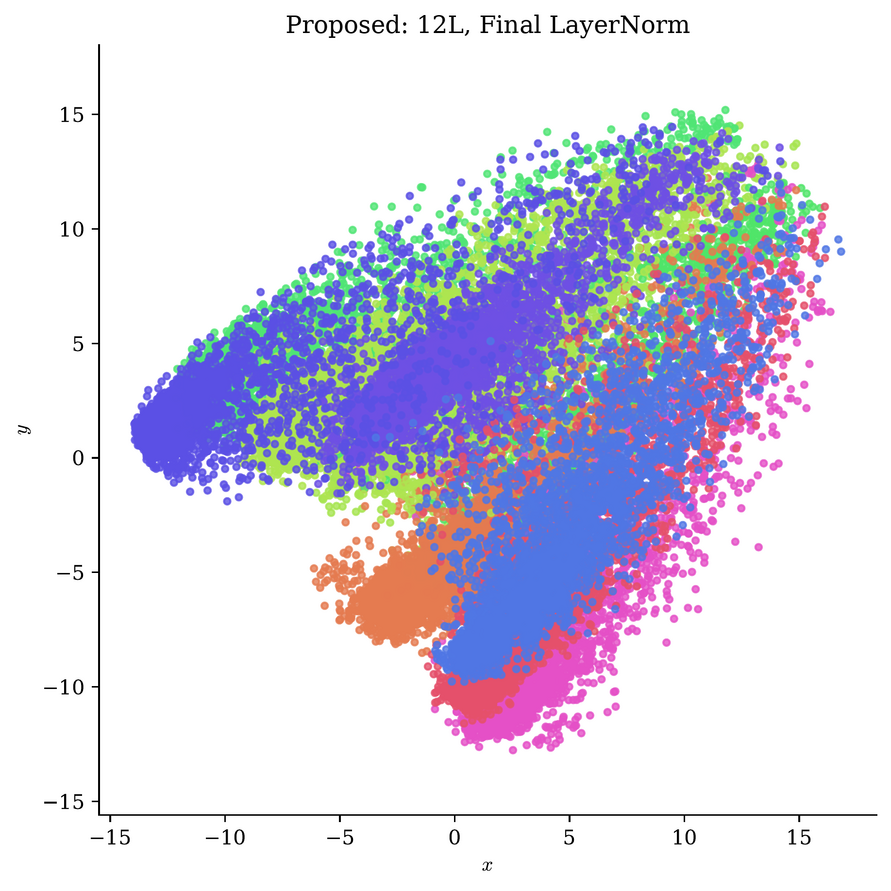}
    \caption{ViT-B-12L (Proposed)}
\end{figure}

\Needspace{0.85\textheight}
\paragraph{PCA (ImageNet)}\leavevmode
\begin{figure}[H]
  \centering  
\includegraphics[width=0.42\linewidth]{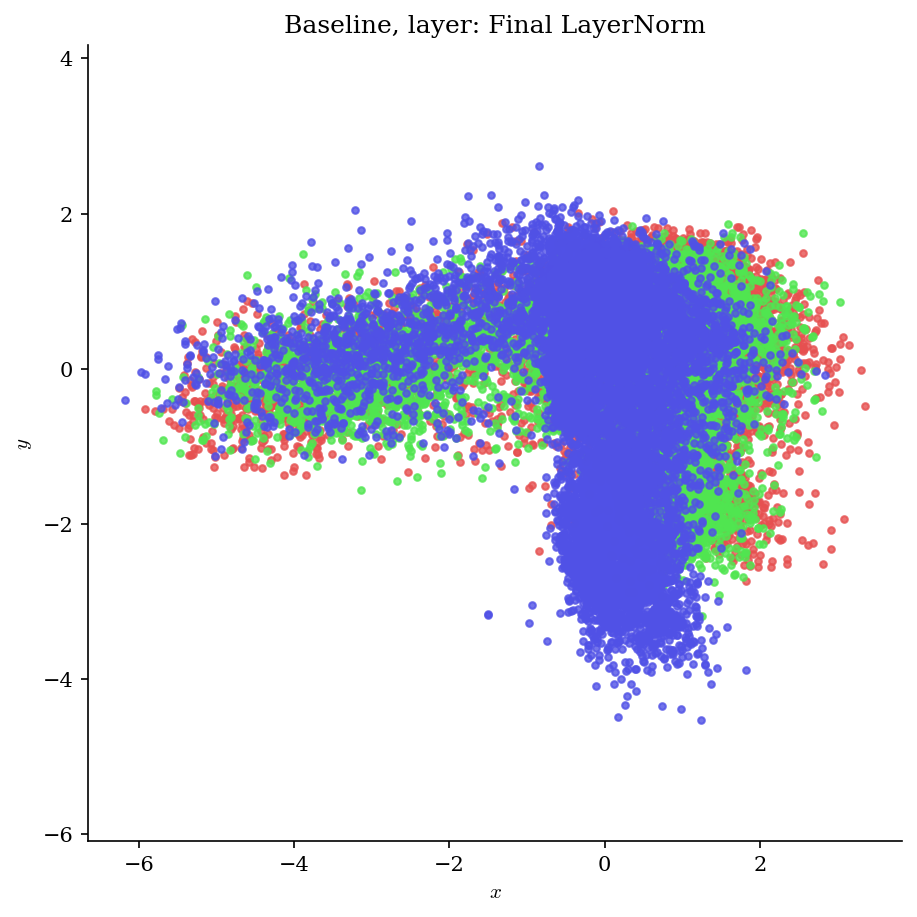}
      \hfill
\includegraphics[width=0.42\linewidth]{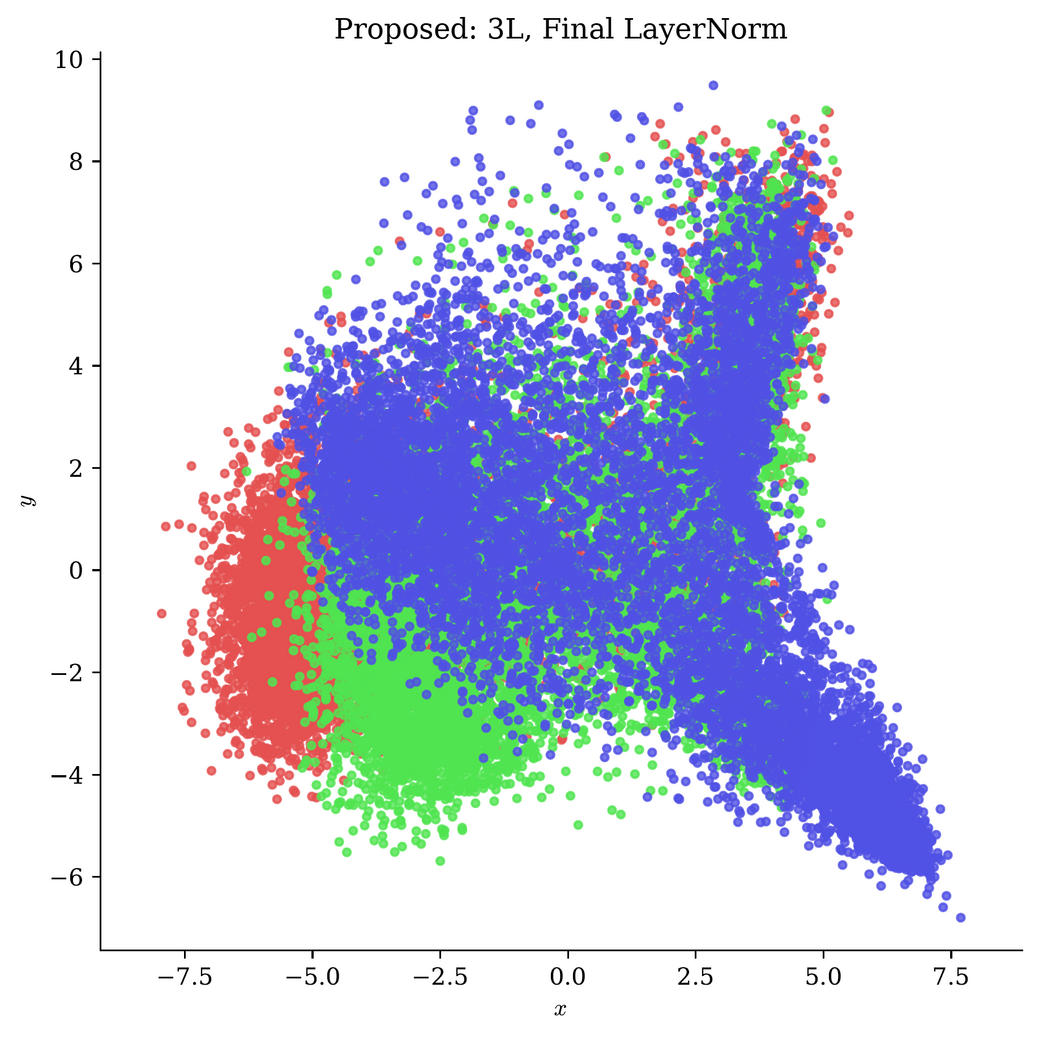}
      \vspace{-0.3em}
\includegraphics[width=0.42\linewidth]{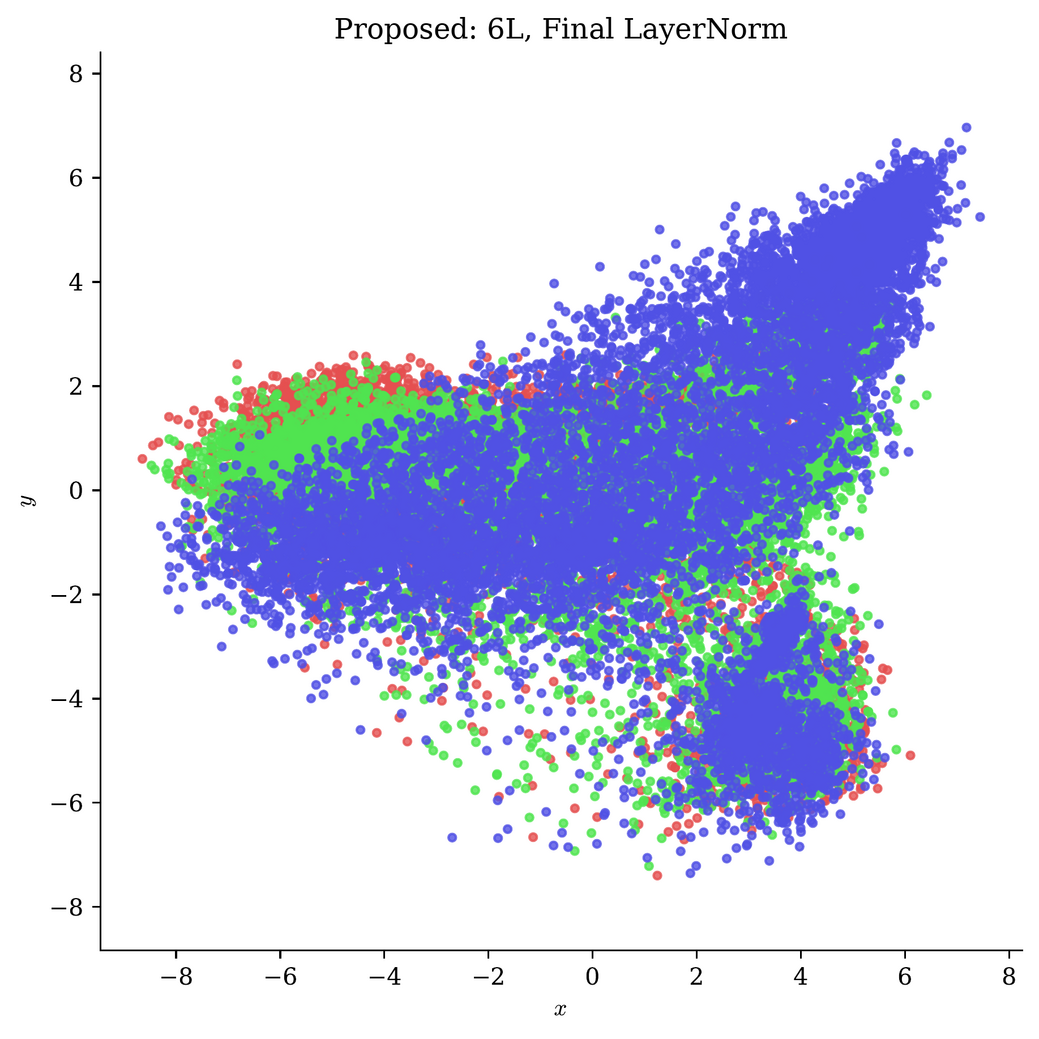}\hfill
    \includegraphics[width=0.42\linewidth]{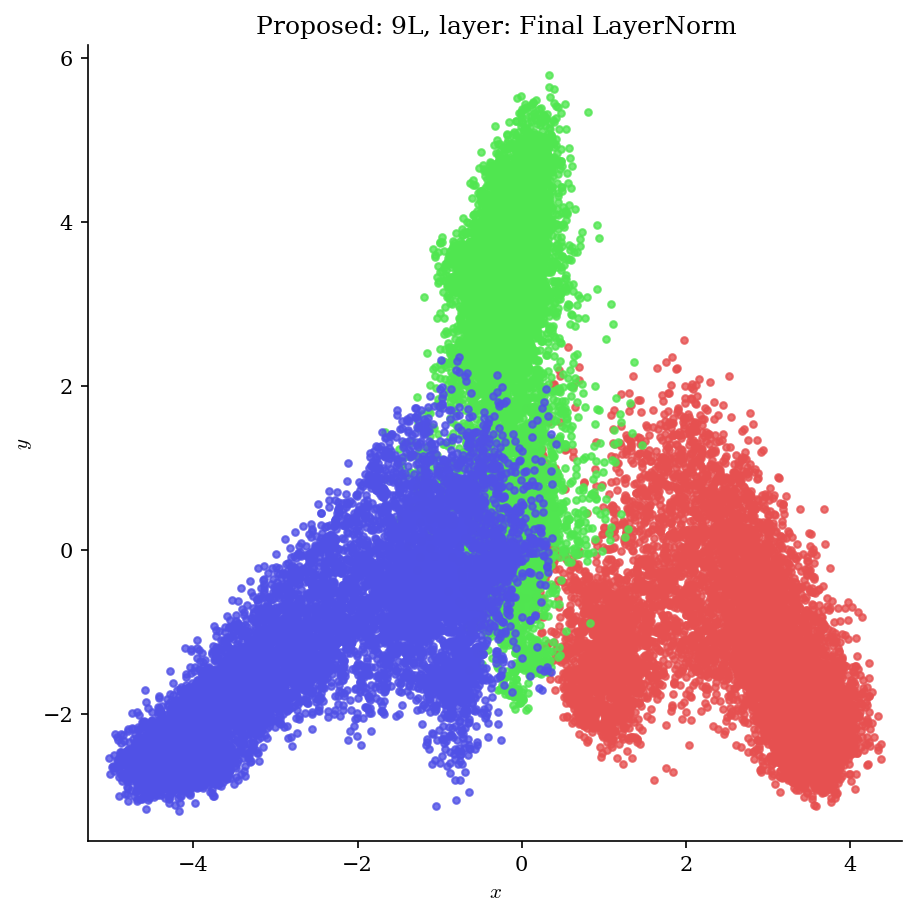}
      \vspace{-0.3em}
    \includegraphics[width=0.42\linewidth]{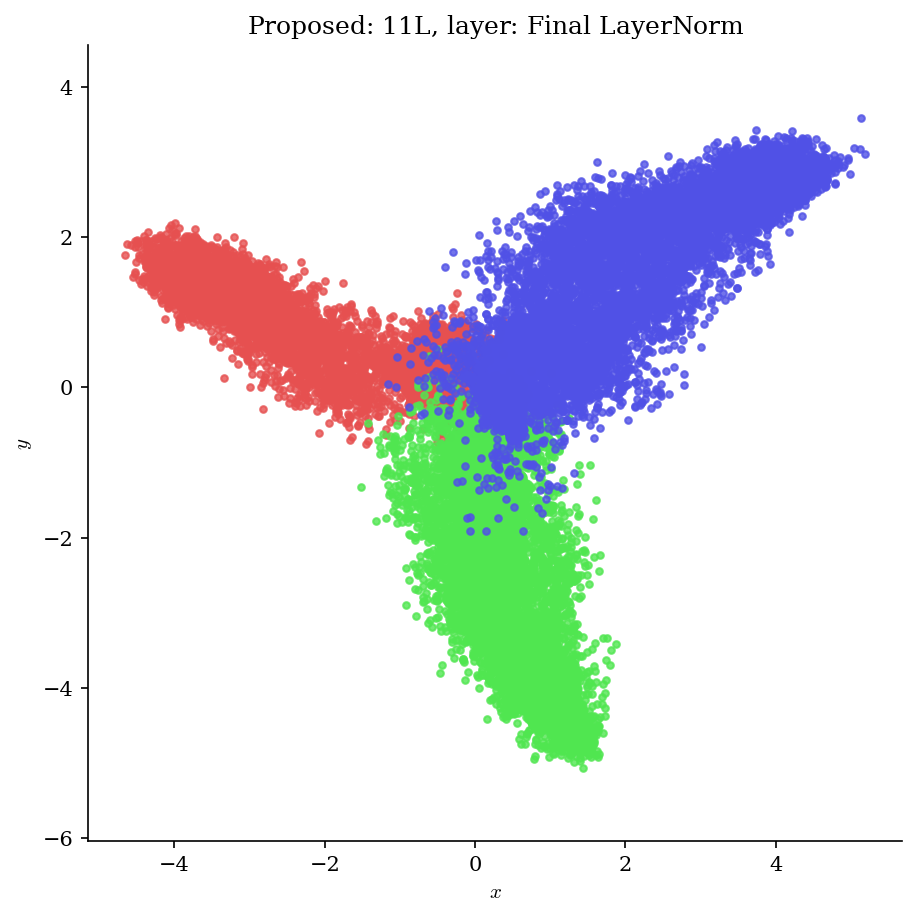}
  \hfill
    \includegraphics[width=0.42\linewidth]{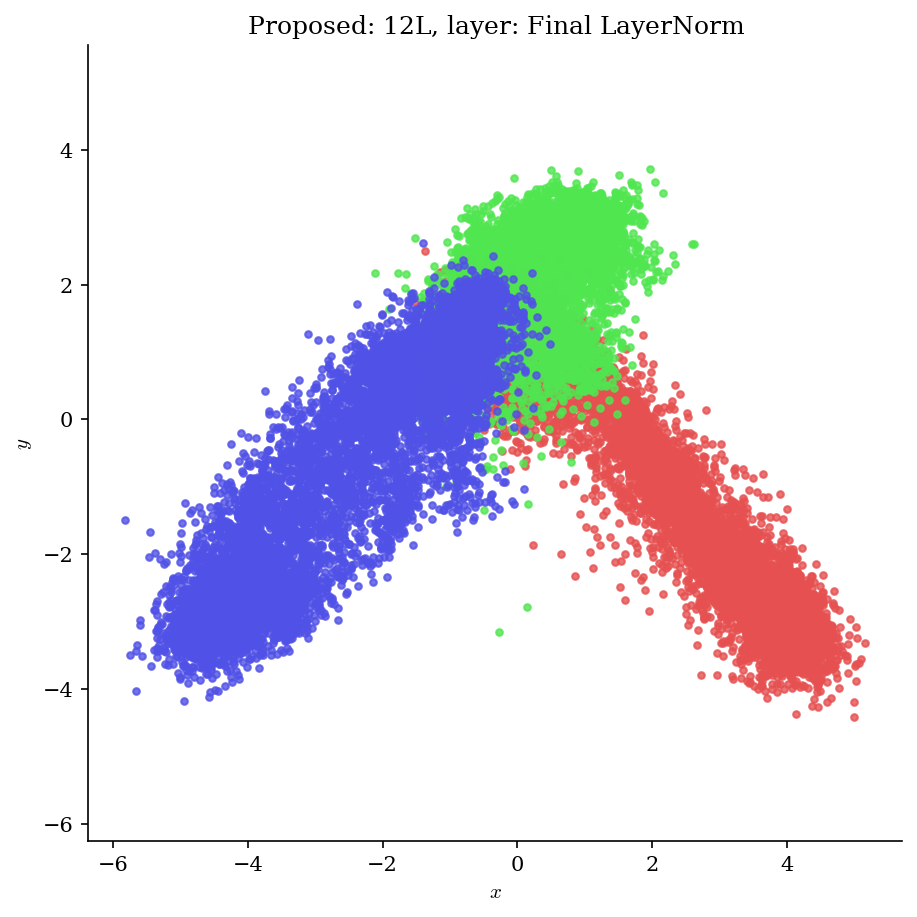}
    \caption{ImageNet-1k PCA projections, as the number of Laplacian heads ranges across $\{0, 3, 6, 9, 11, 12\}$ (from top to bottom, left to right)}
\end{figure}

\Needspace{0.30\textheight}
\paragraph{ANOVA (CIFAR10)}\leavevmode
\begin{figure}[H]
\centering
\includegraphics[width=\textwidth, trim=0 0 0 1cm, clip]{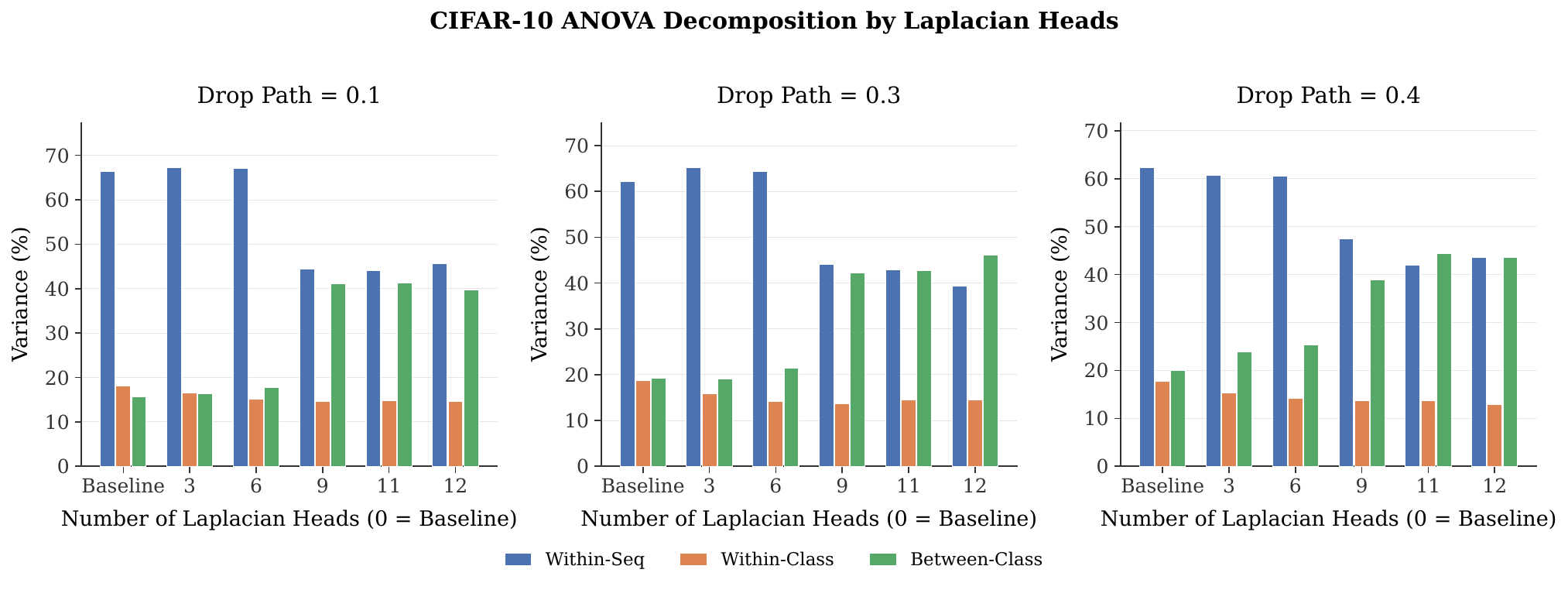}
\caption{
ANOVA decomposition on CIFAR10.}
\vspace{-0.5em}
\end{figure}

\Needspace{0.30\textheight}
\paragraph{ANOVA (CIFAR100)}\leavevmode
\begin{figure}[H]
\centering
\includegraphics[width=\textwidth, trim=0 0 0 1cm, clip]{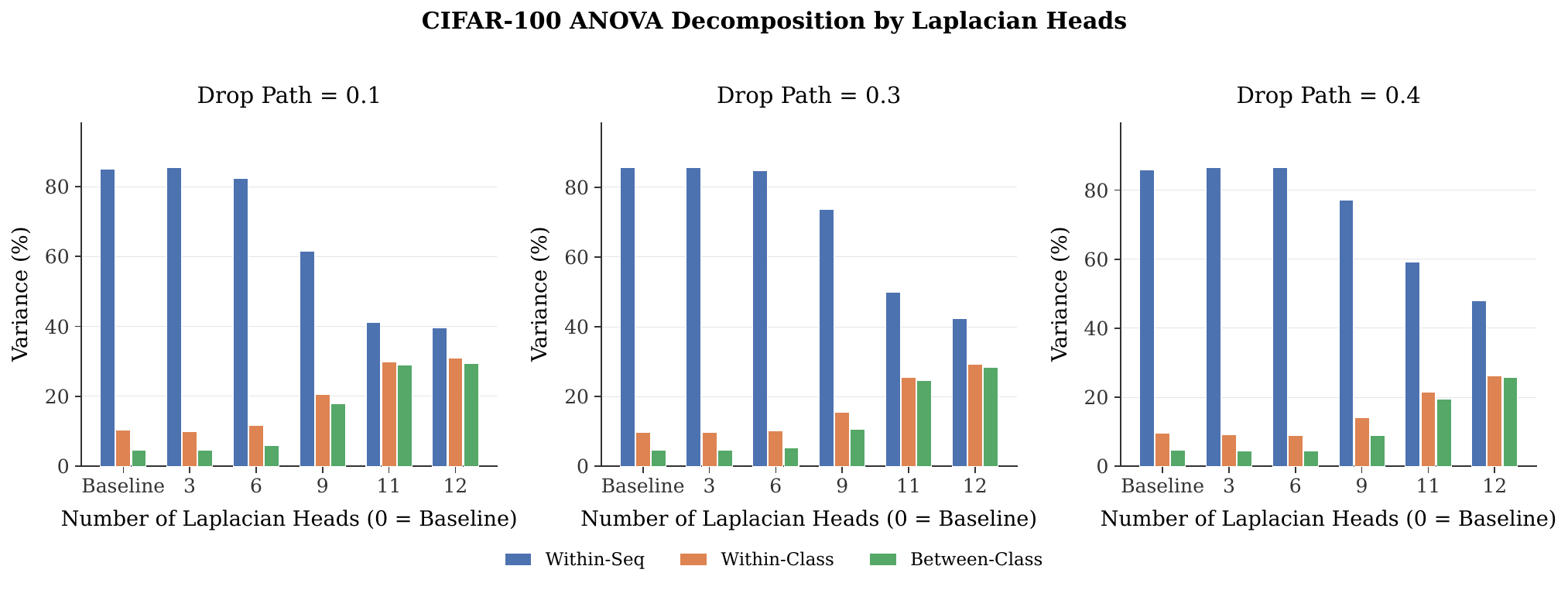}
\caption{
ANOVA decomposition on CIFAR100.}
\vspace{-0.5em}
\end{figure}

\Needspace{0.30\textheight}
\paragraph{ANOVA (ImageNet)}\leavevmode
\begin{figure}[H]
\centering
\includegraphics[width=\textwidth]{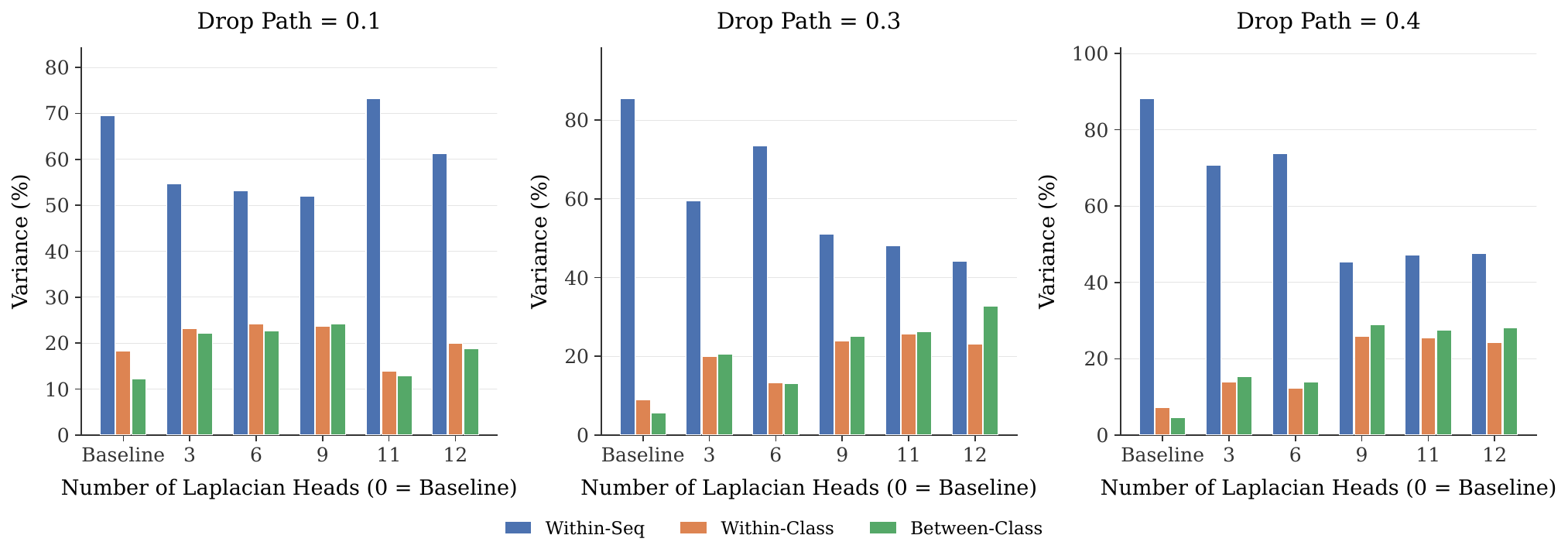}
\caption{
ANOVA decomposition on ImageNet.}
\vspace{-0.5em}
\end{figure}

\Needspace{0.28\textheight}
\paragraph{CosSim (CIFAR10)}\leavevmode
\begin{figure}[H]
    \centering
    \includegraphics[width=0.85\linewidth, trim=0 30 0 20, clip]{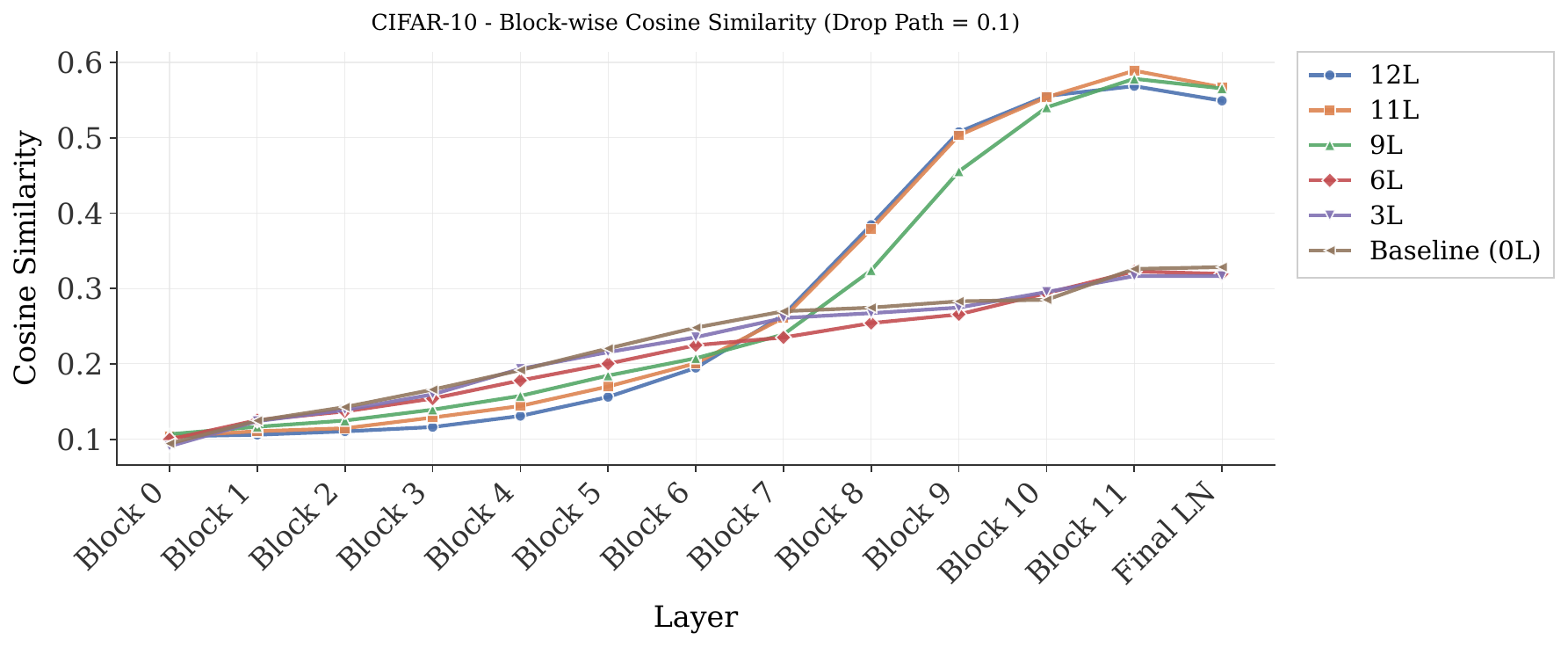}
    \caption{CosSim across depth on CIFAR10.}
    \label{fig:cos_sim_cifar10}
\end{figure}

\Needspace{0.28\textheight}
\paragraph{CosSim (CIFAR100)}\leavevmode
\begin{figure}[H]
    \centering
    \includegraphics[width=0.85\linewidth, trim=0 30 0 20, clip]{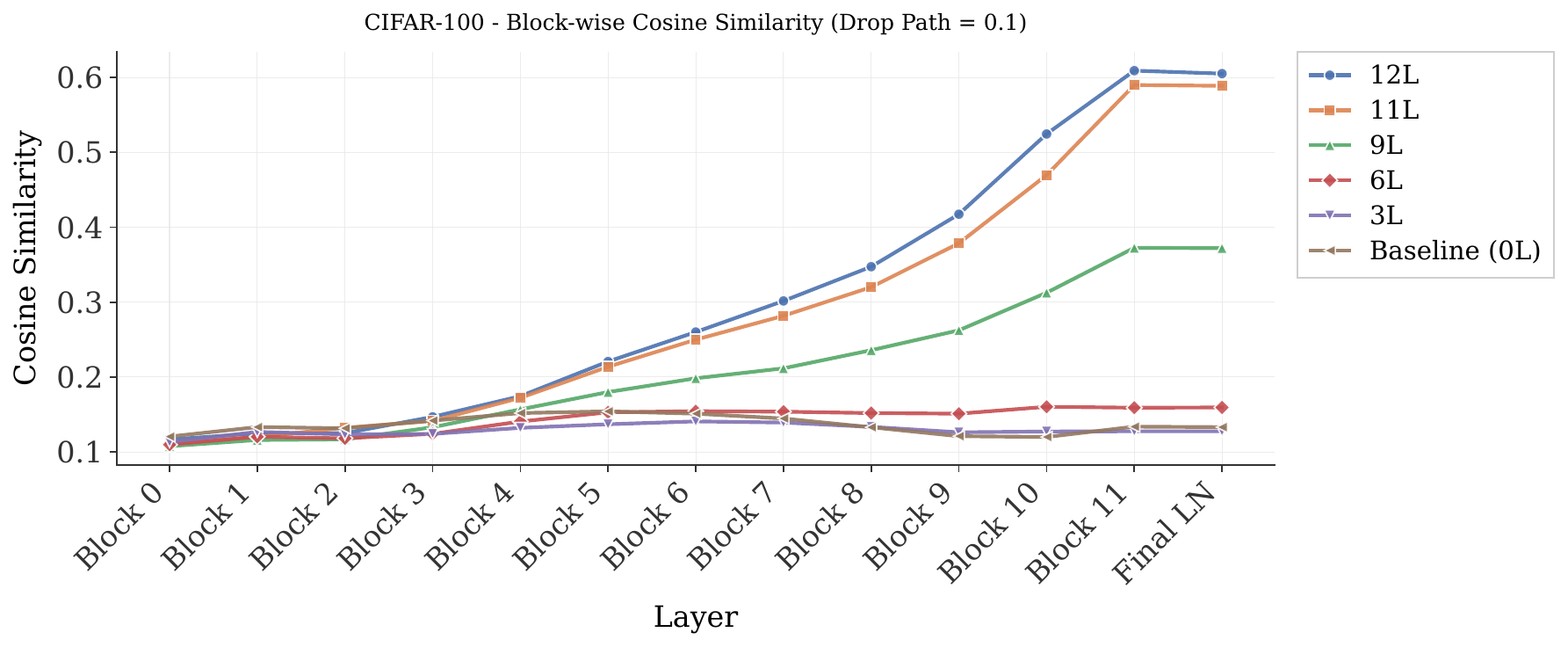}
    \caption{CosSim across depth on CIFAR100.}
    \label{fig:cos_sim_cifar100}
\end{figure}

\clearpage
\subsection{Language Modeling}\label{app:add_results_language}
The d12, d20, and d30 models have 135M, 477M, and 1.45B parameters, respectively.
All language-model representation measurements use
\[
X = \operatorname{RMSNorm}\!\left(
Y + \sum_{i\in\A} P^{(i)} Z W_{V_i} W_{o_i}
  + \sum_{i\in\L} (I-P^{(i)}) Z W_{V_i} W_{o_i}
\right),
\]
where $Y$ is the input to the second-to-last block (zero-based index $N-2$), $Z=\operatorname{RMSNorm}(Y)$.

\begin{figure}[H]
    \centering
    \begin{minipage}[t]{0.48\textwidth}
        \centering
        \includegraphics[width=\linewidth]{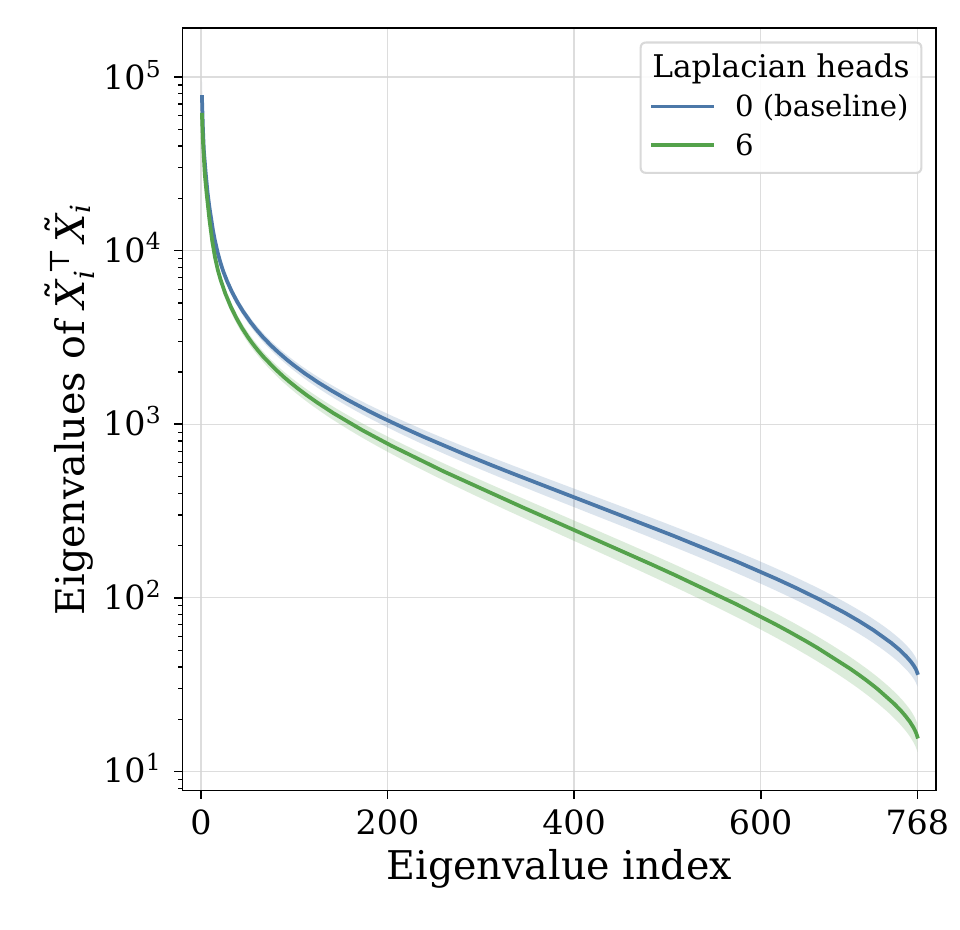}
    \end{minipage}\hfill
    \begin{minipage}[t]{0.48\textwidth}
        \centering
        \includegraphics[width=\linewidth]{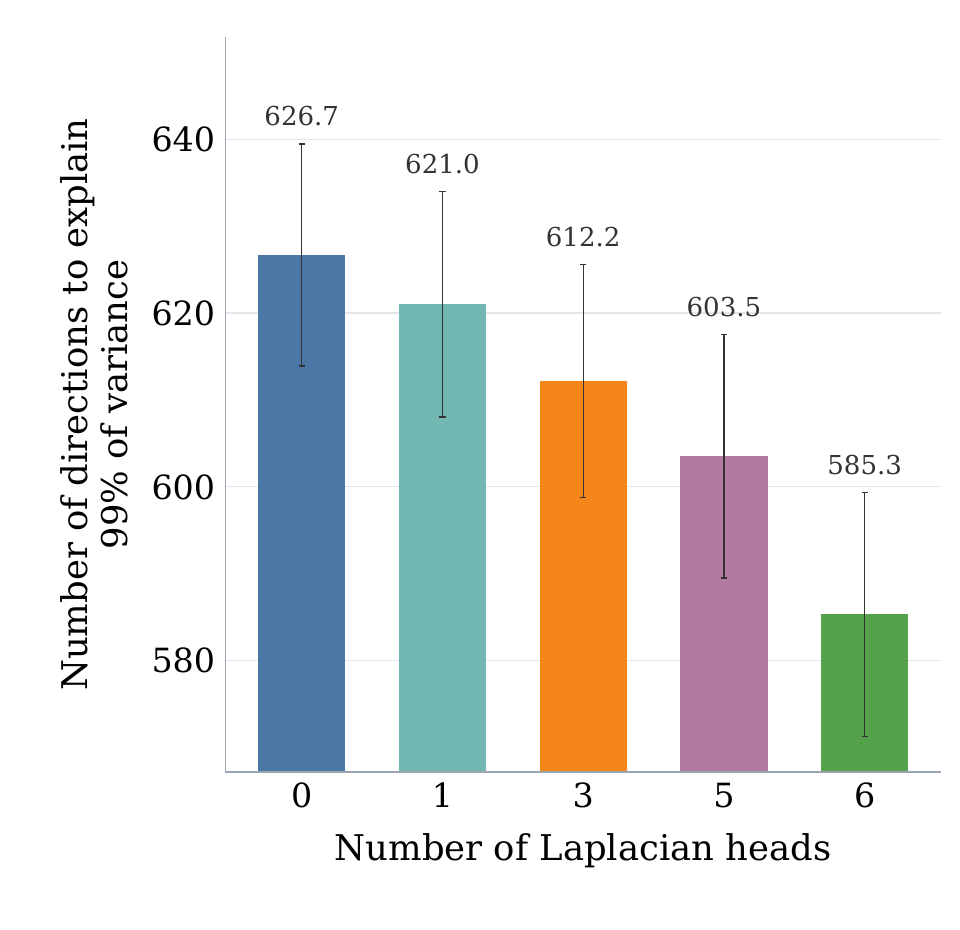}
    \end{minipage}
    \caption{Spectra and number of principal components explaining 99\% of variance for d12 models.}
    \label{fig:language_spectrum_d12}
\end{figure}

\noindent
\begin{minipage}[t]{0.48\textwidth}
    \vspace{0pt}
    \captionsetup{type=figure}
    \centering
    \includegraphics[width=\linewidth]{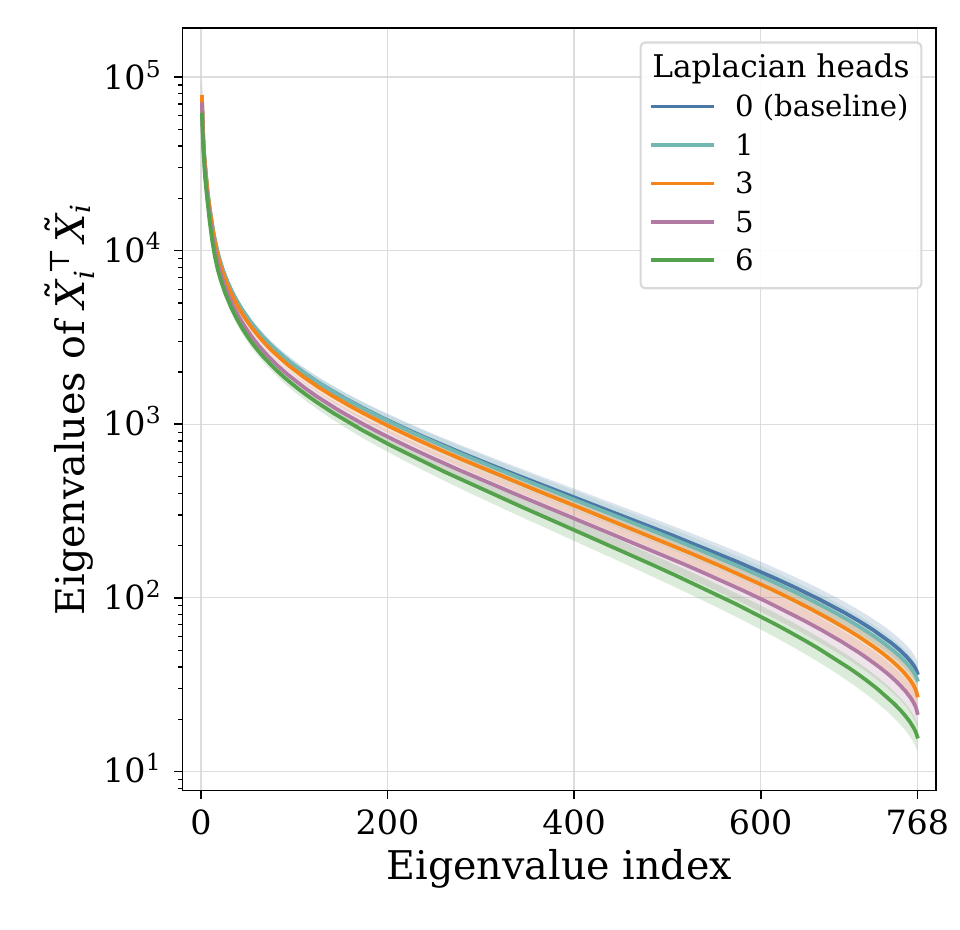}
    \caption{Spectra for d12 models with varying numbers of Laplacian heads.}
    \label{fig:language_spectrum_all_d12}
\end{minipage}\hfill
\begin{minipage}[t]{0.48\textwidth}
    \vspace{0pt}
    \captionsetup{type=figure}
    \centering
    \includegraphics[width=\linewidth]{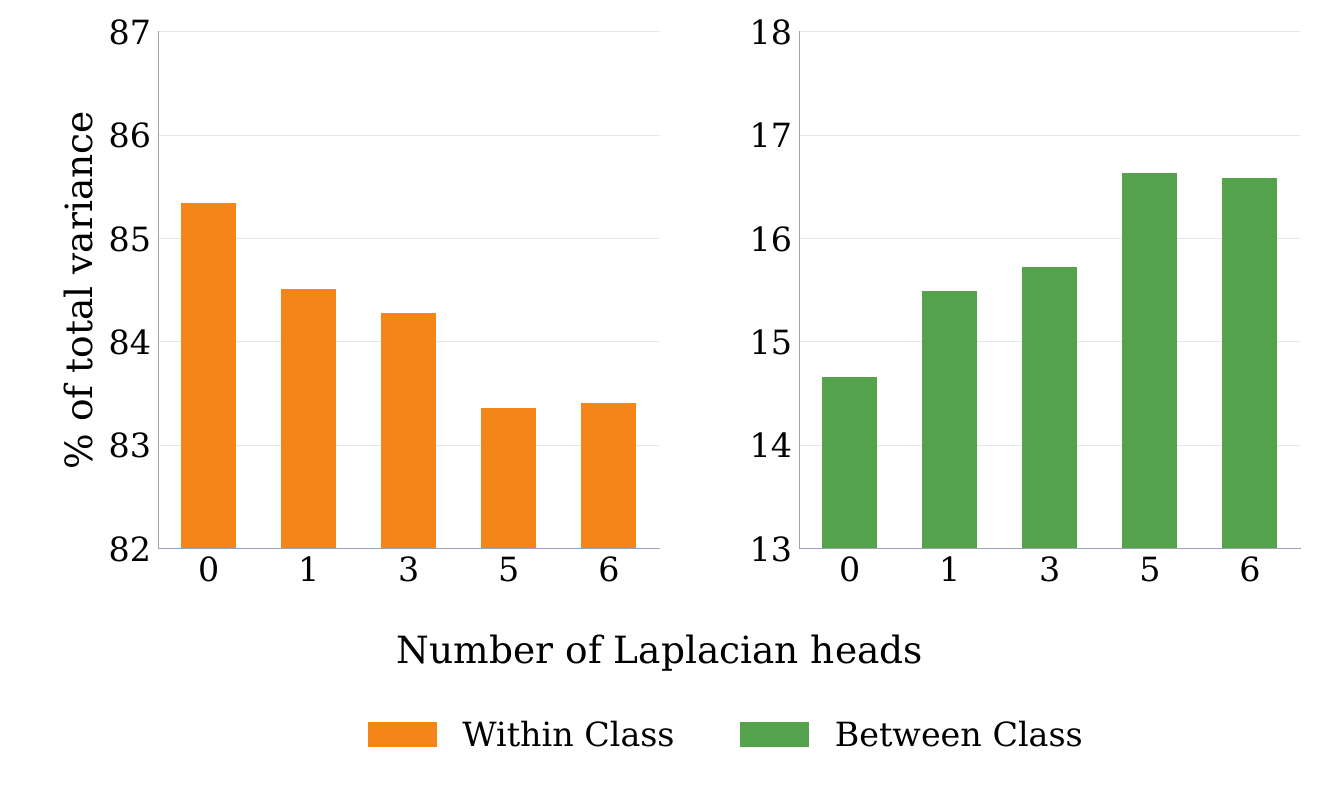}
    \caption{ANOVA for d12 models.}
    \label{fig:language_anova_d12}
\end{minipage}\par

\clearpage
\begin{figure}[H]
    \centering
    \begin{minipage}[t]{0.48\textwidth}
        \centering
        \includegraphics[width=\linewidth]{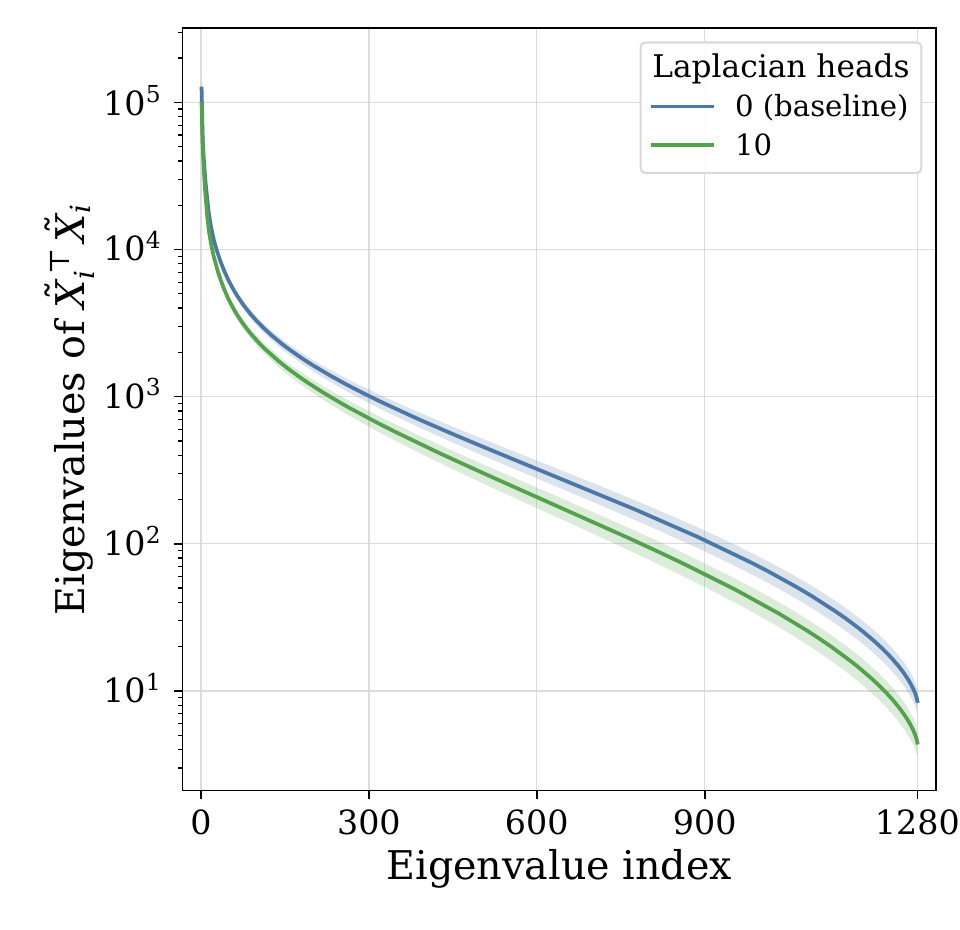}
    \end{minipage}\hfill
    \begin{minipage}[t]{0.48\textwidth}
        \centering
        \includegraphics[width=\linewidth]{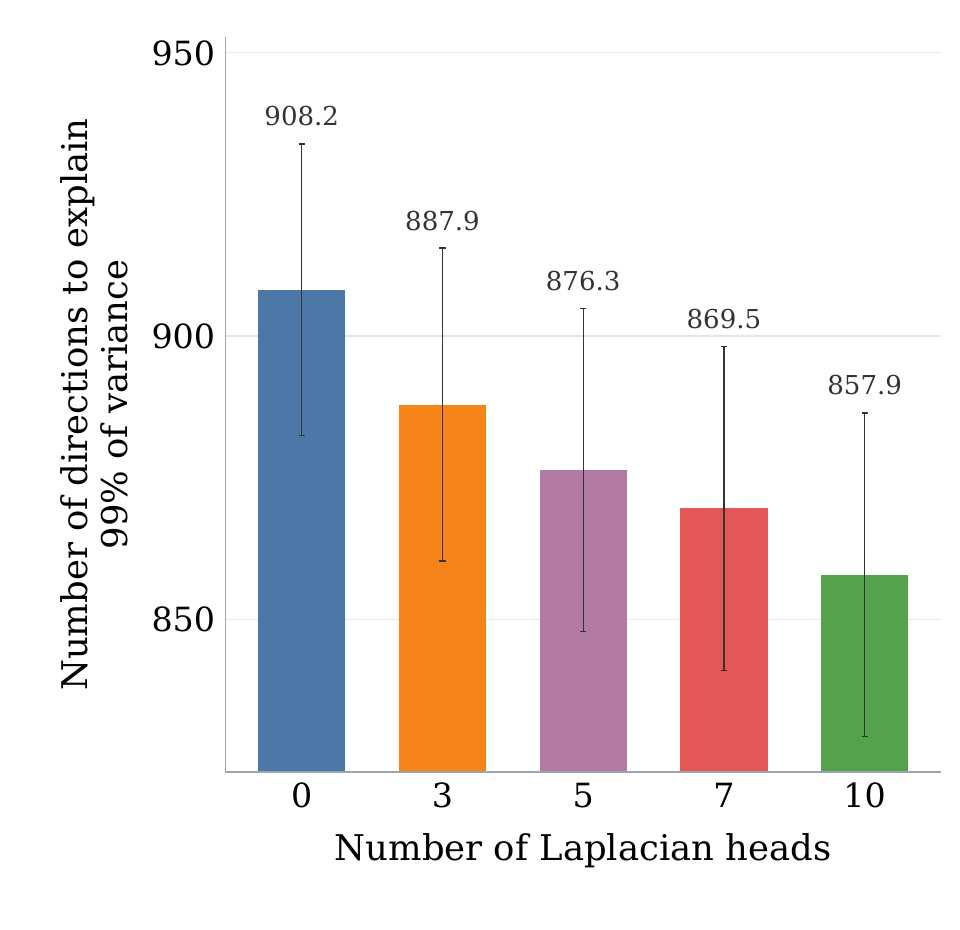}
    \end{minipage}
    \caption{Spectra and number of principal components explaining 99\% of variance for d20 models.}
    \label{fig:language_spectrum_d20}
\end{figure}

\noindent
\begin{minipage}[t]{0.48\textwidth}
    \vspace{0pt}
    \captionsetup{type=figure}
    \centering
    \includegraphics[width=\linewidth]{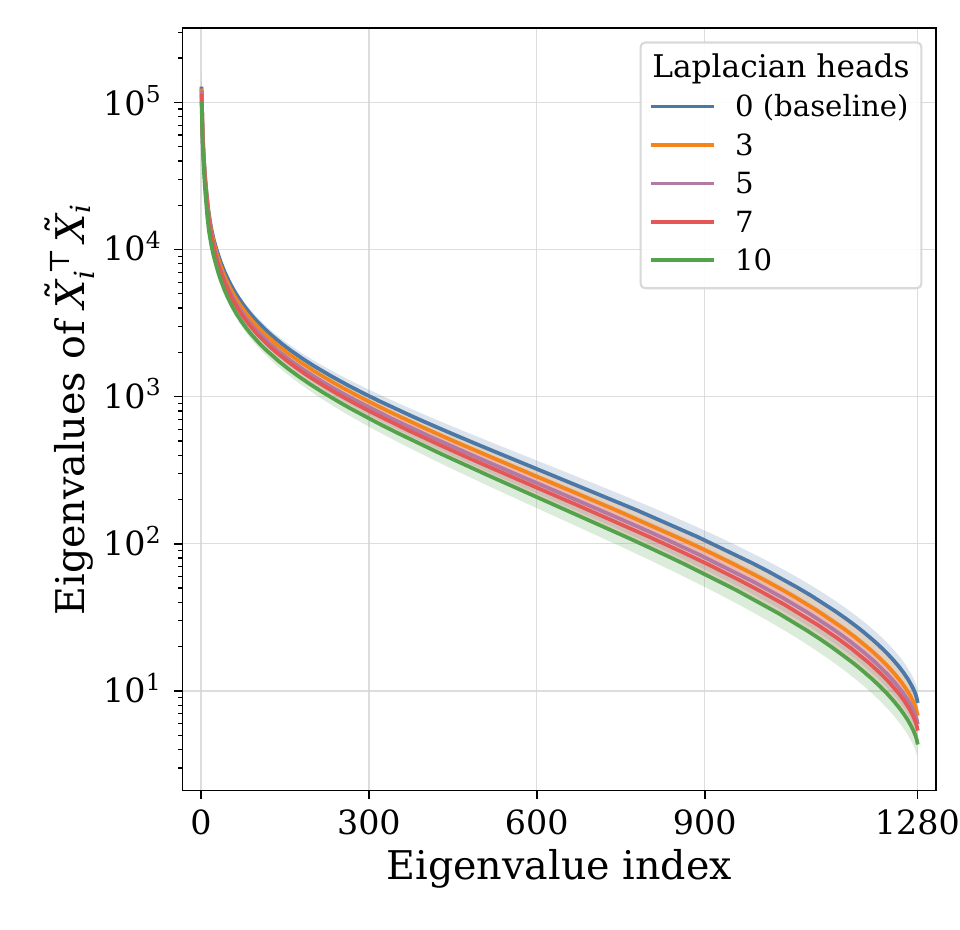}
    \caption{Spectra for d20 models with varying numbers of Laplacian heads.}
    \label{fig:language_spectrum_all_d20}
\end{minipage}\hfill
\begin{minipage}[t]{0.48\textwidth}
    \vspace{0pt}
    \captionsetup{type=figure}
    \centering
    \includegraphics[width=\linewidth]{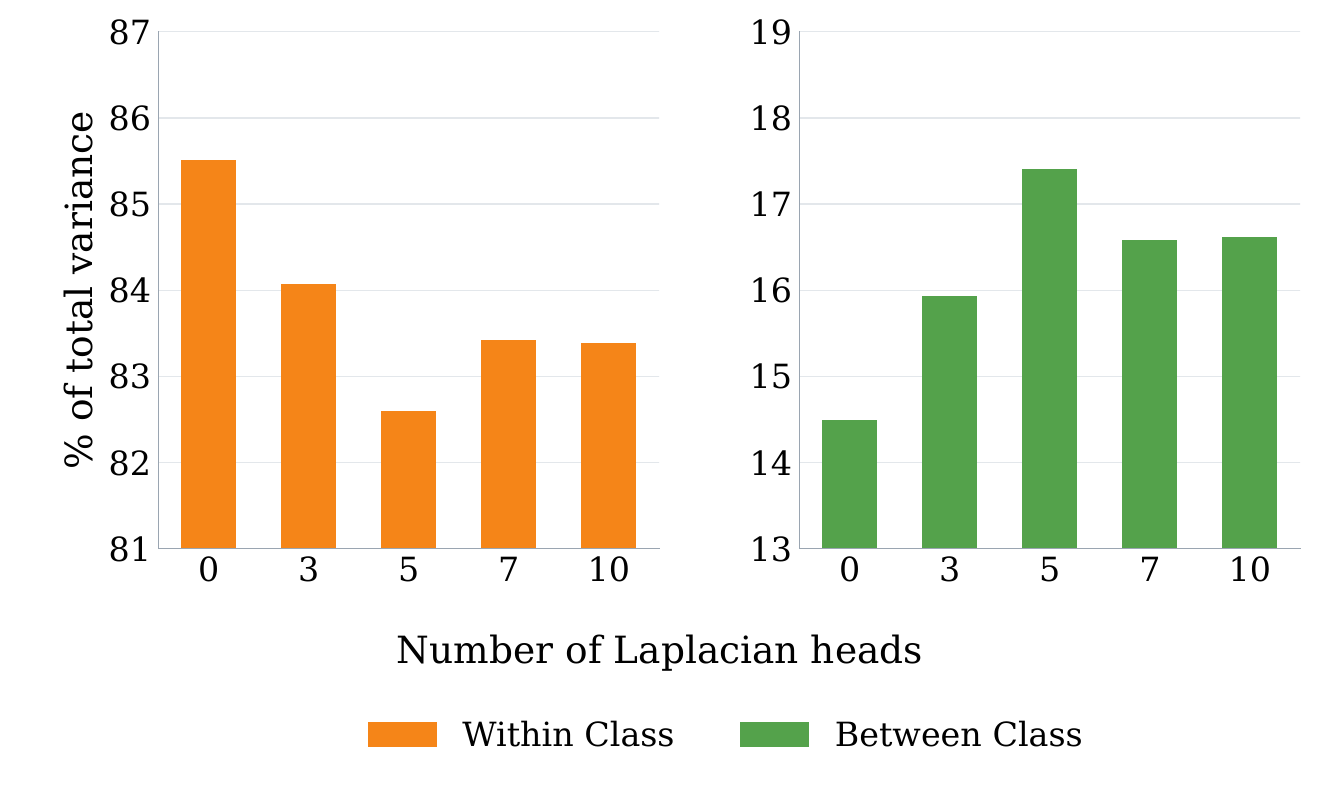}
    \caption{ANOVA for d20 models.}
    \label{fig:language_anova_d20}
\end{minipage}\par

\clearpage
\begin{figure}[H]
    \centering
    \includegraphics[width=0.48\textwidth]{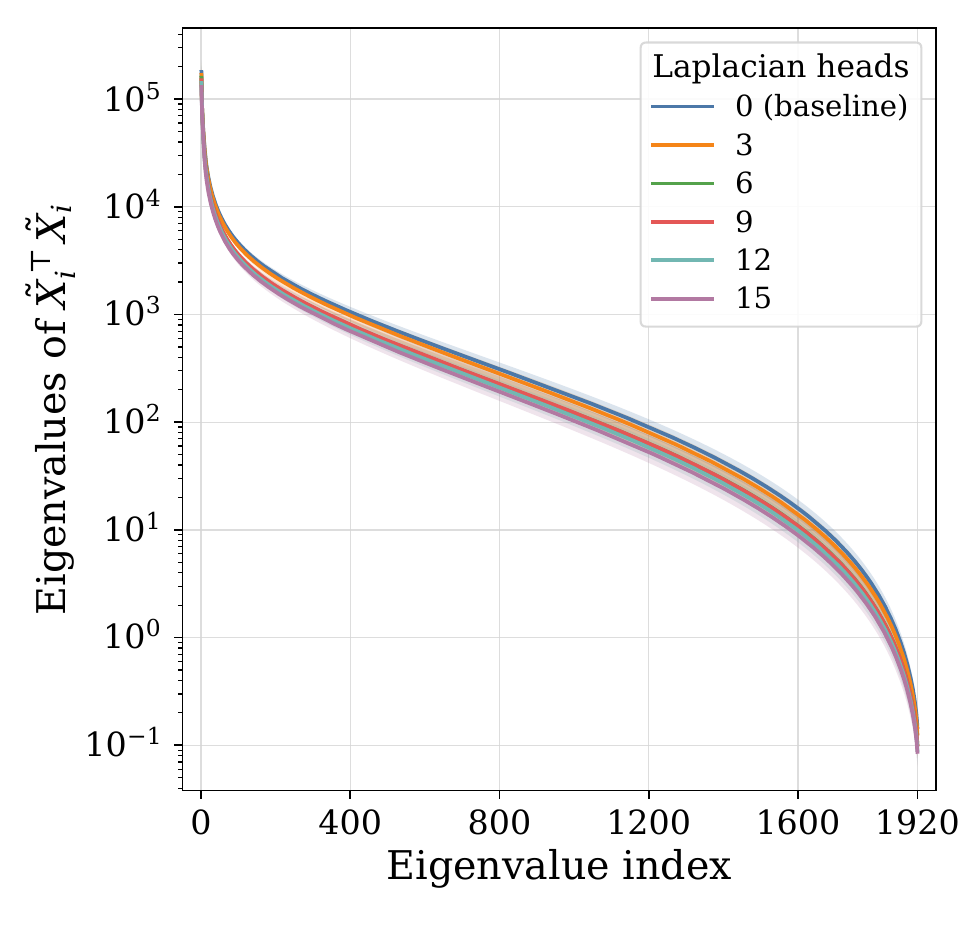}
    \caption{Spectra for d30 models with varying numbers of Laplacian heads.}
    \label{fig:spectrum_language_all_models}
\end{figure}

\begin{figure}[H]
    \centering
    \begin{subfigure}[t]{0.32\textwidth}
        \centering
        \includegraphics[width=\linewidth]{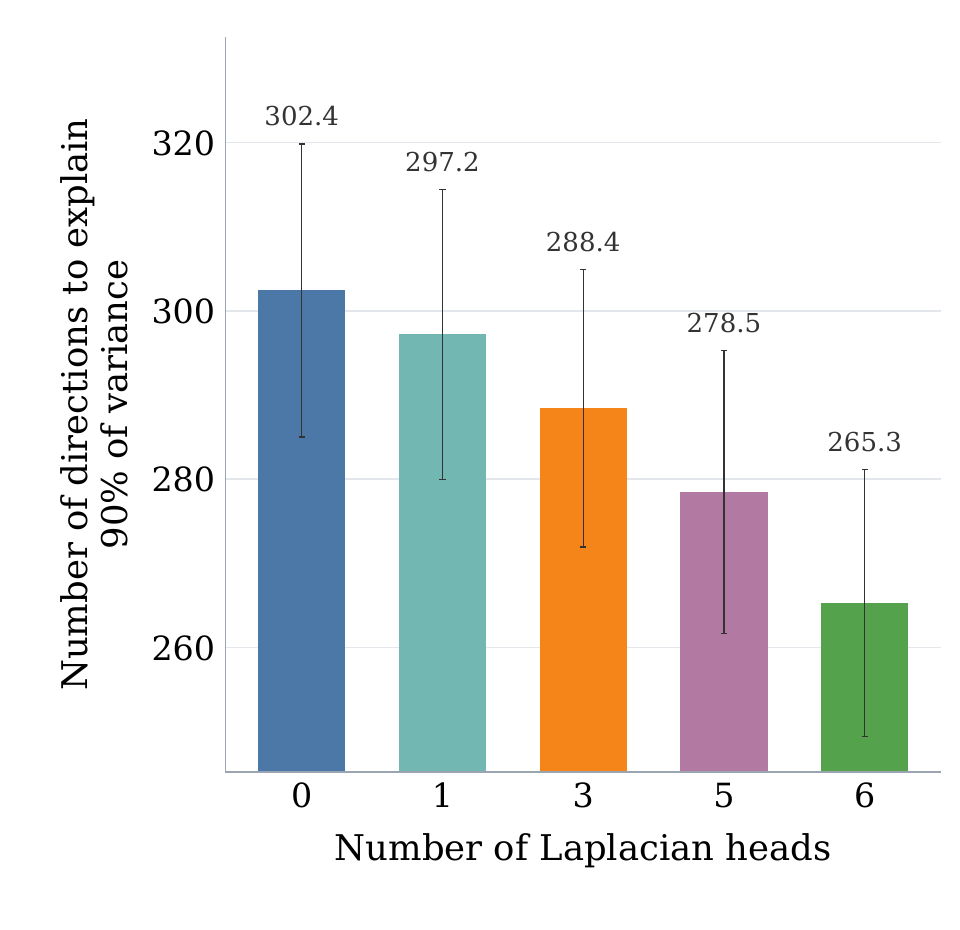}
        \caption{d12 models.}
    \end{subfigure}\hfill
    \begin{subfigure}[t]{0.32\textwidth}
        \centering
        \includegraphics[width=\linewidth]{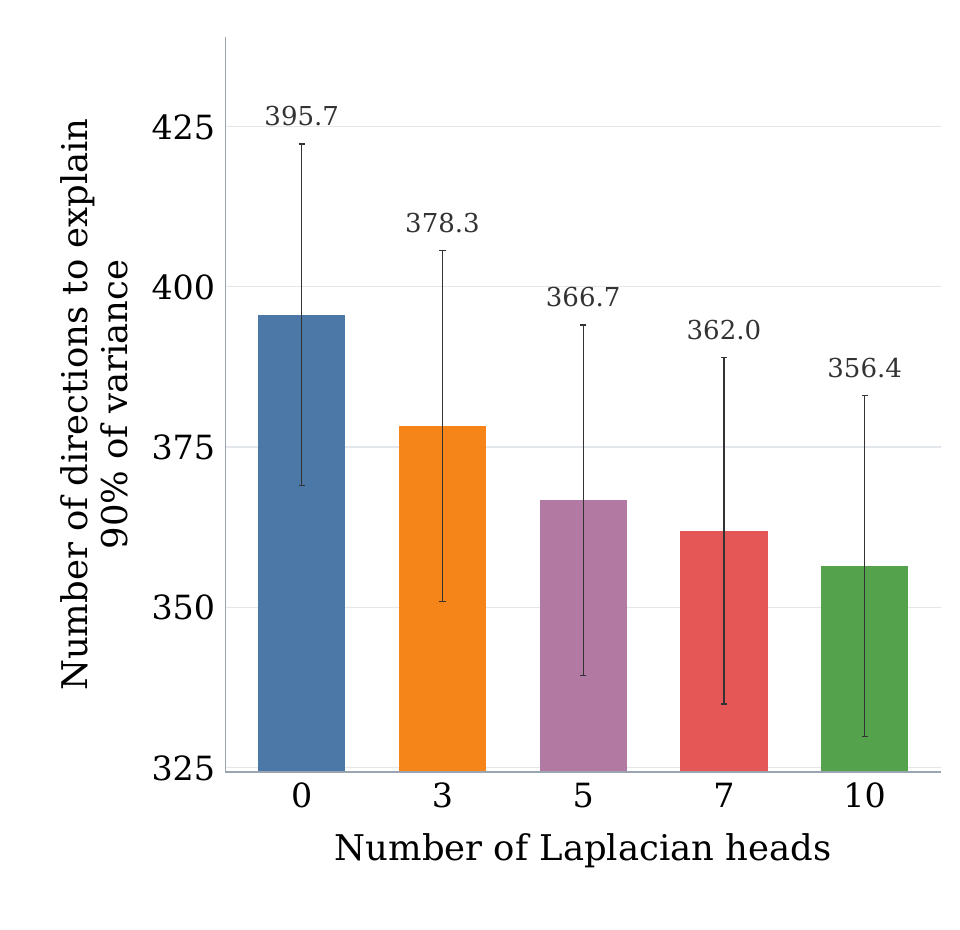}
        \caption{d20 models.}
    \end{subfigure}\hfill
    \begin{subfigure}[t]{0.32\textwidth}
        \centering
        \includegraphics[width=\linewidth]{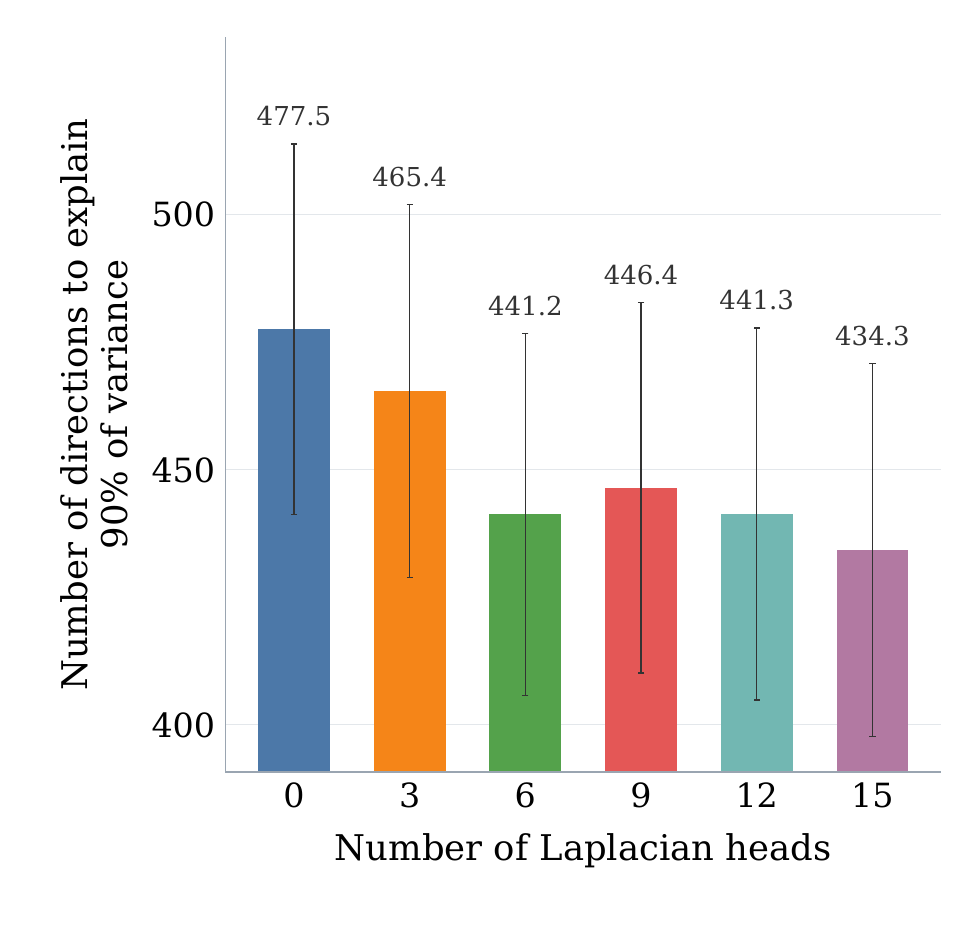}
        \caption{d30 models.}
    \end{subfigure}
    \caption{Number of principal components explaining 90\% of variance for d12, d20, and d30 models.}
    \label{fig:language_pc_90_all_depths}
\end{figure}

\begin{figure}[H]
    \centering
    \begin{subfigure}[t]{0.32\textwidth}
        \centering
        \includegraphics[width=\linewidth]{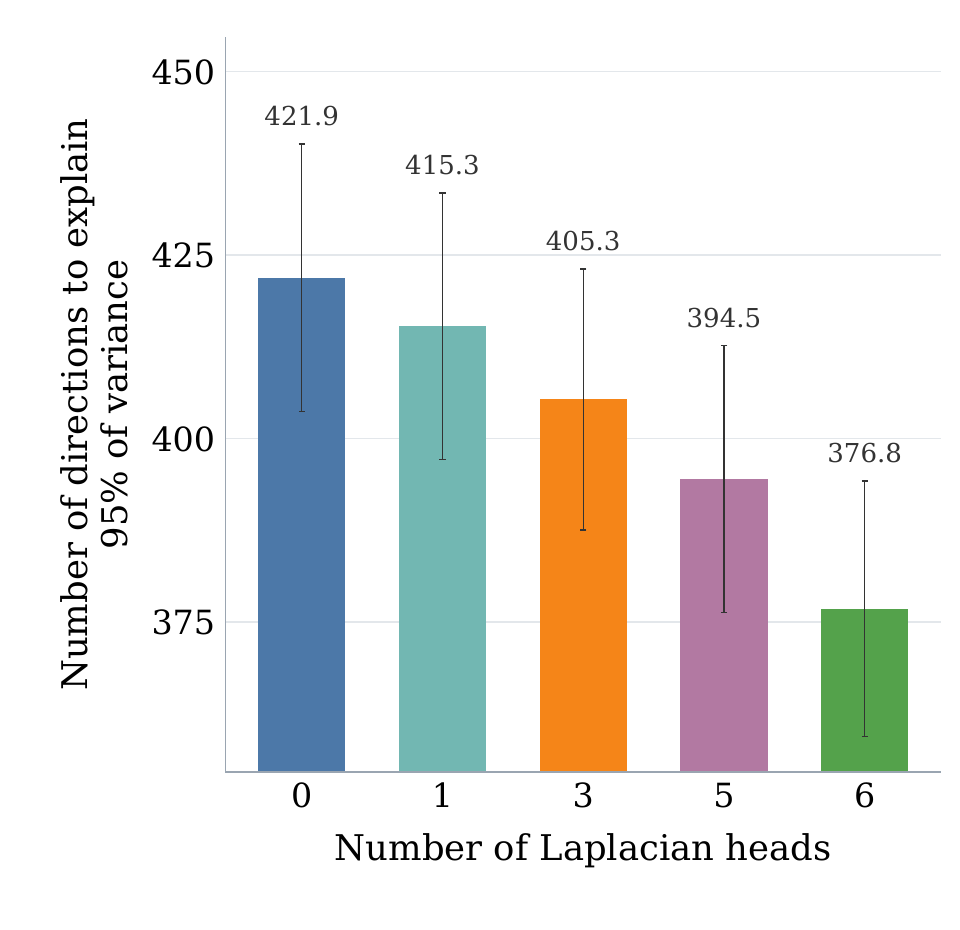}
        \caption{d12 models.}
    \end{subfigure}\hfill
    \begin{subfigure}[t]{0.32\textwidth}
        \centering
        \includegraphics[width=\linewidth]{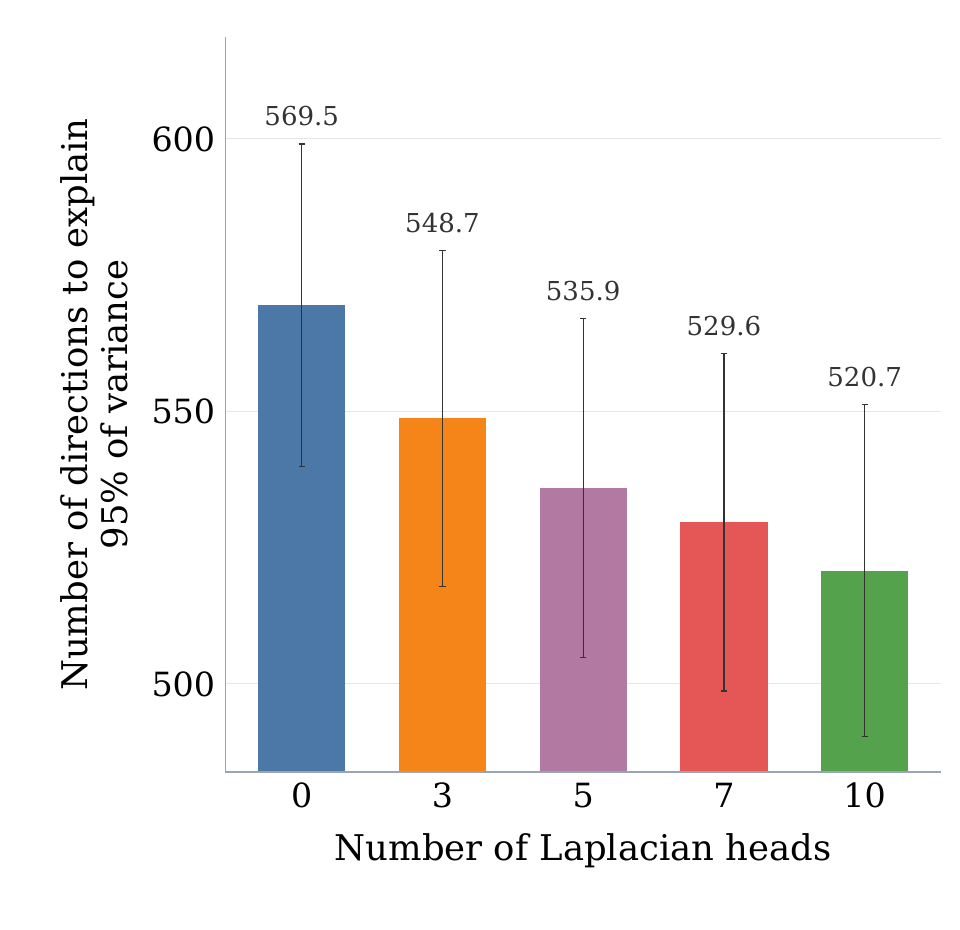}
        \caption{d20 models.}
    \end{subfigure}\hfill
    \begin{subfigure}[t]{0.32\textwidth}
        \centering
        \includegraphics[width=\linewidth]{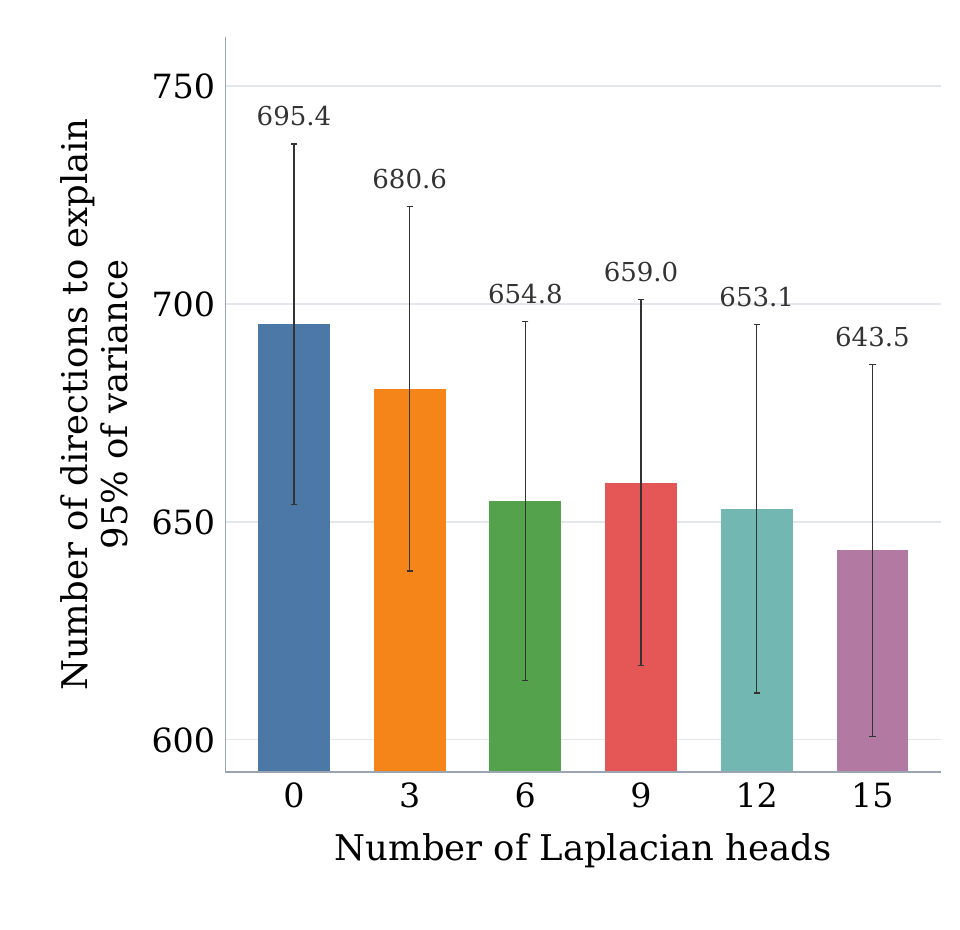}
        \caption{d30 models.}
    \end{subfigure}
    \caption{Number of principal components explaining 95\% of variance for d12, d20, and d30 models.}
    \label{fig:language_pc_95_all_depths}
\end{figure}

\begin{figure}[H]
    \centering
    \begin{subfigure}[t]{0.32\textwidth}
        \centering
        \includegraphics[width=\linewidth]{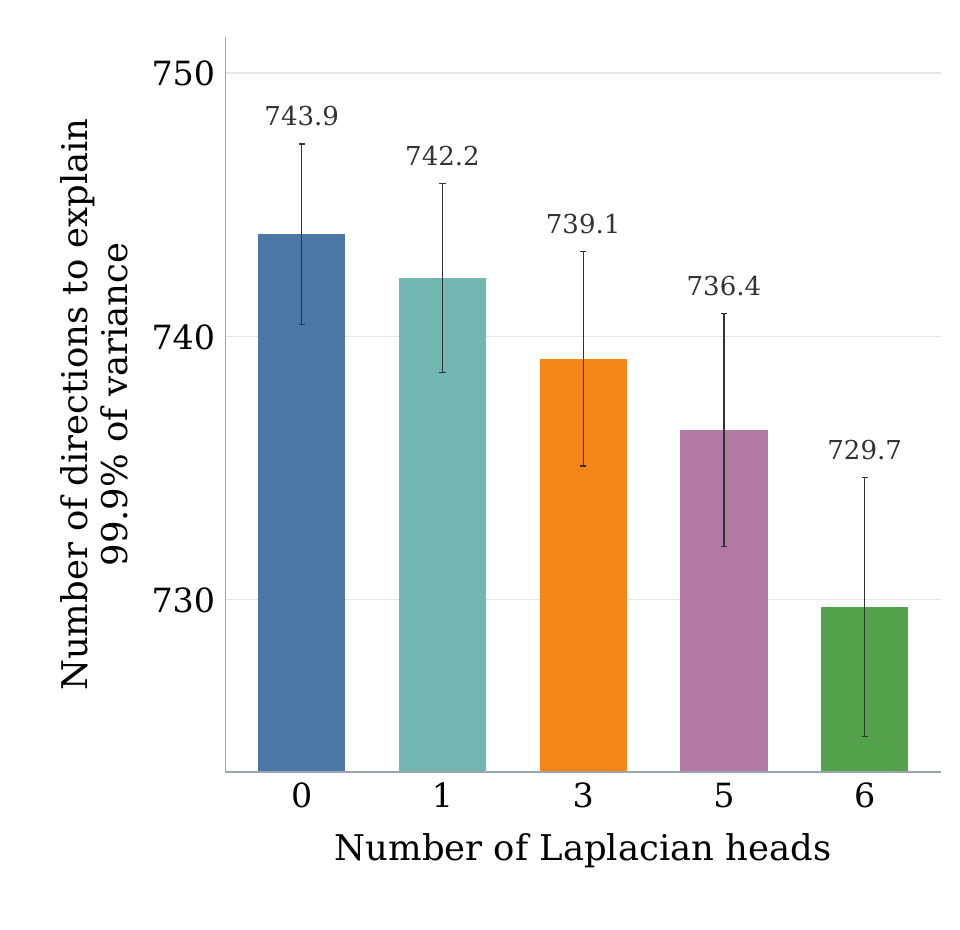}
        \caption{d12 models.}
    \end{subfigure}\hfill
    \begin{subfigure}[t]{0.32\textwidth}
        \centering
        \includegraphics[width=\linewidth]{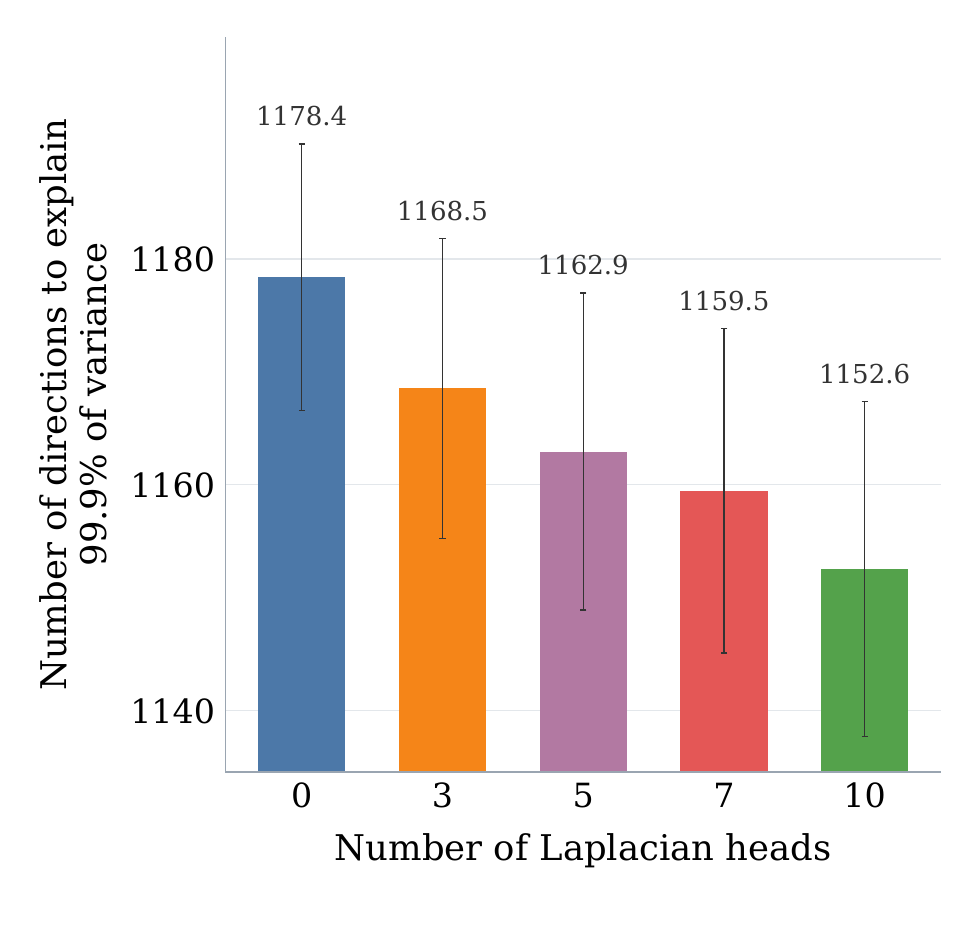}
        \caption{d20 models.}
    \end{subfigure}\hfill
    \begin{subfigure}[t]{0.32\textwidth}
        \centering
        \includegraphics[width=\linewidth]{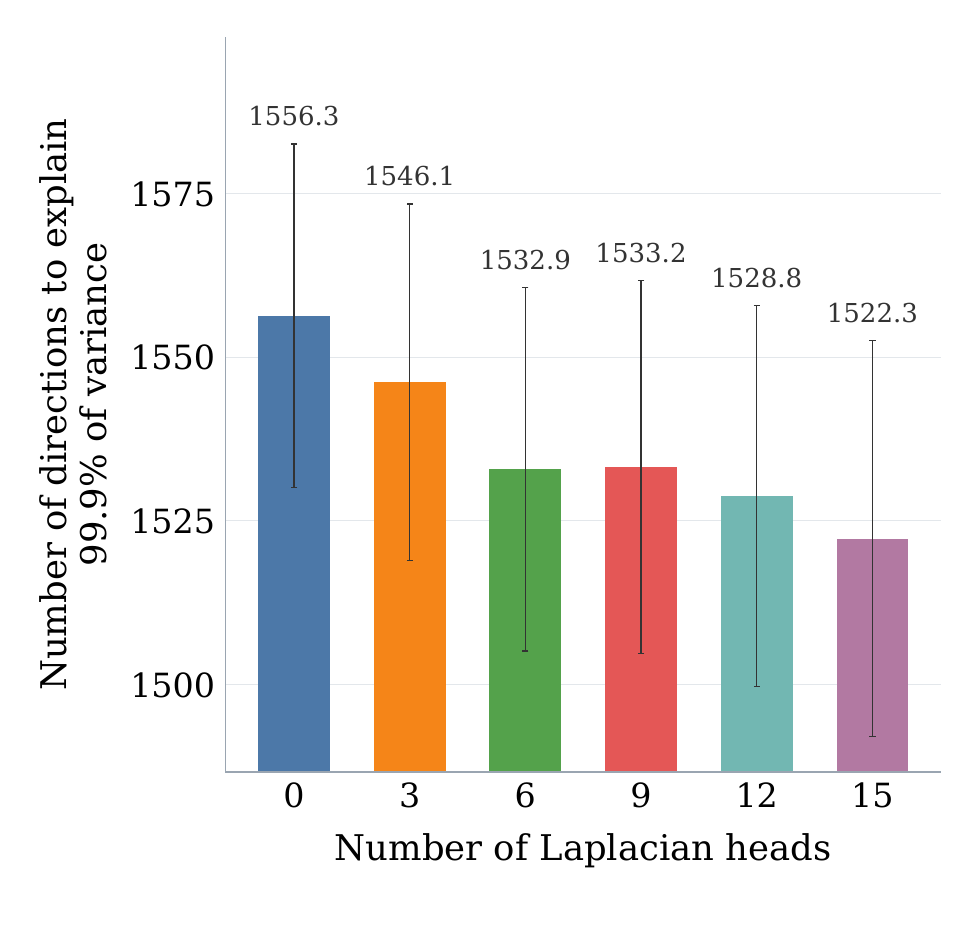}
        \caption{d30 models.}
    \end{subfigure}
    \caption{Number of principal components explaining 99.9\% of variance for d12, d20, and d30 models.}
    \label{fig:language_pc_999_all_depths}
\end{figure}

\clearpage
\paragraph{Uncentered Spectrum Measurements}
The following figures use the same models and sequences without centering: spectra are the eigenvalues of $X_i^\top X_i$, and component counts capture fractions of the total energy $\|X_i\|_F^2$.

\begin{figure}[H]
    \centering
    \begin{subfigure}[t]{0.42\textwidth}
        \centering
        \includegraphics[width=\linewidth]{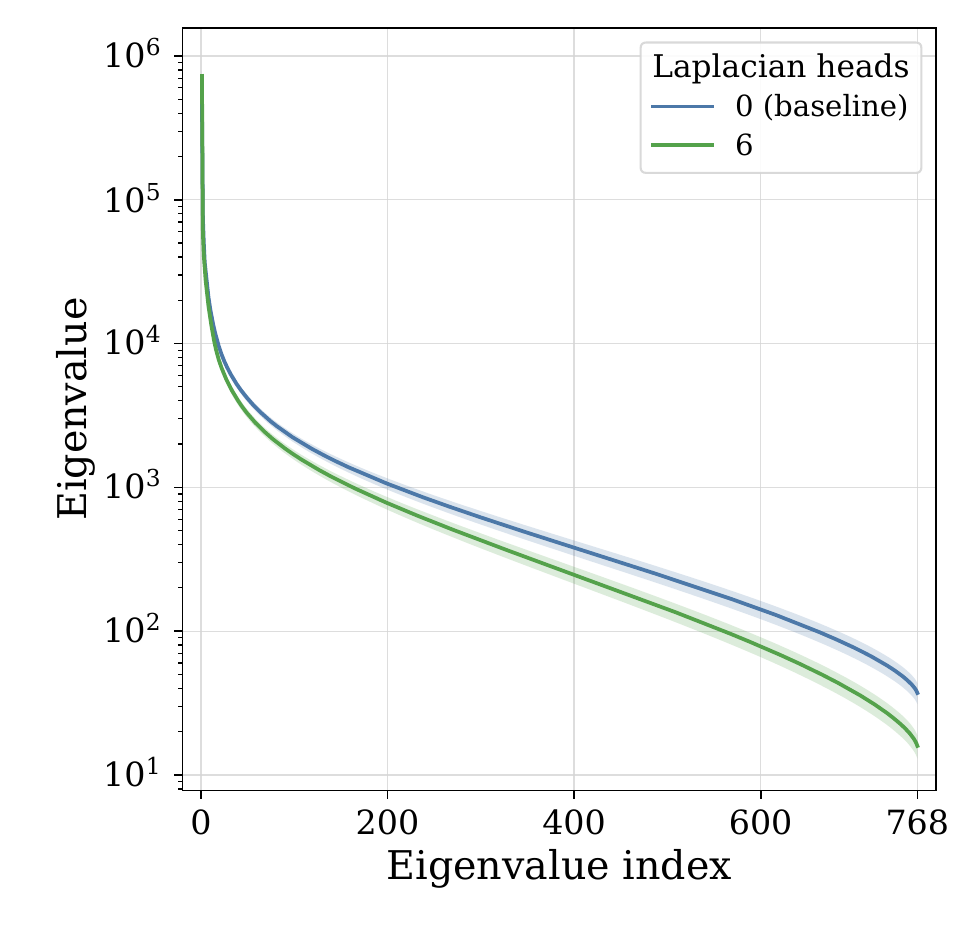}
        \caption{Baseline and 6 Laplacian heads.}
    \end{subfigure}\hfill
    \begin{subfigure}[t]{0.42\textwidth}
        \centering
        \includegraphics[width=\linewidth]{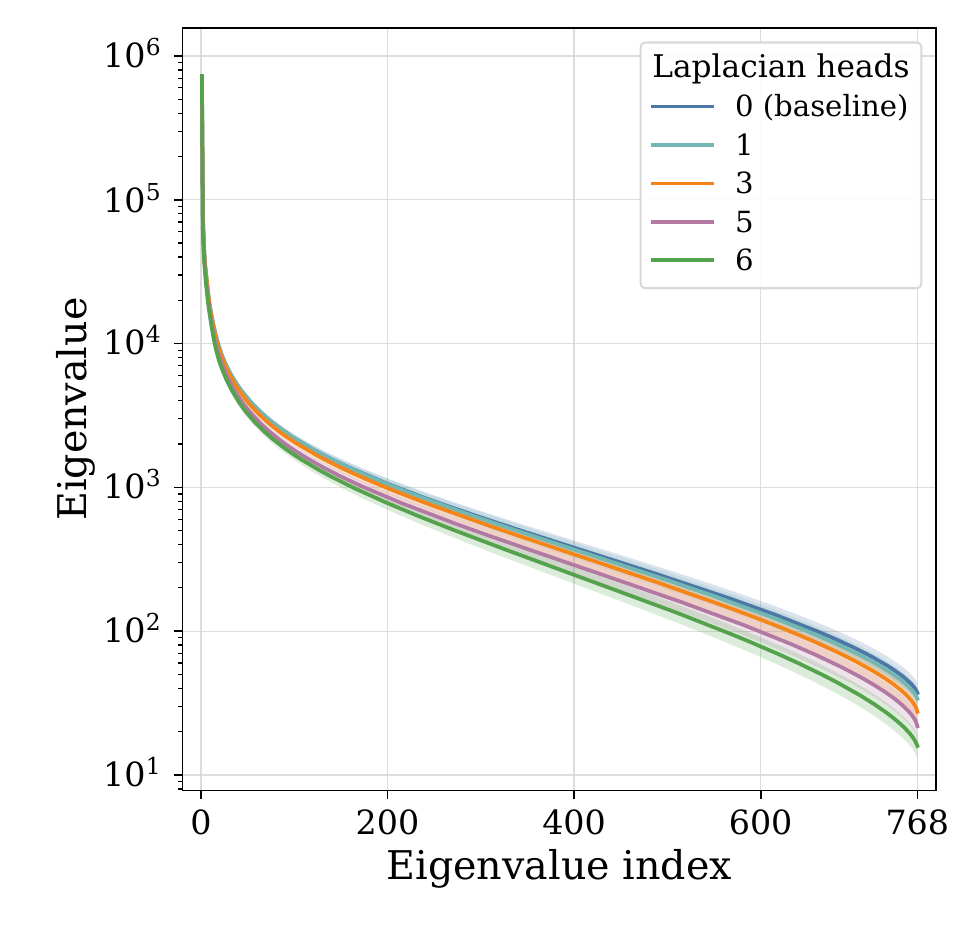}
        \caption{All models.}
    \end{subfigure}
    \par\smallskip
    \begin{subfigure}[t]{0.42\textwidth}
        \centering
        \includegraphics[width=\linewidth]{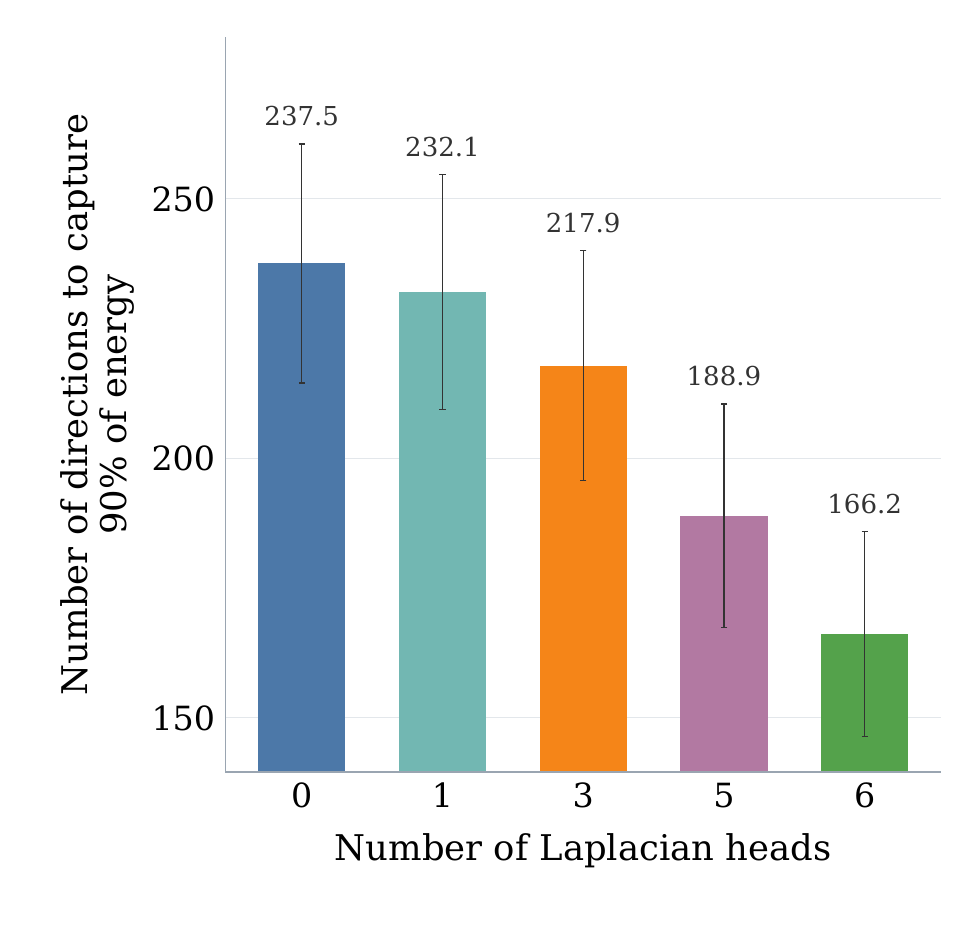}
        \caption{90\% of energy.}
    \end{subfigure}\hfill
    \begin{subfigure}[t]{0.42\textwidth}
        \centering
        \includegraphics[width=\linewidth]{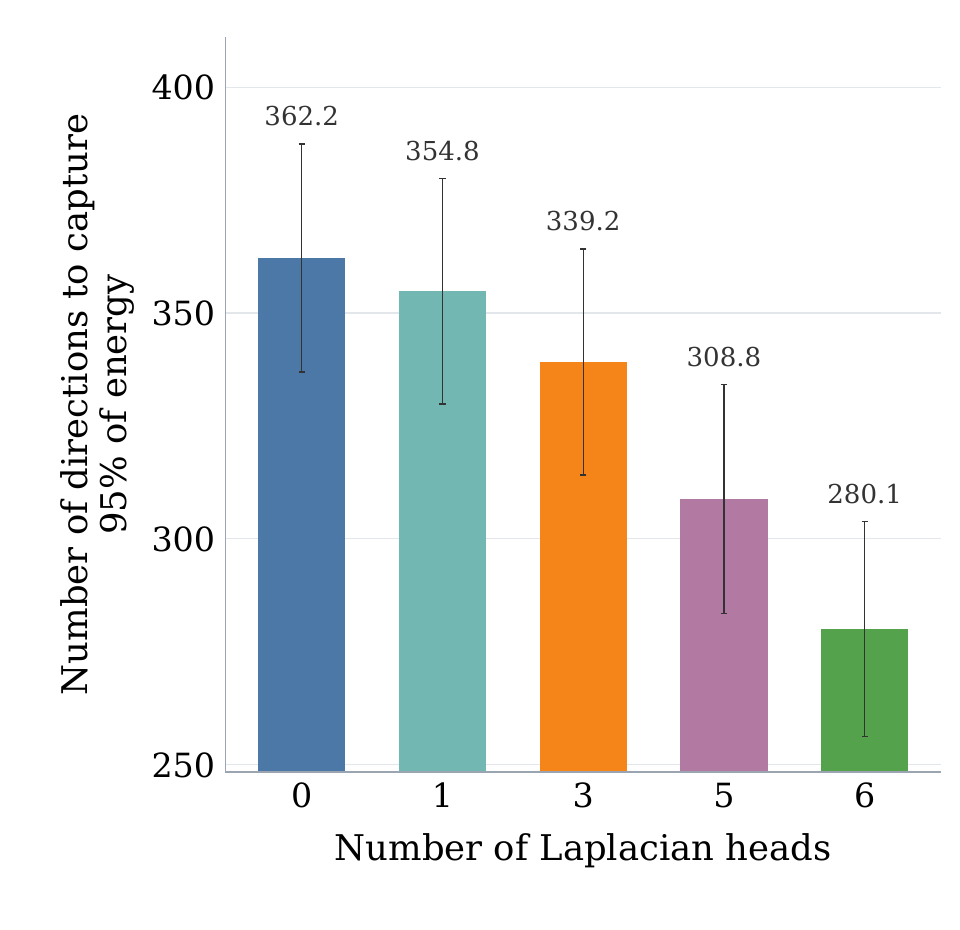}
        \caption{95\% of energy.}
    \end{subfigure}
    \par\smallskip
    \begin{subfigure}[t]{0.42\textwidth}
        \centering
        \includegraphics[width=\linewidth]{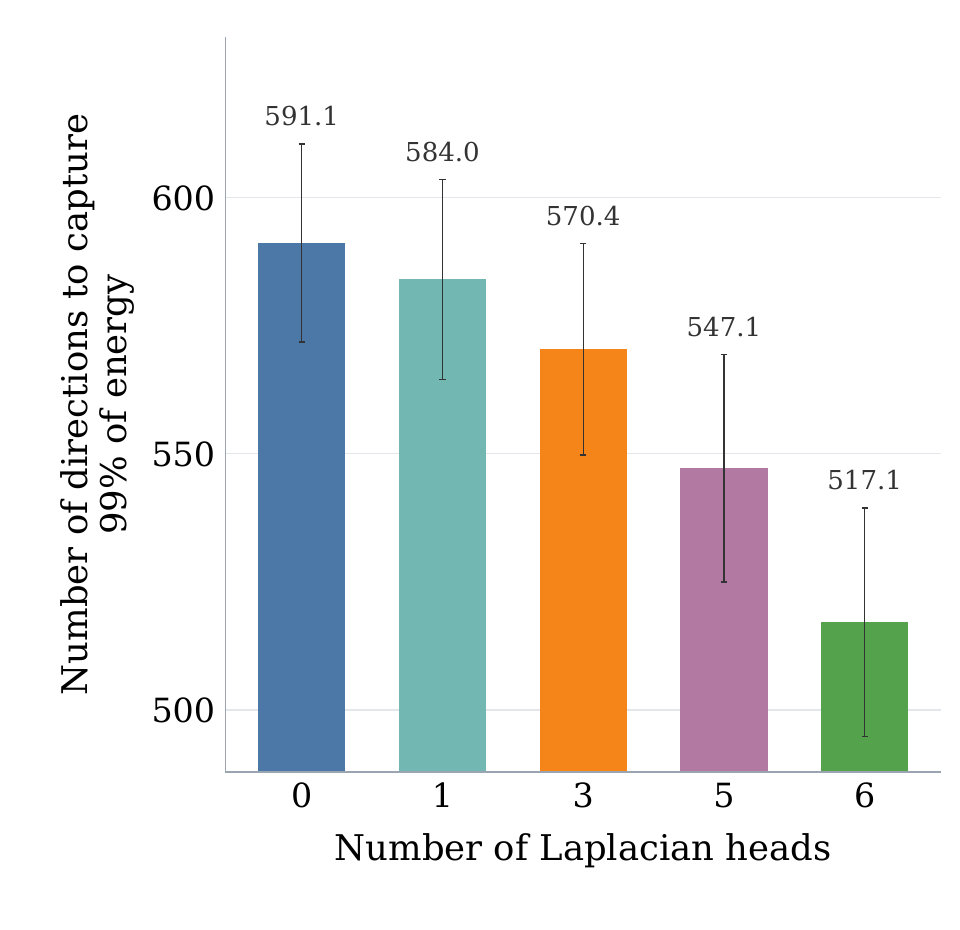}
        \caption{99\% of energy.}
    \end{subfigure}\hfill
    \begin{subfigure}[t]{0.42\textwidth}
        \centering
        \includegraphics[width=\linewidth]{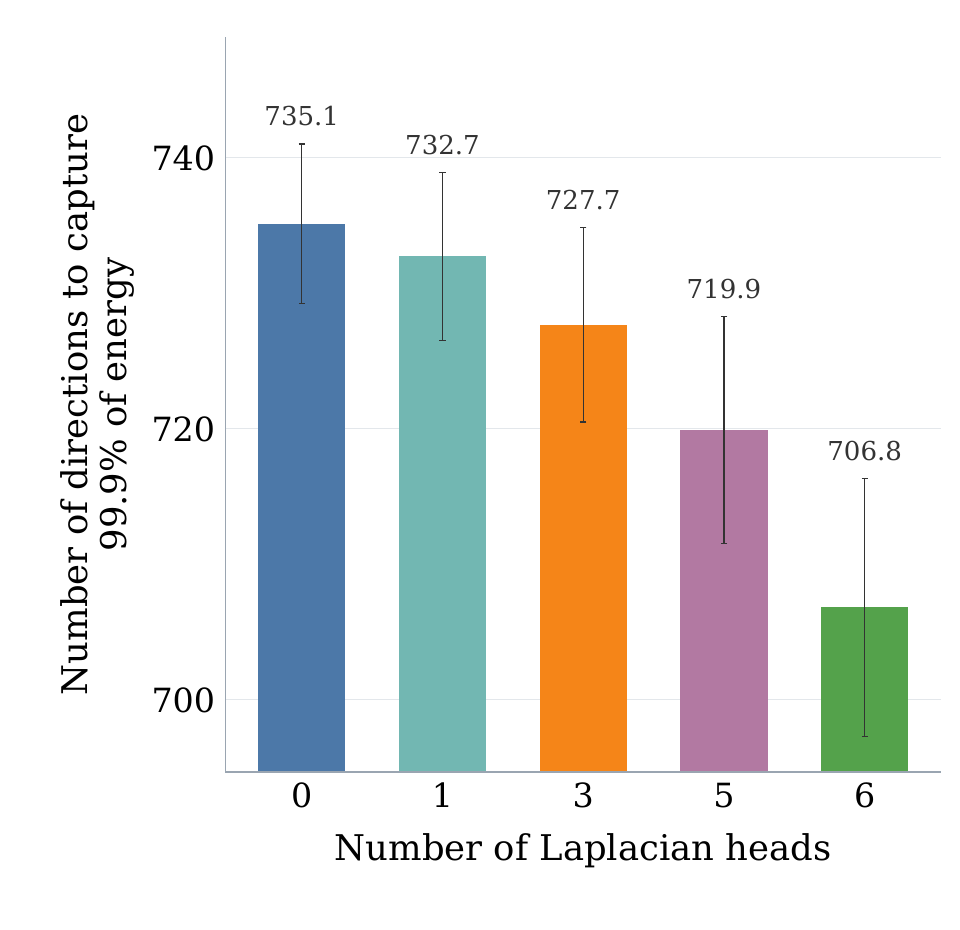}
        \caption{99.9\% of energy.}
    \end{subfigure}
    \caption{Uncentered spectrum measurements for d12 models. Shading and error bars show one standard deviation across sequences.}
    \label{fig:language_uncentered_d12}
\end{figure}

\clearpage

\begin{figure}[H]
    \centering
    \begin{subfigure}[t]{0.42\textwidth}
        \centering
        \includegraphics[width=\linewidth]{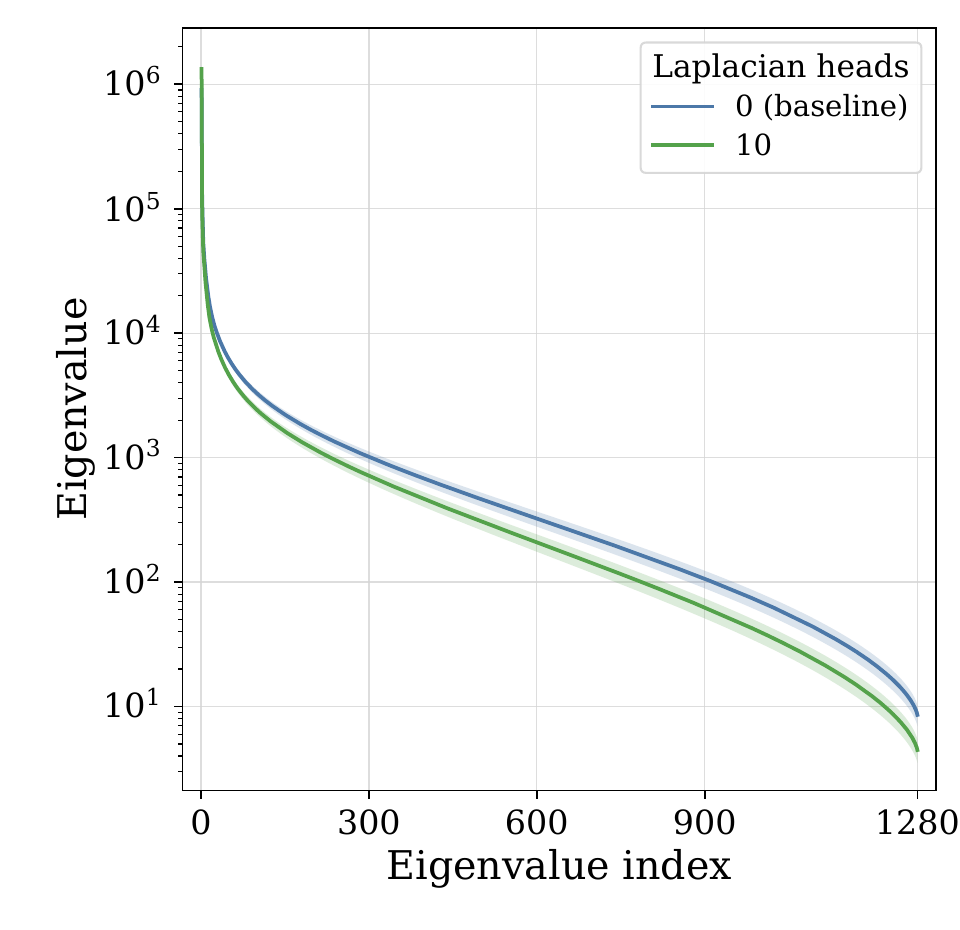}
        \caption{Baseline and 10 Laplacian heads.}
    \end{subfigure}\hfill
    \begin{subfigure}[t]{0.42\textwidth}
        \centering
        \includegraphics[width=\linewidth]{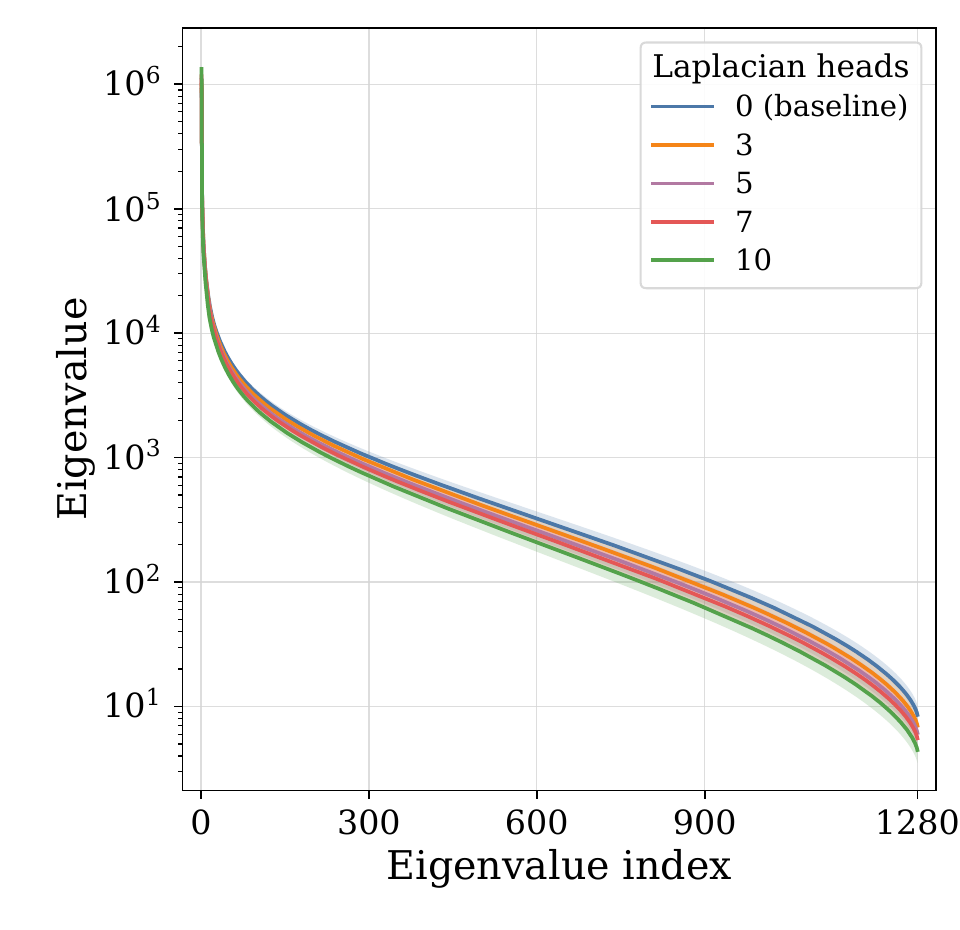}
        \caption{All models.}
    \end{subfigure}
    \par\smallskip
    \begin{subfigure}[t]{0.42\textwidth}
        \centering
        \includegraphics[width=\linewidth]{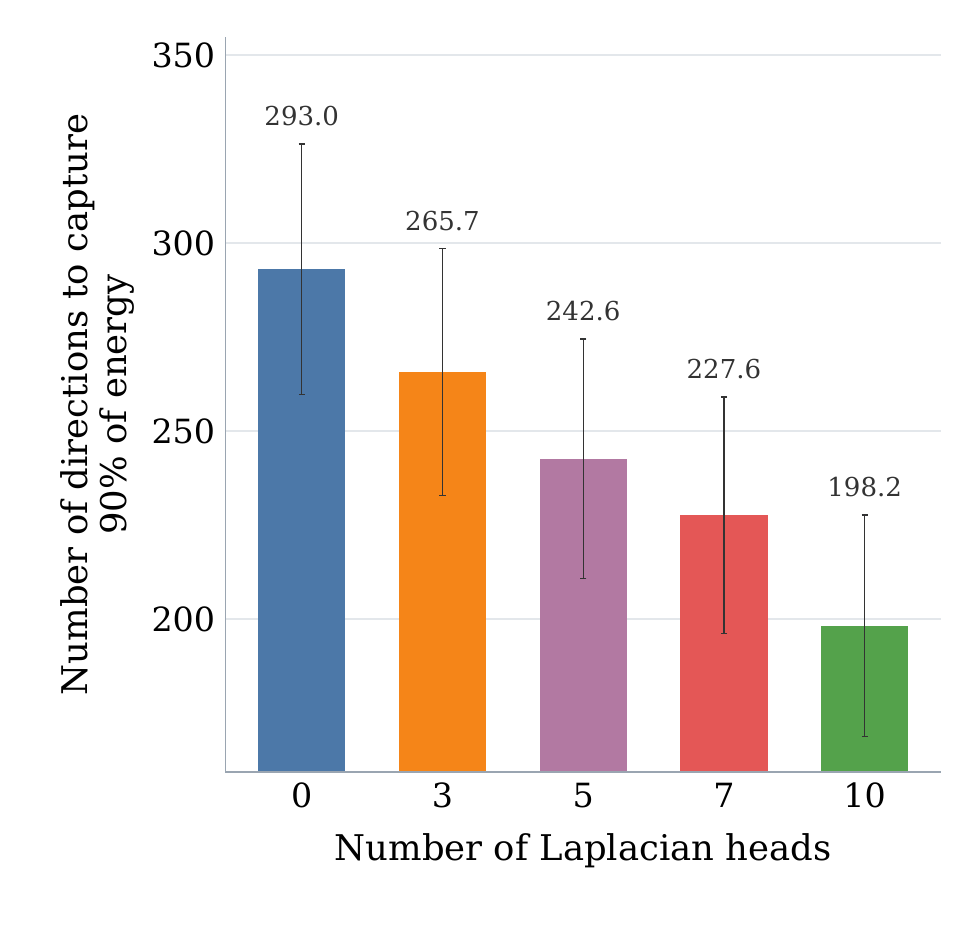}
        \caption{90\% of energy.}
    \end{subfigure}\hfill
    \begin{subfigure}[t]{0.42\textwidth}
        \centering
        \includegraphics[width=\linewidth]{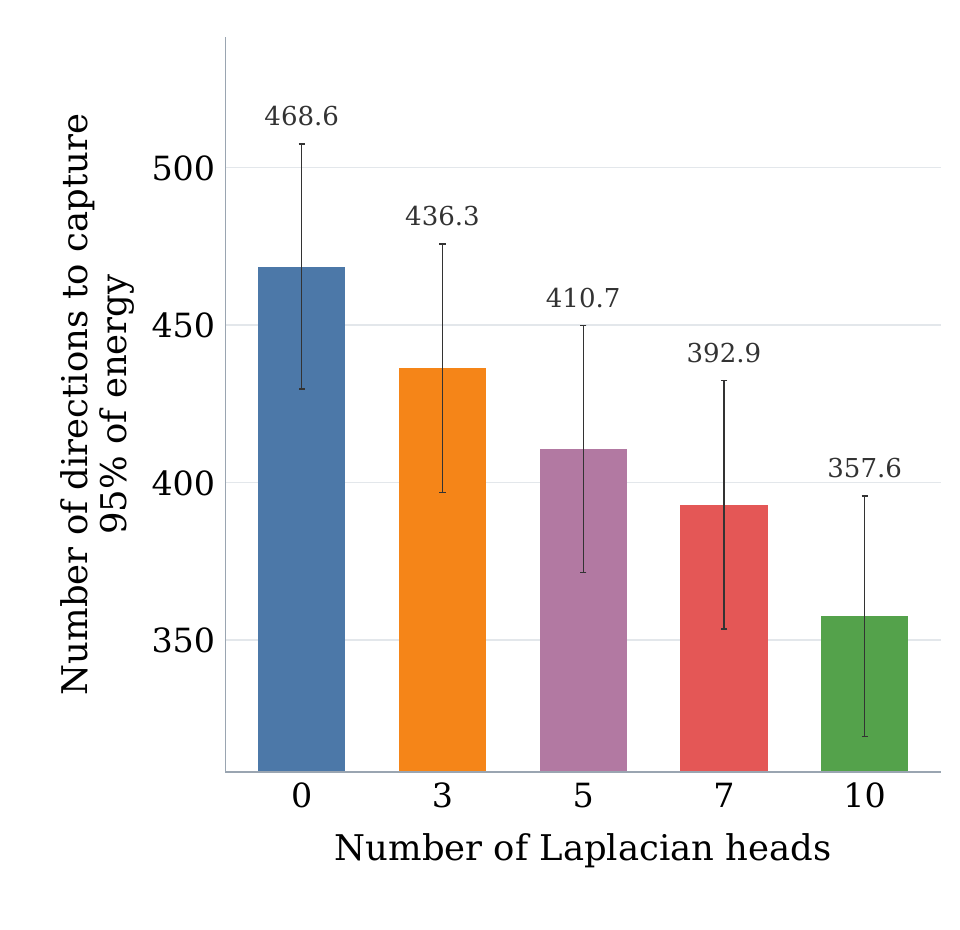}
        \caption{95\% of energy.}
    \end{subfigure}
    \par\smallskip
    \begin{subfigure}[t]{0.42\textwidth}
        \centering
        \includegraphics[width=\linewidth]{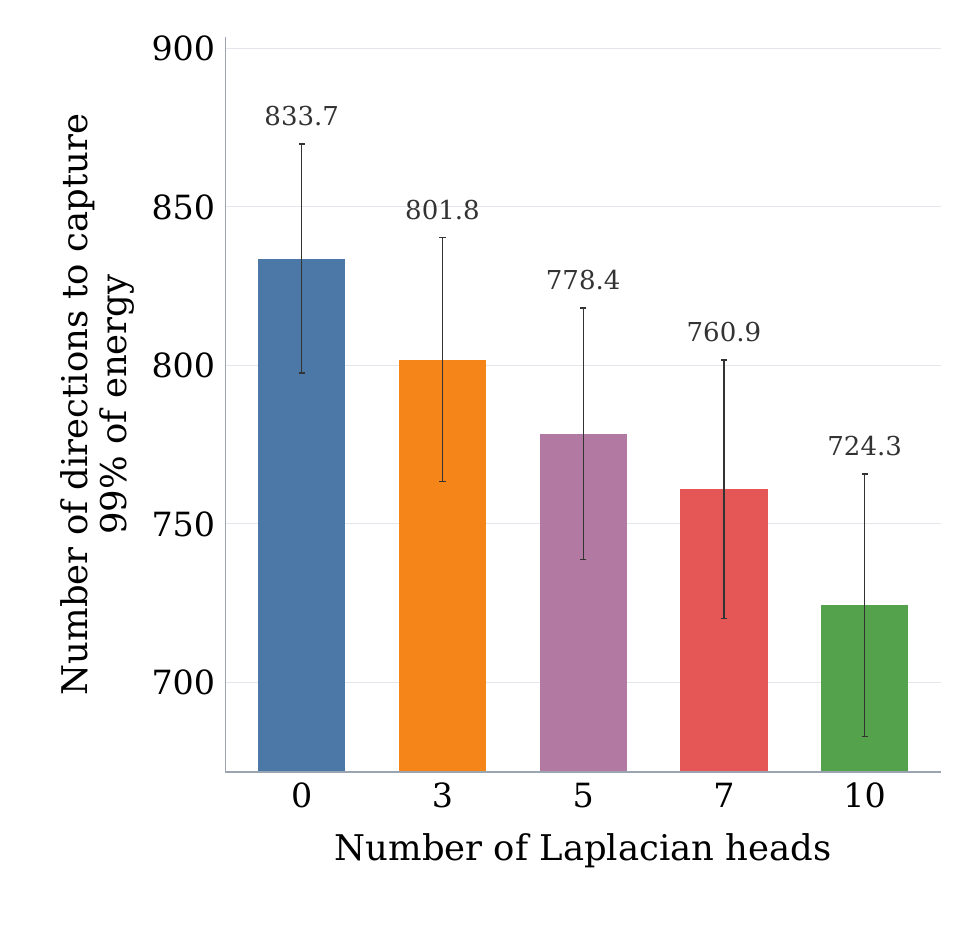}
        \caption{99\% of energy.}
    \end{subfigure}\hfill
    \begin{subfigure}[t]{0.42\textwidth}
        \centering
        \includegraphics[width=\linewidth]{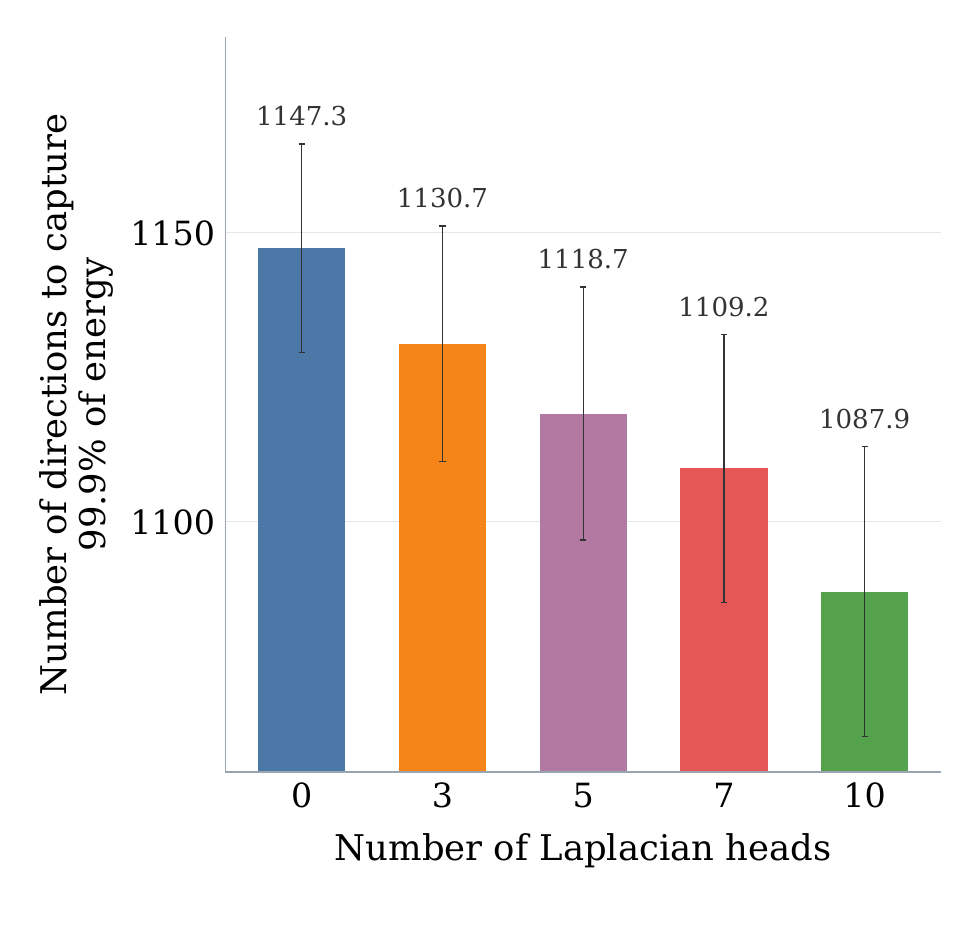}
        \caption{99.9\% of energy.}
    \end{subfigure}
    \caption{Uncentered spectrum measurements for d20 models. Shading and error bars show one standard deviation across sequences.}
    \label{fig:language_uncentered_d20}
\end{figure}

\clearpage

\begin{figure}[H]
    \centering
    \begin{subfigure}[t]{0.42\textwidth}
        \centering
        \includegraphics[width=\linewidth]{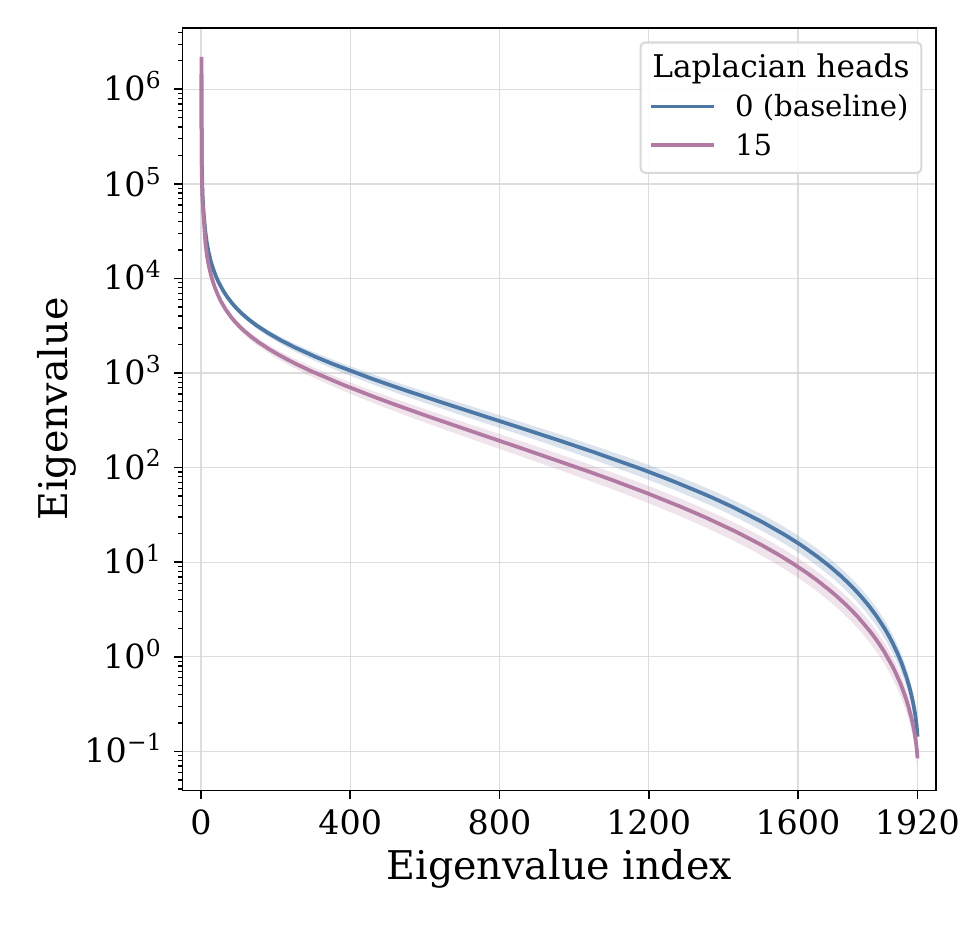}
        \caption{Baseline and 15 Laplacian heads.}
    \end{subfigure}\hfill
    \begin{subfigure}[t]{0.42\textwidth}
        \centering
        \includegraphics[width=\linewidth]{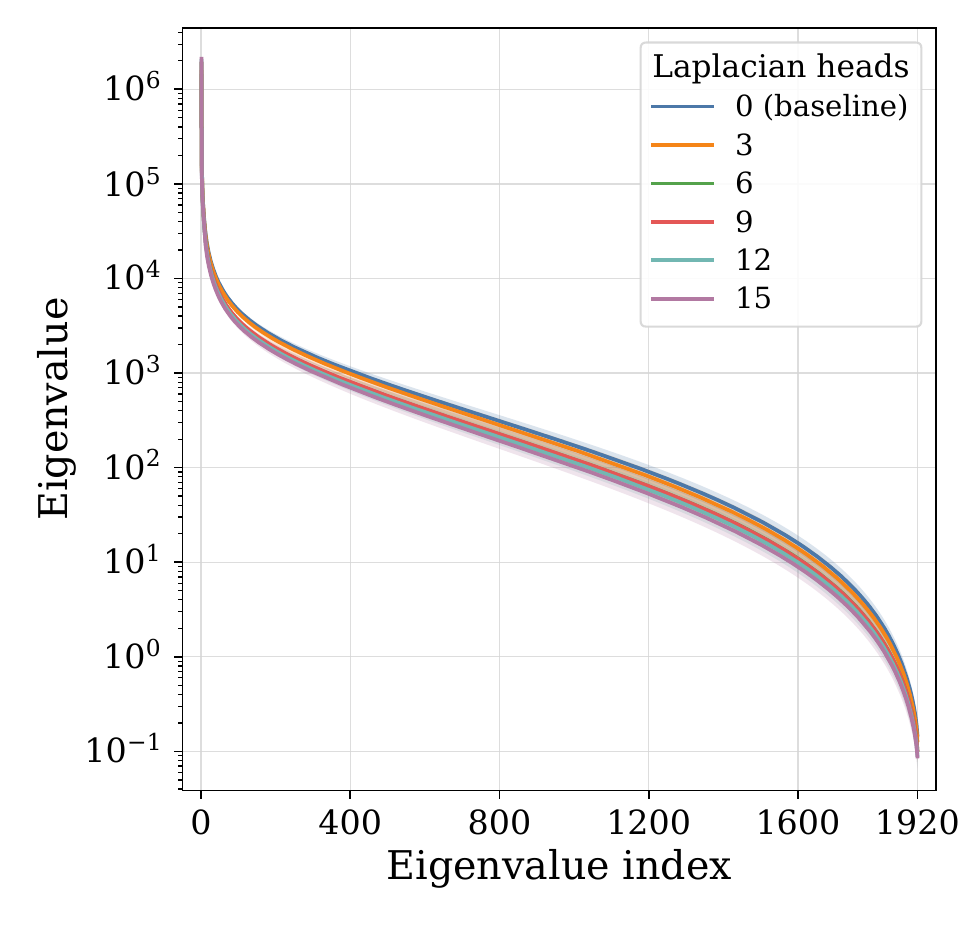}
        \caption{All models.}
    \end{subfigure}
    \par\smallskip
    \begin{subfigure}[t]{0.42\textwidth}
        \centering
        \includegraphics[width=\linewidth]{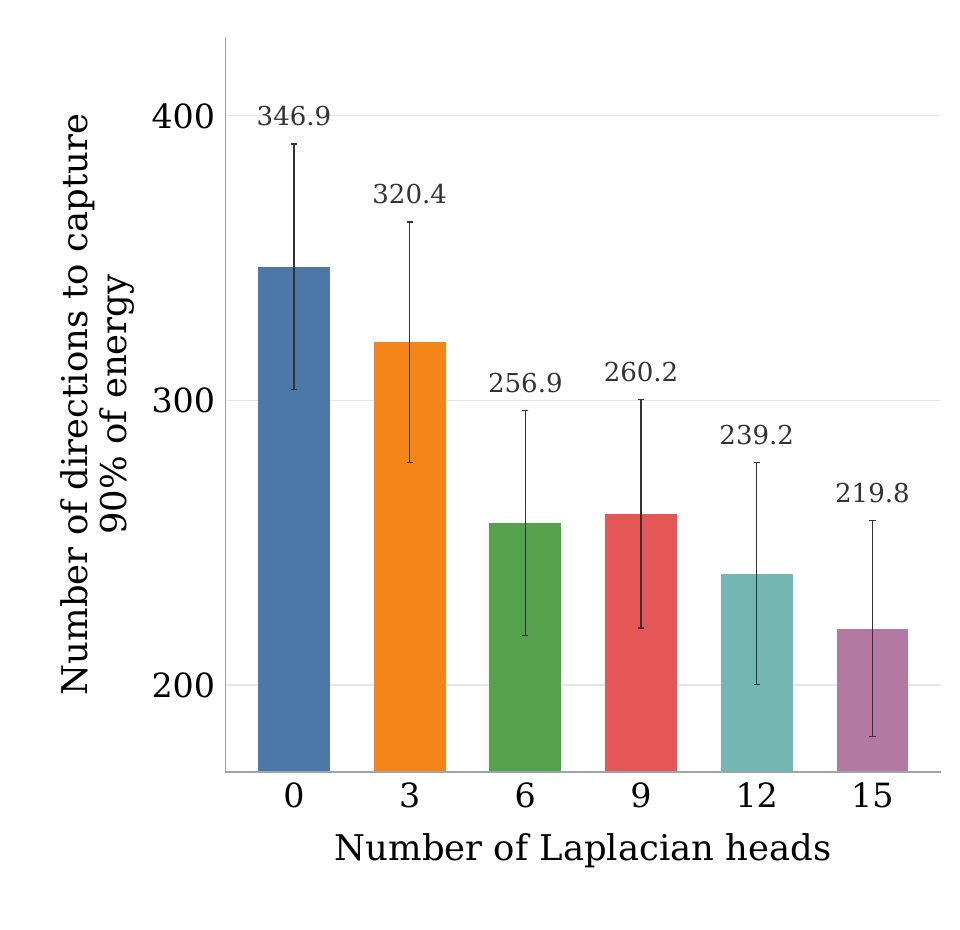}
        \caption{90\% of energy.}
    \end{subfigure}\hfill
    \begin{subfigure}[t]{0.42\textwidth}
        \centering
        \includegraphics[width=\linewidth]{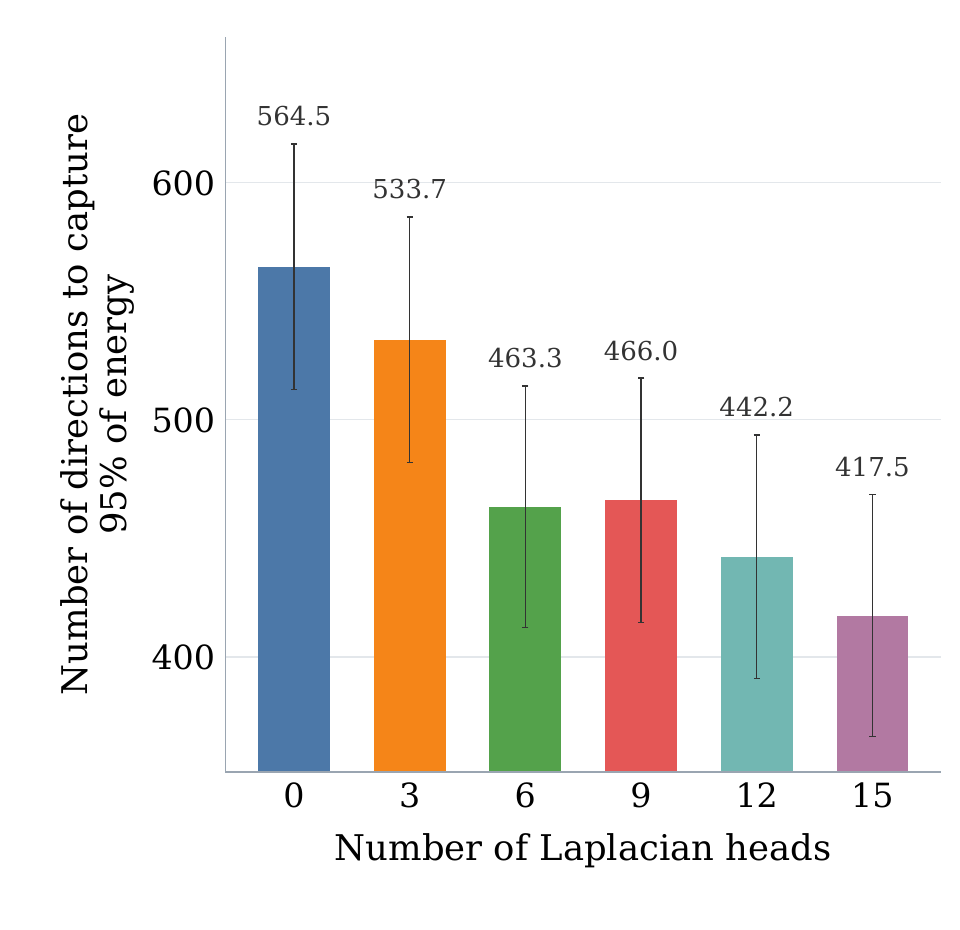}
        \caption{95\% of energy.}
    \end{subfigure}
    \par\smallskip
    \begin{subfigure}[t]{0.42\textwidth}
        \centering
        \includegraphics[width=\linewidth]{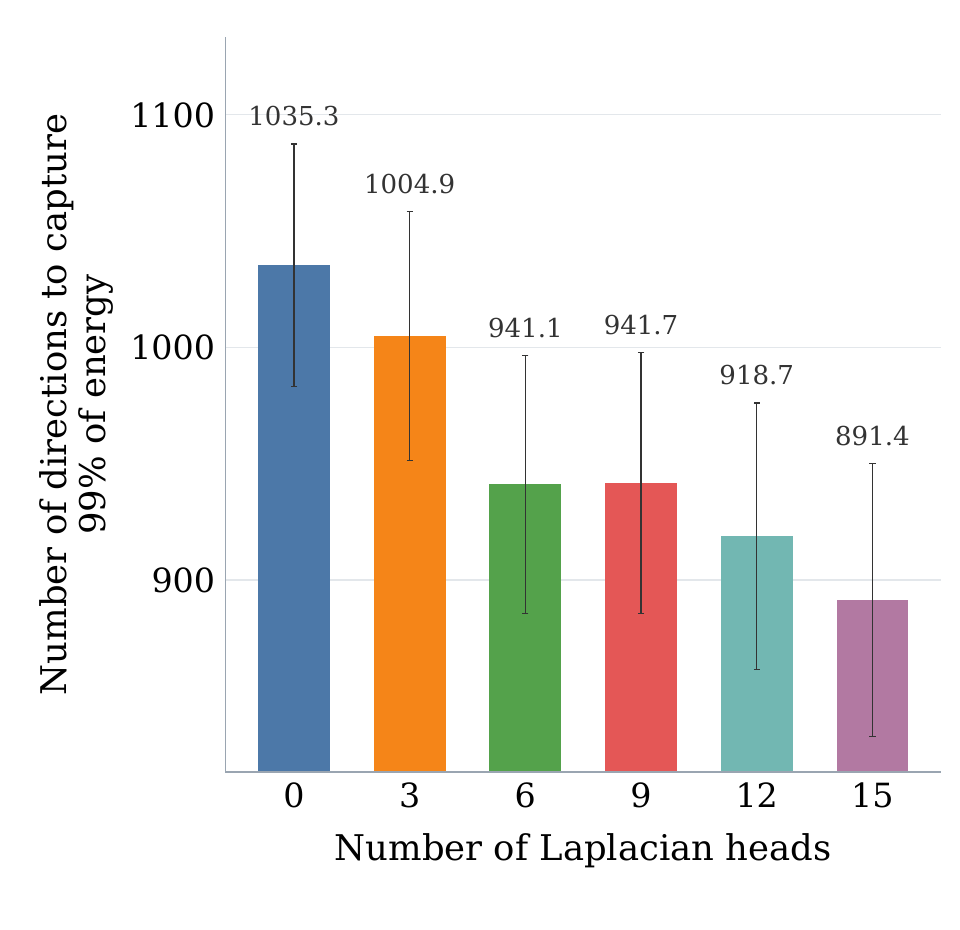}
        \caption{99\% of energy.}
    \end{subfigure}\hfill
    \begin{subfigure}[t]{0.42\textwidth}
        \centering
        \includegraphics[width=\linewidth]{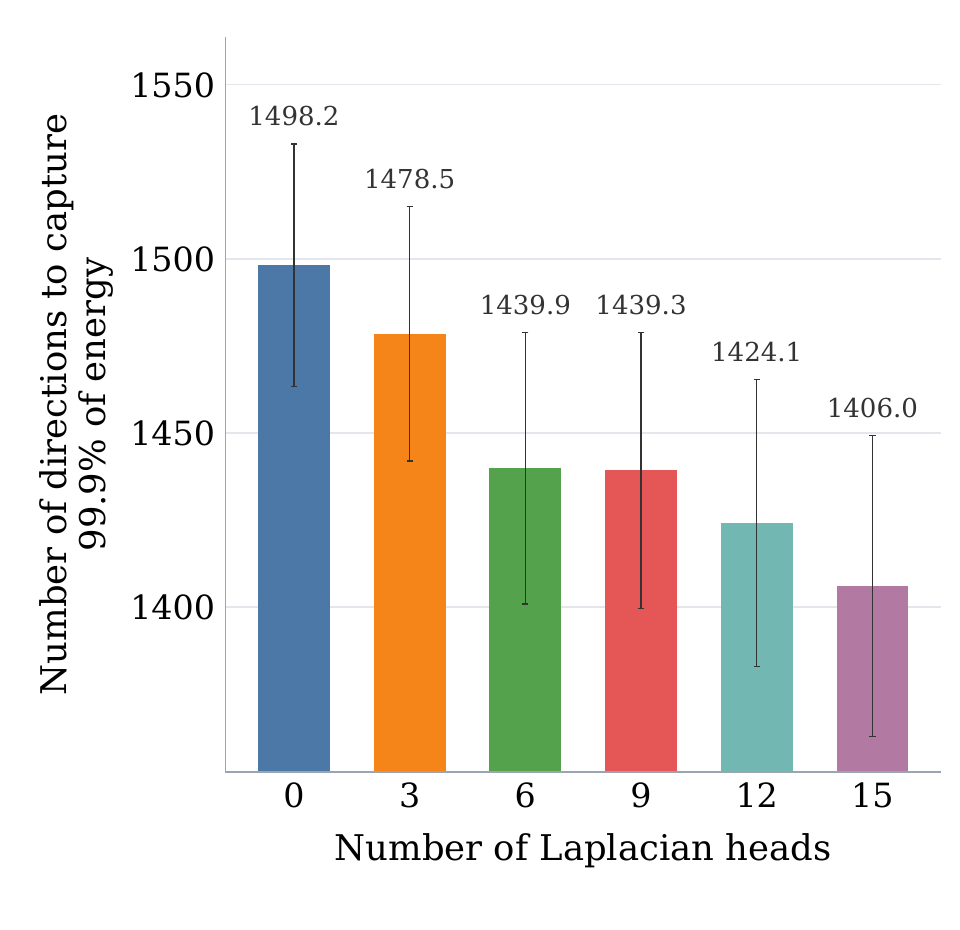}
        \caption{99.9\% of energy.}
    \end{subfigure}
    \caption{Uncentered spectrum measurements for d30 models. Shading and error bars show one standard deviation across sequences.}
    \label{fig:language_uncentered_d30}
\end{figure}


\clearpage
\subsection{DINO}
\paragraph{PCA Feature Maps}\leavevmode\par
\begin{algorithm}[H]
\caption{PCA feature maps}
\label{alg:dino_pca_maps}
\begin{algorithmic}[1]
\REQUIRE One image's final-LayerNorm patch features $X\in\R^{T\times d}$, excluding CLS/register tokens. Apply to ViT-S first.
\STATE Subtract the mean feature vector across patches from $X$.
\STATE Compute the top three PCA scores $Z$, then scale each column to unit sample variance.
\STATE For other models, reorder and sign-flip $Z$'s columns to maximize their mean Pearson correlation with the corresponding ViT-S scores for the same image.
\STATE Display $\operatorname{sigmoid}(2Z)$ as RGB on the patch grid, using all patches without masking or overlays.
\end{algorithmic}
\end{algorithm}

\begin{figure}[H]
    \centering
    \includegraphics[width=\linewidth]{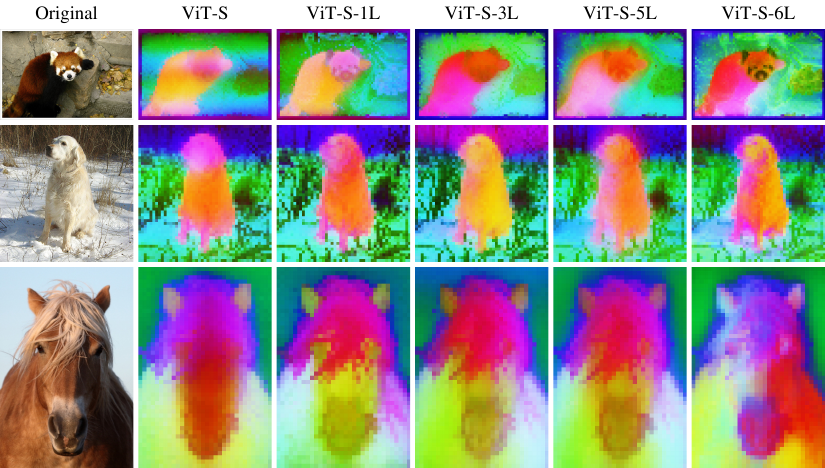}
    \caption{PCA feature maps for red panda, golden retriever, and horse (top to bottom).}
    \label{fig:dino_pca_animals_1_app}
\end{figure}

\begin{figure}[H]
    \centering
    \includegraphics[width=\linewidth]{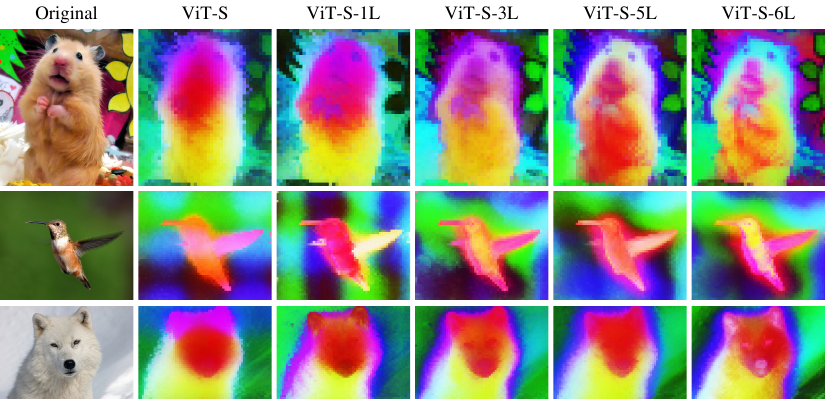}
    \caption{PCA feature maps for hamster, hummingbird, and Arctic wolf in snow (top to bottom).}
    \label{fig:dino_pca_animals_2_app}
\end{figure}

\begin{figure}[H]
    \centering
    \includegraphics[width=\linewidth]{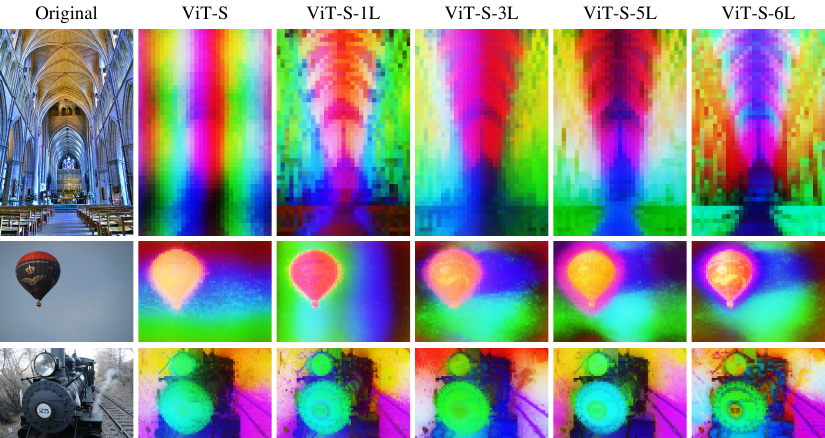}
    \caption{PCA feature maps for a cathedral's vaulted nave, a hot air balloon, and a steam locomotive (top to bottom).}
    \label{fig:dino_pca_scenes_1_app}
\end{figure}

\begin{figure}[H]
    \centering
    \includegraphics[width=\linewidth]{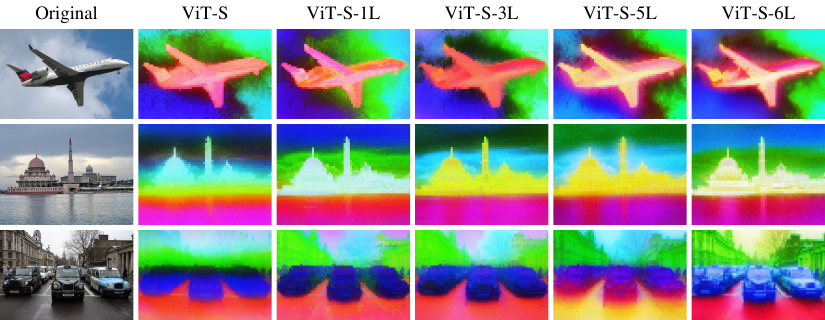}
    \caption{PCA feature maps for an airliner in flight, a waterfront mosque, and taxis on a busy street (top to bottom).}
    \label{fig:dino_pca_scenes_2_app}
\end{figure}

\Needspace{0.46\textheight}
\paragraph{Spectrum Measurements}
All spectrum measurements use the output token representations of the final LayerNorm.
As in Figure~\ref{fig:spectrum_measurements}, we use the same 1,000 ImageNet validation images for all models.

\noindent
\begin{minipage}[t]{0.48\textwidth}
    \vspace{0pt}
    \captionsetup{type=figure}
    \centering
    \includegraphics[width=\linewidth]{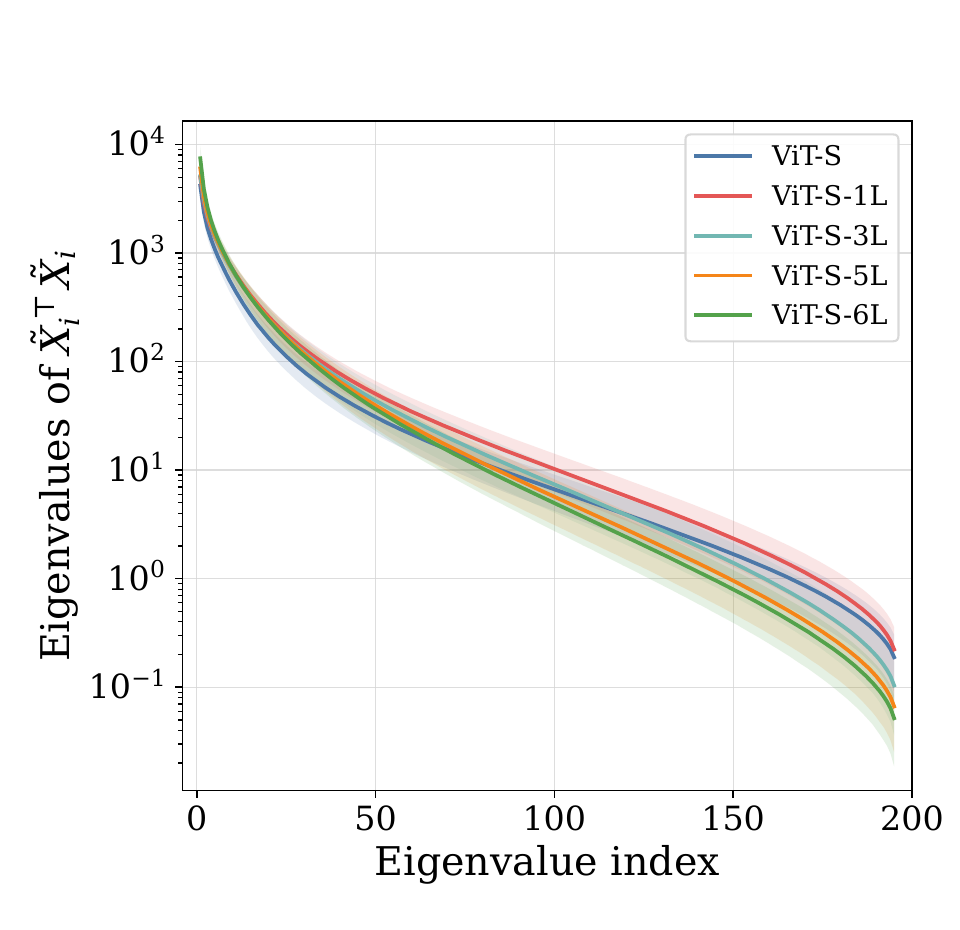}
    \caption{Spectra for DINO models with varying numbers of Laplacian heads.}
    \label{fig:dino_spectrum_all_models_app}
\end{minipage}\hfill
\begin{minipage}[t]{0.48\textwidth}
    \vspace{0pt}
    \captionsetup{type=figure}
    \centering
    \includegraphics[width=\linewidth]{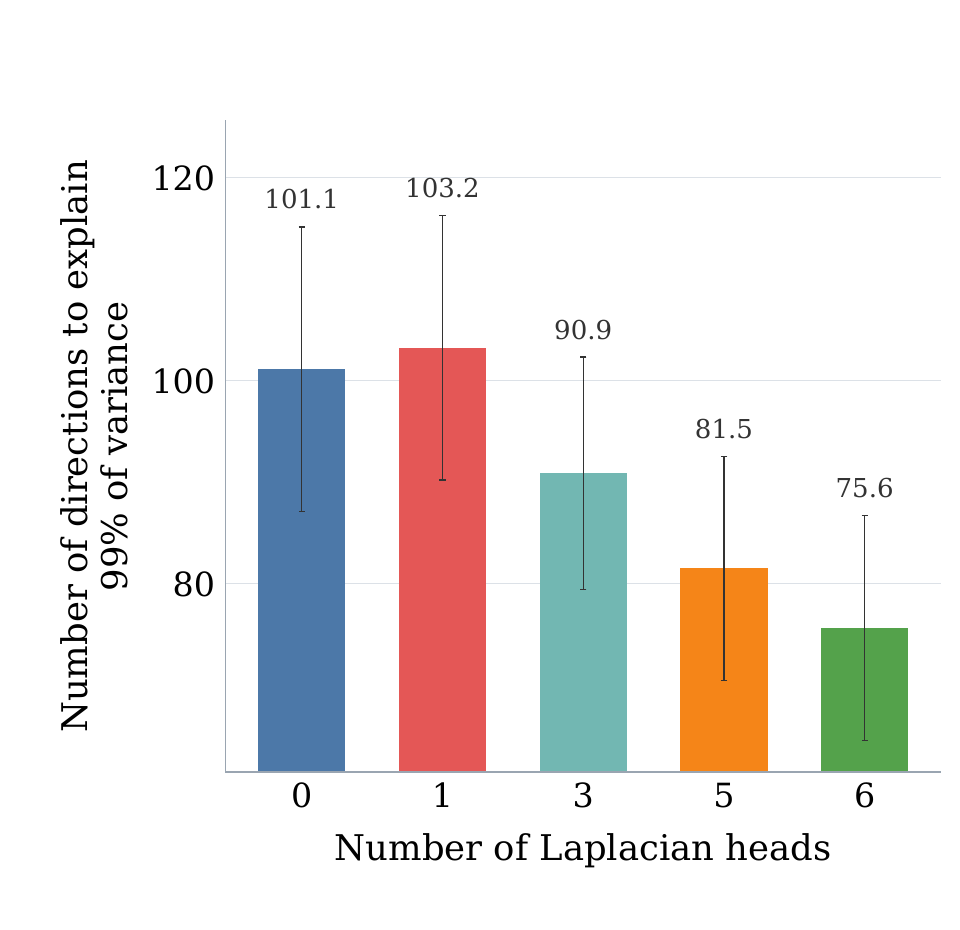}
    \caption{Number of principal components explaining 99\% of variance for DINO models.}
    \label{fig:dino_pc_99_app}
\end{minipage}\par

\clearpage
\paragraph{Uncentered Spectrum Measurements}
These measurements use the same models and images without centering. Spectra are the eigenvalues of $X_i^\top X_i$, and component counts capture fractions of the total energy $\|X_i\|_F^2$.

\begin{figure}[H]
    \centering
    \begin{subfigure}[t]{0.48\textwidth}
        \centering
        \includegraphics[width=\linewidth]{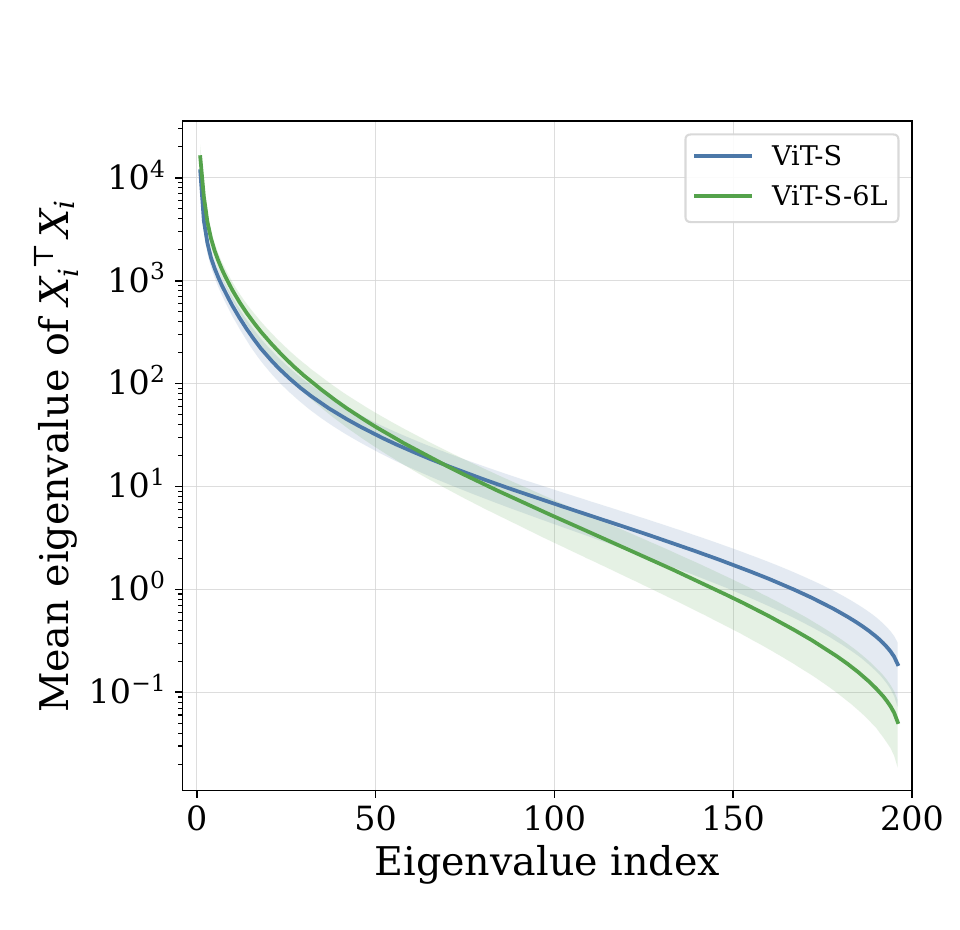}
        \caption{ViT-S and ViT-S-6L.}
    \end{subfigure}\hfill
    \begin{subfigure}[t]{0.48\textwidth}
        \centering
        \includegraphics[width=\linewidth]{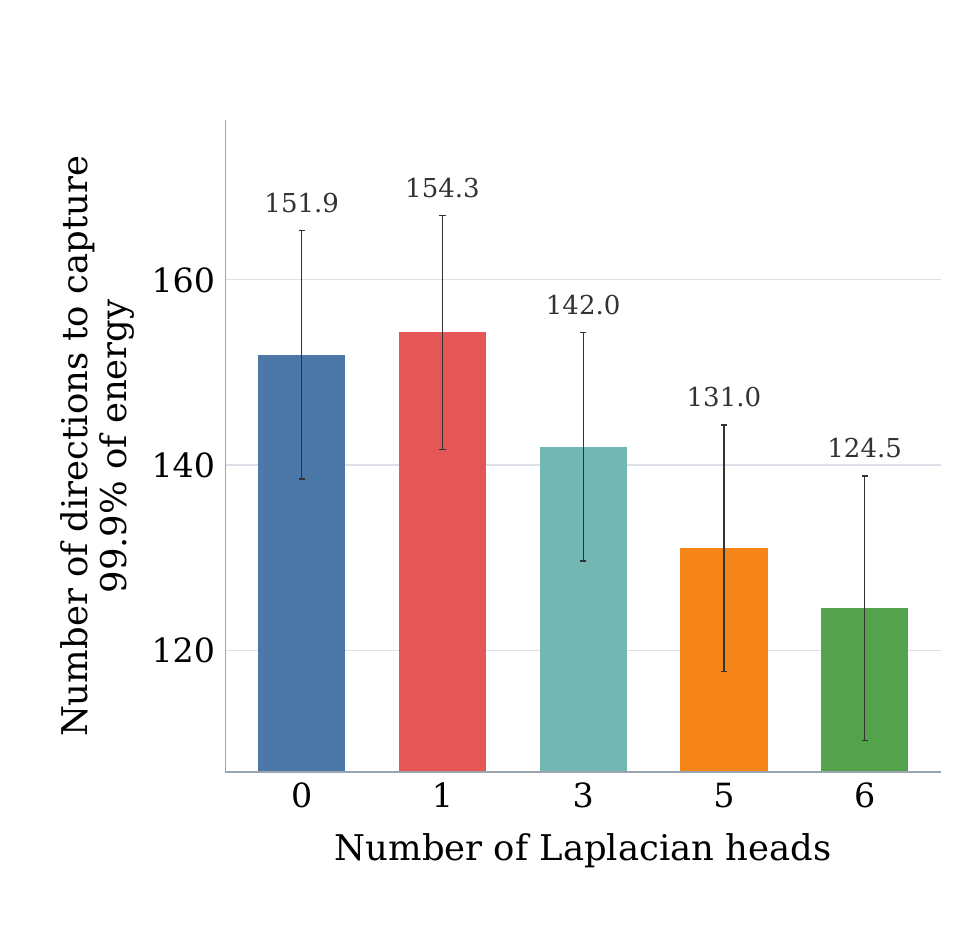}
        \caption{99.9\% of energy.}
    \end{subfigure}
    \par\smallskip
    \begin{subfigure}[t]{0.48\textwidth}
        \centering
        \includegraphics[width=\linewidth]{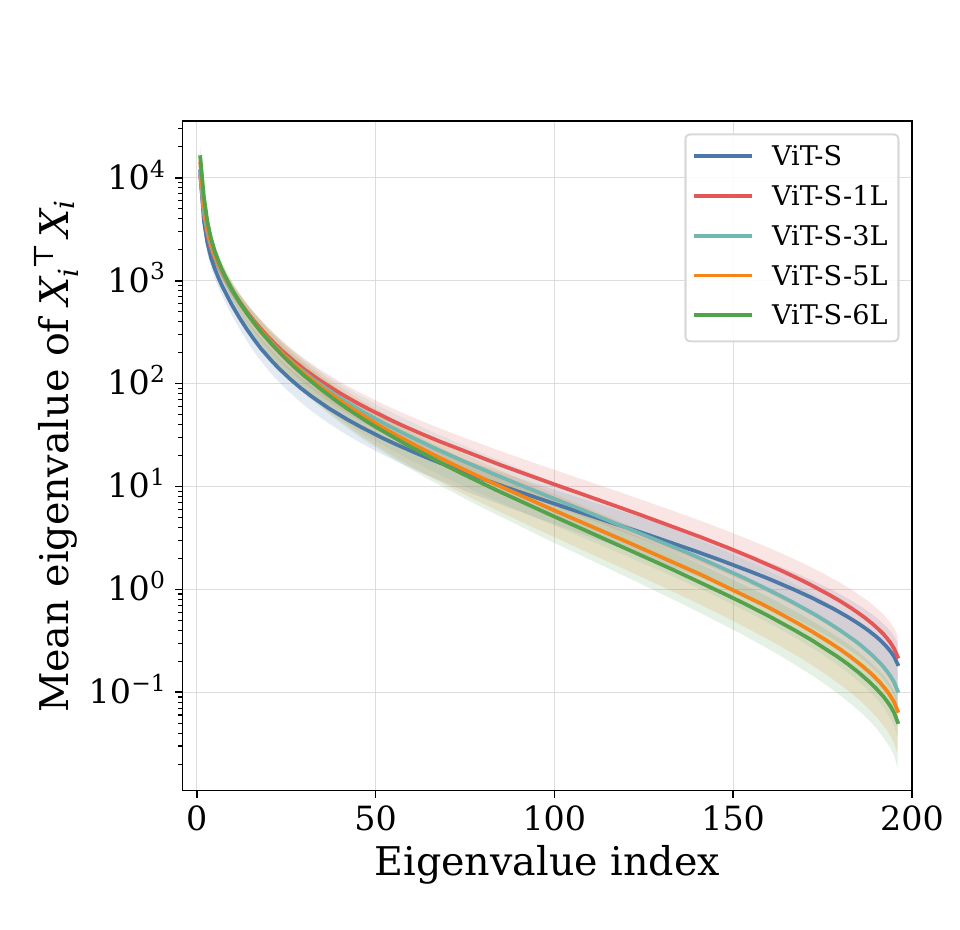}
        \caption{All models.}
    \end{subfigure}\hfill
    \begin{subfigure}[t]{0.48\textwidth}
        \centering
        \includegraphics[width=\linewidth]{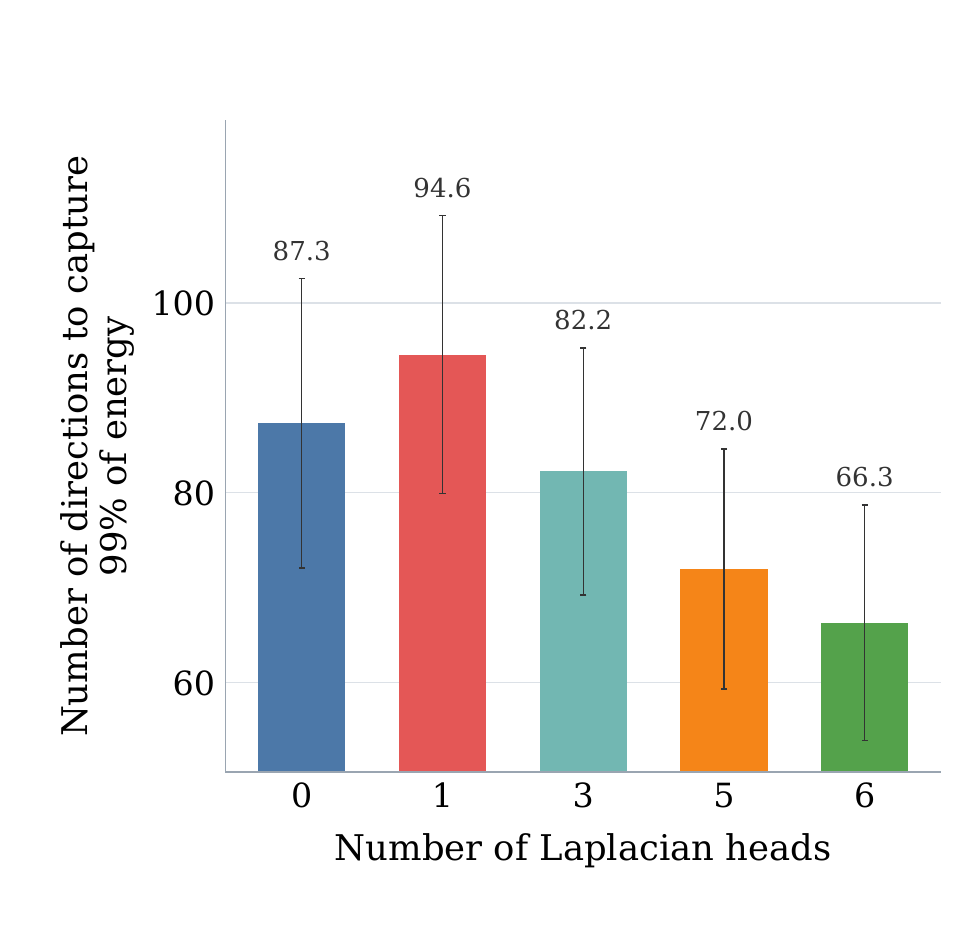}
        \caption{99\% of energy.}
    \end{subfigure}
    \caption{Uncentered spectrum measurements for DINO models.}
    \label{fig:dino_uncentered_supplement}
\end{figure}

\clearpage
\section{Empirical Evidence for Motivation 1}\label{app:out_snr}
Given a batch of sequences of token representations $X \in \R^{B\times T \times d}$, we define the average signal-to-noise ratio (SNR) of $X$ as
\[\text{SNR}(X) = \frac{1}{B} \sum_{b=1}^B \frac{||\text{Mean}(X_b)||_2}{\text{Std}(X_b)},\]
where 
\[\text{Mean}(X_b) = \frac{1}{T}\sum_{i=1}^T X_{b, i}\quad \text{ and } \quad \text{Std}(X_b) = \sqrt{\frac{1}{T} \sum_{i=1}^T ||X_{b, i} - \text{Mean}(X_b)||_2^2}\]
Here, $X_b \in \R^{T\times d}$ denote the $b$th sequence in the batch and $X_{b, i} \in \R^d$ denote the $i$th token representation vector within the sequence. 

\begin{figure}[H]
    \includegraphics[width=\linewidth, trim=0 24 0 24, clip]{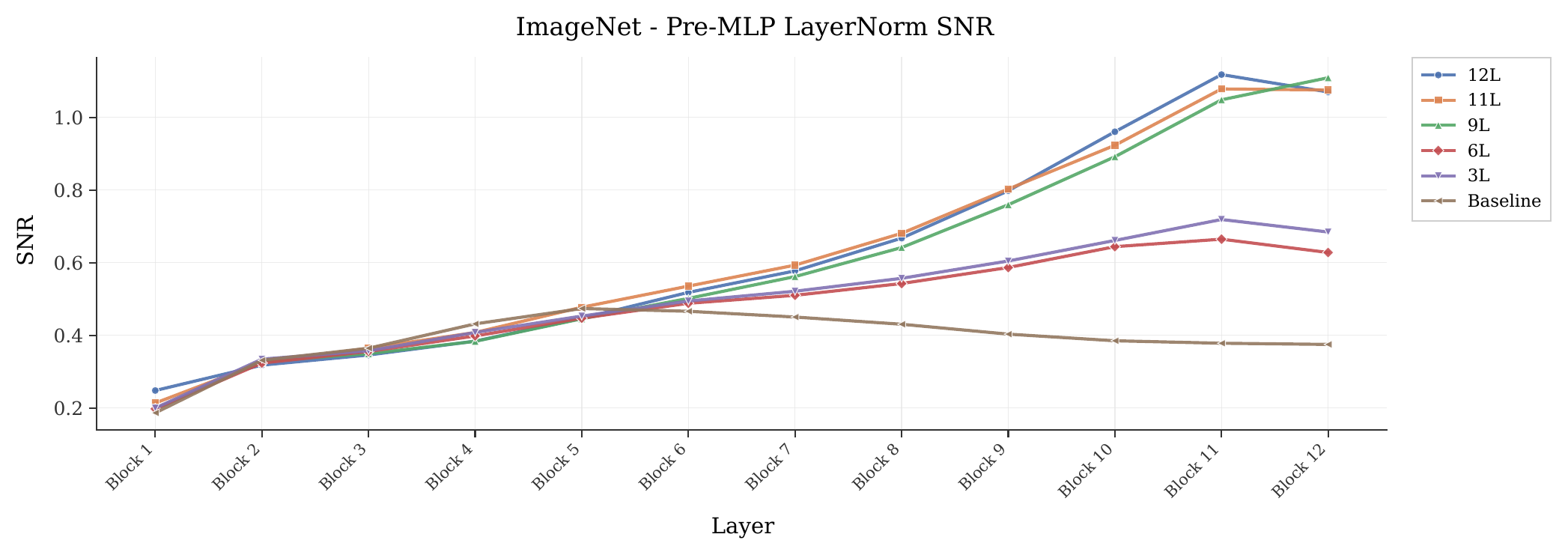}
    \caption{The Laplacian mechanism collapses tokens more effectively.}
    \label{fig:snr}
\end{figure}

The SNR measures how large (measured by the $l_2$ norm) the mean of a sequence of tokens is relative to their variance/standard deviation. To validate the interpretation in Section~\ref{sec:motivation1}, we measure the SNR of the output of the pre-MLP Layer Normalization module. i.e., we measure the SNR of
\[\text{LayerNorm} \bigl(X + \text{Multi-Head Attention}(\text{LayerNorm}(X))\bigr)\]

across layers. Figure~\ref{fig:snr} plots the SNR of the ImageNet token representations for the baseline and the proposed models as a function of depth. It clearly illustrates that for all models that use Laplacian heads, the output of the Pre-MLP LayerNorm has higher SNR than the baseline across layers, growing more drastically as depth increases. Moreover, more Laplacian heads lead to higher SNR and steeper growth. This measurement directly supports the intuition in Section~\ref{sec:motivation1} that Laplacian heads control the within-sequence variance more effectively.

\section{Neural Collapse} \label{appendix_nc}
\subsection{Metrics}
Let $M \in \mathbb{R}^{d \times C}$ be the matrix whose columns are the class means $\{\mu_i: 1 \leq i \leq C\}$ and $W \in \R^{C \times d}$ be the weight matrix of the final-layer classifier. We quantify NC2 - NC4 following \cite{han2022neuralcollapsemseloss}:
\begin{itemize}
    \item \textbf{NC2 (Equinorm and Maximal Equiangularity)}:  
    \begin{itemize}
        \item \textit{Equinorm}: Measures how uniform the vector norms are within the class means or weights, using the coefficient of variation (CoV):
        \[
        \frac{\operatorname{std}(\|\mu_c\|)}{\operatorname{mean}(\|\mu_c\|)} \quad \text{and} \quad \frac{\operatorname{std}(\|w_c\|)}{\operatorname{mean}(\|w_c\|)},
        \]
        where $w_c$ is the classifier weight vector corresponding to class $c$.
        
        \item \textit{Maximal Equiangularity}: Measures how close the vectors are to forming a maximally equiangular tight frame (ETF):
        \[
        \frac{1}{C(C-1)} \sum_{i \ne j} \left| \left\langle \hat{v}_i, \hat{v}_j \right\rangle + \frac{1}{C-1} \right|,
        \]
        where $\hat{v}_i$ and $\hat{v}_j$ are $\ell_2$-normalized class means or weight vectors. A lower value indicates greater conformity to an ETF structure.
    \end{itemize}

    \item \textbf{NC3 (Self-Duality)}:  
    Measures the alignment between the classifier weights and the centered class means:
    \[
    \left\| \frac{W^T}{\|W^T\|_F} - \frac{M'}{\|M'\|_F} \right\|_F^2,
    \]
    where $M' = M - \mu_G \mathbf{1}^T$ is the matrix of class means centered by their global mean $\mu_G$.

    \item \textbf{NC4 (Convergence to NCC)}:  
    Measures how close the learned classifier is to a Nearest Class Center (NCC) classifier:
    \[
    1 - \frac{1}{N} \sum_{i=1}^N \mathds{1} \left[ \arg\max f(x_i) = \arg\min_c \|h_i - \mu_c\| \right],
    \]
    where $f(x_i)$ are the logits, $h_i$ is the feature of sample $x_i$, and $\mu_c$ is the mean feature for class $c$.
\end{itemize}

\Needspace{0.42\textheight}
\paragraph{NC Metrics Results}\leavevmode
\begin{figure}[H]
\centering
\begin{subfigure}[b]{0.48\textwidth}
  \centering
  \includegraphics[width=0.9\linewidth]{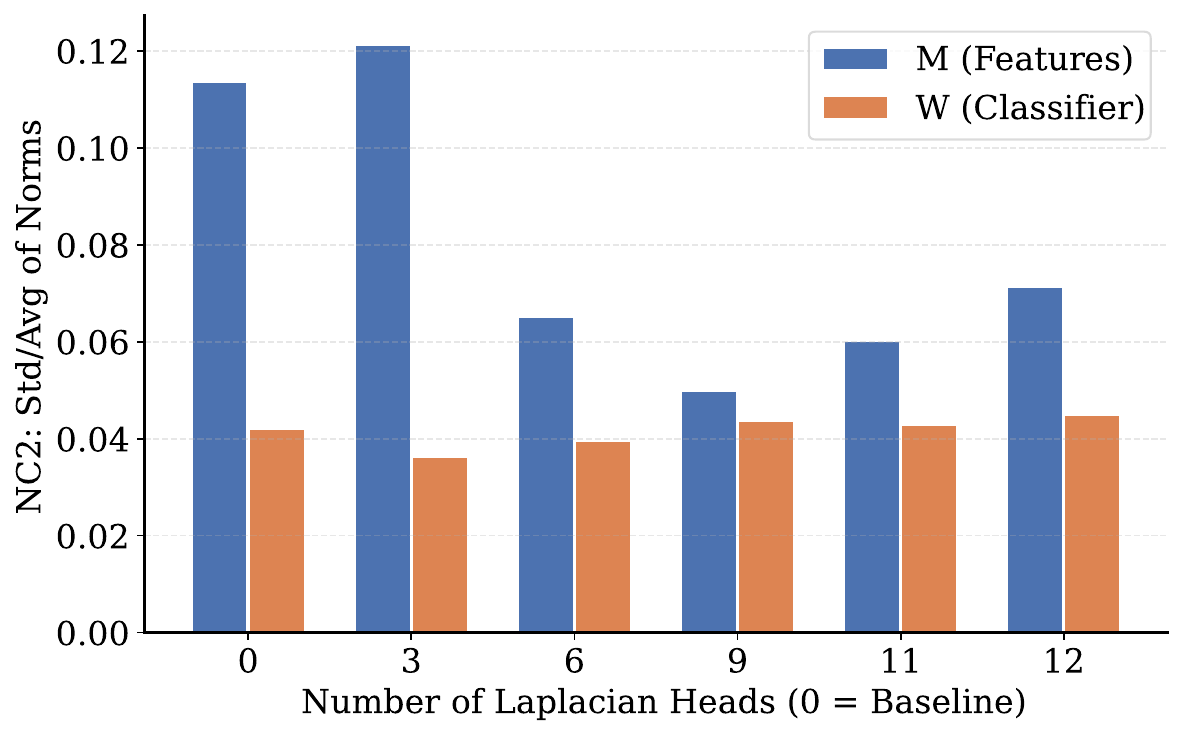}
  \caption{NC2: Equinormness}
  \label{fig:nc2a}
\end{subfigure}
\hfill
\begin{subfigure}[b]{0.48\textwidth}
  \centering
  \includegraphics[width=0.9\linewidth]{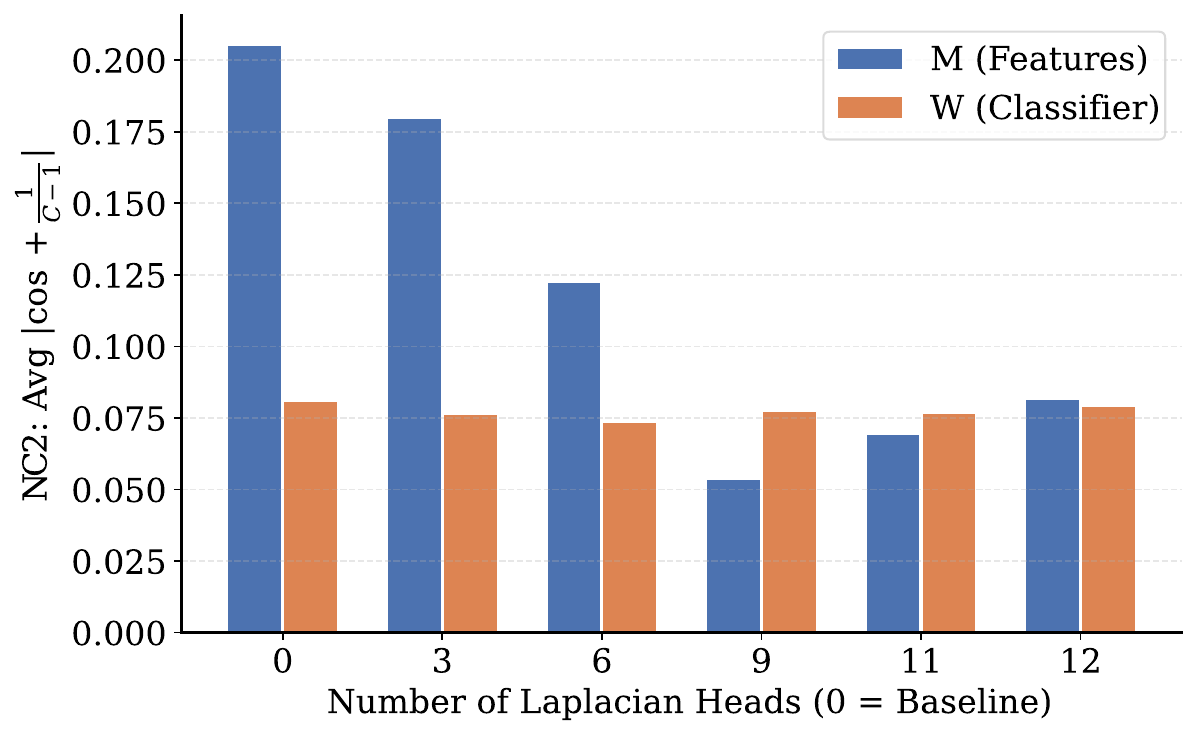}
  \caption{NC2: Max Equiangularity}
  \label{fig:nc2b}
\end{subfigure}

\vspace{-0.1em}

\begin{subfigure}[b]{0.48\textwidth}
  \centering
  \includegraphics[width=0.9\linewidth]{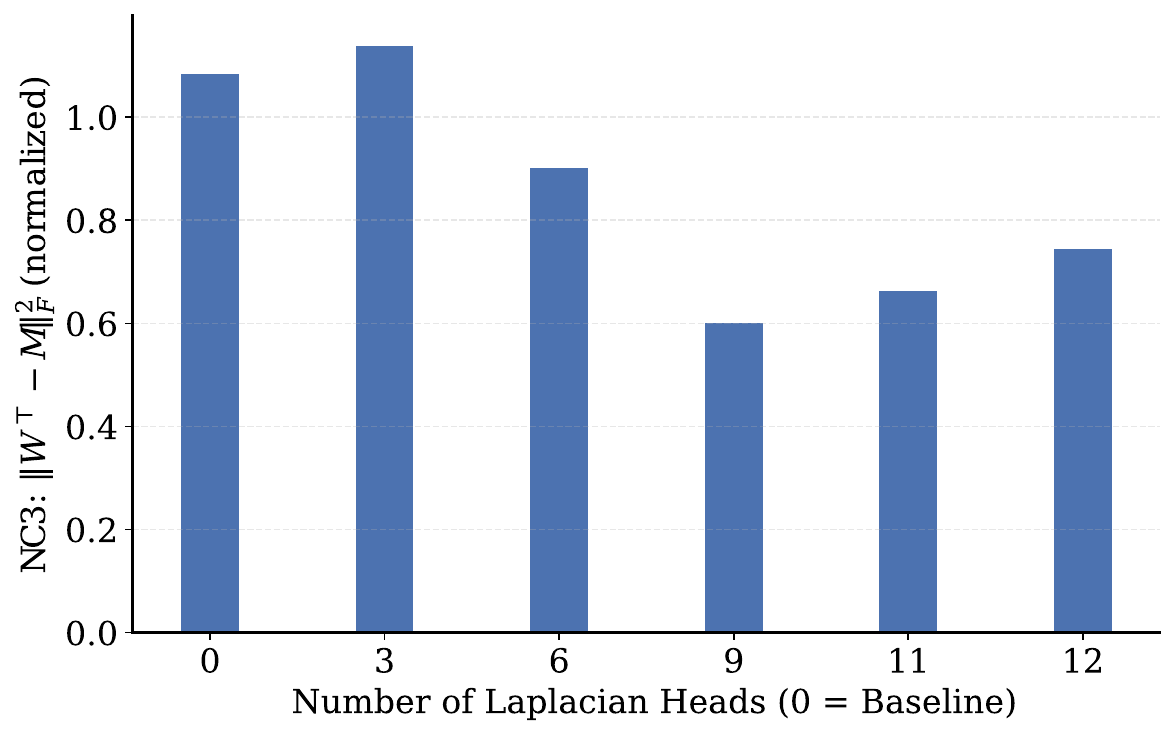}
  \caption{NC3: Self-Duality}
  \label{fig:nc3}
\end{subfigure}
\hfill
\begin{subfigure}[b]{0.48\textwidth}
  \centering
  \includegraphics[width=0.9\linewidth]{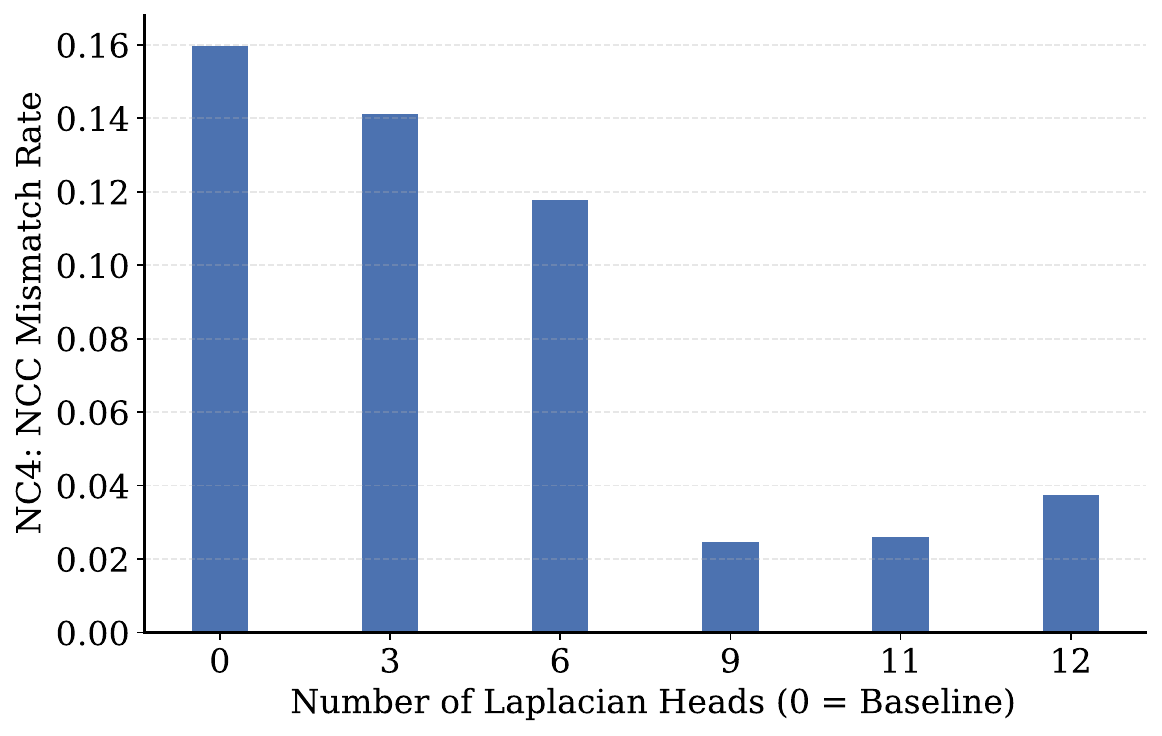}
  \caption{NC4: Convergence to NCC}
  \label{fig:nc4}
\end{subfigure}
\caption{Neural-collapse metrics on CIFAR-10.}
\label{fig:nc_metrics_cifar10}
\end{figure}

\begin{figure}[H]
\centering
\begin{subfigure}[b]{0.48\textwidth}
  \centering
  \includegraphics[width=0.9\linewidth]{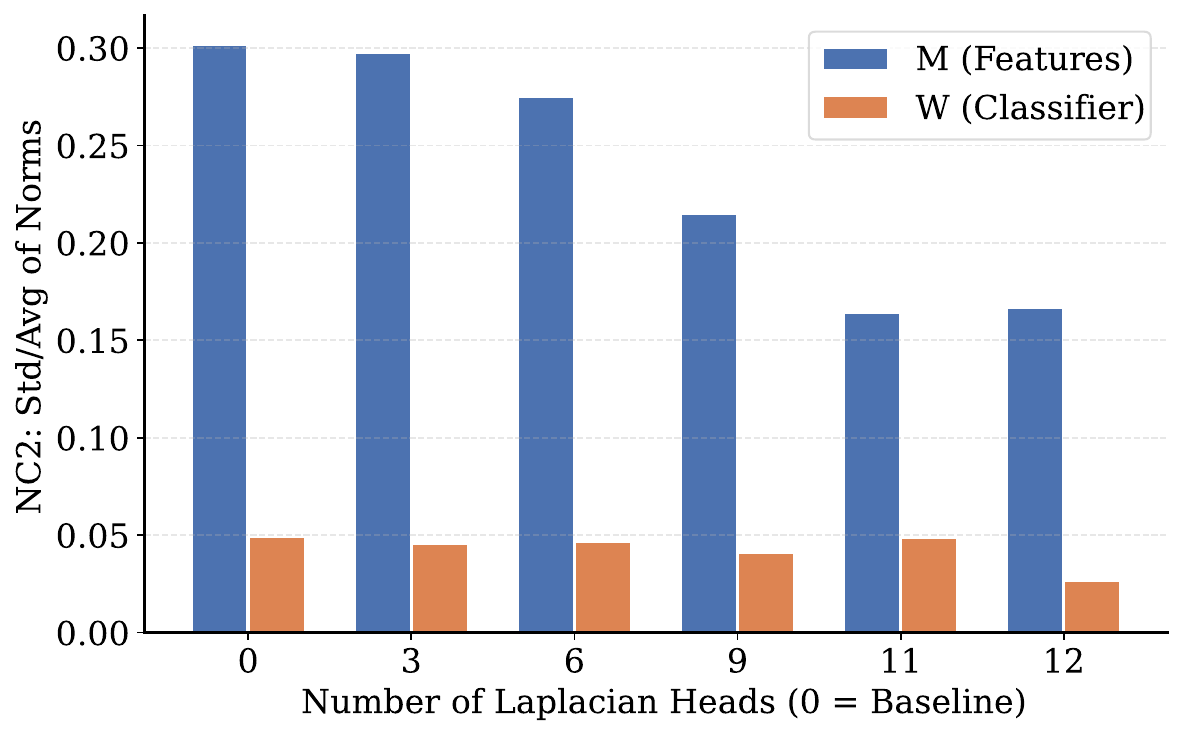}
  \caption{NC2: Equinormness}
\end{subfigure}
\hfill
\begin{subfigure}[b]{0.48\textwidth}
  \centering
  \includegraphics[width=0.9\linewidth]{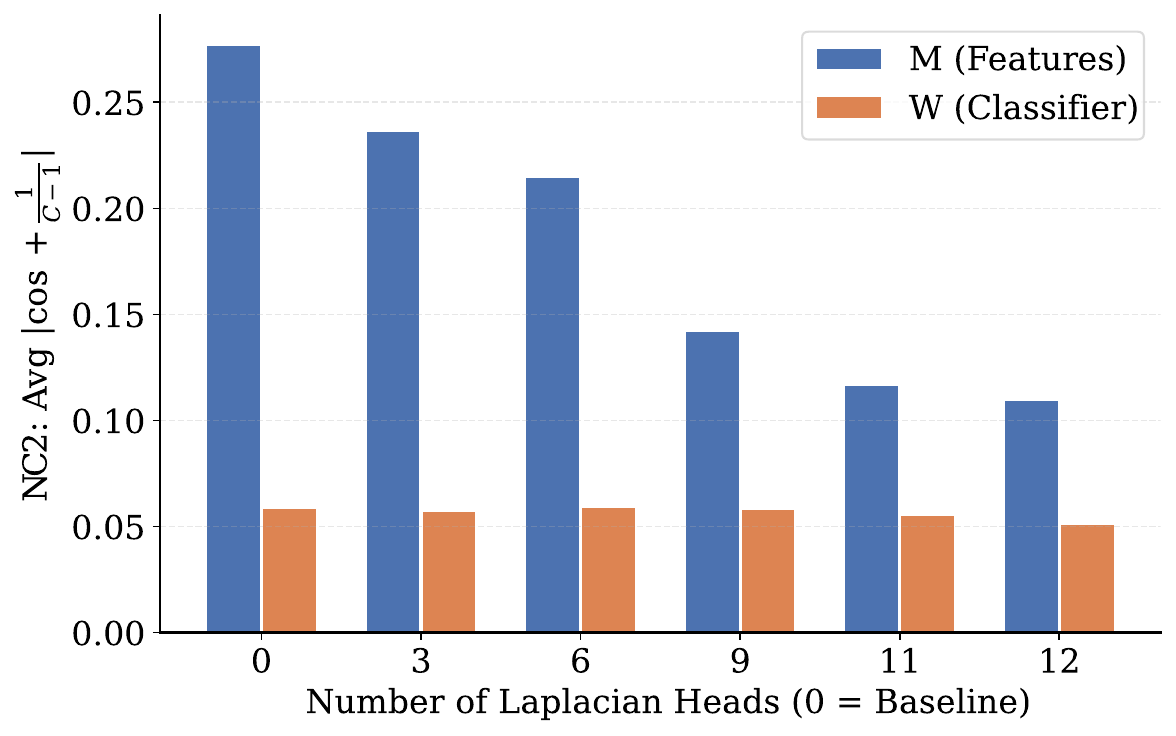}
  \caption{NC2: Max Equiangularity}
\end{subfigure}

\vspace{-0.1em}

\begin{subfigure}[b]{0.48\textwidth}
  \centering
  \includegraphics[width=0.9\linewidth]{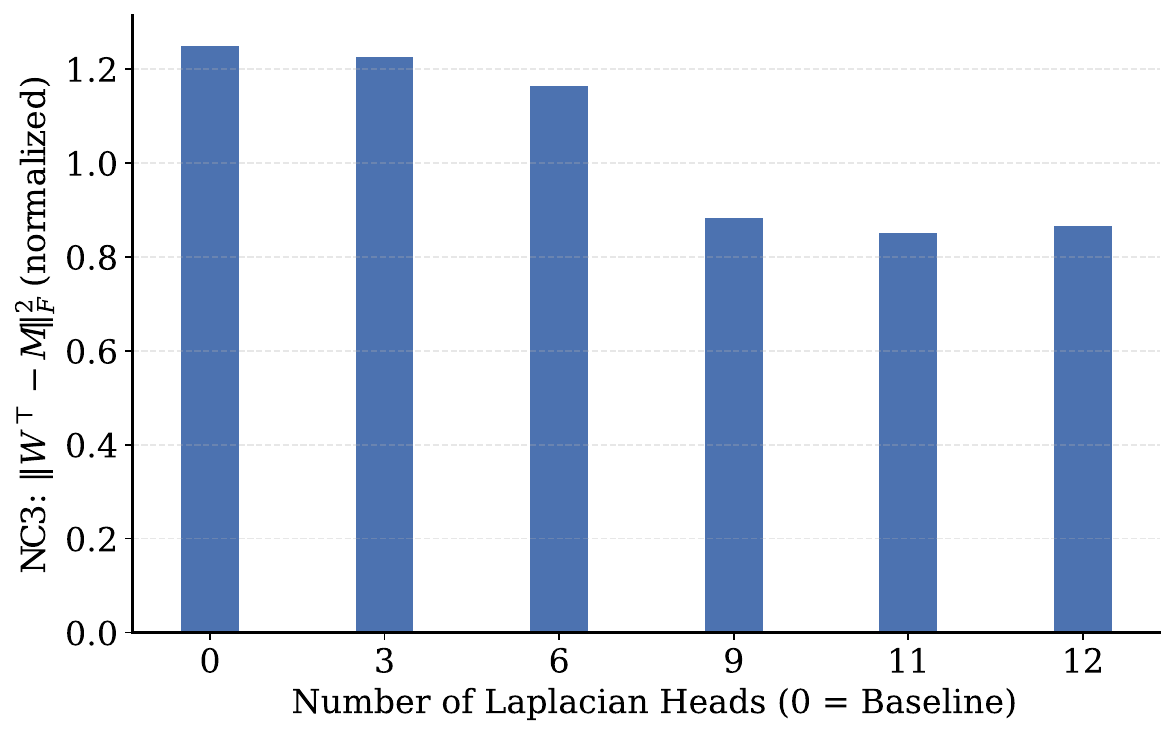}
  \caption{NC3: Self-Duality}
\end{subfigure}
\hfill
\begin{subfigure}[b]{0.48\textwidth}
  \centering
  \includegraphics[width=0.9\linewidth]{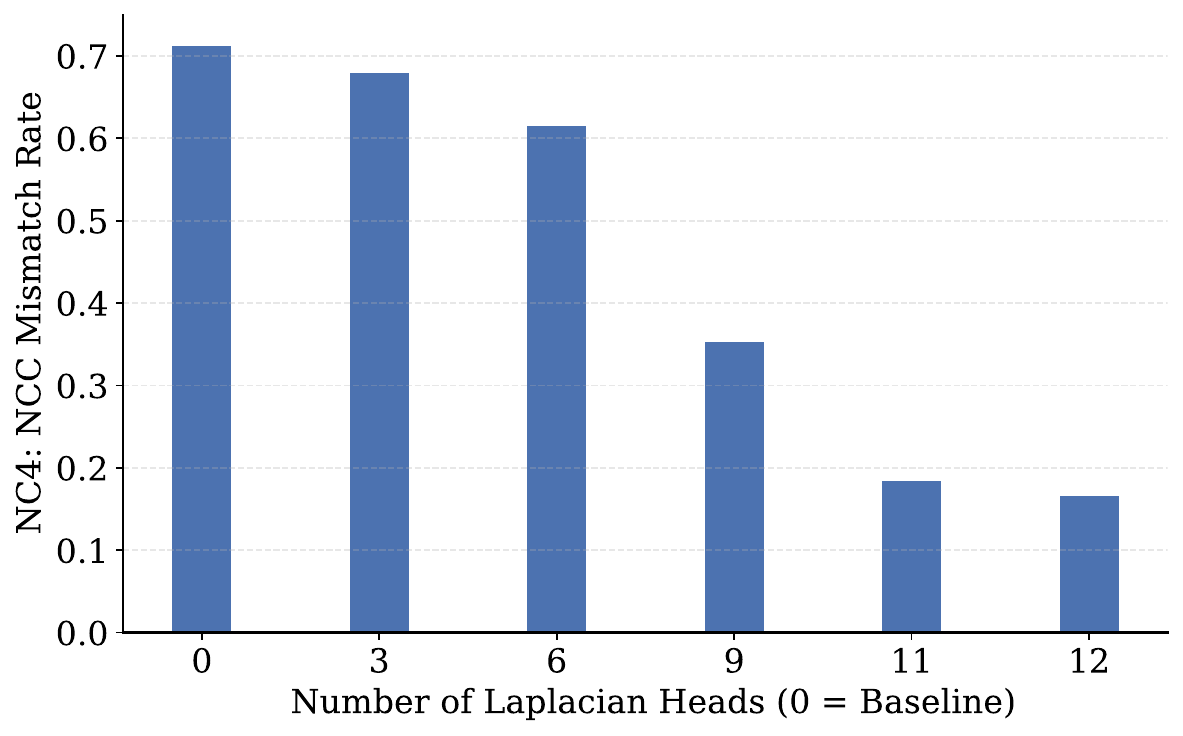}
  \caption{NC4: Convergence to NCC}
\end{subfigure}
\caption{Neural-collapse metrics on CIFAR-100 (a--d). \(M\) denotes the matrix of class means—averaged over all tokens and instances within each class—while \(W\) denotes the classifier weight matrix.}
\label{fig:nc_metrics}
\end{figure}

\begin{figure}[H]
\centering
\begin{subfigure}[b]{0.48\textwidth}
  \centering
  \includegraphics[width=0.9\linewidth]{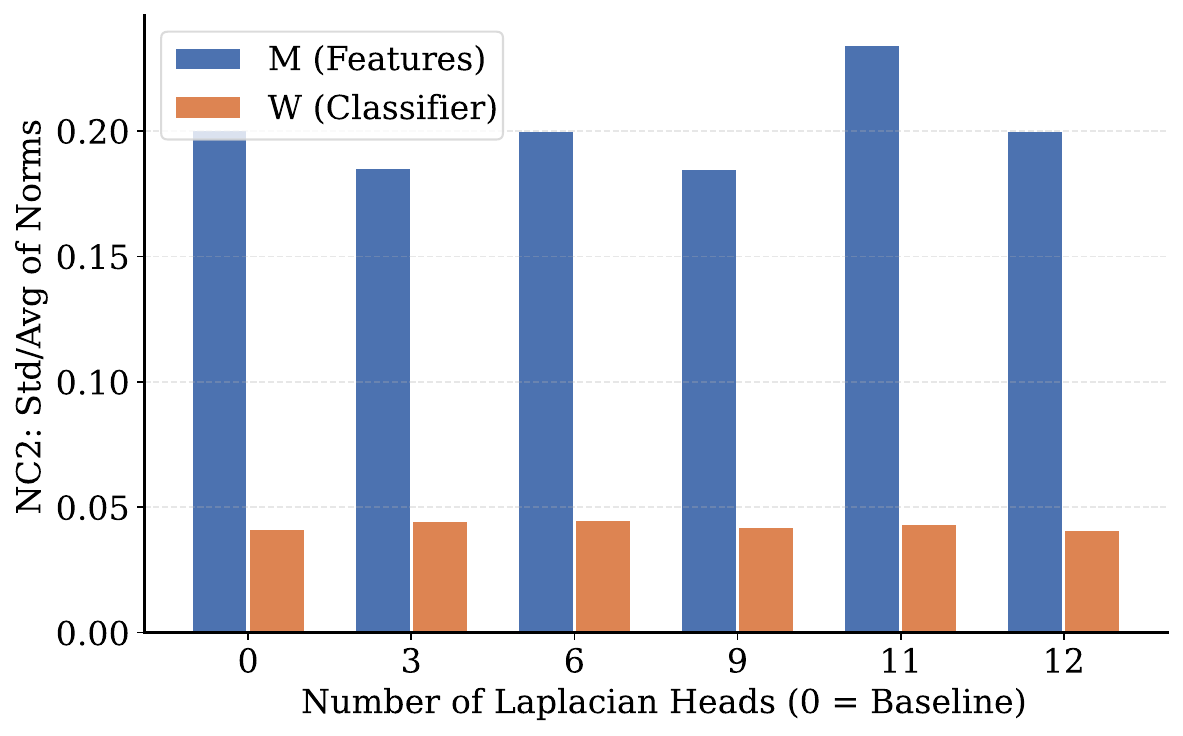}
  \caption{NC2: Equinormness}
\end{subfigure}
\hfill
\begin{subfigure}[b]{0.48\textwidth}
  \centering
  \includegraphics[width=0.9\linewidth]{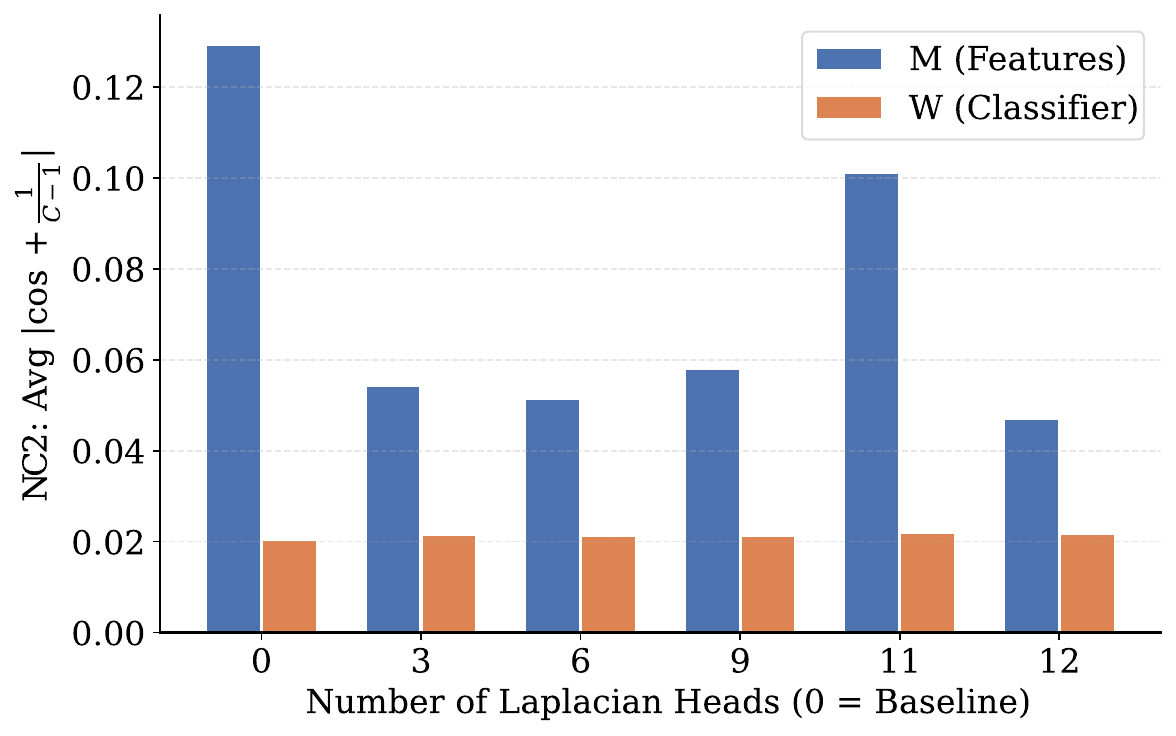}
  \caption{NC2: Max Equiangularity}
\end{subfigure}

\vspace{-0.1em}

\begin{subfigure}[b]{0.48\textwidth}
  \centering
  \includegraphics[width=0.9\linewidth]{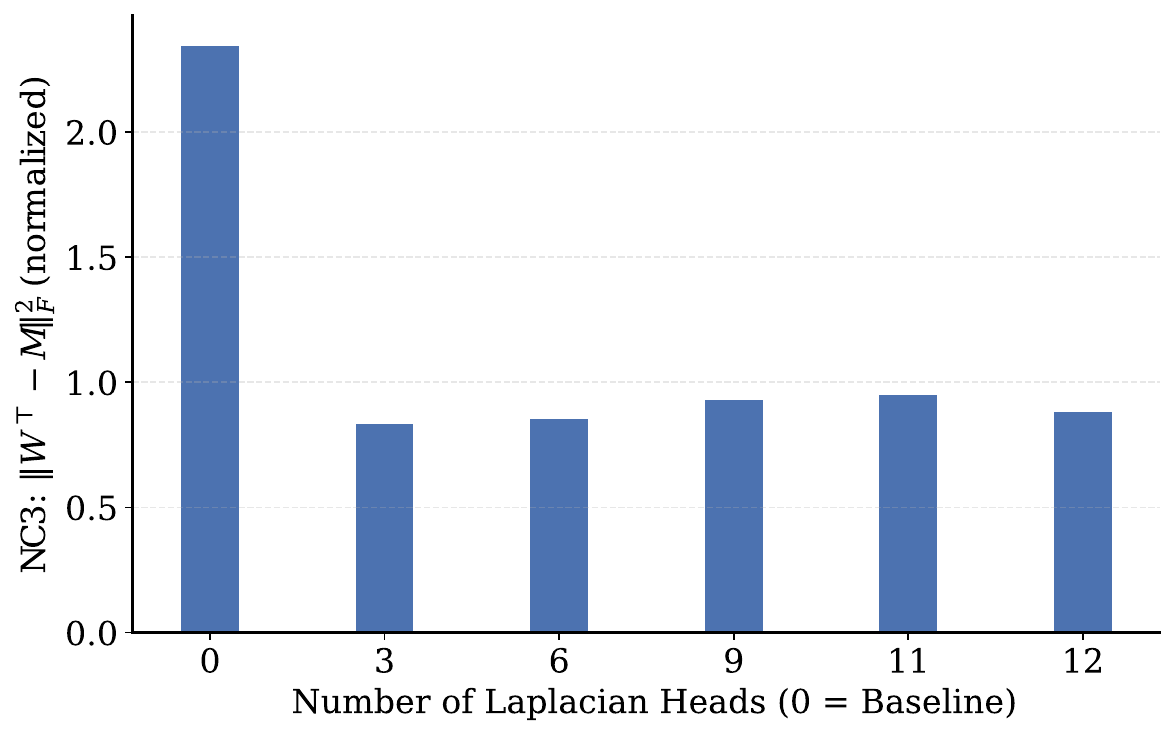}
  \caption{NC3: Self-Duality}
\end{subfigure}
\hfill
\begin{subfigure}[b]{0.48\textwidth}
  \centering
  \includegraphics[width=0.9\linewidth]{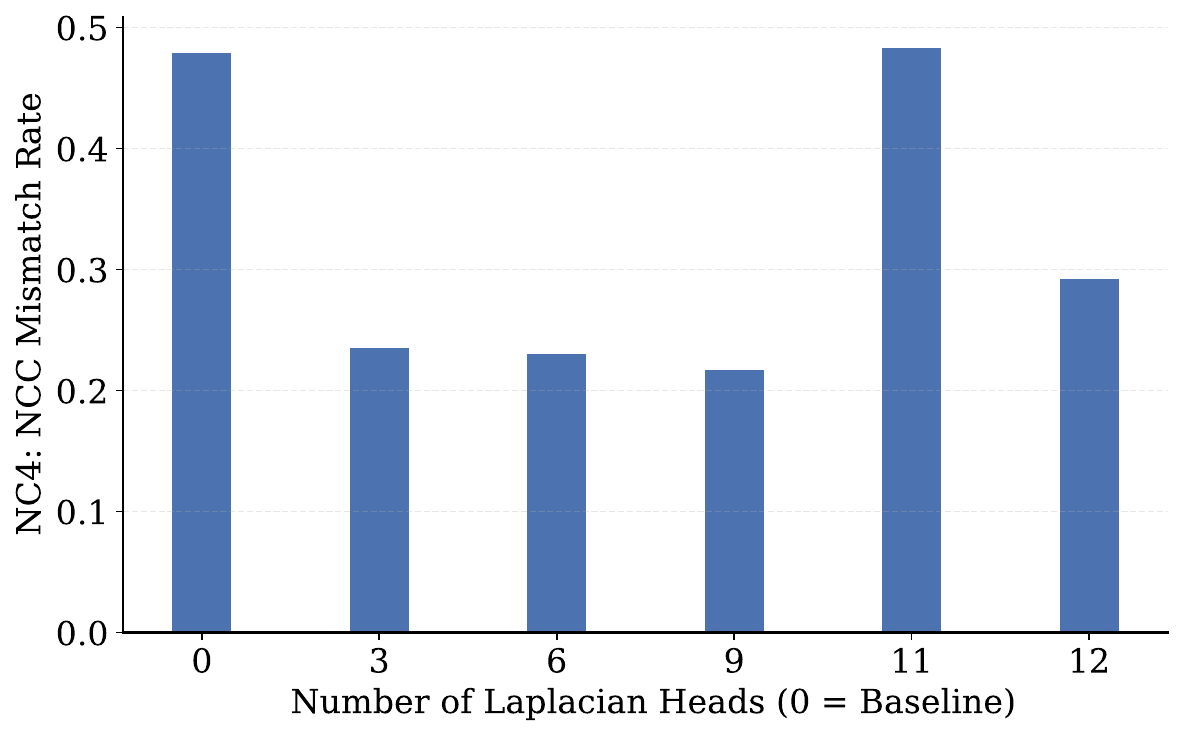}
  \caption{NC4: Convergence to NCC}
\end{subfigure}
\caption{Neural-collapse metrics on ImageNet.}
\label{fig:nc_metrics_imagenet}
\end{figure}

\Needspace{0.30\textheight}
\subsection{Visualization of Projection onto Simplex ETF}\label{app:simplex_proj}
Each token embedding is first projected onto the classifier $W$ for a random subset of three classes, then the result is projected again onto a
two-dimensional representation of a three-dimensional simplex ETF. The result is visualized with each point colored according to its ground truth class. This visualization aims to illustrate the conformity of token representations to a simplex ETF. 
\begin{algorithm}[H]
\caption{Projection of Tokens to a simplex ETF}
\label{alg:simplex_projection}
\begin{algorithmic}[1]
\REQUIRE $X \in \mathbb{R}^{B \times T \times d}, W \in \mathbb{R}^{C\times d}$
\STATE $X \gets \text{reshape}(X, [B \cdot T, d])$, $W'\in \mathbb{R}^{3\times d} \gets \text{random sample}(W)$
\STATE $U, S, V^T = \text{SVD}(\text{normalize($W'$)})$
\STATE $A \gets \sqrt{2} \cdot 
\begin{bmatrix}
\frac{1}{2} & -\frac{1}{2} & 0 \\
0 & 0 & \frac{\sqrt{3}}{2}
\end{bmatrix}
\cdot \left(I_3 - \frac{1}{3} \mathbf{1}\mathbf{1}^\top \right)$
\STATE \textbf{return} $AUV^TX^T$
\end{algorithmic}
\end{algorithm}
\begingroup
\setlength{\intextsep}{5pt}
\setlength{\parskip}{0pt}
\begin{figure}[H]
    \centering
    \includegraphics[width=\linewidth,trim=0 0 0 30,clip]{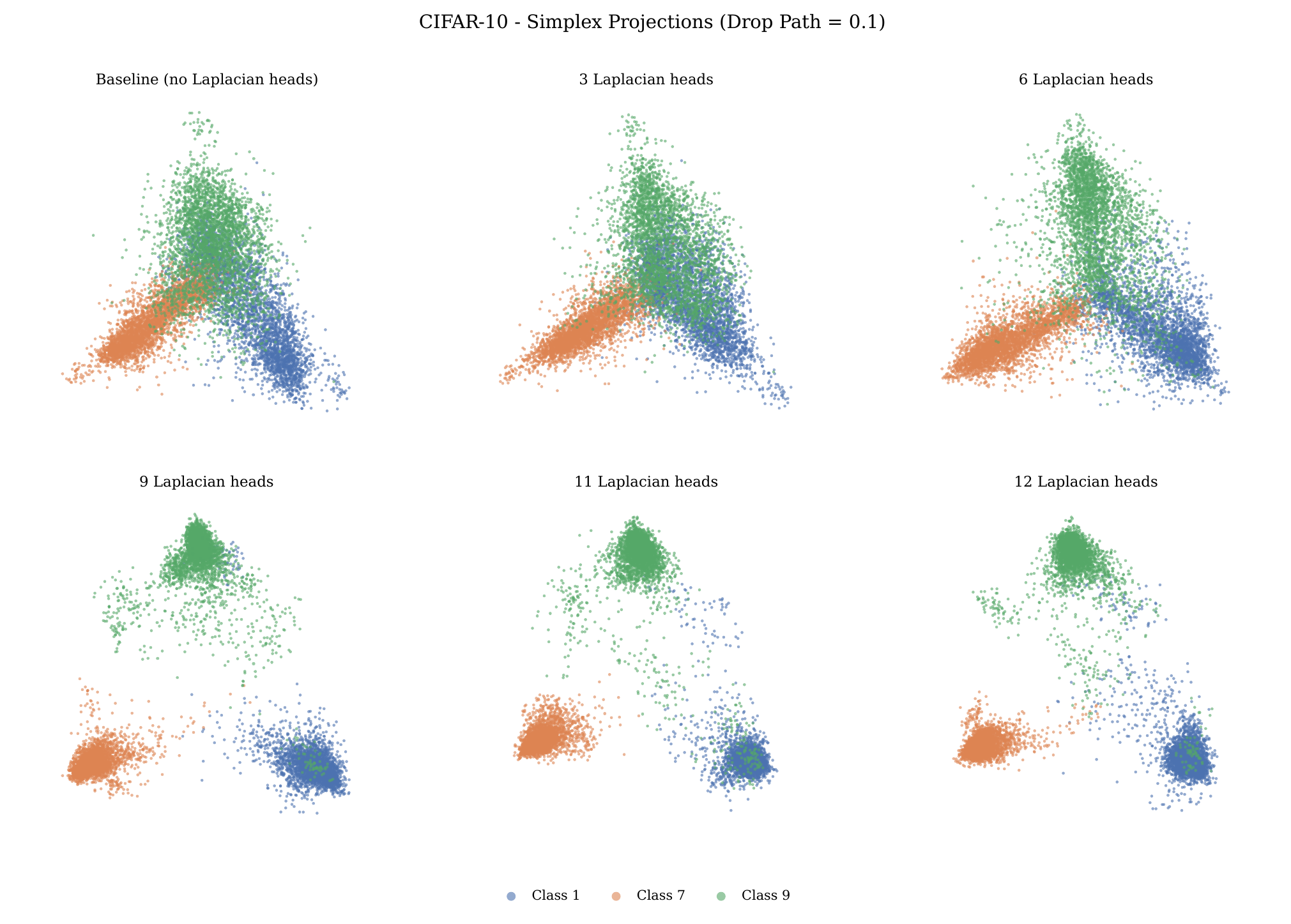}
    \caption{Visualization of Projections on Simplex for CIFAR10}
    \label{fig:simplex_grid_cifar10}
\end{figure}

\begin{figure}[H]
    \centering
    \includegraphics[width=\linewidth,trim=0 0 0 30,clip]{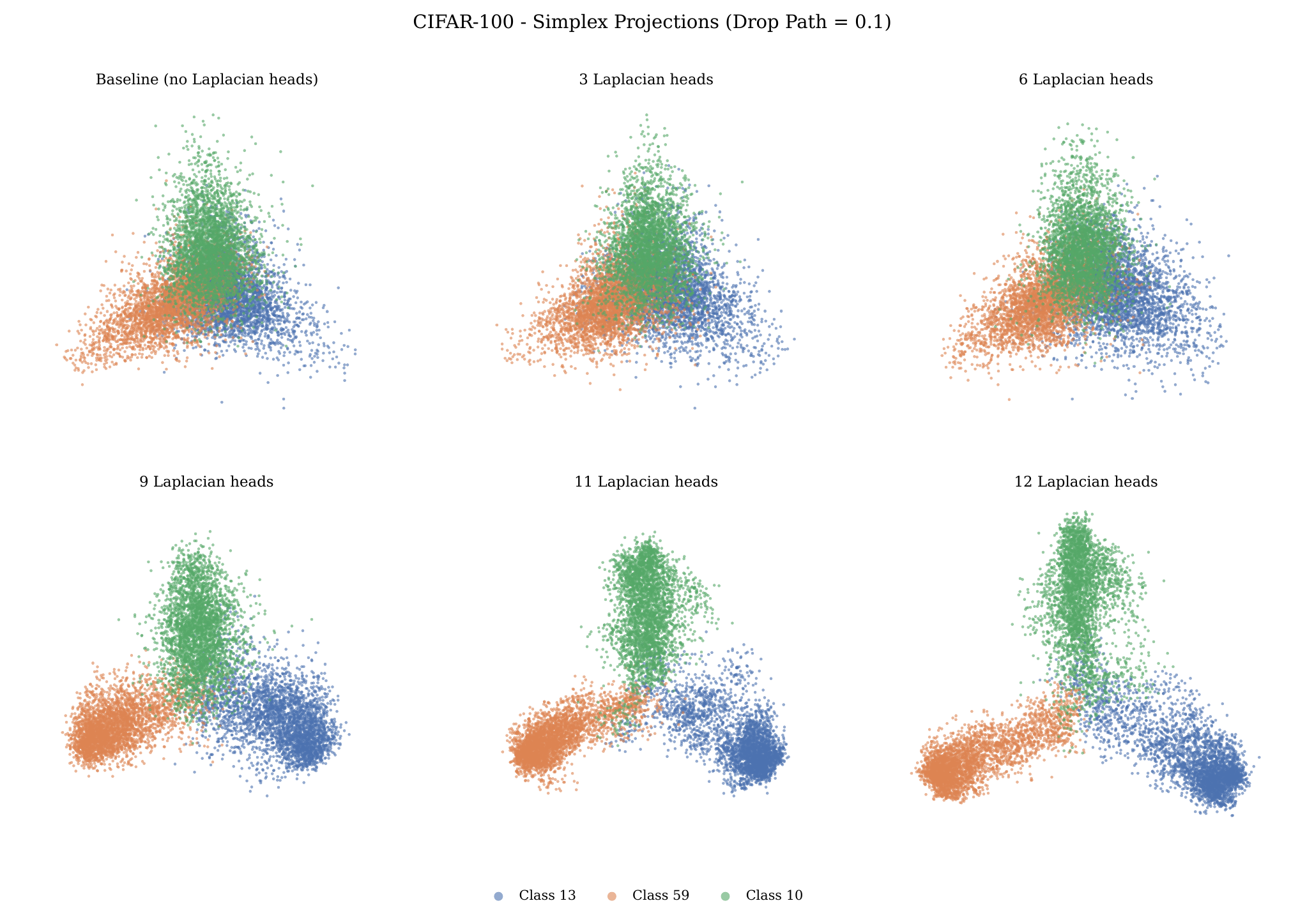}
    \caption{Visualization of Projections on Simplex for CIFAR100}
    \label{fig:simplex_grid_cifar100}
\end{figure}

\begin{figure}[H]
    \centering
    \includegraphics[width=\linewidth,trim=0 0 0 30,clip]{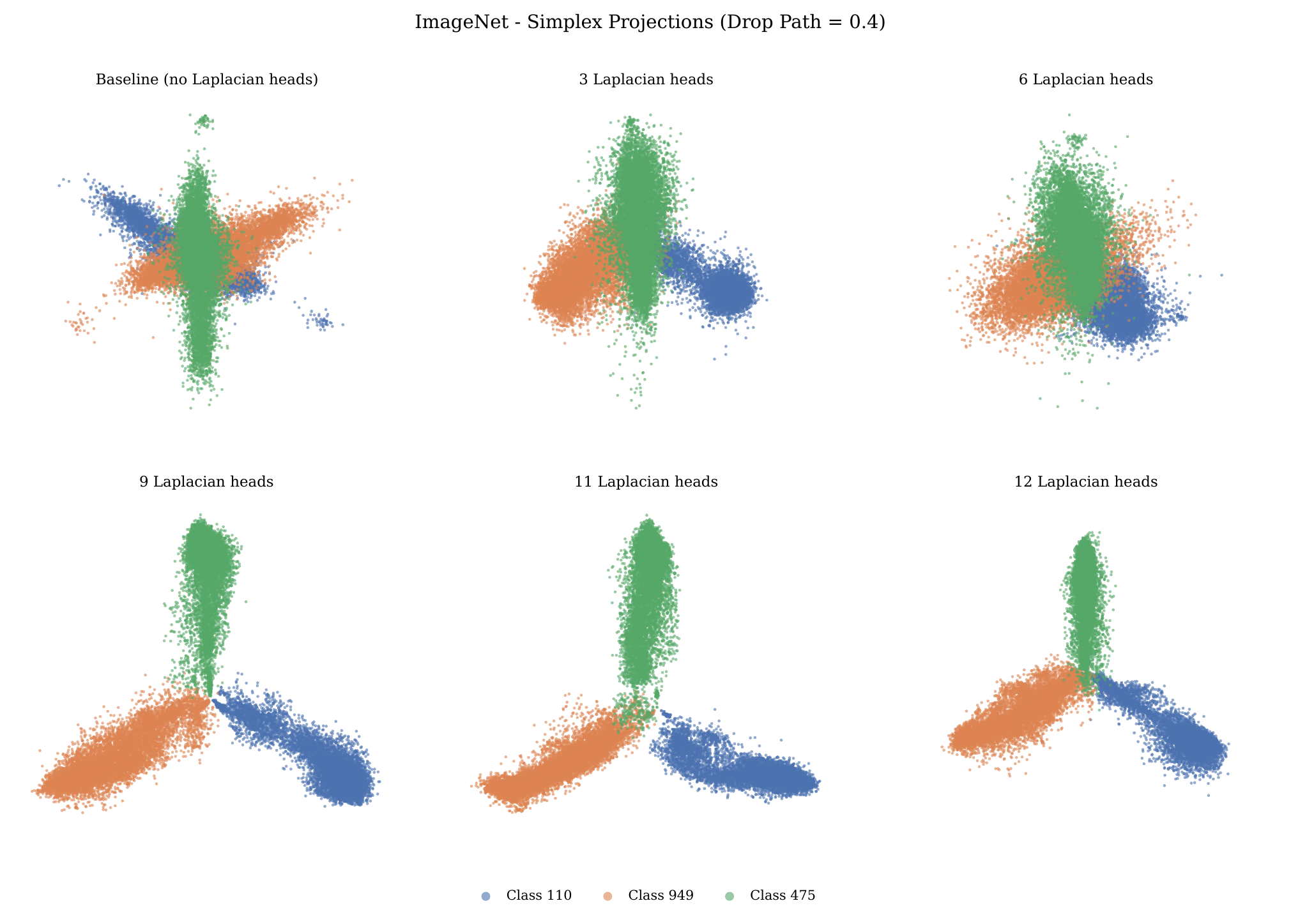}
    \caption{Visualization of Projections on Simplex for ImageNet}
    \label{fig:simplex_grid_imagenet}
\end{figure}
\endgroup


\clearpage

\end{document}